\documentclass[a4paper,12pt,twoside]{report}
\usepackage[left=2cm,right=2cm,top=2cm,bottom=3cm]{geometry}
\usepackage{graphicx}
\usepackage{verbatim}
\usepackage{latexsym}
\usepackage{mathchars}
\usepackage{setspace}
\usepackage[acronym]{glossaries}
\makeglossaries
\newacronym{AI}{AI}{Artificial Intelligence}
\newacronym{COMPAS}{COMPAS}{Correctional Offender Management Profiling for Alternative Sanctions}
\newacronym{EU}{EU}{European Union}
\newacronym{GDPR}{GDPR}{General Data Protection Regulation}
\newacronym{UK}{UK}{United Kingdom}
\newacronym{MCAR}{MCAR}{Missing Completely At Random}
\newacronym{MAR}{MAR}{Missing At Random}
\newacronym{MNAR}{MNAR}{Missing Not At Random}
\newacronym{POP}{POP}{Polynomial Optimisation Problem}
\newacronym{GMP}{GMP}{Generalised Moment Problem}
\newacronym{SDP}{SDP}{Semi-Definite Programming}
\newacronym{NCPOP}{NCPOP}{Non-Commutative Polynomial Optimisation Problem}
\newacronym{NPA}{NPA}{Navascu\'es--Pironio--Ac\'in}
\newacronym{GNS}{GNS}{Gelfand--Naimark--Segal}
\newacronym{LDS}{LDS}{Linear Dynamic System}
\newacronym{AR}{AR}{Auto-Regressive}
\newacronym{ARMA}{ARMA}{Auto-Regressive Moving-Average}
\newacronym{TSSOS}{TSSOS}{Term-Sparsity exploiting moment/SOS}
\newacronym{OT}{OT}{Optimal Transport}
\newacronym{KL}{KL}{Kullback--Leibler}
\newacronym{TV}{TV}{Total Variation}
\newacronym{nrmse}{nrmse}{Normalised Root Mean Square Error}

\usepackage{url}
\usepackage{natbib}

\usepackage{algorithm}
\usepackage{algpseudocode}

\usepackage{amsmath}
\usepackage{amsthm}
\usepackage{amsfonts}
\usepackage{amssymb}
\usepackage{nccmath}
\usepackage{mathrsfs}
\usepackage[bb=boondox]{mathalfa}
\usepackage{enumerate}

\usepackage[short]{optidef}

\usepackage{xcolor}

\usepackage{makecell}
\usepackage{multirow}

\usepackage{appendix}

\usepackage{bibentry}
\nobibliography*

\usepackage{tikz}
\usetikzlibrary{positioning, shapes.geometric}
\usetikzlibrary{calc}
\tikzset{>=latex}

\input{blocked.sty}
\input{uhead.sty}
\input{boxit.sty}
\input{icthesis.sty}

\newcommand{\NN}{{\sf I\kern-0.14emN}}   
\newcommand{\ZZ}{{\sf Z\kern-0.45emZ}}   
\newcommand{\QQQ}{{\sf C\kern-0.48emQ}}   
\newcommand{\RR}{{\sf I\kern-0.14emR}}   

\newcommand{\II}{{\bf I}}

\newcommand{\normallinespacing}{\renewcommand{\baselinestretch}{1.5} \normalsize}

\newcommand{\syncc}{~\stackrel{\textstyle \rhd\kern-0.57em\lhd}{\scriptstyle L}~}

\newtheorem{definition}{Definition}[chapter]
\newtheorem{theorem}{Theorem}[chapter]
\newtheorem{assumption}{Assumption}[chapter]
\newtheorem{lemma}{Lemma}[chapter]
\newtheorem{example}{Example}[chapter]
\newtheorem{remark}{Remark}[chapter]

\DeclareMathOperator{\diagg}{diag}
\newcommand{\diag}[1]{\diagg\left({#1}\right)}

\DeclareMathOperator*{\nrmse}{nrmse}
\DeclareMathOperator*{\mean}{mean}
\DeclareMathOperator*{\tr}{tr}
\DeclareMathOperator*{\trd}{dtr}
\DeclareMathOperator*{\Sym}{Sym}
\DeclareMathOperator*{\cyc}{cyc}
\DeclareMathOperator*{\rk}{rank}

\DeclareMathOperator*{\loss}{loss}

\newcommand{\pp}[1]{\Pr\left[#1\right]}

\newcommand{\pover}[2]{\Pr\left[#1 \mid #2\right]}

\DeclareMathOperator{\suppp}{supp}
\newcommand{\supp}[1]{\suppp(#1)}

\DeclareMathOperator{\kll}{KL}
\DeclareMathOperator{\tvv}{TV}
\newcommand{\kl}[2]{\kll\left({#1}\|{#2}\right)}
\newcommand{\tv}[2]{\tvv\left({#1},{#2}\right)}

\DeclareMathOperator*{\tx}{\Tilde{x}}

\DeclareMathOperator{\prox}{\mathcal{P}}
\DeclareMathOperator{\dom}{dom}

\newcommand{\td}[1]{\Tilde{#1}}

\begin{document}

\title{\LARGE {\bf Optimisation Strategies for Ensuring Fairness in Machine Learning:\\ With and Without Demographics}\\
 \vspace*{6mm}
}

\author{Quan Zhou}
\submitdate{August 1, 2024}
\setcounter{page}{1}
\normallinespacing
\maketitle
\preface   
\pagestyle{plain}
\doublespacing
\begin{center}{
\large \bf Statement of Originality}
\end{center}
\small
\linespread{1.5}
\normalsize
The work is my own and that all else is appropriately referenced.
\par

\vspace{80pt}

\begin{center}{
\large \bf Copyright Declaration}
\end{center}
\small
\linespread{1.5}
\normalsize
The copyright of this thesis rests with the author. Unless otherwise indicated, its contents are licensed under a Creative Commons Attribution-Non Commercial 4.0 International Licence (CC BY-NC).

Under this licence, you may copy and redistribute the material in any medium or format. You may also create and distribute modified versions of the work. This is on the condition that: you credit the author and do not use it, or any derivative works, for a commercial purpose.

When reusing or sharing this work, ensure you make the licence terms clear to others by naming the licence and linking to the licence text. Where a work has been adapted, you should indicate that the work has been changed and describe those changes.

Please seek permission from the copyright holder for uses of this work that are not included in this licence or permitted under UK Copyright Law.
\par

\cleardoublepage 

\begin{acknowledgements}


First and foremost, I am deeply grateful to my PhD advisers, Prof. Robert Shorten and Dr. Jakub Mare\v{c}ek, for their unwavering guidance, encouragement, and expertise. 

I am grateful for the mentorship of Dr. Jakub Mare\v{c}ek, who has been my master thesis advisor at University of Edinburgh, who has shown me the wonders of research during the three-month master thesis project, and who is one of the reasons why I decided to pursue a PhD.
I am likewise grateful for the guidance and support I have received across all aspects of my life during the PhD program. Especially during the challenging period of COVID-19 pandemic, it was often difficult to maintain focus on research and stay positive emotionally, but the mentorship has been a constant source of strength and encouragement. The two-month visit to Czech Technical University in Prague has been really enjoyable.

I am grateful to Prof. Robert Shorten for providing me with the opportunity to pursue a PhD at University College Dublin, as well as the chances to visit and transfer to Imperial College London. 
Thanks to the guidance that I received, I discovered my core area of interest, and constantly found myself ``flowing'' in the research\footnote{Pioneered by Prof. Mihaly Csikszentmihalyi, \textit{Flow} describes a state in which individuals become fully involved in an activity, often to the extent that they are willing to continue despite significant challenges.}, which thrived on the encouragement for independent research, and was nurtured by care for daily life concerns, e.g., housing and health insurance.
My research benefited greatly from the motivation to upskill and the pondering of ``what if'' and ``why'', often ignited by the exceptional vision and insightful comments during our discussions. 
The dedicated attitude has helped me establish the goal of ``going deeper'', and continuous reinforcement of this belief has brought out the best in me.

I extend my heartfelt appreciation to my colleagues and collaborators 
Aida Manzano Kharman, Christian Jursitzky,  Pietro Ferraro, Pierre Pinson, 
Ramen Ghosh,
Mengjia Niu, Xiaoyu He, and Petr Ry\v{s}av\'{y},
for their camaraderie, intellectual discussions, and support. 
I am thankful to the financial support by the Science Foundation Ireland, the European Union’s Horizon Europe research and innovation programme, as well as the Innovate UK under the Horizon Europe Guarantee.
My sincere appreciation goes to my friends and family for their consistent encouragement and love throughout this journey. 

Last but not least, I wish I could contribute something to science.

\end{acknowledgements}
\cleardoublepage 
\addcontentsline{toc}{chapter}{Abstract}

\begin{abstract}
Ensuring fairness has emerged as one of the primary concerns in \acrfull{AI} and its related algorithms. Over time, the field of machine learning fairness has evolved to address these issues. This paper provides an extensive overview of this field and introduces two formal frameworks to tackle open questions in machine learning fairness.

In one framework, operator-valued optimisation and min-max objectives are employed to address unfairness in time-series problems. This approach showcases state-of-the-art performance on the notorious \acrfull{COMPAS} benchmark dataset, demonstrating its effectiveness in real-world scenarios.

In the second framework, the challenge of lacking sensitive attributes, such as gender and race, in commonly used datasets is addressed. This issue is particularly pressing because existing algorithms in this field predominantly rely on the availability or estimations of such attributes to assess and mitigate unfairness. Here, a framework for a group-blind bias-repair is introduced, aiming to mitigate bias without relying on sensitive attributes. The efficacy of this approach is showcased through analyses conducted on the Adult Census Income dataset.

Additionally, detailed algorithmic analyses for both frameworks are provided, accompanied by convergence guarantees, ensuring the robustness and reliability of the proposed methodologies.

\end{abstract}
\cleardoublepage
\listoffigures
\listoftables
\singlespacing
\printglossary[title=List of Acronyms, type=\acronymtype]
\tableofcontents
\cleardoublepage
\pagenumbering{arabic}
\pagestyle{plain}
\chapter{Introduction}
\label{cha:introduction}


\begin{quote}
\textit{This chapter discusses the urgent need for fairness amidst the proliferation of machine-learning applications. 
In particular, we outline the emerging recognition of fairness in both AI regulation and academia.
Then, we put forth the contributions of the thesis within this context.
}
\end{quote}

Citizens across the world are becoming more and more aware of discrimination that can arise due to the poor design of algorithms that guide our everyday lives. Examples of algorithms that affect citizens range from benign recommendation systems, to the less benign machine learning algorithms that make decisions that affect the lives of citizens in profound ways (decisions on bank loans, school admissions, etc.). Consequently, citizens, politicians, and societies, are becoming very concerned with potential algorithmic bias and unfairness, and are driving demand for certification and audibility of algorithmic systems\footnote{See \url{https://fairnessfoundation.com/}}.
Unfairness not only undermines the health of democracies but, from a societal standpoint, emotional reactions to unfairness may engender feelings of exclusion, thus endangering cohesion\footnote{This dynamic is explored further in the context of a fair society in the report at \url{https://op.europa.eu/webpub/jrc/fair-society/}}. This, in turn, can lead to reduced productivity and efficiency in societies.
In light of these arguments, a fair society, which places a premium on diversity and strives for inclusivity, actively promotes the acceptance of individuals from varied backgrounds, cultures, and identities. This commitment plays a vital role in bridging educational gaps and facilitating social mobility. As fairness has been identified as a driver of motivation and performance \citep{hancock2018fairness}, a fair society has the capacity to inspire higher levels of productivity and efficiency.

Fairness will become even more important in the coming years in the aftermath of the global economic crisis, triggered by the COVID-19 pandemic. 
The World Bank\footnote{See the ``World Development Report 2022'' at \url{https://www.worldbank.org/en/publication/wdr2022}.} expresses concerns that the recovery is likely to be uneven as its initial economic impacts, with disadvantaged populations and regions facing a prolonged journey towards recovering pandemic-induced losses of livelihoods, thereby exacerbating existing inequalities and poverty.
The International Monetary Fund\footnote{See the article ``Inequality in the Time of COVID-19'' at \url{https://www.imf.org/external/pubs/ft/fandd/2021/06/inequality-and-covid-19-ferreira.htm}.} states that the pandemic has further intensified preexisting inequalities in the labour market, largely because the ability to work remotely is highly correlated with education, and hence with pre-pandemic earnings. Job and income losses are likely to have disproportionately affected lower-skilled and uneducated workers, particularly those in informal labour without access to furlough programs or unemployment insurance. This has forced hundreds of millions of such workers to make daily trade-offs between staying safe at home and risking infection to provide food for their families.
Moreover, the burden of additional time required for childcare and housework, falling disproportionately on women, has likely widened gender inequality in earnings.

In the \acrfull{UK}, Public Health England\footnote{See the report ``COVID-19: review of disparities in risks and outcomes'' at \url{https://www.gov.uk/government/publications/covid-19-review-of-disparities-in-risks-and-outcomes}.} has confirmed that COVID-19 has not only mirrored existing health inequalities but has, in some cases, worsened them. Individuals from Black, Asian, and other Minority Ethnic groups faced higher exposure, diagnosis rates, and mortality from COVID-19.
As investigated by Local Government Association\footnote{See multiple reports at \url{https://www.local.gov.uk/our-support/safer-and-more-sustainable-communities/health-inequalities-hub}} highlights interconnected factors such as deprivation, low income, minority ethnicity, and poor housing, all contributing to an increased risk of COVID-19, with pre-existing health inequalities unfairly disadvantage individuals in their ability to survive the pandemic.

It is in this context that the research in this thesis began. The focus is on fairness in machine learning, with two main directions of contributions. 

The first set of contributions focuses on the development of technical tools in the optimisation context.
More specifically, we introduce two algorithms to solve fairness problems that arise in applications of machine learning. 
The first algorithm focuses on learning for linear dynamic systems, innovatively eliminating the need for assumptions about the dimension of hidden states. 
The second algorithm introduces a group-blind projection map capable of projecting two distinct datasets into similar ones, innovating by eliminating the requirement for group membership information for each data point.
The second set of contributions is the applications of the aforementioned technical tools to several applications.

\section{Societal Impacts of Machine Learning}

Machine learning, or the broader concept, \acrshort{AI}, is already bringing significant social and economic advantages to individuals, ranging from medical diagnostics, autonomous vehicles to virtual assistants such as Apple Siri and Amazon Alexa. Recent progress in generative AI, exemplified by OpenAI ChatGPT, provides a preview of the vast opportunities awaiting us in the near future. Many of us are starting to grasp the transformative potential of AI as the technology advances rapidly.

However, as machine learning becomes more prevalent, especially in sensitive or human-related applications, it is crucial to carefully examine its impacts on society. 
A notorious example is the \acrshort{COMPAS} system.
It has been used by the U.S. states to assess the likelihood of a defendant re-offending.
According to the COMPAS Practitioner's Guide \footnote{Northpointe (March 2015). ``A Practitioner's Guide to COMPAS Core'' at \url{https://s3.documentcloud.org/documents/2840784/Practitioner-s-Guide-to-COMPAS-Core.pdf}.}, the risk scales for predicting general and violent recidivism, and for pretrial misconduct, were designed using behavioural and psychological constructs ``of very high relevance to recidivism and criminal careers''.
In 2016, a nonprofit organisation, ProPublica investigated the algorithm and found that ``African-American defendants are almost twice as likely as Caucasian defendants to be labelled a higher risk but not actually re-offend,'' whereas COMPAS ``makes the opposite mistake among Caucasian defendants: They are much more likely than African-American defendants to be labelled lower-risk but go on to commit other crimes'' \citep{angwin2016machine}. 

The impact of unfair machine learning applications on society is significant, yet identifying and mitigating the unfairness is a complex challenge.
It has been well understood \citep{mhasawade2021machine,barocas2016big,calders2013unbiased} that seemingly neutral machine learning models and algorithms are capable of disproportionately negative effects on certain individuals or groups based on their membership in a protected class or demographic category, even if no sensitive attribute is used. While their use in other settings can be judged fair and there may be no explicit intention to discriminate, these models can still lead to biased outcomes. 


There is vast array of risks associated with the use of AI, including those associated with unrepresentative training data, including outcomes-biases in the data, bias in the algorithms, and bias in user interactions with algorithms. 
The first-mentioned risk entails biases in data that skew what is learned by machine learning algorithms. These are designed to learn and replicate historical patterns in the data, even if such patterns are biased (e.g., if historically men were hired more frequently than women for technical positions).
For example, if historical data shows a bias in hiring, such as a higher frequency of men being hired for technical positions than women, the algorithm may perpetuate this bias in its predictions or decisions.
Secondly, algorithms may work to prevent themselves from making fair decisions, even when the data is unbiased. 
Consider two sub-populations: a privileged one and an unprivileged one, where unprivileged candidates may outperform privileged candidates with similar scores, possibly due to facing greater challenges. Relying solely on SAT scores for hiring decisions without considering candidates' backgrounds in this scenario could lead to excluding high-potential unprivileged candidates. 
Thirdly, biased outcomes might impact user experience, thus generating a feedback loop between data, algorithms, and users that can reinforce or even amplify existing bias. For instance, if additional police officers are deployed to an area predicted as high-risk for crime, the subsequent increase in arrests within that area will further heighten the risk prediction according to the model \citep{ensign2018runaway}.

\section{Initiatives to Address Bias}


To address the potential societal impact of AI technologies, there has been a remarkable surge of interest in studying fairness in machine learning within the academic community. Once a niche topic, fairness has evolved into a major subfield of machine learning \citep{chouldechova2020snapshot}.
Fair algorithms are designed to prevent biased and discriminatory results, safeguard individuals and groups, foster trust, adhere to legal and ethical standards, and ultimately enhance the impact of machine learning systems \citep{mehrabi2021survey, osoba2017intelligence}. This commitment to fairness is not merely a short-term goal; it is essential for the long-term sustainability of machine learning applications. In the next chapter, we will discuss the current state of the art in academia.

At the same time, regulatory policies have emerged to address AI fairness in the \acrfull{EU}, as well as in the United States of America and other regions, following the success of ChatGPT.
Regulatory awareness plays a crucial role in establishing trust in AI among users, stakeholders, and the general public.
When people believe that systems treat them fairly, they are more likely to accept and  embrace these technologies.

The \acrfull{GDPR}\footnote{See ``General Data Protection Regulation'' at \url{https://gdpr-info.eu/}}, effective since May 25, 2018, is a cornerstone of EU privacy and human-rights law. Enshrined in Article 5, it establishes fundamental principles such as lawfulness, fairness, transparency, purpose limitation, data minimisation, accuracy, storage limitation, integrity and confidentiality, and accountability.

In March 2023, the \acrshort{UK} Government introduced its AI White Paper\footnote{See ``A pro-innovation approach to AI regulation'' at \url{https://www.gov.uk/government/publications/ai-regulation-a-pro-innovation-approach/white-paper}}. This emphasises setting expectations for AI development and usage. The government empowers existing regulators to issue guidance and regulate AI within their jurisdiction.
The approach is rooted in five principles for responsible AI development across all sectors: safety, security, and robustness; appropriate transparency and explainability; fairness; accountability and governance; and contestability and redress.

Policy attention on fairness in AI is evident in the recently published AI Act of the \acrshort{EU}.
In December 2023, the EU Parliament provisionally agreed with the Council on the AI Act, marking the first regulation on artificial intelligence
\footnote{See the ``Artificial Intelligence Act'' at \url{https://www.europarl.europa.eu/news/en/headlines/society/20230601STO93804/eu-ai-act-first-regulation-on-artificial-intelligence}}.
The AI Act adopts a risk-based approach, categorising AI systems based on risk levels and imposing regulations accordingly. The categories include unacceptable risk, high risk, limited risk, and minimal risk, as well as categories specific for general-purpose AI systems.  
The legislation prioritises five aspects: safety, transparency, traceability, non-discrimination, and environmental impact.


Furthermore, in numerous applications, such as those involving AI's impact on financial markets, machine learning is subject to specific regulations tailored to the field. For instance, in financial services, fairness is pivotal to consumer-credit regulations and ensuring compliance with environmental, social, and governance criteria \citep[ESG]{ESG2015}. This alignment, in turn, promotes sustainable business practices and contributes to a positive societal impact.

\section{Contributions of the Thesis}
This section outlines the research conducted on fairness in machine learning throughout the course of this PhD program. Each point of contribution is then positioned within the framework of machine learning fairness (provided later in Chapter~\ref{cha:overview-MLfairness}), as outlined below.

\paragraph{Proper Learning of Linear Dynamic Systems (Chapter~\ref{cha:tac}):} There has been much recent progress in time series forecasting and estimation of system matrices of \acrfull{LDS}. 
We present an approach to both problems based on an asymptotically convergent hierarchy of convexifications of a certain non-convex operator-valued problem, which is known as \acrfull{NCPOP}. We present promising computational results,
including a comparison with methods implemented in Matlab System Identification Toolbox.
This is published in the IEEE Transactions on Automatic Control \citep{zhou2023learning}.
        \begin{itemize}
            \item \textbf{Problem:} Identification of \acrshort{LDS}.
            \item \textbf{Innovation:} This method operates without assumptions about the dimensions of hidden dynamics, accommodating polynomial shape constraints with global optimality guarantees. The complexity can be mitigated through the exploitation of sparsity.
            \item \textbf{Methodology:} \acrshort{NCPOP} (cf. Appendix~\ref{sec:ncpop}).
        \end{itemize}

\paragraph{Fairness in Forecast for Imbalanced Data (Chapter~\ref{cha:jair}):} In machine learning, training data often capture the behaviour of multiple subgroups of some underlying human population.
This behaviour can often be modelled as observations of an unknown dynamical system with an unobserved state.
When the training data for the subgroups are not controlled carefully,  however, under-representation bias arises.
To counter under-representation bias, we introduce two natural notions of fairness in time-series forecasting problems: subgroup fairness and instantaneous fairness.
These notions extend predictive parity to the learning of dynamical systems.
We also show globally convergent methods for the fairness-constrained learning problems using hierarchies of convexifications of non-commutative polynomial optimisation problems. 
We also show that by exploiting sparsity in the convexifications, we can reduce the run time of our methods considerably.
Our empirical results on a biased dataset motivated by insurance applications and the well-known COMPAS dataset demonstrate the efficacy of our methods.
This is published in the Journal of Artificial Intelligence Research \citep{zhou2023fairness}, with a preliminary version presented in the AAAI Conference on Artificial Intelligence \citep{zhou2021fairness}.
        \begin{itemize}
            \item \textbf{Problem:} Imbalanced data (cf. Section~\ref{sec:imbalanced data}).
            \item \textbf{Innovation:} We present two fairness notions within the forecasting context. The global optimality proposed formulations rely on the technologies of \acrshort{NCPOP}.
            \item \textbf{Methodology:} Proper Learning of \acrshort{LDS} (cf. Chapter~\ref{cha:tac}).
        \end{itemize}

\paragraph{Group-Blind Transport Map for Unavailable Sensitive Attributes (Chapter~\ref{cha:ot}):}
Fairness plays a pivotal role in the realm of machine learning, particularly when it comes to addressing groups categorised by sensitive attributes, e.g., gender, race. 
Prevailing algorithms in fair learning predominantly hinge on accessibility or estimations of these sensitive attributes.
We design a single group-blind projection map that aligns the feature distributions of both groups in the source data, achieving (demographic) group parity, without requiring values of the sensitive attribute for individual samples in the computation of the map, as well as its use.
Instead, our approach utilises the feature distributions of the privileged and unprivileged groups in a boarder population and the essential assumption that the source data are unbiased representation of the population.
We present numerical results on synthetic data and the well-known Adult Census Income dataset. This is under review  and can be found on arXiv \citep{zhou2023group}.
        \begin{itemize}
            \item \textbf{Problem:} Unavailability of sensitive attributes (cf. Section~\ref{sec:unvailable sa}).
            \item \textbf{Innovation:}
            We design a single group-blind projection map. The map aligns feature distributions of both majority and minority groups in source data, achieving demographic parity \citep{dwork2012fairness} without requiring values of the protected attribute for individual samples in the computation or use of the map. 
            \item \textbf{Methodology:} Dykstra's algorithm with Bregman projections \citep{bauschke2000dykstras}.
        \end{itemize}

\section{Structure of the Thesis}

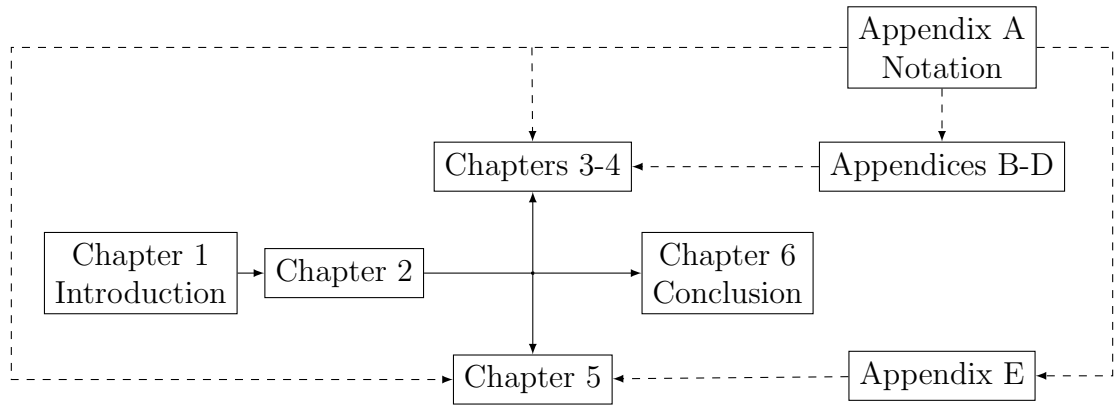
\begin{figure}[htp]
\centering
\begin{tikzpicture}[node distance=10pt]
  \node[draw,align=center] (start)   {Chapter 1\\ Introduction};
  \node[draw, right=of start] (cha 2)  {Chapter 2};
  \node[draw, circle,fill,inner sep=0.1pt, right=40pt of cha 2] (middle)  {};
  \node[draw, above=30pt of middle] (cha 3)  {Chapters 3-4};
  \node[draw, below=30pt of middle] (cha 5)  {Chapter 5};
  \node[draw, right=40pt of middle, align=center] (cha 6)  {Chapter 6\\Conclusion};
  \node[draw, right=70pt of cha 3] (app B)  {Appendices B-D};
  \node[draw, below=60pt of app B] (app E)  {Appendix E};
  \node[draw, above=20pt of app B,align=center] (app A)   {Appendix A\\Notation};
  
  \draw[->] (start)  -- (cha 2);
  \draw[-] (cha 2) -- (middle);
  \draw[->] (middle) -- (cha 3);
  \draw[->] (middle) -- (cha 5);
  \draw[->] (middle) -- (cha 6);
  \draw[->,dashed] (app B) -- (cha 3);
  \draw[->,dashed] (app E) -- (cha 5);
  \draw[->,dashed] (app A) -- (app B);
\draw[->,dashed] (app A) -- ($(app A.west) + (-11,0)$) node[anchor=south] {} |- (cha 5);
\draw[->,dashed] ($(app A.west) + (-4.15,0)$) -- (cha 3);
\draw[->,dashed] (app A) -- ($(app A.east) + (1,0)$) node[anchor=south] {} |- (app E);
\end{tikzpicture}
    \caption{Structure of the chapters in the thesis}
    \label{fig:chapter-structure}
\end{figure}

\begin{itemize}
\item\textbf{Chapter~\ref{cha:introduction}: Introduction} 
\begin{itemize}
\item Initiate a discussion on the societal impact of AI's popularity, and present the efforts of governments and academia to address the associated concerns.
\item Outline the contributions of the thesis within this context.
\item Include a publication list.
\end{itemize}
\item\textbf{Chapter~\ref{cha:overview-MLfairness}: Preliminaries in Machine Learning Fairness} 
\begin{itemize}
\item Introduce the state-of-the-art of machine learning fairness.
\item Outline the current open questions and common strategies.
\end{itemize}

\item\textbf{Chapters~\ref{cha:tac}-\ref{cha:jair}: Learning from Imbalanced Data} 
\begin{itemize}
\item Chapter~\ref{cha:tac}: Introduce the method for proper learning of \acrshort{LDS}, i.e., applying \acrshort{NCPOP} to system identification.
\item Chapter~\ref{cha:jair}: Present the method for fairness in forecast from imbalanced data, i.e., applying proper learning of \acrshort{LDS} to fair learning.
\item Reference Section~\ref{sec:imbalanced data} for a discussion on imbalanced data.
\end{itemize}

\item\textbf{Chapter~\ref{cha:ot}: Learning with Unavailability of Sensitive Attributes} 
\begin{itemize}
\item Present the work for learning with the unavailability of sensitive attributes.
\item Reference Section~\ref{sec:unvailable sa} for a discussion on unavailable sensitive attributes.
\end{itemize}
\item \textbf{Chapter~\ref{cha:conclusions}:} Concluding Remarks and Future Plans

\item \textbf{Appendices:} 
\begin{itemize}
\item{Appendix~\ref{app:notation}: Notation lists.}
\item{Appendix~\ref{app:sdp}:} Provide the fundamentals of \acrfull{SDP}, serving as a preliminary for \acrfull{GMP}.
\item{Appendix~\ref{cha:mom_app}-\ref{cha:ncpop}:} Provide the fundamentals of \acrshort{GMP}, its relation to \acrfull{POP} and their extension to \acrshort{NCPOP}, serving as a preliminary for Chapters~\ref{cha:tac}-\ref{cha:jair}.
\item{Appendix~\ref{app:ot}:} Include the proofs of some lemmas used in Chapter~\ref{cha:ot}.
\end{itemize}
\end{itemize}

\section{Publications}

\begin{itemize}
\item \bibentry{zhou2023learning}
\item \bibentry{zhou2023fairness}
\item \bibentry{zhou2023subgroup}
\item \bibentry{zhou2021fairness}
\end{itemize}

\color{black}
\cleardoublepage
\chapter{Preliminaries in Machine Learning Fairness}
\label{cha:overview-MLfairness}

\begin{quote}
\textit{After discussing fairness in broad terms, this chapter undertakes a tailored survey of fairness focused on the machine learning domain. It introduces the primary categories of fairness definitions, while also discussing prevalent open questions and common approaches within this field.
}
\end{quote}

As mentioned in the previous chapter, the issue of fairness is very topical, not only in the machine learning community, but also in related areas, such as computer networking, smart cities, and behavioural economics.
While the topic of fairness is broad, encompassing many subject domains, much of recent activities have emerged from the field of machine learning. The objective of this chapter is to provide a formal setting for the discussion of fairness in this machine learning context, and to provide a snapshot of available results and some of the most pressing open questions. 

Generally speaking, unfairness in machine learning, can be attributed to various factors such as historical and systematic discrimination, conscious or unconscious prejudices, diverse forms of statistical bias, human negligence, and insufficient objective conditions. 
This multitude of factors lead to several problems when formulation a precise mathematical definition of fairness.

Fairness is also inherently subjective. While most individuals care deeply about fairness, they vary significantly in their definitions due to differing views on its relative importance. The diversity in perspectives and values within different communities adds complexity in establishing a universal definition of machine learning fairness.
Furthermore, individuals not only care about ``distributive justice'', which pertains to the allocation of resources or opportunities, but also about ``procedural justice'', which involves how decisions are made.

Consequently, a proliferation of proposed definitions of fairness has emerged, with each seeking to confront a variety of potential biases in machine learning practices and design solutions to specific settings within real-life applications.
There has been a widely-recognised foundational fairness framework within the machine learning fairness community \citep{caton2023fairness,pessach2022review,mehrabi2021survey}, which serves as a basis for categorising these definitions.
Building upon this foundational framework, which we will discuss in the sequel, machine learning fairness concepts can be broadened through various applications, i.e., fair classification \citep{zafar2019fairness,barocas2023fairness}, fair prediction \citep{chouldechova2017fair,plevcko2024causal}, fair ranking \citep{10.1145/3533380,10.1145/3533379,yang2016measuring,feldman2015certifying}, fair policy \citep{plecko2023causal,chzhen2021unified,nabi2019learning}, and fair forecasting \citep{chouldechova2017fair,gajane2017formalizing,NIPS20199603,Jeong2021FairnessWI}.
One can also mention in the passing that most problems of operations research have fairness-aware variants formulated, or to be formulated. 

Although the foundational fairness framework offers practical guidance for promoting fairness, real-world scenarios can potentially undermine its effectiveness or even render it obsolete. The encouraging aspect is that, despite the vast range of applications, there are many similarities in the challenges encountered across these diverse scenarios. Later in this chapter, we will explore common challenges that cut across various applications and suggest prevalent methods to address them, with a focus on issues relevant to the upcoming chapters.
\section{Fairness Definitions}

\subsection{Group Fairness Definitions}
\label{sec:group-fairness}


Group fairness is the most commonly utilised category of a notion of fairness and typically involves dividing the population into groups based on sensitive attributes, such as gender, sexual orientation, and race. 
The definitions in this category can be roughly divided into three principles \citep{barocas2023fairness}.
To expound upon these principles, we make the assumption, without loss of generality, that there are sensitive attributes ($S$), ground-truth of target variables ($Y$), and the output of machine learning models ($\hat{Y}$). 
In this COMPAS example mentioned above, $S,Y,\hat{Y}$ could refer to race, actual re-offend status, and predicted probability assigned by the COMPAS model to each defendant.

In the context of group fairness:

\textbf{Independence} requires the output to be independent of sensitive attributes, i.e., $\hat{Y}\perp S$, where the symbol $\perp$ denotes statistically independence.
An illustrative instance is the concept of ``demographic parity'' \citep{dwork2012fairness}, which demands the output of a model is consistent across different demographic groups.
In this principle, half of Caucasian defendants being labelled as low risk would also entail the same proportion for African-American defendants.
The opposite term, ``disparate impact'' \citep{barocas2016big}
denotes divergent outcomes that disproportionately benefit or harm a group of individuals sharing a value of a sensitive attribute more frequently than other groups.

\textbf{Separation} requires the output to be unrelated to sensitive attributes, but conditional on the target variables, i.e., $\hat{Y}\perp S\mid Y$.
For example, ``equal opportunity'' \citep{hardt2016equality} demands whether individuals who are qualified for an opportunity are equally likely to access it regardless of their group membership.
In this principle, if half of Caucasian defendants without recidivism are labelled as low risk, the same proportion applies to African-American defendants without recidivism.
The apparent advantage of this principle lies in the fact that when the output is absolutely accurate, it is also inherently fair in terms of the notion of separation. This implies that there is no trade-off between accuracy and fairness.

\textbf{Sufficiency} is derived from calibration and requires the target variables be independent from sensitive attributes conditional on the model output, i.e., $Y\perp S\mid \hat{Y}$. 
For instance, sufficiency would require that for any given COMPAS score, the recidivism rates in different groups are similar.
We would consider the COMPAS system fair, following this principle, if the proportion of defendants who actually re-offend among those predicted to re-offend by the COMPAS system is equalised across different groups.


\subsection{Individual Fairness Definitions}

Individual fairness focuses on specific pairs of individuals, rather than on a quantity that is averaged over groups \citep{chouldechova2020snapshot}. 

In the context of individual fairness:

\textbf{Fairness through Unawareness} of \cite{dwork2012fairness} requires to not explicitly employ sensitive features when making decisions, following the principle that if we are unaware of sensitive attributes while making decisions, our decisions will be fair. 
It aims to mitigate disparate treatment \citep{barocas2016big}, where deliberate inclusion of variables associated with sensitive attributes leads to disproportionate favouring of a specific class. 
However, this principle has been shown to be insufficient in many cases \citep{cornacchia2023auditing,awwad2020exploring}.

\textbf{Fairness through Awareness} of \cite{dwork2012fairness} requires ``similar individuals should be treated similarly''.
This principle necessitates a similarity metric that accurately reflects the ground truth \citep{petersen2021post,sharifi2019average}, and could be addressed through fair representation learning \citep{mukherjee2020two}.

\textbf{Counterfactual Fairness} operates at the individual level, utilising causal methods to scrutinise whether a decision remains consistent when an individual's sensitive attributes are altered \citep{kusner2017counterfactual}. This concept distinguishes itself from notions based solely on correlations of statistical measures.
Its generalised variant, path-specific fairness \citep{chiappa2019path,nabi2018fair}, specifies the effects of sensitive attributes along certain path in a causal directed acyclic graph.
Causal approaches go beyond the limitations of observational data, facilitating a better understanding of unfairness and its underlying reasons.
A classic example of Simpson's paradox involves a study on gender bias in graduate school admissions at the University of California, Berkeley, in 1973 \citep{bickel1977sex}. 
While the overall admission rates were about 44\% for men and 35\% for women, analysing decisions made by each department separately revealed equalised rates for both groups. The paradox emerged as women tended to apply to more competitive departments with lower admission rates, while men applied to less competitive departments with higher admission rates, resulting in the observed difference in admission rates.

\section{Common Approaches and Open Questions}

Next, we discuss typical obstacles to the fundamental framework and explore commonly employed strategies to overcome them. 
We start from the most direct source of unfairness -- the data itself. 
The bias present in the training data has the potential to permeate into the trained model. This may arise from improper associations between sensitive attributes and other model inputs, the absence of sensitive attributes, or incomplete representation of all protected classes. 
Moreover, if not carefully examined, trained models may introduce new biases or magnify pre-existing ones.
The pursuit of fairness also needs to strike a balance among multiple fairness definitions, as well as finding equilibrium with considerations, such as accuracy and privacy.
While acknowledging that there are many further obstacles to be tackled, we focus on the challenges we address later in the thesis.

\subsection{Proxies of Sensitive Attributes}

Sensitive attributes can exhibit correlations with other attributes (proxies) in the data. For instance, race or religion might be associated with a specific city or neighbourhood within a city. ``Fairness through unawareness'' proves ineffective in practice because bias can persist through these proxies. Even if sensitive attributes are excluded, the model may still attempt to discern patterns connecting the data labels, that express this bias, through the remaining features that correlate with the removed sensitive attribute.

To confront this highly prevalent issue, there are three approaches.
The first approach is fair representation learning, dating back to at least \cite{zemel2013learning}.
This approach seeks intermediate data representations that optimally encode the data (preserving maximal information about individual attributes) while concurrently eliminating any information about membership with regard to sensitive attributes \citep{shui2022fair,zhao2022inherent,creager2019flexibly}.
The second approach focuses on bias repairing schemes \citep{feldman2015certifying}, that involve the modification of features to align the distributions from different groups, making it harder for the algorithm to differentiate among groups. 
Another approach is fair feature selection, which examines the input features used in the decision-making process and assesses how including or excluding certain features would impact outcomes. This aligns with the concept of procedural fairness \citep{grgic2018beyond}.

\subsection{(Partial) Unavailability of Sensitive Attributes}
\label{sec:unvailable sa}
Sensitive attributes are often not readily available due to privacy concerns, ethical considerations, difficulties in measurement\footnote{Some sensitive attributes, e.g., gender identity, are more of a spectrum, can be fluid and influenced by context.}, and legal regulations, as discussed in \cite{andrus2022demographic}.
Many data protection laws, like the General Data Protection Regulation (GDPR) in Europe, impose restrictions on collecting and processing sensitive information.
The unavailability of sensitive attributes, compounded by the prevalence of proxies, poses a key challenge. In contrast, most fairness-aware algorithms assume accessibility to individual-level sensitive attributes, creating a mismatch with real-world data limitations.

This challenge has been partially relaxed to considering sensitive attributes only at training time \citep{oneto2020fairness,jiang2020wasserstein,quadrianto2017recycling,zafar2017fairnessBeyong}. This approach trains a model that is blind to sensitive attributes by employing regularisation during training to penalise performance discrepancies across different groups.
This line of research has recently been extended to require sensitive attributes only in a validation set \citep{elzayn2023estimating, chai2022self} to guide the training process. In the two-stage framework of \cite{liu2021just}, sensitive attributes in the validation set are employed to compute the worst-group validation error, which is then utilised for hyperparameter tuning.

Standard definitions of group fairness based on specific sensitive attributes, possess the practical benefit of being extensively recognised, researched, and are readily accessible in open-source toolkits.
In practical scenarios, certain approaches aim to ensure fairness towards unspecified minority groups, often using proxies, or some noisy estimates.
\cite{amini2019uncovering,oneto2019taking} introduce the latent variables as the proxies of unknown sensitive attributes.
Additionally, \cite{sohoni2020no} explore clustering for a similar purpose. 
\cite{zhao2022towards} minimises the correlation between proxies and model predictions to achieve fair classifier learning. 
Some methods employ distributional robust optimisation techniques to minimise the worst-case expected loss for groups identified by ambiguous sensitive attributes \citep{wang2020robust,sohoni2020no}, or by regions where the model incurs (computationally-identifiable) errors \citep{lahoti2020fairness,veldanda2023hyper}.


Further along the road, some fairness definitions that does not focus on groups may better suit specific contexts, while necessitating more specialised expertise. 
``Fairness through awareness'' could be one example that requires context-specific definitions of similarity among individuals.
\cite{liu2023group} avoids any form of group memberships and uses only pairwise similarities between individuals to define inequality in outcomes, using the common property of social networks that individuals sharing similar (sensitive as well as non-sensitive) attributes are more likely to be connected to each other than individuals that are dissimilar.

Additionally, a survey in ``Fairness Without Demographic Data'' \citep{ashurst2023fairness} also discusses trusted third parties and cryptographic solutions to encourage the collection of sensitive attributes. These methods, e.g., \cite{veale2017fairer}, focus on protecting sensitive attributes from misuse or use without informed consent, while also offering guidance or assessment for AI developers.

\subsection{Imbalanced Data}
\label{sec:imbalanced data}

In the domain of imbalanced learning \citep{kaur2019systematic,he2013imbalanced,hoens2013imbalanced}, the presence of imbalanced data poses challenges across various realms of real-world research within machine learning. The inherent skewness in the distribution of primary data often introduces a bias favouring one group over another, resulting in representation bias. When the primary training objective prioritises overall accuracy without considering groups, established models tend to optimise for better accuracy in the majority group, consequently leading to sub-optimal accuracy for the minority group.

To address this issue, standard methods can be broadly categorised into two approaches. The first involves pre-processing techniques, which commonly encompass re-sampling and re-weighting strategies. Re-sampling may include methods such as over-sampling the minority or under-sampling the majority, while re-weighting assigns different weights to classes during the training process to rectify imbalances. Notably, \cite{rolf2021representation,chen2018my} suggest promoting a more diverse representation in the training data as part of these pre-processing efforts.
The second approach entails integrating regularises or constraints derived from group fairness definitions into the training process. These modifications are designed to discourage the model from exhibiting a bias toward the majority group and promote fair treatment of all groups during the learning process.

Imbalanced data may stem from mishandling missingness linked to sensitive attributes. The three missing data categories---\acrfull{MCAR}, \acrfull{MAR}, and \acrfull{MNAR} \citep{rubin1976inference}---pose challenges, as seen in instances where men exhibit reluctance in disclosing income (MAR missingness) and gender itself has missing entries (MNAR missingness) \citep{tu2019causal}. While removing samples with missingness might seem tempting, it can lead to imbalanced data, disproportionately representing groups, especially in MAR or MNAR cases. In fact, missing data and fairness are intertwined, and empirical evidence supports imputing missing values instead of discarding rows \citep{fernando2021missing}. Literature, such as \cite{caton2022impact,fernando2021missing}, explores imputation strategies for fairness with respect to missing data. Additionally, \cite{tu2019causal} provides valuable strategies for causal-based fairness in the presence of missing data.

\subsection{Dynamic Nature of Data Collection}

In numerous machine learning scenarios, there is a dynamic nature where data are collected over time and decisions based on models shape subsequent data collection. This phenomenon is often referred to as a feedback loop or a self-reinforcing cycle, and it can lead to selection bias and the amplification of existing bias \citep{d2020fairness,liu2018delayed}.
For example, students from unprivileged backgrounds, facing additional challenges in achieving higher scores, are more likely to be classified as ``low-performing'' in early assessments. Subsequently, they may be placed in lower-level classes with limited resources. This situation impedes their educational experience, and their performance may conform to the initial classification, given the restricted opportunities available to them.

Fair sequential learning emphasises decisions made for short-term fairness can have repercussions on long-term fairness outcomes (long-term fairness). Striking a balance between exploiting existing knowledge and exploring sub-optimal solutions to gather additional data is crucial in navigating this complex landscape. The aim is to ensure that the learning process remains fair not only in immediate outcomes but also over an extended period, taking into account the evolving nature of data and decision-making contexts.
Research studies such as \cite{joseph2016fairness} delve into fairness within bandits, while \cite{wen2021algorithms,jabbari2017fairness} extends this exploration to fairness in reinforcement learning. These studies, based on long-term rewards, prioritise one decision over another if the former's long-term reward surpasses the latter's.
Further extensions incorporate considerations of causal structures into the fairness framework \citep{creager2020causal,d2020fairness,nabi2019learning,hu2022achieving}.

\subsection{Trade-offs}

There is an inherent incompatibility among different concepts of group fairness \citep{mashiat2022trade,kleinberg2017inherent} and between group fairness and individual fairness \citep{binns2020apparent,dwork2012fairness}.
This tension often makes it inherently challenging, or impossible, to concurrently fulfil multiple seemingly natural fairness criteria \citep{awasthi2020beyond,kleinberg2017inherent}.
With the surge of numerous fairness notions, there have been efforts to design systems that address the challenge of balancing multiple potentially conflicting fairness measures \citep{awasthi2020beyond,kim2020fact,lohia2019bias}. This becomes particularly relevant when there is no widely-recognised fairness notion that can be universally applied.

Trade-offs between fairness and accuracy have been well-identified \citep{sun2024optimizing,liu2022accuracy,mandal2020ensuring} while an empirical study on real-world problems by \citep{rodolfa2021empirical} challenges the presumed existence or magnitude of the trade-off between accuracy and fairness. Furthermore, maintaining accuracy introduces tension between fairness and interpretability (feature deduction) \cite{agarwal2021trade}.
Inspired by concepts from fair division and envy-freeness in economics and game theory, a line of work has emerged focusing on achieving envy-free outcomes \citep{ustun2019fairness,zafar2017parity}. This is based on the observation that, in certain decision-making scenarios, existing parity-based fairness notions may be too stringent. This stringency can potentially hinder the attainment of more accurate decisions that are also desired by every sensitive attribute group.




Dwork et al. \citep{dwork2012fairness} initiated a discussion about the relationship between privacy and fairness, that ``Fairness Through Awareness'' is generalisation of differential privacy \citep{Dwork2006differential}.
Both fairness and privacy can be improved by obfuscating sensitive information. Whereas, there are discussions about trade-offs between fairness and privacy in the literature \citep{tran2021differentially, pinzon2021impossibility, chang2021privacy, cummings2019compatibility}.

\cleardoublepage
\chapter{Non-Commutative Polynomial Optimisation to System Identification}\label{cha:tac}




\begin{quote}
\textit{There has been much recent progress in time series forecasting and estimation of system matrices of linear dynamical systems \citep{WestHarrison}. 
We present an approach to both problems based on an asymptotically convergent hierarchy of convexifications of non-commutative polynomial optimisation. 
We present promising computational results, including a comparison with methods implemented in Matlab System Identification Toolbox.
This chapter is based on the joint work with Dr. Jakub Mare\v{c}ek, and has been published in the IEEE Transactions on Automatic Control \citep{zhou2023learning}.}
\end{quote}

We consider the identification of 
vector autoregressive processes with hidden components from time series of observations, which is a key problem in system identification \citep{Ljung1999}.
Its applications range from the identification of parameters in epidemiological models \citep{anderson1992infectious} and reconstruction of reaction pathways in other biomedical applications \citep{chou2009recent}, to identification of models of quantum systems \citep{bondar2023globally,bondar2022recovering}. 
Beyond this, one encounters either partially observable processes or questions of causality \citep{9363924,geiger2015causal} in almost any application domain.
In the ``prediction-error'' approach to forecasting \citep{Ljung1999}, it allows the estimation of subsequent observations in a time series.


To state the problem formally, let us define a \acrfull{LDS}, represented by the tuple $(G,F,V,W)$, as in \cite{WestHarrison} 
\begin{equation}
\begin{split}
 \theta_{t} &= G \theta_{t-1} + \omega_t, \\ 
Y_t &= F' \theta_t + \nu_t,    
\end{split}
\tag{LDS}
\label{equ:LDS}
\end{equation}
where $\theta_t\in\mathbb{R}^{k\times 1}$ is the hidden state, $Y_t \in\mathbb{R}$ is the observed output (measurements, observations), 
$G\in \mathbb{R}^{k\times k}$ and $F\in\mathbb{R}^{k\times 1}$ are system matrices, and $\{\omega_t,\nu_t \}_{t\in\mathbb{N}}$ are normally distributed process and observation noises with zero mean and covariance of $W$ and $V$ respectively.
The transpose of $F$ is denoted as $F'$.
Learning (or proper learning) refers to identifying the quadruple $(G,F,V,W)$ given the output $\{Y_t\}_{t\in\mathbb{N}}$. 
We assume that the linear dynamical system $(G,F,V,W)$ is observable \citep{van}, i.e., its observability matrix \citep{WestHarrison}
has full rank. Note that a minimal representation is necessarily observable and controllable, cf. Theorem 4.1 in \cite{tangirala2014principles}, so the assumption is not too strong.


There are three complications. 
First, the dimension $k$ of the hidden state $\theta_t$ is not known, in general. Although \cite{TuhinJMLR} have shown that a lower-dimensional model can approximate a higher-dimensional one rather well, in many cases, it is hard to choose $k$ in practice.
Second, the corresponding optimisation problem is non-convex, and guarantees of global convergence have been available only for certain special cases.
Finally, the operators-valued optimisation problem is non-commutative, and hence much work on general-purpose commutative non-convex optimisation is not applicable without making assumptions \cite[cf.]{bondar2023globally} on the dimension of the hidden state.  

Here, we aim to develop a method for proper learning of LDS that could also estimate the dimension of the hidden state and that would do so with guarantees of global convergence to the best possible estimate, given the observations. 
This would promote explainability beyond what forecasting methods without global convergence guarantees allow for.
In particular, our contributions are the following.

\begin{itemize}
    \item We cast learning of a \acrshort{LDS} with an unknown dimension of the hidden state as a \acrshort{NCPOP}, explained in Appendix~\ref{sec:ncpop}. This also makes it possible to utilise prior information as shape constraints in the NCPOP.
    \item We show how to use \acrfull{NPA} hierarchy \citep{pironio2010convergent,navascues2012sdp} of convexifications of the NCPOP to obtain bounds and guarantees of global convergence. The runtime is independent of the (unknown) dimension of the hidden state.

    \item In two well-established small examples of \cite{arima_aaai,hazan2017learning,Jakub}, our approach outperforms standard subspace and least squares methods, as implemented in Matlab\texttrademark{ }  System Identification Toolbox\texttrademark. 
\end{itemize}

This chapter is organised as follows. 
Section~\ref{sec:tac_related_work} set our work in the context of related work.
Section~\ref{sec:tac_main} introduces our formulation of proper learning of LDS with an unknown dimension of the hidden state, following the discussion of NCPOP in Appendix~\ref{sec:ncpop}. Eventually, Section~\ref{sec:tac_numerical} presents promising computational results, including a comparison with methods implemented in Matlab\texttrademark{ }  System Identification Toolbox\texttrademark.

\section{Related Work in System Identification and Control}
\label{sec:tac_related_work}

There is a long history of research within system identification \citep{Ljung1999}. 
In forecasting under LDS assumptions (improper learning of LDS), a considerable progress has been made in the analysis of predictions for the expectation of the next measurement using \acrfull{AR} processes in Statistics and Machine Learning. In \cite{anava13}, first guarantees were presented for \acrfull{ARMA} processes. In \cite{arima_aaai}, these results were extended to a subset of autoregressive integrated moving average (ARIMA) processes. \cite{Jakub} have shown that up to an arbitrarily small error given in advance, AR processes will perform as well as \emph{any} Kalman filter on any bounded sequence. 
This has been extended by \cite{tsiamis2020online} to Kalman filtering with logarithmic regret.

Another stream of work within improper learning focuses on sub-space methods \citep{katayama2006subspace,van} and spectral methods \cite{hazan2017learning,hazan2018spectral}. 
\cite{tsiamis2019sample,tsiamis2019finite} presented the present-best guarantees for traditional sub-space methods.
\cite{sun2020finite} utilise regularisations to improve sample complexity. 
Within spectral methods, \cite{hazan2017learning} and \cite{hazan2018spectral} have considered learning LDS with input, employing certain eigenvalue-decay estimates of Hankel matrices in the analyses of an AR process in a dimension increasing over time.
We stress that none of these approaches to improper learning  are ``prediction-error'' methods; namely, they do \emph{not} estimate the system matrices.

In proper learning of LDS, many state-of-the-art approaches consider the least squares method, despite complications encountered in unstable systems \citep{faradonbeh2018finite}. \cite{simchowitz2018learning} have  
provided non-trivial guarantees for the Ordinary Least Squares (OLS) estimator 
in the case of stable $G$ and there being no hidden component, i.e., $F'$ being the identity matrix and $Y_t = \theta_t$. 
Surprisingly, they have also shown that more unstable linear systems are easier to estimate than less unstable ones, in some sense. 
\cite{simchowitz2019learning} extended the results to allow for a certain pre-filtering procedure.
\cite{SarkarRakhlin,TuhinJMLR} extended the results to cover stable,
marginally stable, and explosive regimes.
\cite{oymak2019non} provide a finite-horizon analysis of the Ho-Kalman algorithm.
Most recently, \cite{bakshi2023new} provided a detailed analysis of the use of the method of moments in learning linear dynamical systems, which could be seen as a polynomial-time algorithm for learning a LDS from a trajectory of polynomial length up to a polynomial error.
Our work could be seen as a continuation of the work on the least squares method, with guarantees of global convergence.

\section{The Model}
\label{sec:tac_main}

Given a trajectory of observations $Y_1,\dots,Y_{t-1}$, 
loss is a one-step error function at time $t$ that compares an estimate $\hat{Y}_t$ with the actual observation $Y_t$. 
Within the least squares estimator, we aim to minimise the sum of quadratic loss functions, i.e., 
\begin{equation*}
    \min_{\hat{Y}_{t,t\geq 1}} \sum_{t\geq 1} \|Y_{t}-\hat{Y}_{t}\|^2,
\end{equation*}
where the estimates $\hat{Y}_t,t\geq 1$ are decision variables.
The properties of the optimal least squares estimate are well understood: it is consistent, cf. Mann and Wald \citep{mann1943statistical} and Ljung \citep{ljung1976consistency}, and has favourable sample complexity, cf. Theorem 4.2 of Campi and Weyer \cite{campi2002finite} in the general case, and to Jedra and Proutiere \cite{jedra2020finite} for the latest result parameterised by the size of a certain epsilon net.
We stress, however, that \emph{it has not been understood} how to solve the non-convex optimisation problem, in general, outside of some special cases \cite{Hardt} and recent, concurrent work of \cite{bakshi2023new}. In contrast to \cite{Hardt}, we focus on a method achieving global convergence under mild assumptions, and specifically without assuming the dimension of the hidden state is known. 

When the dimension of the hidden state is not known, we need operator-valued variables $\hat{\theta}_t$ to model the state evolution, and some additional scalar-valued variables. We denote the process noise and the observation noise at time $t$ by $\omega_t$ and $\nu_t$, respectively. We also denote as such  
the decision variables corresponding to the estimates thereof, if there is no risk of confusion.   
If we add the sum of the squares of $\omega_t$ and the sum of the squares of $\nu_t$ as regularisers to the objective function with sufficiently large multipliers and minimise the resulting objective, we 
should reach a feasible solution with respect to the system matrices with the process noise $\omega_t$ and observation noise $\nu_t$ being close to zero.


Overall, such a formulation has the form in Equations~\eqref{obj_tac} subject to (\ref{NFF_1}--\ref{NFF_2}).
The inputs are $Y_{t},t\geq 1$, i.e., the time series of the actual measurements, of a time window $T$ thereof, and multipliers $\lambda_1, \lambda_2$.
Decision variables are system matrices $G$, $F$; noisy estimates $\hat{Y}_t$, realisations $\omega_t$, $\nu_t$ of noise, for $t\geq 1$;
and state estimates $\hat{\theta}_t$, for $t\geq 0$, which include the initial state $\hat{\theta}_0$. 
We minimise the objective function:
\begin{equation}
\min_{\hat{Y}_t, \hat{\theta}_t, G, F, \omega_t, \nu_t} \sum_{t\geq 1} \| Y_{t}-\hat{Y}_{t} \|^2  + \lambda_1 \sum_{t\geq 1} \|\nu_t\|^2 + \lambda_2 \sum_{t\geq 1} \|\omega_t\|^2 \label{obj_tac} 
\end{equation}
for a $l_2$-norm $\| \cdot \|$ over the feasible set given by constraints for $t\geq 1$:
\begin{align}
\hat{\theta}_t & = G \hat{\theta}_{t-1} + \omega_t \label{NFF_1} \\
\hat{Y}_{t} & = F' \hat{\theta}_t + \nu_t. \label{NFF_2}
\end{align}
We call the term $F'\hat{\theta}_t$ noise-free estimates, which are regarded as our simulated/ predicted outputs.
Equations~\eqref{obj_tac} subject to (\ref{NFF_1}--\ref{NFF_2}) give us the least squares model. We can now apply the techniques of non-commutative polynomial optimisation to the model so as to recover the system matrices of the underlying linear system.

\begin{theorem}
\label{T1}
For any observable linear system $(G,F,V,W)$, 
for any length $T$ of a time window,
and any error $\epsilon > 0$, 
under the Archimedean assumption, detailed in Theorem~\ref{the:positivetrace2cyclic},
there is a convex optimisation problem whose objective function value is at most $\epsilon$ away from Equations \eqref{obj_tac} subject to (\ref{NFF_1}--\ref{NFF_2}). 
Furthermore, an estimate of $(G,F,V,W)$ can be extracted from the solution of the same convex optimisation problem.
\begin{proof}
First, we need to show the existence of a sequence of convex optimisation problems, whose objective function approaches the optimum of the non-commutative polynomial optimisation problem.
As explained in Appendix~\ref{sec:ncpop} above, there is a sequence of \acrshort{SDP} relaxations of Equation~\eqref{equ:trace-NCPO-con-sup}  \citep{pironio2010convergent,burgdorf2016optimization}.
The convergence of the sequence of their objective-function values is shown by 
Theorem~\ref{the:convergence-ncpop}, which requires the Archimedean assumption in Theorem~\ref{the:positivetrace2cyclic}.
The translation of a problem involving multiple scalar- and operator-valued variables $\hat{Y}_t, \hat{\theta}_t, G, F, \omega_t, \nu_t$ in (\ref{obj_tac}--\ref{NFF_2}) to the form of Equation~\eqref{equ:trace-NCPO-con-sup}, also known as the product-of-cones construction, is somewhat tedious, but routine and implemented in multiple software packages, e.g., \cite{wittek2015algorithm,wang2021exploiting}.
Second, we need to show that extraction of an estimate of $(G,F,V,W)$ from the SDP relaxation of order $k(\epsilon)$ in the series is possible. 
As explained in Section 2.2 of \cite{klep2018minimizer} or in Theorem~\ref{the:finite-convergence-ncpop}, when flatness condition is satisfied,  one utilises the \acrfull{GNS} construction \citep{gelfand1943imbedding,segal1947irreducible}. Notice that \cite[cf.]{lee2023computability} the estimate of $(G,F,V,W)$ may have a higher error than $\epsilon$.
\end{proof}
\end{theorem}

This reasoning can be applied to more 
complicated formulations, involving shape constraints.
For instance, in quantum systems \cite{bondar2023globally}, density operators are Hermitian and this constraint can be added to the least squares formulation.

Crucially for the practical applicability of the method, one should like to exploit the sparsity in the NCPOP (\ref{obj_tac}--\ref{NFF_2}), as in Section~\ref{sec:sparse_ncpop}. Notice that one can decompose the problem (\ref{obj_tac}--\ref{NFF_2}) into $t$ subsets of variables involving 
$\hat{Y}_t, \hat{\theta}_{t-1}, \hat{\theta}_t, G, F, \omega_t, \nu_t$, which satisfy the running intersection property in Assumption~\ref{ass:rip}.

Also, note that the extraction of the minimiser using the \acrfull{GNS} construction, as explained in Theorem~\ref{the:GNS}, is stable to errors in the moment matrix, for \emph{any} NCPOP, including the pre-processing above.
See Theorem 4.1 in \cite{klep2018minimizer}.
That is: it suffices to solve the \acrshort{SDP} relaxation with a fixed error, in order to extract the minimiser. 

One can also utilise a wide array of reduction techniques on the resulting SDP relaxations. Notable examples include facial reduction \citep{borwein1981facial,permenter2018partial} and exploiting sparsity \citep{fukuda2001exploiting}. Clearly, these can be applied to any SDP, irrespective of the non-commutative nature of the original problem, but can also introduce \citep{kungurtsev2018two} numerical issues.
We refer to \cite{majumdar2019recent} for an up-to-date discussion.

\section{Numerical Illustrations}
\label{sec:tac_numerical}

Let us now present the implementation of the approach using the techniques of NCPOP  \citep{pironio2010convergent,burgdorf2016optimization} and to compare the results with traditional system identification methods.
Our implementation is available online \footnote{\url{https://github.com/Quan-Zhou/Proper-Learning-of-LDS}}.

\subsection{The General Setting}
\label{sec:settings}


\paragraph{Our formulation and solvers}

For our formulation, we use Equations~\eqref{obj_tac} subject to (\ref{NFF_1}--\ref{NFF_2}), where we need to specify the values of $\lambda_1$ and $\lambda_2$.
To generate the SDP relaxation of this formulation as in Equation~\eqref{equ:trace-NCPO-dual}, we need to specify the moment order $d$. Because the degrees of objective \eqref{obj_tac} and constraints in (\ref{NFF_1}--\ref{NFF_2}) are all less than or equal to $2$, the moment order $d$ within the respective hierarchy can start from $d=1$. (The reasoning of choosing moment order $d$ can be found in the definition of SDP relaxations in Equation~\eqref{equ:trace-NCPO-sup-Theta} or~\eqref{equ:trace-NCPO-dual}.)

In our implementation, we use a globally convergent \acrshort{NPA} hierarchy \citep{pironio2010convergent} of SDP relaxations, 
and its sparsity-exploiting variant, known as the non-commutative variant of the \acrfull{TSSOS} hierarchy \citep{wang2021tssos,wang2020chordal,wang2021exploiting}.
The SDP of a given moment order within the NPA hierarchy is constructed using \texttt{ncpol2sdpa} 1.12.2\footnote{\url{https://github.com/peterwittek/ncpol2sdpa}} of Wittek \citep{wittek2015algorithm}.
The SDP of a given moment order within the non-commutative variant of the TSSOS hirarchy is constructed using the \texttt{nctssos}\footnote{\url{https://github.com/wangjie212/NCTSSOS}} of Wang et al. \citep{wang2021exploiting}.
Both SDP relaxations are then solved by \texttt{mosek} 9.2 \cite{mosek2020mosek}.

\paragraph{Baselines}
We compare our method against leading methods for estimating state-space models, as implemented in Matlab\texttrademark{ } System Identification Toolbox\texttrademark. Specifically, we test against a combination of least squares algorithms implemented in routine \texttt{ssest}  (``least squares auto''), subspace methods of \cite{van} implemented in routine \texttt{n4sid} (``subspace auto''), and a subspace identification method of \cite{jansson2003subspace} with an ARX-based algorithm to compute the weighting, again utilised via \texttt{n4sid} (``ssarx''). 

To parameterise the three baselines, we need to specify the estimated dimension $\hat{k}$ of the state-space model. We would set $\hat{k}=k$ directly or alternatively, iterate from $1$ to the highest number allowed in the toolbox when the underlying system is unknown, e.g., in real-world stock-market data.
Then, we need to specify the error to be minimised in the loss function during estimation.
In fairness to the baselines, we use the one-step ahead prediction error when comparing prediction performance and simulation error between measured and simulated outputs when comparing simulation performance.

\paragraph{The performance index}
To measure the goodness of fit between the ground truth $\{Y_t\}_{t=1}^T$ (actual measurements) and the noise-free simulated/ predicted outputs $\{F'\hat{\theta}_t\}_{t=1}^T$, using different system identification methods, we introduce the fit value\footnote{This definition follows the one used in Matlab\texttrademark{ } System Identification Toolbox\texttrademark, at \url{https://uk.mathworks.com/help/ident/ref/compare.html}}
\begin{equation}
\mathrm{fit}:=\left(1-\frac{\left\|Y-F'm\right\|}{\left\|Y-\mean(Y)\right\|}\right)\times 100\%,\label{NRMSE}
\end{equation}
where $Y$ and $F'm$ are the vectors consisting of the sequence $\{Y_t\}_{t=1}^T$ and $\{F'\hat{\theta}_t\}_{t=1}^T$ respectively.
A higher fit value indicates better simulation or prediction performance.

\subsection{Experiments on the Example of Hazan et al.}
\label{sec:exp:hazan}

Experiments in Sections \ref{sec:exp:hazan}--\ref{sec:exp:baselines} utilise synthetic time series of $T$ observations generated using LDS of the form in Equation~\eqref{equ:LDS}, with the tuple $(G,F,V,W)$ and the initial hidden state $\theta_0$ 
 detailed next.
We use the dimension $k$ to indicate that the time series of observations were generated using $k \times k$ system matrices,
while we use operator-valued variables to estimate these. 
The standard deviations of process noise and observation noise $W,V$ are chosen from $0.1,0.2,\dots,0.9$. 
Note that $W$ is an $k\times k$ matrix in general, while we consider the spherical case of $W=0.1\times I_{k}$, where $I_k$ is the $k$-dimensional identity matrix, which we denote by $W=0.1$.

In our first experiment, we explore the statistical performance of feasible solutions of the \acrshort{SDP} relaxations using the example of Hazan et al.\citep{hazan2017learning,Jakub}.
We performed one experiment on each combination of standard deviations of process $W$ and observation noise $V$ from the discrete set $0.1,0.2,\dots,0.9$, i.e., 81 runs in total.

\begin{figure}[t!]
\centering
\includegraphics[width=0.6\textwidth]{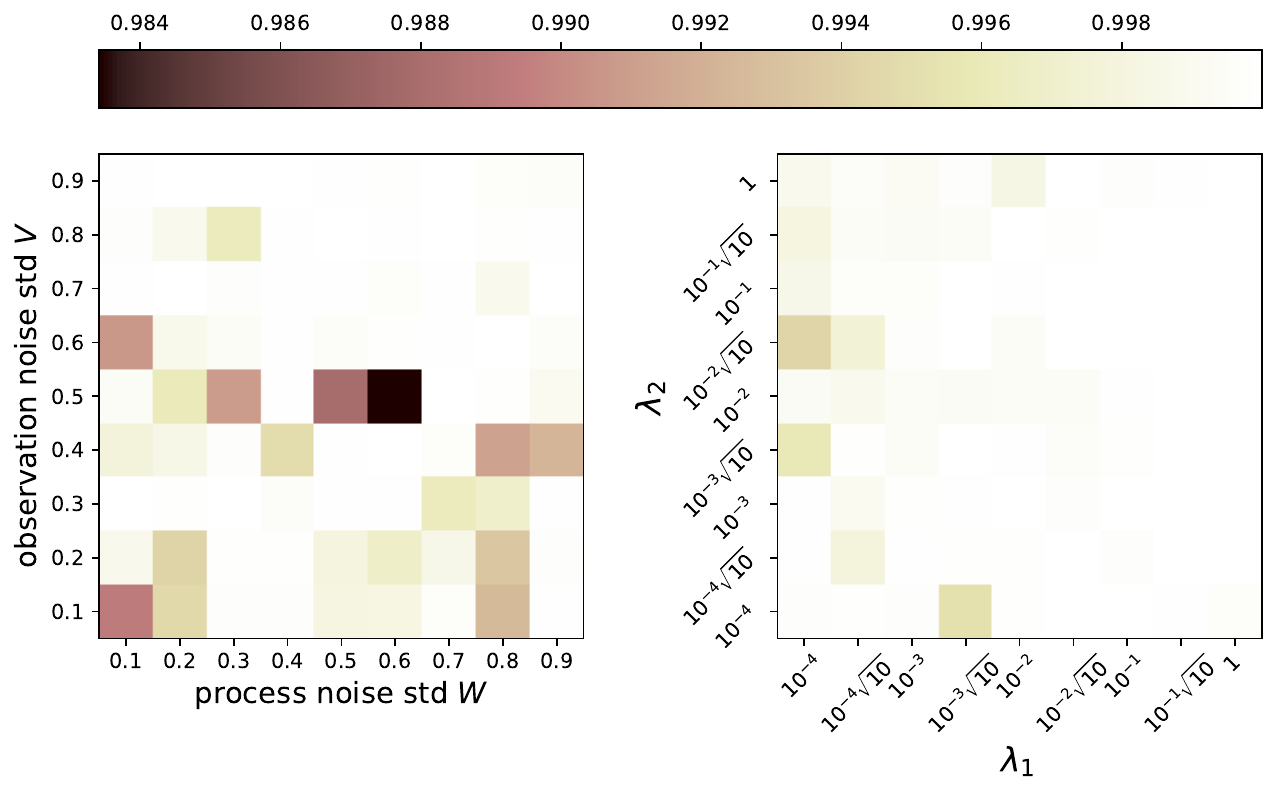}
\caption[The fit values of our method at different combinations of noise standard deviations of noises, as well as parameters]{\textbf{Left:} The fit values \eqref{NRMSE} of $81$ experiments of our method at different combinations of noise standard deviations of process noise $W$ and observation noise $V$ and \textbf{Right:} at different combinations of parameters $\lambda_1$ and $\lambda_2$. Both use the data generated from systems in \eqref{equ:LDS}.
Lighter colours indicate higher fit values and thus better simulation performance.
}\label{fig:NCPO100}
\end{figure}

Figure~\ref{fig:NCPO100} illustrates the fit values of the $81$ runs of our method in different combinations of standard deviations of process noise $W$ and observation noise $V$ (left), and another $81$ experiments in different combinations of $\lambda_1$ and $\lambda_2$ (right). 
In the left subplot of Figure~\ref{fig:NCPO100}, we consider: $k=2$, $G=\bigl(\begin{smallmatrix} 0.9&0.2\\0.1&0.1\end{smallmatrix}\bigr)$, $F'=\bigl(\begin{smallmatrix} 1&0.8\end{smallmatrix}\bigr)$, the starting point $\theta_0'=\bigl(\begin{smallmatrix} 1 & 1 \end{smallmatrix}\bigr)$, and $T=20$. 
In the right subplot of Figure~\ref{fig:NCPO100}, we have the same settings as in the left one, except for $W=V=0.5$ and the parameters $\lambda_1,\lambda_2$ being chosen from $10^{-4},\dots,1$. 
It seems clear the highest fit value is to be observed for the standard deviation of both process and observation noises close to $0.5$.
While this may seem puzzling at first, notice that higher standard deviations of noise make it possible to approximate the observations by an AR process with low regression depth \cite[Theorem 2]{Jakub}. 
The observed behaviour is therefore in line with previous results \cite[e.g., Figure 3]{Jakub}. 

\subsection{Comparisons against the Baselines}
\label{sec:exp:baselines}

Next, we investigate the simulation performance of our method in comparison with other system identification methods, for varying LDS used to generate the time series. 
Our method and the three baselines described in Section~\ref{sec:settings} are run $30$ times for each choice of the standard deviations of the noise, 
with all methods using the same time series.

\begin{figure}[t!]
\centering
\includegraphics[width=0.4\textwidth]{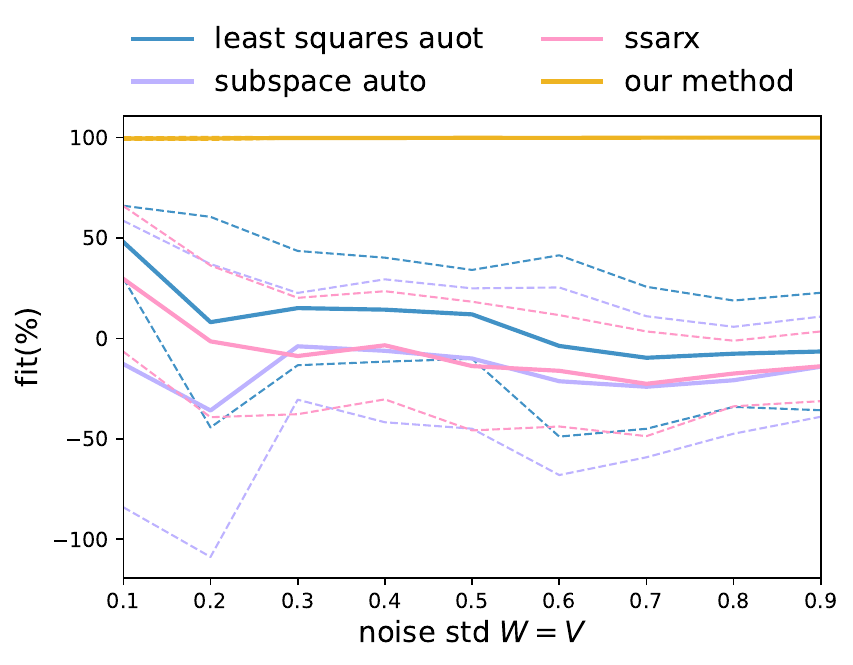}
\includegraphics[width=0.42\textwidth]{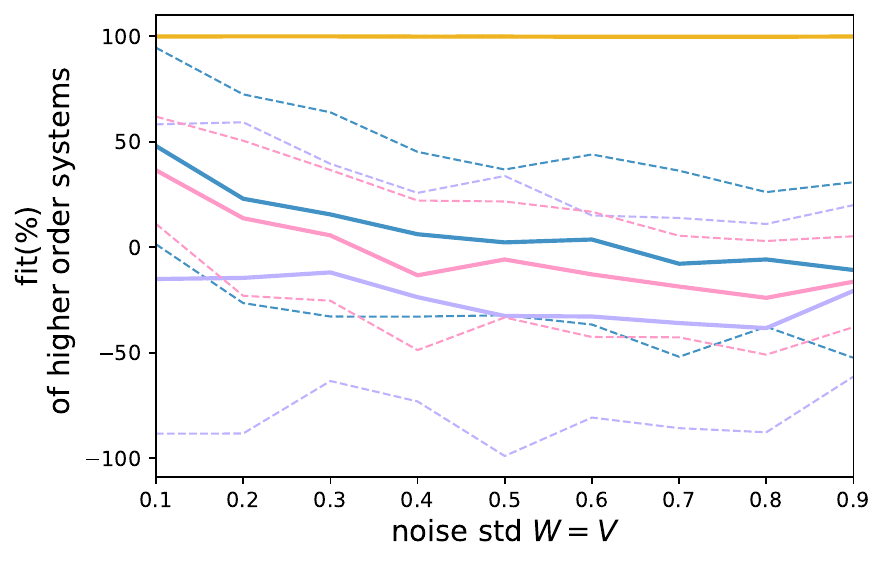}
\includegraphics[width=0.4\textwidth]{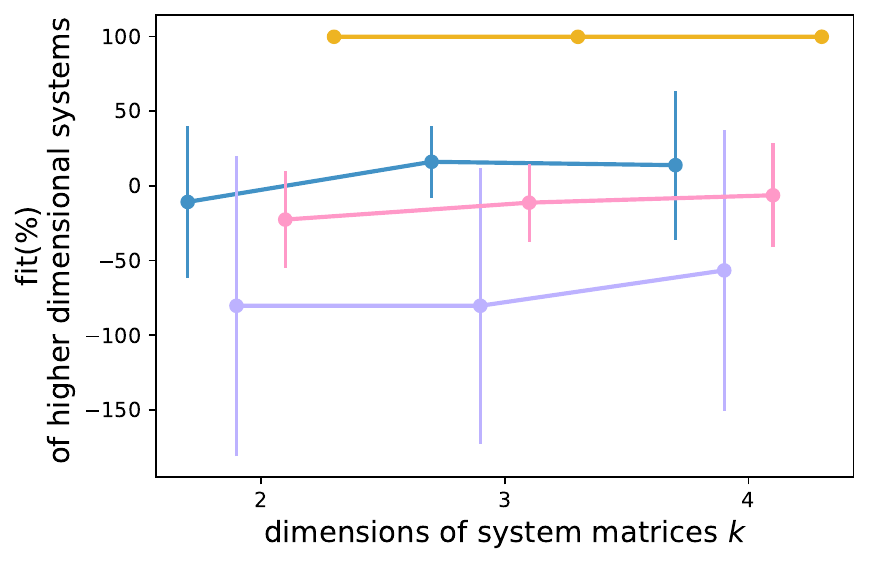}
\caption[The fit values of our method compared to the leading system identification methods implemented in Matlab\texttrademark{ } System Identification Toolbox\texttrademark.]{The fit values \eqref{NRMSE} of our method compared to the leading system identification methods implemented in Matlab\texttrademark{ } System Identification Toolbox\texttrademark.
\textbf{Upper:} the mean (solid lines) and mean $\pm$ one standard deviations (dashed lines) of fit values as standard deviation of both process noise and observation noise increasing in lockstep from $0.1$ to $0.9$.
The time series used for simulation are generated from systems in \eqref{equ:LDS} (left) and higher differential order systems in \eqref{equ:LDS-higher} (right), with the dimensions $k$ of both systems being $2$. 
\textbf{Lower:} the mean (solid dots) and mean $\pm$ one standard deviations (vertical error bars) of fit values at different dimensions $k$ of the underlying systems in \eqref{equ:LDS}.
Higher fit values indicate better simulation performance.
}
\label{fig:CompareSim}
\end{figure}

Figure~\ref{fig:CompareSim} illustrates the results, with methods distinguished by colours: 
blue for ``least squares auto'', purple for ``subspace auto'', pink for ``ssarx'', and yellow for our method.
The upper left subplot presents the mean (solid lines) and mean $\pm$ one standard deviations (dashed lines) of fit values as standard deviation of both process noise and observation noise (``noise std'') increasing in lockstep from $0.1$ to $0.9$. The underlying system is the same as in the left subplot of Figure~\ref{fig:NCPO100}, except for $W=V=0.1,0.2,\dots,0.9$.
The upper right subplot is similar, except the time series are generated by systems of a higher differential order:
\begin{equation}
\begin{split}
\theta_t & = G \theta_{t-1} + \omega_t  \\
Y_{t} & = F_1' \theta_t + F_2'(\theta_t-\theta_{t-1}) + \nu_t,
\end{split}
\label{equ:LDS-higher}
\end{equation}
and the formulation of our method is changed accordingly.
In the lower subplot of Figure~\ref{fig:CompareSim},
we consider the mean (solid dots) and mean $\pm$ one standard deviations (vertical error bars) of fit values at different dimensions $k=2,3,4$ of the underlying system in Equation~\eqref{equ:LDS}.

\textcolor{black}{
As Figure~\ref{fig:CompareSim} suggests, it is expected that other methods would perform better when the noise standard deviations approach zero, as shown in the upper subplots of Figure~\ref{fig:CompareSim}. 
However, despite the fact that the dimensions used by the baseline methods are the true dimensions of the underlying system ($\hat{k}=k$), the fit values of these methods rarely reach 50\%, even when the noise standard deviations are very small. This could be due to the small number of observations ($T=20$).} (We will use ``least squares auto'', which seems to work best within the other methods, in the following experiment on stock-market data.)

\textcolor{black}{
Our method outperforms primarily because the technologies of NCPOP do not assume the dimension of the variables, allowing it to handle the noise very well. However, as explained in Appendix~\ref{cha:ncpop}, due to the high complexity of iteratively increasing the moment order until the global optimality of NCPOP is detected, we cannot compute the exact optimal solution of NCPOP. Instead, the results of our method presented here are not the exact optimal solution of NCPOP but the solutions of its SDP relaxation at moment order $d=1$.} Furthermore, our method shows better stability; the gap between the yellow dashed lines in the upper or middle subplot, which suggests the width of two standard deviations, is relatively small. 


\subsection{Experiments with Stock-Market Data}

Our approach to proper learning of LDS could also be used in a ``prediction-error'' method for improper learning of LDS, i.e., forecasting its next observation (output, measurement). As such, it can be applied to any time series. 
To exhibit this, we consider real-world stock-market data first used in \cite{arima_aaai}.
In particular, we predict the evolution of the stock price from the 21$^\textrm{st}$ period to the 121$^\textrm{st}$ period, where each prediction is based on the 20 immediately preceding observations ($T=20$). 
For our method, 
we use the same formulation \eqref{obj_tac} subject to \eqref{NFF_1}--\eqref{NFF_2}, but with the variable $F'$ removed. 
For comparison,
the combination of least squares algorithms ``least squares auto'' is used again.
Since we are using the stock-market data, the dimension $k$ of the underlying system is unknown.
Hence, the estimated dimensions $\hat{k}$ of the ``least squares auto'' are iterated from $1$ to $4$, wherein $4$ is the highest setting allowed in the toolbox for $20$-period observations.

\begin{figure}[t!]
\centering
\includegraphics[width=0.78\textwidth]{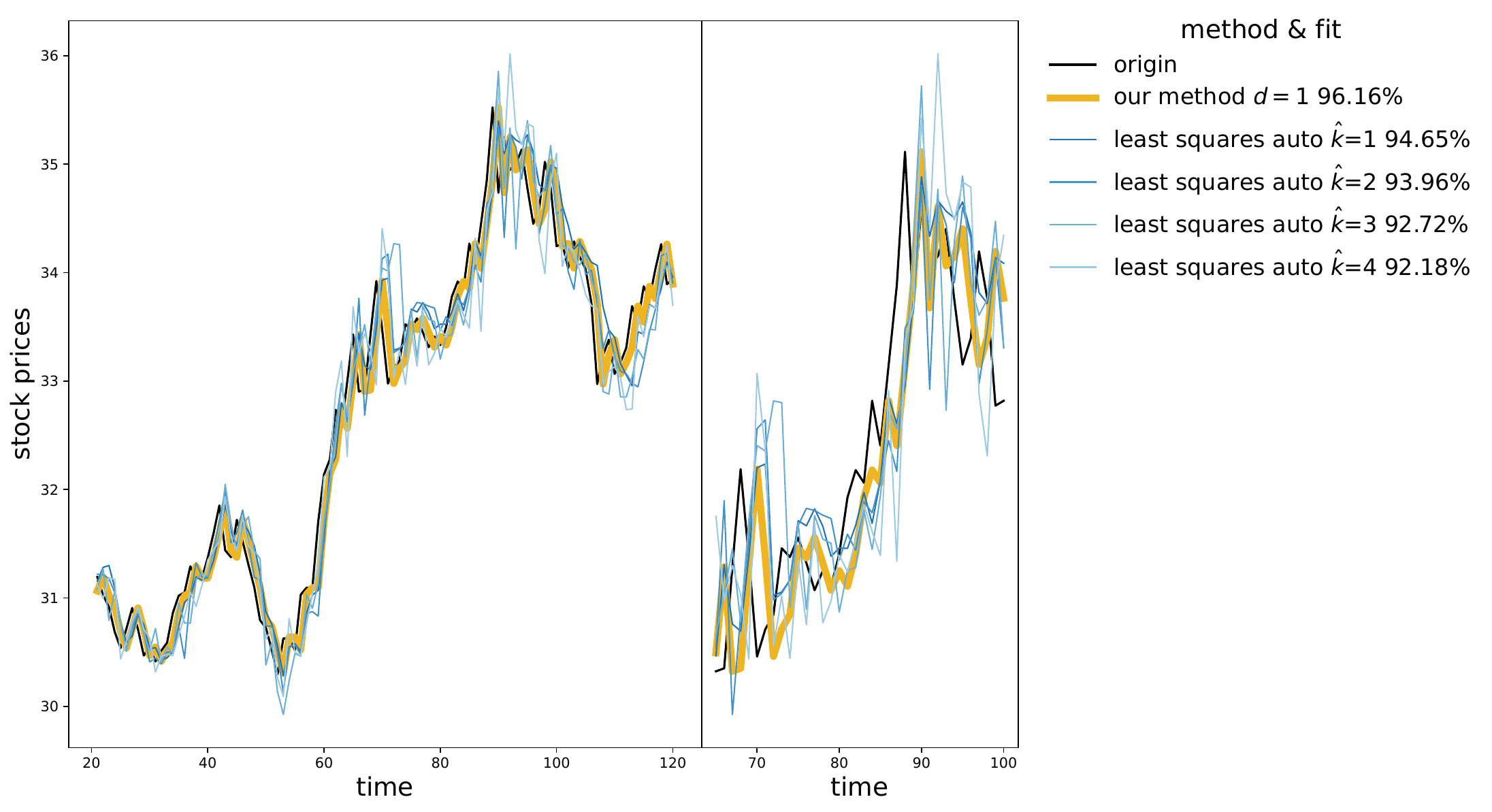}
\caption[The time series of stock price used in \cite{arima_aaai}, and 
the predicted outputs of our method compared against Matlab\texttrademark{} System Identification Toolbox\texttrademark.]{\textbf{Left:} 
The time series of stock price (dark) for the 21$^\textrm{st}$-121$^\textrm{st}$ period used in \cite{arima_aaai}, and 
the predicted outputs of our method (yellow) compared against ``least squares auto'' (blue) implemented in Matlab\texttrademark{} System Identification Toolbox\texttrademark.
The estimated dimension $\hat{k}$ of ``least squares auto'' is iterated from $1$ to the highest number of $4$. 
The percentages in legend are corresponding fit values of one-step predictions. \textbf{Right:} a zoom-in for the 66$^\textrm{th}$-101$^\textrm{st}$ period.}
\label{fig:TimePlot}
\end{figure}

Figure~\ref{fig:TimePlot} shows in the left subplot the results obtained by our method (a yellow curve), and the ``least squares auto'' of varying estimated dimensions $\hat{k}=1,2,3,4$ (four blue curves).
The true stock price ``origin'' is displayed by a dark curve.
The percentages in the legend correspond to fit values \eqref{NRMSE}. Both from the fit values and the shape of these curves, we notice that ``least squares auto'' performs poorly when the stock prices are volatile.
This is highlighted in the right subplot, which zooms in on the 66$^\textrm{th}$-101$^\textrm{st}$ period.

\begin{figure}[!t]
\centering
\includegraphics[width=0.8\textwidth]{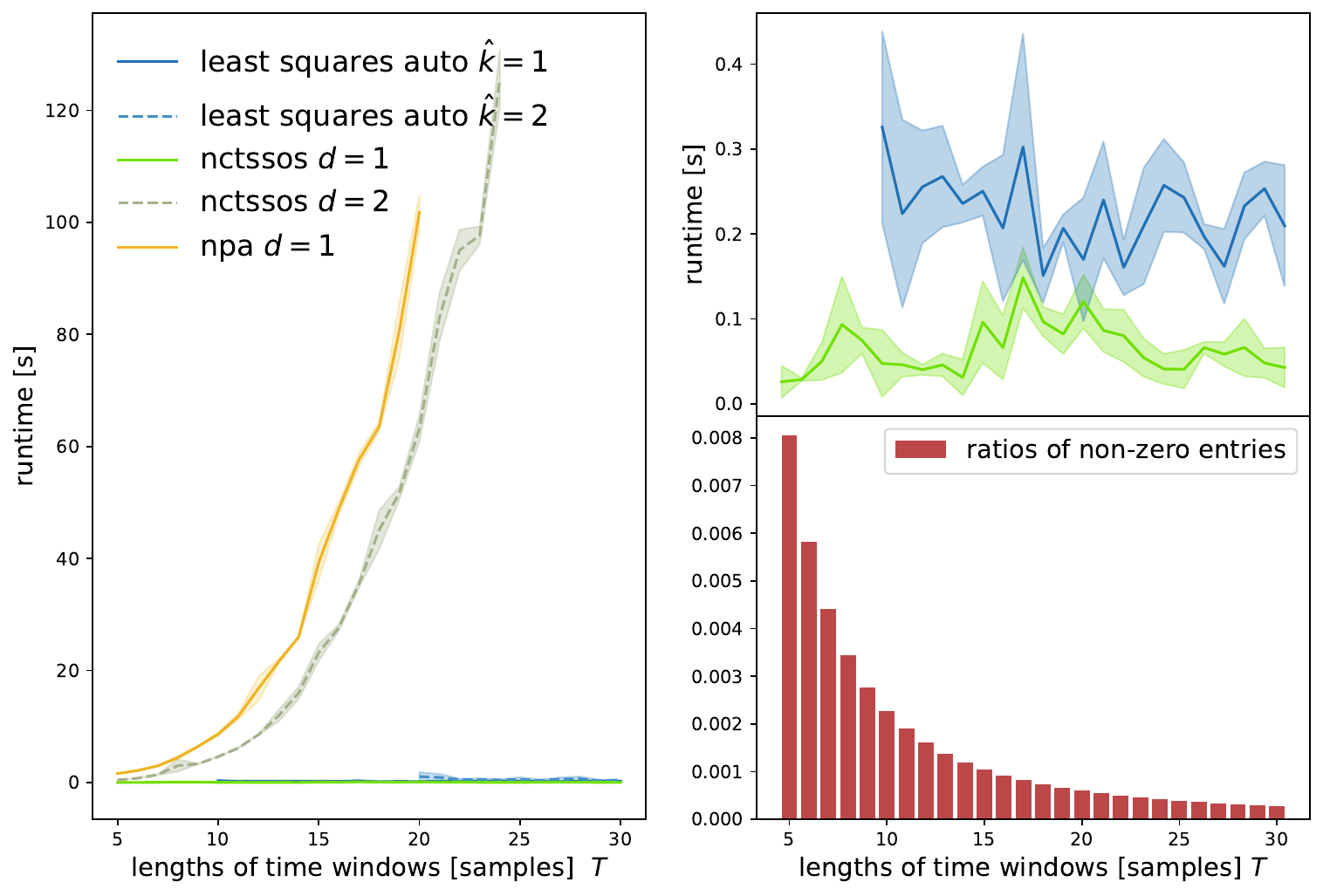}  
\hspace{0in}
\caption[The runtime of our method implemented via the TSSOS hierarchy and the NPA hierarchy, at different moment orders or estimated dimensions of hidden states, compared against Matlab\texttrademark{} System Identification Toolbox\texttrademark.]{\textbf{Left:} The (solid or dashed) curves show the mean runtime of the SDP relaxation of the baseline ``least squares auto'' (blue), the TSSOS hierarchy (green) and the NPA hierarchy (yellow), at different moment orders $d$ or estimated dimensions $\hat{k}$.
The mean $\pm$ one standard deviation of runtime is displayed by shaded error bands.
 \textbf{Upper-right:} The mean and mean $\pm$ one standard deviation of runtime of the SDP relaxation of TSSOS hierarchy at moment order $d=1$ and the ``least squares auto'' with dimension $\hat{k}=1$.
\textbf{Lower-right:} 
The red bars display the sparsity of NPA hierarchy of the experiment on stock-market data against the length of time window, by ratios of non-zero coefficients out of all coefficients in the SDP relaxations}
\label{fig:runtime}
\end{figure}



\subsection{Runtime}

As discussed in Appendix~\ref{sec:sparse_ncpop}, the computational complexity of \acrshort{NCPOP} is substantial. For $n$ variables in NCPOP formulation and a moment order $d$, the cardinality of the monomial basis, is $\frac{n^{d+1}-1}{n-1}$. Then, the number of variables in the corresponding SDP relaxation at the moment degree $d$, is its square.
The computation becomes impractical as the length of time windows, related to the number of variables in NCPOP formulation, increases.
Efforts to exploit sparsity patterns in these \acrshort{SDP} relaxations are also discussed in Appendix~\ref{sec:sparse_ncpop}.

To illustrate the effects, consider the runtime of two implementations of solvers for \eqref{obj_tac} subject to \eqref{NFF_1}--\eqref{NFF_2}. 
The first implementation constructs the \acrshort{SDP} relaxation of \acrshort{NPA} hierarchy via \texttt{ncpol2sdpa} 1.12.2 with moment order $d=1$. 
The second implementation constructs the non-commutative variant of the \acrshort{TSSOS} hierarchy via \texttt{nctssos}, with moment order $d=1,2$. 
For comparison purposes, we include the baseline ``least squares auto'' with estimated dimensions $\hat{k}=1,2$. 
We randomly select a time series from the stock-market data, with the length of time window $T$ chosen from $5,6,\dots,30$, and run these three methods three times for each $T$.

Figure~\ref{fig:runtime} illustrates the runtime of the \acrshort{SDP} relaxations and the baseline ``least squares auto'' as a function of the length of the time window. 
These implemented methods are distinguished by colours: blue for ``least squares auto'', green for
the non-commutative variant of the \acrshort{TSSOS} hierarchy (``nctssos''), and yellow for the \acrshort{NPA} hierarchy (``npa'').
The mean and mean $\pm$ one standard deviation of runtime are displayed by (solid or dashed) curves and shaded error bands.
The upper-right subplot compares the runtime of our method with ``nctssos'' at moment order $d=1$ against ``least squares auto'' with estimated dimension $\hat{k}=1$.
The red bars in the lower-right subplot display the sparsity of \acrshort{NPA} hierarchy of the experiment on stock-market data against the length of time window, by ratios of non-zero coefficients out of all coefficients in the \acrshort{SDP} relaxations.

As in most primal-dual interior-point methods \citep{Tuncel2000}, runtime of solving the relaxation to $\epsilon$ error is polynomial in its dimension and logarithmic in $1/\epsilon$, but it should be noted that the dimension of the relaxation grows fast in the length $T$ of the time window and the moment order $d$. 
It is clear that the runtime of solvers for \acrshort{SDP} relaxations within the non-commutative variant of the \acrshort{TSSOS} hierarchy exhibits a modest growth with the length of time window, much slower than that of the plain-vanilla \acrshort{NPA} hierarchy.

\section{Conclusion}

We have presented an alternative approach to the recovery of hidden dynamic underlying a time series, without assumptions on the dimension of the hidden state. 
For the first time in system identification and machine learning, this approach utilises NCPOP, which has been recently developed within mathematical optimisation \citep{pironio2010convergent,wittek2015algorithm,klep2018minimizer,wang2021exploiting}. 
This can accommodate a variety of other objectives and constraints, as we shall see in Chapter~\ref{cha:jair}.

\cleardoublepage
\chapter{Non-Commutative Polynomial Optimisation to Fairness in Forecasting}\label{cha:jair}


\begin{quote}
\textit{In machine learning, training data often capture the behaviour of multiple subgroups of some underlying human population.
This behaviour can often be modelled as observations of an unknown dynamical system with an unobserved state. When the training data for the subgroups are not controlled carefully, however, under-representation bias arises.
To counter this bias in learning dynamical systems, we introduce two natural notions of fairness in time-series forecasting problems: subgroup fairness and instantaneous fairness.
Our empirical results on the well-known COMPAS dataset \citep{angwin2016machine} demonstrate the efficacy of our methods.
This chapter is a collaboration with Dr. Jakub Mare\v{c}ek and Prof. Robert Shorten, and has been published in the Journal of Artificial Intelligence Research \citep{zhou2023fairness}.}
\end{quote}

Forecasts affect almost all aspects of our daily life, as a basis for access-control mechanisms. 
As the quality of the forecasts impacts our lives, for better or worse, it is becoming more and more apparent that many of the tools that produce the forecasts seem to be, or indeed are, unfair, in a sense we formalise below.
If the tools used to produce the forecasts are unfair, society suffers.

One such example of a forecasting tool that shapes the very pillars of our society is the \emph{FICO Score} \citep{FICO}. FICO is a measure of an individual's creditworthiness (or conversely, credit-default risk), computed by the Fair Isaac Corporation.
It has been suggested \citep{nickerson2016asset,delis2019mortgage,Hegarty9152,vogel2021learning} that the FICO Score may be unfair to certain minorities, although this has been disputed \citep{avery2012does}.

As another example, consider college admissions, where results of standardised tests were often presented as forecasts of potential academic success.
In the context of COVID-19, forecasting algorithms utilising previous grades and input from teachers replaced standardised tests in determining the satisfaction of college-admission requirements in many jurisdictions. 

Other forecasting tools are perhaps less well known, but perhaps even more alarming, as they have the potential to shape the very core of society. One striking example is Northpointe's COMPAS. COMPAS is a criminal-risk assessment tool that is widely used in pretrial, parole, and sentencing decisions at courts in New York, Wisconsin, California, and Florida.
COMPAS forecasts the likelihood that an individual will re-offend within two years. 
It has been suggested \citep{angwin2016machine,dressel2021dangers} that COMPAS under-predicts recidivism for Caucasian defendants, and over-predicts recidivism for African-American defendants.\footnote{
We note that this has been disputed \citep{kleinberg2017inherent,dressel2021dangers}, and that it has been suggested that such forecasts \citep{Neile2107020118} are difficult to make, in general, due to the cohort differences in group-based arrest trajectories.}

\label{page:define-s}
The applications, where fairness seems most important, often capture the behaviour of multiple subgroups of some underlying human population in the training data.
Let us consider a model, where there are a number of individuals within a population. 
The population is partitioned into subgroups indexed by $\mathcal{S}$. 
For each subgroup $s\in\mathcal{S}$, there is a set $\mathcal{I}^{(s)}$ of trajectories of observations available and each trajectory $i \in \mathcal{I}^{(s)}$ has observations for periods $\mathcal{T}^{(i,s)}$, possibly of varying cardinality $\lvert\mathcal{T}^{(i,s)}\rvert$.
Each subgroup $s \in\mathcal{S}$ is associated with a model, $\mathcal{L}^{(s)}$.
For all $i \in \mathcal{I}^{(s)}$, $s\in\mathcal{S}$, the trajectory $\{Y_t\}^{(i,s)}$, for $t\in \mathcal{T}^{(i,s)}$, is hence generated by precisely one model $\mathcal{L}^{(s)}$. 
Throughout, the superscripts distinguish the trajectories and subgroups, while subscripts indicate the periods.    
    

    

\label{page:define-L}
In this setting, under-representation bias \cite[cf. Section 2.2]{blum2019recovering} arises, where the trajectories of observations from one (``disadvantaged'') subgroup are under-represented in the training data.
This is particularly important if the forecasting is constrained to be subgroup-blind, i.e., 
we wish to learn a single subgroup-blind model $\mathcal{L}$.
This is the case when the use of sensitive attributes distinguishing each subgroup can be 
regarded as discriminatory, such as in the case of gender and race \citep{gajane2017formalizing,kleinberg2018discrimination}. 
Notice that such anti-discrimination measures are increasingly stipulated legally, e.g., within insurance pricing, where the sex of the applicant cannot be used, despite being known.
More broadly, under-representation bias harms both the accuracy of the forecast and fairness in the sense of varying accuracy across the subgroups.

%



To address under-representation bias in the training of a forecasting model, it is natural to seek a notion of fairness 
that captures the overall behaviour across all subgroups, 
while taking into account the varying amounts of training data for the individual subgroups. 
To formalise this, suppose that we learn one model $\mathcal{L}$ from the multiple trajectories and define a loss function that measures the loss of accuracy for a certain observation $Y_t^{(i,s)}$ when adopting the forecast $\hat{Y}_t$ for the overall population. For $t\in \mathcal{T}^{(i,s)}$, $i \in \mathcal{I}^{(s)}$, $s\in\mathcal{S}$, we have
\begin{equation}
    \loss^{(i,s)}(\hat{Y}_t):=||Y_t^{(i,s)}-\hat{Y}_t||^2. \label{equ:loss}
\end{equation}
Let $\mathcal{T}^+:=\cup_{i\in\mathcal{I}^{(s)},s\in\mathcal{S}}\mathcal{T}^{(i,s)}$. Following the definitions of $\mathcal{T}^{(i,s)}$, there is no observation for $t\notin \mathcal{T}^+$, in which case, loss in Equation~\eqref{equ:loss} does not exist.
To evaluate the performance of the forecasts, we only consider $\hat{Y}_t$ made in periods $t\in\mathcal{T}^+$. 
Note that, since each trajectory is of varying length, it is possible that for a certain triple $(t,i,s)$, there is no observation $Y_t^{(i,s)}$.


Following much recent work on fairness in classification, e.g., \cite{zliobaite2015relation,hardt2016equality,kilbertus2017avoiding,kusner2017counterfactual,chouldechova2020snapshot,aghaei2019learning},
we propose two objectives to address the under-representation bias, which extend group fairness \citep{feldman2015certifying} to time series are the following.

\begin{enumerate}
    \item \textbf{Subgroup Fairness}. 
    The objective seeks to equalise, across all  
    subgroups, the sum of losses for the subgroup.
    Considering the number of trajectories in each subgroup and the 
    number of observations across the trajectories may differ, 
    we include cardinality $\lvert\mathcal{I}^{(s)}\rvert,\lvert\mathcal{T}^{(i,s)}\rvert$ as weights:
    \begin{equation}
        \min_{\hat{Y}_t,t\in\mathcal{T}^+} \max_{s\in\mathcal{S}} \left \{
        \frac{1}{\lvert\mathcal{I}^{(s)}\rvert}
        \sum_{i \in \mathcal{I}^{(s)}} \frac{1}{\lvert\mathcal{T}^{(i,s)}\rvert}
        \sum_{t\in \mathcal{T}^{(i,s)}} \loss^{(i,s)}(\hat{Y}_t) \right \} \label{equ:obj-Subgroup-Fair}
    \end{equation}
    \item \textbf{Instantaneous Fairness}. The objective seeks to equalise 
    the instantaneous loss, by minimising the 
    maximum of the losses across all subgroups and all times: 
    \begin{equation}
        \min_{\hat{Y}_t,t\in\mathcal{T}^+}
        \left \{ 
        \max_{t\in\mathcal{T}^{(i,s)},i\in\mathcal{I}^{(s)},s\in\mathcal{S}} \left \{ \loss^{(i,s)}(\hat{Y}_t) \right \} \right \} \label{equ:obj-Instant-Fair}
    \end{equation}

\end{enumerate}

We then cast the learning of a LDS with such fairness considerations as a NCPOP, which can be solved efficiently using a globally-convergent hierarchy of SDP relaxations, which can be of independent interest.
A comprehensive comparison is given to illustrate the efficacy of our approach.


\section{The Fair-Forecasting Framework}
\label{sec:models}

As the simplest example of the use of subgroup fairness and instantaneous fairness, cf. Equations~\eqref{equ:obj-Subgroup-Fair} and \eqref{equ:obj-Instant-Fair}, consider their applications in linear regression. For simplicity, let us assume that the cardinality of each subgroup is the same and the lengths of all trajectories are equal. Then:

\begin{equation*}
\begin{split}
    &\min_{a,A,\hat{Y}_t,t\in\mathcal{T}^+} a\qquad\qquad\qquad\qquad\quad\textrm{Subgroup Fairness in LR} \\
     &\textrm{s.t.} \quad a \geq \sum_{i\in\mathcal{I}^{(s)} ,t\in\mathcal{T}^+}\lvert Y^{(i,s)}_t - \hat{Y}_t\rvert,\;\forall s\in\mathcal{S}\\ 
     &\qquad \hat{Y}_t = A X_t,  \\
    \vspace{10pt}
    &\min_{a,A,\hat{Y}_t,t\in\mathcal{T}^+} a \qquad\qquad\qquad\quad\textrm{Instantaneous Fairness in LR}\\ 
    &\textrm{s.t.} \quad a \geq \lvert Y^{(i,s)}_t - \hat{Y}_t \rvert,\;\forall i\in\mathcal{I}^{(s)},s\in\mathcal{S},t\in\mathcal{T}^+\\
    &\qquad \hat{Y}_t = A X_t,     
\end{split}
\end{equation*}
where $A$ concatenates the regression coefficients.
$X_t$ concatenates explanatory variables. $\hat{Y}_t$ is the dependent variable and $Y_t^{(i,s)}$ is the actual observation in a compatible fashion. 
\label{page:define-a}
The auxiliary scalar variable $a$ is used to reformulate ``$\max$'' in the objective in Equations~\eqref{equ:obj-Subgroup-Fair} and \eqref{equ:obj-Instant-Fair}.

Next, let us consider more elaborate models, which assume that there exists a LDS 
corresponding to each subgroup $s\in\mathcal{S}$.
A discrete-time model of a LDS $\mathcal{L}=(G,F,V,W)$, as in \cite{WestHarrison} or Chapter~\ref{cha:tac}, suggests that the random variable $Y_t \in\mathbb{R}$ capturing the observed component (i.e., output, observations or measurements) evolves over time $t\geq 1$ according to
\begin{equation}
\begin{split}
    \theta_{t}  &= G \theta_{t-1} + w_t, \\
    Y_t &= F' \theta_t + v_t, 
\end{split}
\tag{LDS}
\end{equation}
where $\theta_t \in \mathbb{R}^{k}$ is the  hidden component (state) and $G \in \mathbb{R}^{k\times k}$ and $F\in\mathbb{R}^{k}$ are compatible system matrices. Random variables $w_t,v_t$ capture normally-distributed process noise and observation noise, with zero means and covariance matrices $W \in\mathbb{R}^{k\times k}$ and $V \in\mathbb{R}$, respectively.



The objectives in Equations~\eqref{equ:obj-Subgroup-Fair} and \eqref{equ:obj-Instant-Fair}, subject to the state-evolution and observation equations, in Equation~\eqref{equ:LDS}, yield two operator-valued optimisation problems. 
Their inputs are $Y_{t}^{(i,s)},t\in\mathcal{T}^{(i,s)},i\in\mathcal{I}^{(s)},s\in\mathcal{S}$, i.e., the observations of multiple trajectories and the multipliers $\lambda_1,\lambda_2>0$. The operator-valued decision variables $\mathcal{O}$ include operators $F, G$, vectors $\hat{\theta}_t, \omega_t$, and scalars $\hat{Y}_t, \nu_t, a$. Notice that $t$ ranges over $t\in\mathcal{T}^+$, except for $\hat{\theta}_t$, where $t\in\mathcal{T}^+\cup\{0\}$. The auxiliary scalar variable $a$ is used to reformulate ``$\max$'' in the objective in Equations~\eqref{equ:obj-Subgroup-Fair} and \eqref{equ:obj-Instant-Fair}.
Since the process noise and observation noise are assumed to be samples of mean-zero normally-distributed random variables, we add the sum of squares of $\omega_t$ (resp. $\nu_t$) to the objective with the positive multiplier $\lambda_1$ (resp. $\lambda_2$), seeking a solution with $\omega_t$ (resp. $\nu_t$) close to zero.
Overall, the subgroup-fair and instant-fair formulations read:
\begin{mini} 
	  {\mathcal{O}}{a + \lambda_1 \sum_{t\geq 1} \|\omega_t\|^2+\lambda_2 \sum_{t\geq 1} \|\nu_t\|^2}{\label{min:FairA}}{ 
	  }{\quad\textrm{Subgroup-Fair}}
	  \addConstraint{a}{\geq
	  \frac{1}{\lvert\mathcal{I}^{(s)}\rvert}
        \sum_{i \in \mathcal{I}^{(s)}} \frac{1}{\lvert\mathcal{T}^{(i,s)}\rvert}
        \sum_{t\in \mathcal{T}^{(i,s)}} \loss^{(i,s)}(\hat{Y}_t)}
        {, s\in\mathcal{S}}
	  \addConstraint{\hat{\theta}_t}{= G \hat{\theta}_{t-1}+\omega_t}{, t\in\mathcal{T}^+}
	  \addConstraint{\hat{Y}_t}{= F' \hat{\theta}_{t}+\nu_t}{, t\in\mathcal{T}^+.}
\end{mini}
\begin{mini} 
	  {\mathcal{O}}{a  + \lambda_1 \sum_{t\geq 1} \|\omega_t\|^2+\lambda_2 \sum_{t\geq 1} \|\nu_t\|^2}{\label{min:FairB}}{
	  }{\quad\textrm{Instant-Fair}}
	  \addConstraint{a}{\geq
	  \loss^{(i,s)}(\hat{Y}_t)}{, t\in\mathcal{T}^{(i,s)},i\in\mathcal{I}^{(s)},s\in\mathcal{S}}
	  \addConstraint{\hat{\theta}_t}{= G \hat{\theta}_{t-1}+\omega_t}{, t\in\mathcal{T}^+}
	  \addConstraint{\hat{Y}_t}{= F' \hat{\theta}_{t}+\nu_t}{, t\in\mathcal{T}^+.}
\end{mini}

For comparison, we use a traditional formulation that focuses on minimising the overall loss: 
\begin{mini}
	  {\mathcal{O}}{
	  \sum_{s \in \mathcal{S}} \quad
	  \sum_{i \in \mathcal{I}^{(s)}} 
	  \sum_{t\in \mathcal{T}^{(i,s)}} \loss^{(i,s)}(\hat{Y}_t) + \lambda_1 \sum_{t\geq 1} \|\omega_t\|^2+\lambda_2 \sum_{t\geq 1} \|\nu_t\|^2}{\label{min:Unfair}}{ 	  }{\quad\textrm{Unfair}} 
	  \addConstraint{\qquad\qquad \hat{\theta}_t}{= G \hat{\theta}_{t-1}+\omega_t}{, t\in\mathcal{T}^+}
	  \addConstraint{\qquad\qquad \hat{Y}_t}{= F' \hat{\theta}_{t}+\nu_t}{, t\in\mathcal{T}^+.}
\end{mini}


As discussed in Appendix~\ref{cha:ncpop} and Chapter~\ref{cha:tac}, these operator-valued optimisation problems (i.e., ``Unfair'', ``Instant-Fair'', and ``Subgroup-Fair'') can be convexified to any given accuracy, and thence solved efficiently, under the Archimedean assumption related to stability of the LDS, 
which entails that the estimates of states and observations remain bounded, and thus all operator-valued decision variables remain bounded.

\section{Numerical Illustrations}

Under-representation bias considers the situation where some subgroups would be given unfair treatments, either due to the varying numbers or lengths of trajectories across subgroups.
In response to under-representation bias, we have introduced two natural fairness notions for forecasting.
We use a LDS to predict the next observation, as in Chapter~\ref{cha:tac}.
Then, we have given two formulations associated with each notion. We tested our formulations on the famous COMPAS dataset. 
Our implementation is available online \footnote{\url{https://github.com/Quan-Zhou/Fairness-in-Learning-of-LDS}}.



\subsection{Generation of Biased Training Data}
\label{sec:Biased Training Data Generalisation}

To illustrate the impact of our models on data with varying degrees of under-representation bias, we consider a method to generate data with a given degree of bias, which is based on \cite[cf. Section 2.2]{blum2019recovering}.
Suppose that there is one advantaged subgroup ($s_0$) and one disadvantaged subgroup ($s_1$), i.e., $S=\{s_0,s_1\}$, with the set of trajectories $\mathcal{I}^{(s_0)}$ and $\mathcal{I}^{(s_1)}$ for each subgroup. Under-representation bias enters the training set in the following steps.
\begin{enumerate}
\item Consider that the LDS for both subgroups $\mathcal{L}^{(s)},s\in\mathcal{S}$ have the same system matrices: $$G^{(s)}=\begin{bmatrix}
0.99 & 0 \\
1.0 & 0.2 \end{bmatrix},F^{(s)}=\begin{bmatrix}
1.1 \\ 0.8 
\end{bmatrix},$$
while the covariance matrices $V^{(s)},W^{(s)},s\in\mathcal{S}$ are sampled randomly from a uniform distribution over $[0,1)$ and $[0,0.1)$, respectively. The initial states $\hat{\theta}_0^{(s)}$ of both subgroups are $5$ and $7$.
    \item Observations $Y^{(i,s)}_t$ are sampled from the corresponding LDS $\mathcal{L}^{(s)}$, thus $Y^{(i,s)}_t\sim\mathcal{L}^{(s)}$.
    \label{page:define-beta}
    \item Let $\mathcal{\beta}^{(s_1)}$ denote the probability that an observation from subgroup $s_1$ stays in the training data, and $0\leq \mathcal{\beta}^{(s_1)} \leq 1$. It can be seen as the ratio of the number of observations in disadvantaged subgroup to that of advantaged subgroup. The degree of under-representation bias can be controlled by simply adjusting $\mathcal{\beta}^{(s_1)}$.
    Smaller values of $\mathcal{\beta}^{(s_1)}$ correspond to higher level of bias in the training set.
\end{enumerate}
The last step makes the number of observations of the disadvantaged subgroup less than that of the advantaged subgroup when $0\leq \mathcal{\beta}^{(s_1)} <1$.
Hence, the advantaged subgroup becomes over-represented. 
Note that for a small sample size, it is necessary to make sure that there is at least one observation in each subgroup at each period. 

\subsection{Effects of Under-Representation Bias on Forecast}
\begin{figure}[htp]
    \centering
\includegraphics[width=0.5\textwidth]{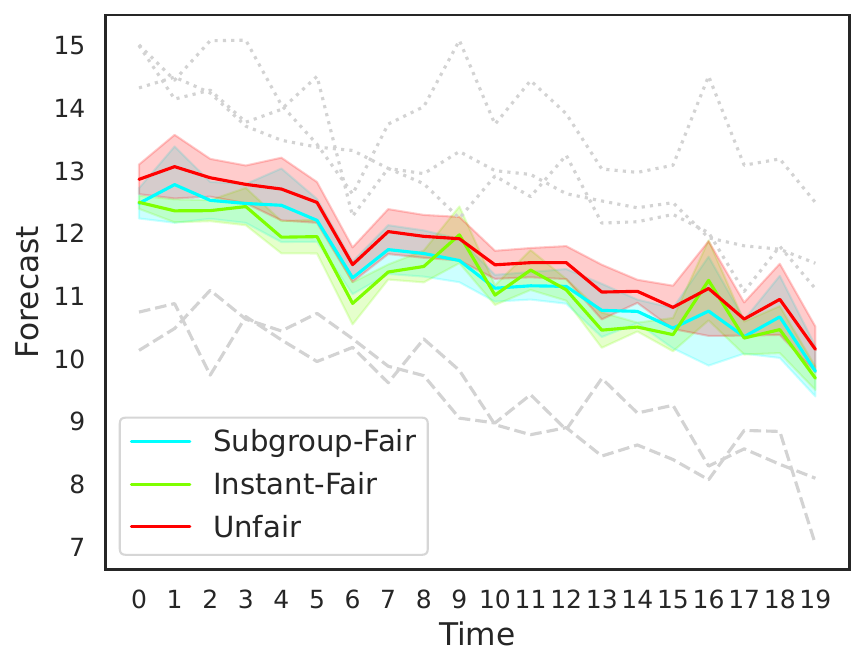} 
\caption[Prediction output of our fairness-aware methods compared against a traditional formulation]{Forecast obtained using Equations~(\ref{min:FairA}--\ref{min:Unfair}): the solid lines in primary colours with error bands display the mean and standard deviation of the forecasts over 10 experiments. For reference, dotted lines and dashed lines in grey denote the trajectories of observations of advantaged and disadvantaged subgroups, respectively, before discarding any observations.
}
\label{fig:LinePlot}
\end{figure}

To display the impact of our models on data with varying degrees of under-representation bias, 
Figure~\ref{fig:LinePlot} illustrates 10 experiments with general forecasting procedures. For each experiment, the same set of observations $Y_t^{(i,s)},t\in \mathcal{T}^{(i,s)}$, $i \in \mathcal{I}^{(s)}$, $s\in\mathcal{S}$ is reused, and the trajectories of advantaged and disadvantaged subgroups are denoted by dotted curves and dashed curves, respectively. However, in each experiment, a subset of observations with the same cardinality is randomly selected and discarded and thus a new biased training set is generated, albeit based on the same ``ground set'' of observations. The three models in Equations~(\ref{min:FairA}--\ref{min:Unfair}) are applied in each experiment with $\lambda_1$ of 1, 3, and 5, respectively, as chosen by iterating over integers 1 to 10, while $\lambda_2$ remains 0.01, 
The mean of forecast $\hat{Y}_t$ across 10 experiments and its standard deviation are shown as solid curves with error bands. The red curve gives an overview of how a prediction without considering fairness would cause an unevenly distributed prediction loss for each subgroup. This is simply because the advantaged subgroup is of larger cardinality, and the overall loss would decrease more steeply if the predicted trajectory gets closer to the advantaged subgroup.

\subsection{Fairness as a Function of Bias}
\begin{figure}[htp]    
    \centering
\includegraphics[width=0.6\textwidth]{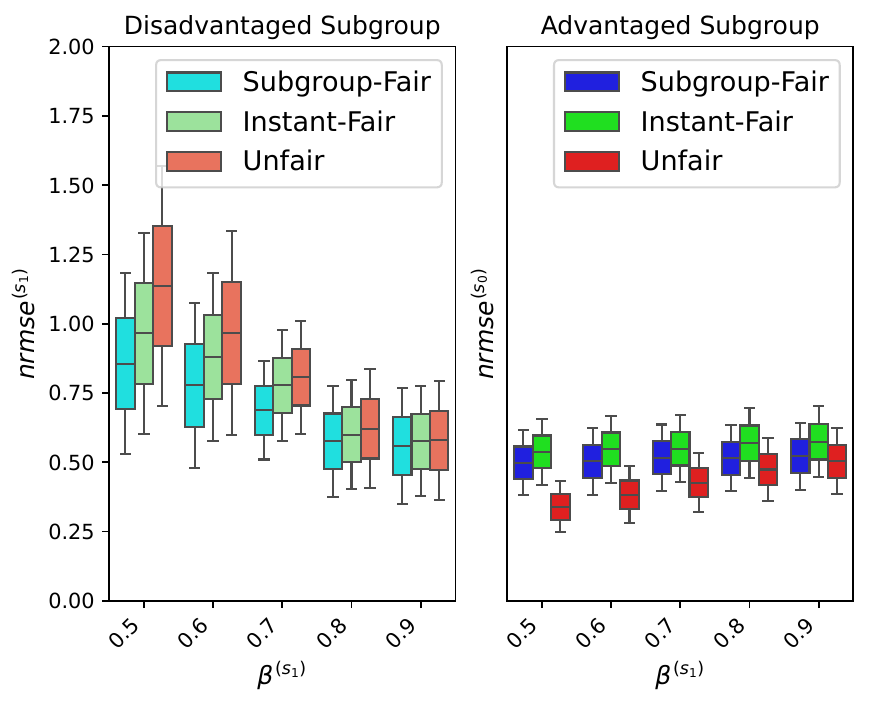}
\caption[Accuracy as a function of the degree of under-representation bias]{Accuracy as a function of the degree of under-representation bias: the boxplot of $\nrmse^{(s)},s\in\mathcal{S}$ against $\mathcal{\beta}^{(s_1)}$, where $\mathcal{\beta}^{(s_1)}=[0.5,0.55,\dots,0.9]$. Each box for the quartiles of $\nrmse^{(s)}$ is obtained from $10$ experiments, with observations generated in the same way as the ones in Figure~\ref{fig:LinePlot}.}
\label{fig:BetaPlot}
\end{figure}

Figure~\ref{fig:BetaPlot} suggests how the degree of bias affects accuracy in each subgroup with and without considering fairness. 
With the number of trajectories in both subgroups set to two, i.e., $|\mathcal{I}_{s_0}|=|\mathcal{I}_{s_1}|=2$, we vary the degree of bias by adjusting $\mathcal{\beta}^{(s_1)}$ within the range of $[0.5, 0.9]$. 
To measure the effect of the degree on accuracy, we introduce the \acrfull{nrmse} fitness value for each subgroup $s\in\mathcal{S}$: 
\begin{equation}
\mathrm{\nrmse^{(s)}}:=\sqrt{\frac{\sum_{i\in\mathcal{I}^{(s)}}\sum_{t\in\mathcal{T}^{(i,s)}}\left( Y_t^{(i,s)}-\hat{Y}_t\right)^2}{\sum_{i\in\mathcal{I}^{(s)}}\sum_{t\in\mathcal{T}^{(i,s)}}\left( Y_t^{(i,s)}-\mean^{(s)}\right)^2} },
\label{equ:NRMSE_s}
\end{equation}
where $\mean^{(s)}:=\frac{1}{\lvert\mathcal{I}^{(s)}\rvert} \sum_{i\in\mathcal{I}^{(s)}}\frac{1}{\lvert\mathcal{T}^{(i,s)}\rvert}\sum_{t\in\mathcal{T}^{(i,s)}} Y_t^{(i,s)}$. Higher $\nrmse^{(s)}$ indicates lower accuracy for subgroup $s$, i.e., the predicted trajectory of subgroup-blind $\mathcal{L}$ is further away from this subgroup. 

The training data are generated in the same way as the set of observations used in Figure~\ref{fig:LinePlot}, but with two trajectories in each subgroup ($|\mathcal{I}_{s_0}|=|\mathcal{I}_{s_1}|=2$).
Then, the biased training data generalisation process (described in Section~\ref{sec:Biased Training Data Generalisation}) is applied in each experiment with the value of $\mathcal{\beta}^{(s_1)}$ selecting from $0.5$ to $0.9$ at the step of $0.1$.
For each value of $\mathcal{\beta}^{(s_1)}$, three models in Equations~(\ref{min:FairA}--\ref{min:Unfair}) are conducted for $10$ experiments with a new biased training set in each experiment.
Therefore, the quartiles of $\nrmse^{(s)}$ across $10$ experiments for each subgroup are shown as boxes in Figure~\ref{fig:BetaPlot}.

One could expect that nrmse fitness values of the advantaged subgroup in Figure~\ref{fig:BetaPlot} to be generally lower than those of the disadvantaged subgroup ($\nrmse^{(s_1)}\geq\nrmse^{(s_0)}$), leaving a gap. Those gaps narrow down as $\mathcal{\beta}^{(s_1)}$ increases, simply because more observations of disadvantaged subgroup remain in the training data. Compared the to ``Unfair'', models with fairness constraints, i.e., ``Subgroup-Fair'' and ``Instant-Fair'', show narrower gaps and higher fairness between two subgroups. More surprisingly, when $\nrmse^{(s_0)}$ decreases as $\mathcal{\beta}^{(s_1)}$ gets close to $0.5$, ``Subgroup-Fair'' model still can keep $\nrmse^{(s_1)}$ at almost the same level, indicating a rise in overall accuracy. This is in contrast to the results of \cite{zliobaite2015relation,dutta2019information} in classification, but in line with recent work \cite{maity2021does}.

\subsection{Runtime}

Minimising multivariate operator-valued polynomial optimisation problems (\ref{min:FairA}--\ref{min:Unfair}) is a known non-trivial problem. The method is tested on globally \acrshort{NPA} hierarchy \citep{navascues2008convergent} of SDP relaxations, and its term-sparsity exploiting variant \acrshort{TSSOS} to develop fast computational methods.
See \cite{klep2022sparse,wang2021tssos,wang2020chordal,wang2020exploiting}.
The SDP of a given moment order $d$ in the respective hierarchy can be constructed using \texttt{ncpol2sdpa} 1.12.2\footnote{\url{https://github.com/peterwittek/ncpol2sdpa}} of \cite{wittek2015algorithm} or the tools of \cite{wang2020exploiting} \footnote{\url{https://github.com/wangjie212/TSSOS}} and then solved by \texttt{mosek} 9.2 of \cite{mosek2020mosek}.

\begin{figure}
\centering{
\includegraphics[width=0.6\textwidth]{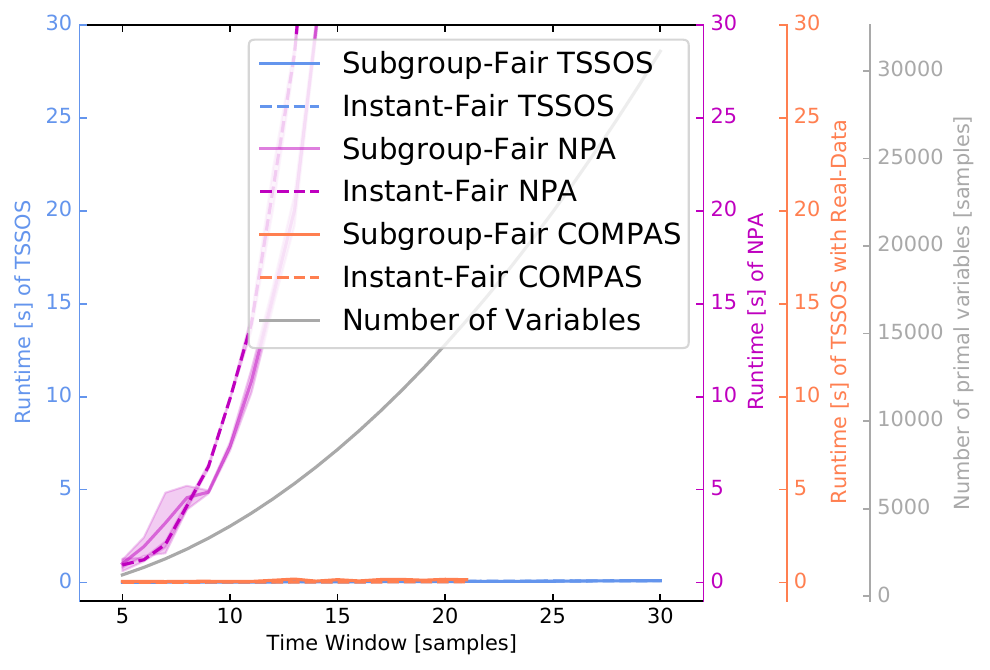} }
\caption[The dimensions of relaxations and the runtime of SDP thereupon as a function of the length of time window]{The dimensions of relaxations and the runtime of SDP thereupon as a function of the length of time window. Runtime of TSSOS and NPA is displayed in cornflower-blue and deep-pink curves, respectively, while the grey curve shows the number of variables in relaxations. Additionally, the runtime of TSSOS using the COMPAS dataset in Figure~\ref{fig:COMPASPlot}, is also displayed as coral-coloured curves. 
For runtime, the mean and mean $\pm$ one standard deviations across three experimental runs are presented by curves with shaded error bands.}
\label{fig:TimePlot-jair}
\end{figure}

In Figure~\ref{fig:TimePlot-jair}, we illustrate the runtime and size of the relaxations as a function of the length of the time window.
Models ``Subgroup-Fair'' in Equation~\eqref{min:FairA} and ``Instant-Fair'' in Equation~\eqref{min:FairB} are implemented three times for each length of the time window, with the same dataset used in Figure~\ref{fig:BetaPlot}.
The type of models, i.e., ``Subgroup-Fair'' (solid curves) and ``Instant-Fair'' (dashed curves), is distinguished by line styles.
The deep-pink and cornflower-blue curves show the runtime of the first-order SDP relaxation of NPA and the second-order SDP relaxation of TSSOS hierarchy, respectively, implemented with five CPUs and 64GB of memory per CPU.
The mean and mean $\pm$ 1 standard deviation of runtime across three experimental runs are presented by curves with shaded error bands.
The grey curve displays the number of variables in the first-order SDP relaxation of our models in Equations~\eqref{min:FairA} against the length of time window.
Further, models ``Subgroup-Fair'' in Equation~\eqref{min:FairA} and ``Instant-Fair'' in Equation~\eqref{min:FairB} are implemented once via TSSOS for each length of the time window, using COMPAS dataset, as the experiment in Figure~\ref{fig:COMPASPlot}, with the runtime displayed by a coral solid curve and a coral dashed curve, respectively.
It is clear that the runtime of TSSOS exhibits a modest growth with the length of time window, while that of the plain-vanilla NPA hierarchy grows much faster.

\section{Numerical Results of COMPAS Dataset}
Finally, we wish to suggest the broader applicability of the two notions of subgroup fairness and instantaneous fairness. We use the well-known dataset \citep{angwin2016machine} of estimates of the likelihood of recidivism made by the Correctional Offender Management Profiling for Alternative Sanctions (COMPAS), as used by courts in the United States. 
The COMPAS dataset, analysed by ProPublica, comprises of defendants' gender, race, age, charge degree, COMPAS recidivism scores, two-year recidivism label, as well as information on prior incidents. 
The COMPAS recidivism scores, ranging from 1 to 10, are positively related to the estimated likelihood of recidivism, given by the COMPAS system.
The two-year recidivism label denotes whether a person actually got rearrested within two years (label 1) or not (label 0). If the two-year recidivism label is $1$, there is also information concerning the recharge degree and the number of days until the person gets rearrested.
The dataset also consists of information on ``Days before Re-offending'', which is the date difference between the defendant's crime offend date and recharge offend date. It could be negatively correlated to the defendant's actual risk level while the COMPAS recidivism scores would be the estimated risk level.
\subsection{An Alternative Approach to COMPAS Dataset}
From the COMPAS dataset, 
we choose $119$ defendants with recidivism label being 1, who are either African-American or Caucasian, male, within the age range of 25-45, and with prior crime counts less than two, with charge degree M and recharge degree M1 or M2.
The defendants are partitioned into two subgroups by their ethnicity and then partitioned by the type of their recharge degree (M1 or M2). Hence, we obtain the $4$ sub-samples.

\begin{figure*}[htp]
\centering
\centering{
\includegraphics[width=0.5\textwidth]{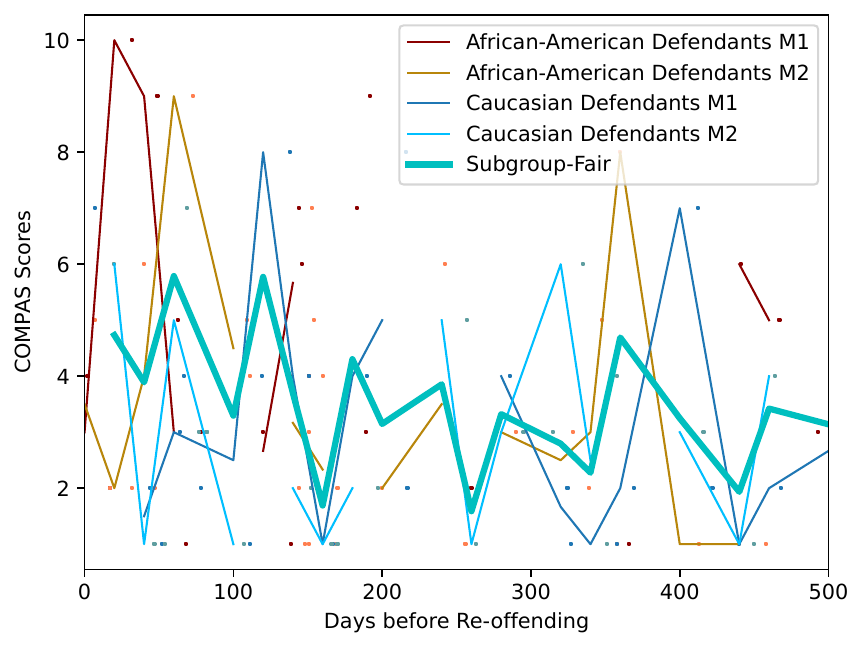}}
\caption[COMPAS recidivism scores of African-American and Caucasian defendants against the actual days before their re-offending]{COMPAS recidivism scores of African-American and Caucasian defendants against the actual days before their re-offending. The sample of defendants' scores is divided into four sub-samples based on race and the type of re-offending, distinguished by colours. Dots and curves with the same colour denote the scores of one sub-sample and the trajectory extracted from the scores, respectively. The cyan curve displays the result of ``Subgroup-Fair'' model with four trajectories.}
\label{fig:COMPASPlot}
\end{figure*}

In the days-to-reoffend-vs-score plot, such as Figure~\ref{fig:COMPASPlot}, dots suggest COMPAS recidivism scores of the four sub-samples against the days before rearrest.
Each curve represents one model, either subgroup-dependent (plotted thin) or Subgroup-Fair (plotted thick).
The thick cyan curve is the race-blind prediction from our Subgroup-Fair method, which equalises scores across the two subgroups.
Ideally, one should like to see smooth, monotonically decreasing curves, overlapping across all subgroup-dependent models. For each sub-sample, the aggregate deviation from the Subgroup-Fair curve would be similar to the aggregate deviations of other sub-samples. 

In Figure~\ref{fig:COMPASPlot}, the dots are far from the ideal monotonically decreasing curve. Furthermore, the subgroup-specific curves (plotted thin) are very different from each other (``subgroup-specific models are unfair''). Specifically, the red and yellow curves are above the sky blue and cornflower blue curves (``at the same risk level, Caucasian defendants get lower COMPAS scores''). 
Notice that the subgroup-dependent models are obtained as follows: we discretise time to $20$-day periods. 
For each subgroup, we check if anyone re-offends within $20$ days (the first period). If so, the (average) COMPAS score (for all cases within the 20 days) is recorded as the observation of the first period of the trajectory of the sub-sample. If not, there is no observation of this period. We repeat this for the subsequent periods and for the three other sub-samples. 



\subsection{A Comparison Against the State of the Art on COMPAS Dataset}
\label{sec:the-State-of-the-Art}

Generally speaking, fairness objectives or constraints might not be easily applied to models that are already in use in applications. For such systems, revision of the model output with some post-processing tools would be a widely applicable and practical solution. 
Since we have shown the existence of unfairness in COMPAS recidivism scores, we now illustrate this approach to improve upon the COMPAS scores by using post-processing methods that embed our fairness notions. We then compare our methods using the AI Fairness 360 toolkit \texttt{AIF360}\footnote{\textcolor{black}{An open-source toolkit created by IBM Research, designed to examine, report, and mitigate discrimination and bias in machine learning models, is available at \url{https://github.com/Trusted-AI/AIF360}.}} of \cite{aif360-oct-2018}.

\paragraph{The training set and test sets: }
The sample set contains $1005$ defendants, whose race is either African-American or Caucasian, selected from the first $1200$ rows of the COMPAS dataset.
For a single trial, we randomly pick $80\%$ of samples as the training set then test the output on the rest $20\%$ of the samples. Each trial uses a new batch of the training set and the test set generated from the same sample set of $1005$ defendants.


Notice that the sample set is biased as there are only $403$ Caucasian defendants.
Since existing data may generally contain biases, stemming for example from poor information acquisition process \citep{bertail2021learning}, due to historical and social injustices \citep{ferrer2021bias}, we seek other methods to validate our approaches.
To this end we randomly remove some observations of African-American defendants from the original test set, such that the number of defendants in both subgroups are the same.
The resulted subset is called the re-weighted test set.

\paragraph{Performance indices: }
We use three baseline unfairness metrics (i.e., independence, separation, and sufficiency), as well as prediction inaccuracy, to measure the performance of post-processing models.
Essentially, the sample set includes two race subgroups $\mathcal{S}=\{$African-American defendants (AA), Caucasian defendants (C)$\}$. The recidivism label and the prediction of the recidivism label outputted from a model, are denoted by binary variables $Y$ and $\hat{Y}$ respectively. 
Let $\Pr(\hat{Y}|Y,s)$ be the probability of a defendant from subgroup $s$ with recidivism label $Y$
being predicted to recidivism label $\hat{Y}$.
We set $Y=1$ and $\hat{Y}=1$ to be a defendant re-offending and being predicted to re-offend, thus they are negative events.
Further, we define the indices of three baseline unfairness metrics (IND, SP, SF), inaccuracy (INA) and their re-weighted versions (i.e., INDrw, SPrw, SFrw, INArw) in Equation~\eqref{equ:indices}:
\begin{equation}
\begin{split}
    \textrm{IND(rw)}&:=\bigg\lvert \pp{\hat{Y}=1\mid s=\textrm{AA}}-\pp{\hat{Y}=1\mid s=\textrm{C}} \bigg\rvert,\\
    \textrm{SP(rw)}&:=\bigg\lvert \pp{\hat{Y}=0\mid Y=1,s=\textrm{AA}}-\pp{\hat{Y}=0\mid Y=1,s=\textrm{C}} \bigg\rvert \\
    &\; +\bigg\lvert \pp{\hat{Y}=1\mid Y=0,s=\textrm{AA}}-\pp{\hat{Y}=1\mid Y=0,s=\textrm{C}} \bigg\rvert, \\
    \textrm{SF(rw)} &:=\bigg\lvert \pp{Y=1\mid \hat{Y}=1,s=\textrm{AA}}-\pp{Y=1\mid \hat{Y}=1,s=\textrm{C}} \bigg\rvert\\
    &\; +\bigg\lvert \pp{\hat{Y}=0\mid Y=0,s=\textrm{AA}}-\pp{\hat{Y}=0\mid Y=0,s=\textrm{C}} \bigg\rvert, \\
    \textrm{INA(rw)} &:=\pp{Y\neq \hat{Y}},
\end{split}
\label{equ:indices}
\end{equation}
where, for example, IND(rw) implies both indices IND and INDrw. The difference between IND and INDrw is that IND measures the performance of a model on the original test set while INDrw on the re-weighted test set. The same applies for SP(rw), SF(rw), INA(rw).
To interpret the definitions in Equation~\eqref{equ:indices}: IND(rw) are race-wise absolute difference of negative rate; SP(rw) combine race-wise absolute difference of false false positive and false negative rates; SF(rw) captures the race-wise absolute difference of positive predictive value and negative predictive value; INA(rw) measure inaccuracy of test set.
In our setting, smaller values of IND(rw), SP(rw), SF(rw), INA(rw) indicate better performance in terms of independence, separation, sufficiency and accuracy, respectively. 

\paragraph{Classification thresholds: }
\label{page:define-yhat}
Since the outputs of all post-processing tools implemented in this paper and COMPAS system are 
(probability) scores, denoted by $\hat{y}$, from varying intervals and, to transfer these (probability) scores to binary labels, we would need a threshold such that $\hat{Y}=1$ when the score $\hat{y}$ is higher than this threshold, and $\hat{Y}=0$ otherwise. 
For ease of comparison, we define uni-race thresholds which differ across different models but all of them are defined as the $p^{\textrm{th}}$ percentile of all scores outputted by the corresponding model, where $p\in[0,100]$ is fixed.
Notice that there is a gap between the percentage of recidivism in African-American defendants ($46\%$) and Caucasian defendants ($59\%$) in terms of the sample set, and we call those percentages base rates, as in \cite{pleiss2017fairness}.
In fairness to African-American defendants, we introduce race-wise thresholds using base rates: for each model, the percentage of defendants in a subgroup whose scores are higher than the subgroup's threshold needs to be the same as the subgroup's base rate.

\paragraph{Post-processing methods: }

Associated with our fairness notions, we propose two post-processing methods. 
Both methods use simple race-wise linear regression models 
\begin{equation}
\hat{y}^{(i,s)}=A^{(s)} X^{(i,s)} + e^{(s)}, i\in\mathcal{I}^{(s)}, s\in\mathcal{S},
\label{equ:post-process-constraints}
\end{equation}
where the subscript $t$ is removed such that we only consider prediction in one period. In other words, we cast the problem of prediction into classification.
$A^{(s)}$ concatenates the regression coefficients. $X^{(i,s)}$ concatenates explanatory variables, including COMPAS recidivism score, prior incidents (i.e., the sum of ``prior counts'', ``juv\_ fel\_ count'' and ``juv\_ misd\_ count''), age category (i.e., 1 if age is less than 25 and 0 otherwise), and recidivism label.
$e^{(s)}$ corresponds to a noise to the linear relationship.
$\hat{y}^{(i,s)}$ is the post-processed recidivism score of the defendant $i$ in subgroup $s$, and $Y^{(i,s)}$ is the actual recidivism label (i.e., the ground truth).
Let $\loss^{(i,s)}(\hat{y}):=\|Y^{(i,s)}-\hat{y}^{(i,s)}\|$, our post-processing methods are Equation~\eqref{equ:post-process-formulations} subject to Equation~\eqref{equ:post-process-constraints}, with $\lambda_3=0.05$. Further, the score $\hat{y}^{(i,s)}$ would be mapped to the binary prediction of recidivism label $\hat{Y}^{(i,s)}$ using a threshold. 

\begin{equation}
\begin{array}{rl}
\textrm{\textbf{Subgroup-Fair}}
&\min_{\hat{y},A,e} \left \{ \max_{s\in\mathcal{S}} \left \{
\frac{1}{\lvert\mathcal{I}^{(s)}\rvert}
\sum_{i \in \mathcal{I}^{(s)}} \loss^{(i,s)}(\hat{y}) \right \} + \lambda_3 \sum_{s\in\mathcal{S}}  \left(e^{(s)}\right)^2 \right\}\\
\textrm{\textbf{Instant-Fair}}
&\min_{\hat{y},A,e}\left \{ \max_{i\in\mathcal{I}^{(s)},s\in\mathcal{S}} \left \{ \loss^{(i,s)}(\hat{y}) \right \}  + \lambda_3 \sum_{s\in\mathcal{S}}  \left(e^{(s)}\right)^2\right \}
\end{array}
\label{equ:post-process-formulations}
\end{equation}

\begin{figure*}[tbp]
\includegraphics[width=0.49\textwidth]{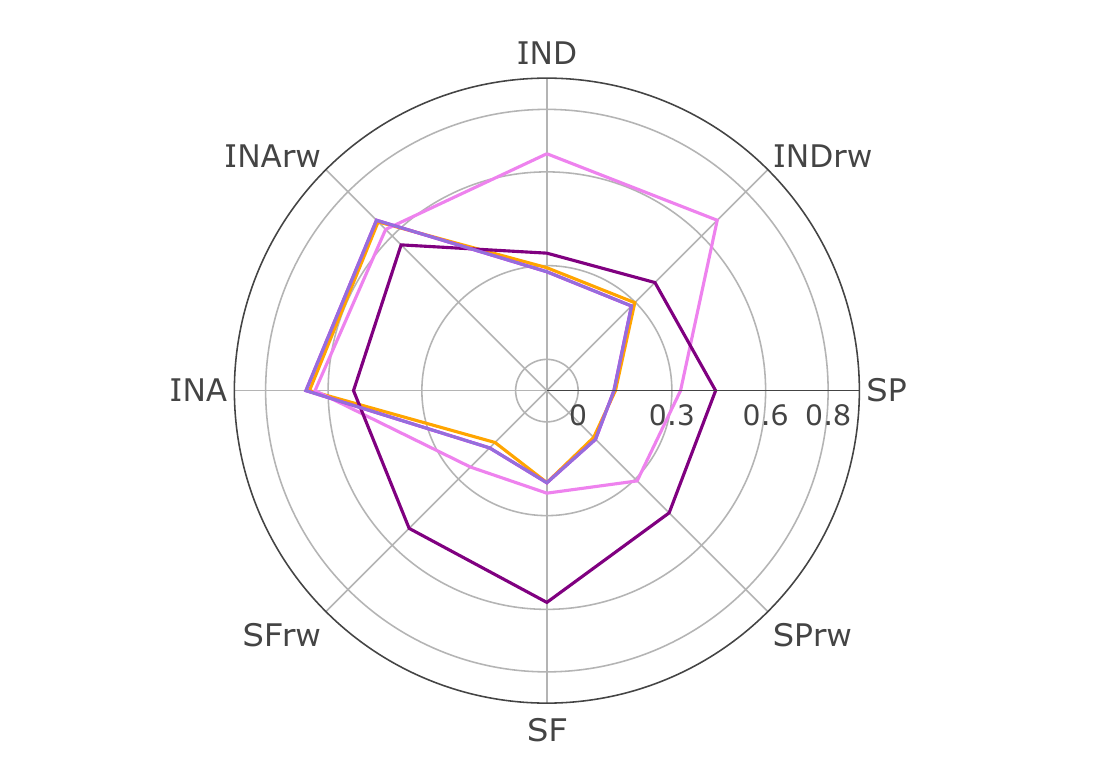}
\includegraphics[width=0.49\textwidth]{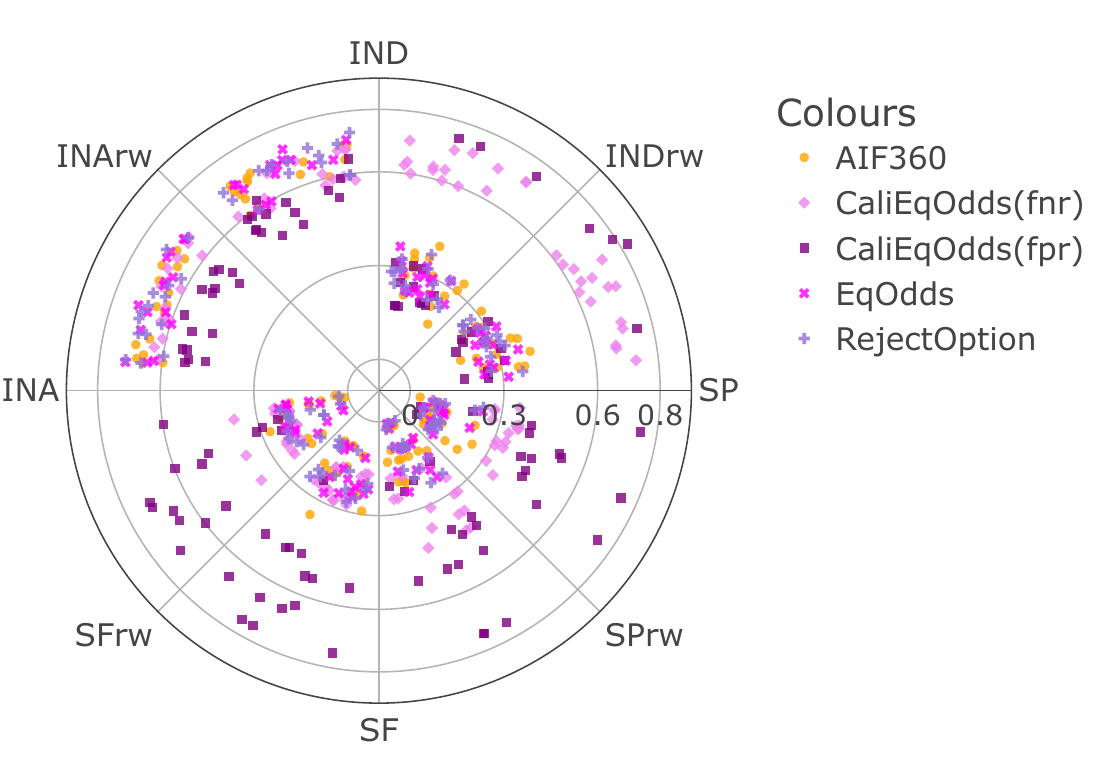}
\caption[The state of the art in post-processing for improving fairness]{The state of the art in post-processing for improving fairness. 
Five post-processing methods implemented in AI Fairness 360 toolkit, i.e., ``AIF360'' (yellow), ``CaliEqOdds(fnr)'' (violet), ``CaliEqOdds(fpr)'' (purple), ``EqOdds'' (fuchsia), and ``RejectOption'' (light purple), are run in five trials per each of three uni-race thresholds $p=[47, 53, 60]$ ($5\times 5\times 3$ runs).
Left: The mean values of eight indices on eight angular axes. 
Right: 
Each sector is labelled with the index immediately counter-clockwise to it.
Dots in one sector denote the values of the corresponding index of all experimental runs.}
\label{fig:AIF360}
\end{figure*}

\paragraph{}
In Figure~\ref{fig:AIF360}, we test the performance of all post-processing methods implemented in AI Fairness 360 toolkit: 
\begin{itemize}
    \item ``AIF360'': calibrated equalised odds post-processing with cost constraint being a combination of both false negative rate and false positive rate, as suggested by the authors of the AI Fairness 360 toolkit \cite{aif360-oct-2018};
    \item ``CaliEqOdds(fnr)'': calibrated equalised odds post-processing with cost constraint being the false negative rate;
    \item ``CaliEqOdds(fpr)'': calibrated equalised odds post-processing with cost constraint being the false positive rate;
    \item ``EqOdds'': equalised odds post-processing;
    \item ``RejectOption'': reject option classification.
\end{itemize}
Note that ``AIF360'', ``CaliEqOdds(fnr)'', ``CaliEqOdds(fpr)'' are based on the fairness notion of ``calibrated equalised odds'' in \cite{pleiss2017fairness}. 
``EqOdds'' is derived from the fairness notion of ``equalised odds'' in \cite{hardt2016equality}.
``RejectOption'' comes from \cite{kamiran2012decision}, which is rooted in the fairness notion of ``demographic parity''.
Those five methods are implemented in five trials for each of three uni-race thresholds $p=[47, 53, 60]$ ($5\times 5\times 3$ runs). 
The left subplot displays mean values of eight indices across all five trials and three thresholds, with each angular axis representing one index, and each colour denoting one post-processing method.
The right subplot represents the values of all experimental runs as dots in a circular sector, with each sector representing one index.
Each sector is labelled with the index immediately counter-clockwise to it. For instance, the sector between the labels ``IND'' and ``INDrw'' displays the ``IND'' values of all experimental runs.
In both subplots, the value represented by a dot is displayed by its distance from the original point, with shorter distances indicating better performance.

Figure~\ref{fig:AIF360} depicts a summary of the state of the art on COMPAS dataset, and how we select the appropriate method to benchmark our own algorithms.
Referring to this figure, since in both subplots, most of yellow (``AIF360'') dots are relatively closer to the origin than other dots, it seems fair to consider ``AIF360'' as the state of the art post-processing method, at least within those implemented in AI Fairness 360 toolkit, and to compare our methods against it in the following.

\begin{figure*}[tbp]
\centering
\begin{tabular}{p{0.4\textwidth} p{0.6\textwidth}} \vspace{0pt}\includegraphics[width=0.39\textwidth]{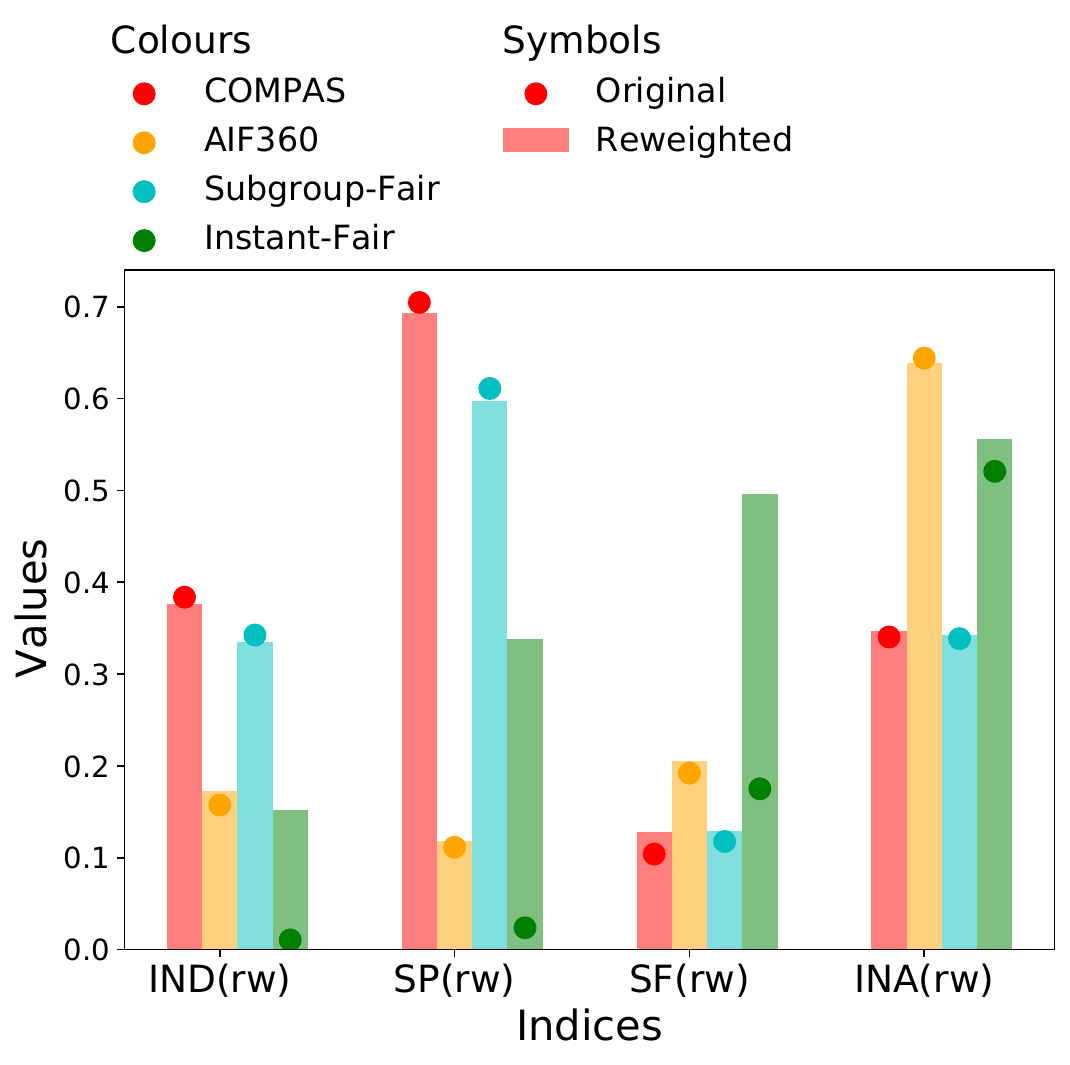} & \vspace{0pt}\includegraphics[width=0.4\textwidth]{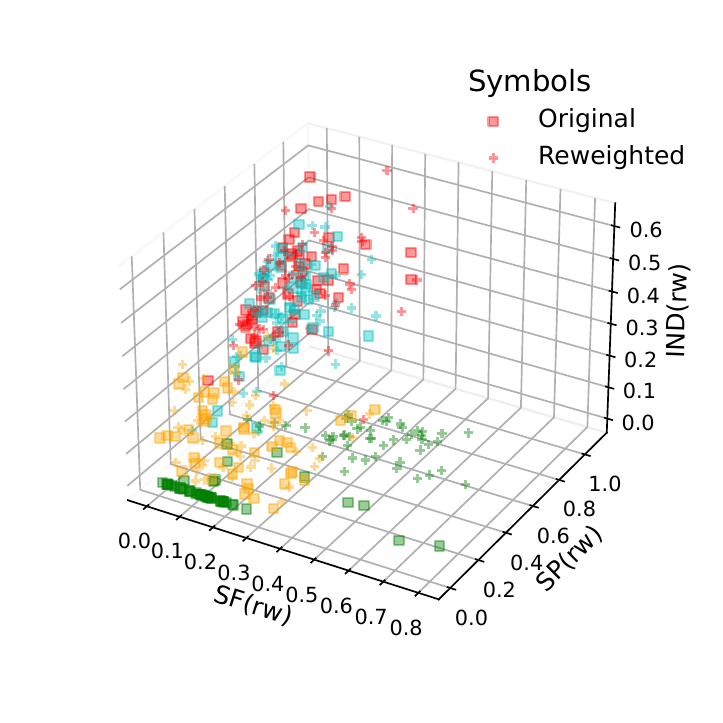} 
\end{tabular}
\caption[the COMPAS scores, and our fairness-aware post-processing methods compared with state of the art, at a fixed threshold]{The effects of our methods, when applied as post-processing. Three post-processing methods ``subgroup-fair'' (blue), ``instant-fair'' (green), ``AIF360'' (yellow) are run in $50$ trials, and their outputs are compared with the COMPAS scores (red) with race-wise thresholds being base rates.
Left: The mean values of original (dots) and re-weighted (bars) indices associated with outputs from four models.
Right: The original (squares) and re-weighted (crosses) unfairness metrics (i.e., INDrw, SPrw, SFrw) of all experimental runs.}
\label{fig:postprocess}
\end{figure*}

In Figure~\ref{fig:postprocess}, COMPAS scores (``COMPAS'' red), the state of the art (``AIF360'', yellow), and the outputs of our methods ``Subgroup-Fair'' (blue) and ``Instant-Fair'' (green) are evaluated across 50 trials using base rates as race-wise thresholds ($50\times 4$ runs).
The left subplot illustrates the average performance of four methods, where dots represent the mean values of original indices, and bars are those of re-weighted indices.
The right subplot displays fairness performance of all experimental runs in a triangular area.
For a single run, the original unfairness metrics (i.e., IND, SP, SF) are denoted by one square and re-weighted ones (i.e., INDrw, SPrw, SFrw) are shown as one cross. 
A marker, that is, a dot or a square, represents the value of IND(rw), SP(rw), SF(rw), by its positions along the left, right, and bottom axes in ternary coordinates.
The colour of this marker denotes the method used in this run.

Figure~\ref{fig:postprocess} illustrates the performance of our methods compared with the state of the art and COMPAS scores when using base rates as race-wise thresholds. 
As we can see on the left, there is not much difference between the values of original indices and their re-weighted versions, except that ``Instant-Fair'' generally performs worse in re-weighted version than in original version. 
It implies that ``Instant-Fair'' might not be appropriate to use in this case because its performance varies with the test set being re-weighted or not.
The performance of ``Subgroup Fair'' is similar to that of ``COMPAS'', but with a slight improvement in IND(rw) and SP(rw).
``AIF360'' seems to sacrifice a lot of accuracy for lower average fairness indices, while its fairness performance shows a lot of variability, as shown in  the right subplot.
On the contrary, the concentration of blue and green markers (i.e., dots and squares) indicates less variability of our methods.

\begin{figure*}[tbp]
\centering
\includegraphics[width=0.6\textwidth]{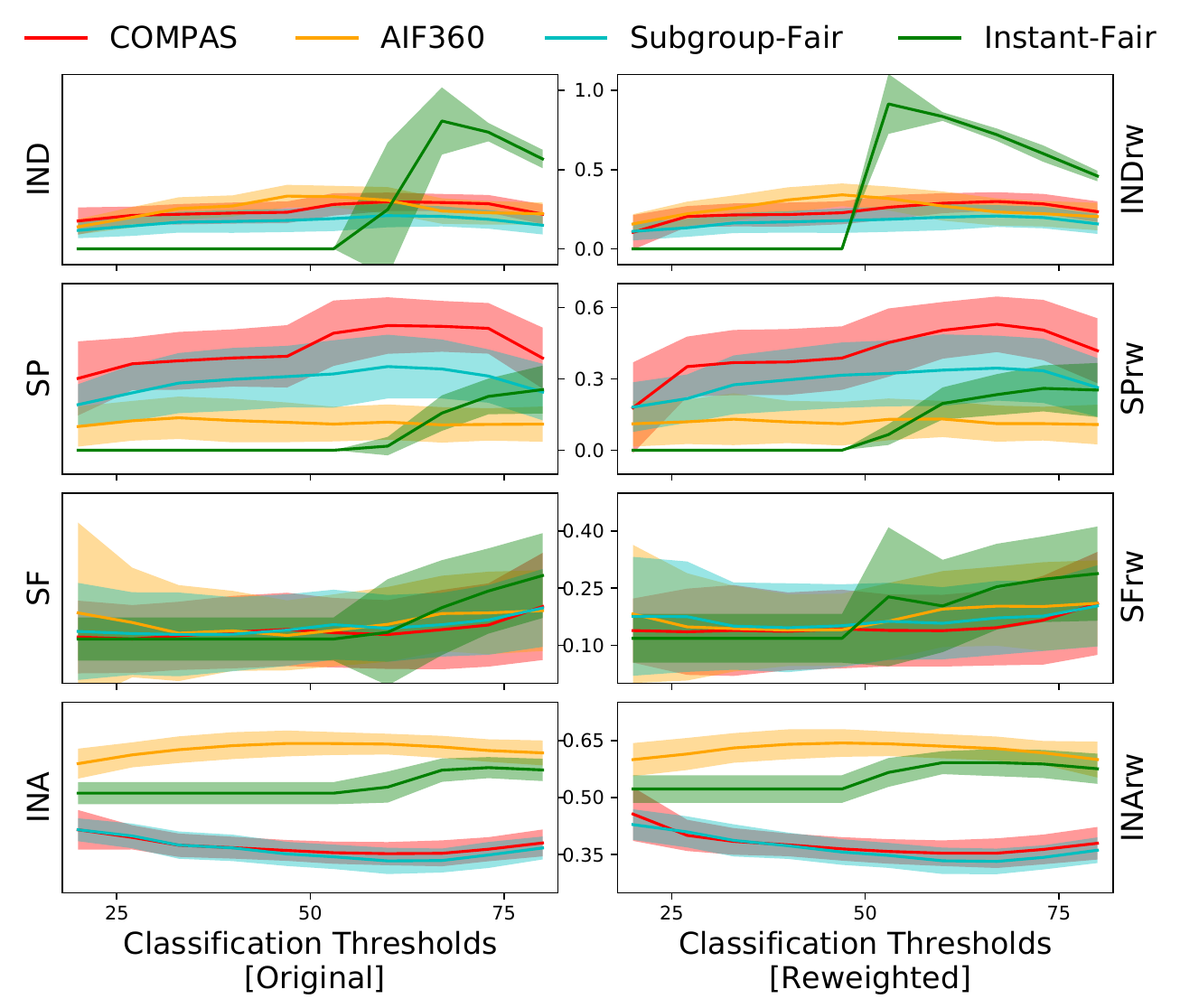}
\caption[the COMPAS scores, and our fairness-aware post-processing methods compared with state of the art, at varying thresholds]{Three post-processing methods ``subgroup-fair'' (blue), ``instant-fair'' (green), ``AIF360'' (yellow) are run in $50$ trials, and their outputs are compared with the COMPAS scores (red), with 10 uni-race thresholds $p=[20,26.7,\dots,80]$.
Each subplot displays the mean (curves) and mean $\pm$ one standard deviations (shaded error bands) of the corresponding index across the $50$ trials, against $10$ uni-race thresholds.
}
\label{fig:thresholds}
\end{figure*}

In Figure~\ref{fig:thresholds}, 
COMPAS scores (``COMPAS'' red), the state of the art (``AIF360'', yellow), and the outputs of our methods ``Subgroup-Fair'' (blue) and  ``Instant-Fair'' (green) are evaluated across 50 trials, with 10 different uni-race thresholds $p=[20,27,\dots,80]$ ($50\times 10\times 4$ runs).
Each subplot represents the mean (curves) and mean $\pm$ one standard deviations (shaded error bands) of the corresponding index across 50 trials, against 10 different uni-race thresholds $p=[20,27,\dots,80]$ with four methods distinguished by the same palette as in Figure~\ref{fig:postprocess}.

Figure~\ref{fig:thresholds} depicts an investigation of the performance of ``Subgroup-Fair'', ``Instant-Fair'', ``AIF360'' and ``COMPAS'' for different uni-race thresholds. 
Notice that the error bands generally overlap each other in subplots of the first and third rows (which represent the indices IND(rw) and SF(rw)). 
One potential implication of this work is that there might not be significant differences amongst the four methods, in terms of IND(rw) and SF(rw), when using uni-race thresholds. 
If we look at the remaining indices, ``Subgroup-Fair'' (blue) surpasses ``COMPAS'' (red) in terms of the performance indexes SP(rw), and both achieve the best in INA(rw).
Furthermore, ``AIF360'' has relatively low values of SP(rw), but at a large expense of INA(rw). 

\section{Conclusion}
We have introduced the two natural notions of fairness in forecasting. 
When the corresponding optimisation problems are solved to global optimality, the solutions outperform the COMPAS system in terms of independence and separation indices IND(rw) and SP(rw), cf. Equation~\eqref{equ:indices}.

As a further technical contribution, we have presented globally convergent methods for solving the optimisation problems arising from the two notions of fairness using hierarchies of convexifications of non-commutative polynomial optimisation problems. Also, we have shown that the runtime of standard solvers for the convexifications is independent of the dimension of the hidden state. This provides a technical tool in machine learning and statistics that is of independent interest, and that can also be applied in other settings.

\cleardoublepage
\chapter{Optimal Transport to Fairness without Demographic Membership}\label{cha:ot}

\begin{quote}
\textit{When it comes to groups categorised by sensitive attributes, prevalent fair learning algorithms primarily rely on accessibility, estimations of these attributes.
We propose a group-blind projection map aligning feature distributions of two groups in the source data, achieving group parity without needing individuals' sensitive attribute values for computation or usage.
Instead, our method utilises feature distributions of privileged and unprivileged groups in a broader population, the assumption that source data are unbiased representation of the population.
We present numerical results on the Adult Census Income dataset \citep{misc_adult_2}. 
This chapter is a collaboration with Prof. Robert Shorten and  Dr. Jakub Mare\v{c}ek, and has been submitted for publication.}
\end{quote}

To mitigate the disparate impact in classification, two common strategies have been employed in transfer learning and beyond.
The first strategy directly adjusts the values of features, labels, model outputs (estimated labels) or any combination thereof.
Initially, 
\cite{zemel2013learning} proposed a mapping to some fair representations of features.
\cite{feldman2015certifying} suggested modifying features so that the distributions for privileged and unprivileged groups become similar to a ``median'' distribution, making it harder for the algorithm to differentiate between the two groups.
The principle has been expanded to adjusting both features and the label \citep{calmon2017optimized} and further to projecting distributions of features to a barycenter \citep{gordaliza2019obtaining,yang2022obtaining}, which introduces the least data distortion.
\cite{oneto2020fairness,chzhen2020fair,gouic2020projection} performed post-processing of the model outputs by transporting the distributions of the outputs of each group to a barycenter. 
\cite{jiang2020identifying} corrected for a range of biases by re-weighting the training data.

The second strategy incorporates a regularisation term to penalise discriminatory effects whilst building some classification or prediction models.
\cite{quadrianto2017recycling} designed regularisation components using the maximum mean discrepancy criterion of \cite{gretton2012kernel} to encourage the similarity of the distributions of the prediction outputs across the privileged and unprivileged groups, with the assumption of sensitive attributes only available at training time.
\cite{jiang2020wasserstein} used Wasserstein distance of the distributions of outputs between privileged and unprivileged groups as the regularisation in logistic regression, again without any requirements on the availability of the sensitive attribute at test time. 
The Wassersterin regularisation is also used in neural-network classifiers \citep{risser2022tackling} and various applications \citep{jourdan2023optimal} thereof.
\cite{buyl2022optimal} design a regularisation term by the minimum cost to transport a classifier's score function to a set of fair score functions, and generalise the fairness measures allowed.


A key challenge, which has not been explored in depth, is the availability of sensitive attributes (e.g., gender, race). In practise, 
sensitive attributes are generally unavailable or inaccessible. In many jurisdictions, businesses cannot collect data on the race of their customers, for example, for any purpose whatsoever. 
In the EU, the AI act will make it possible to collect sensitive attributes in so-called sandboxes to audit the fairness of algorithms, but not to train the algorithms.
In contrast, most fairness-aware algorithms and techniques that incorporate group fairness are based on the assumption that sensitive attributes are accessible.
All the methods mentioned above use the sensitive attribute of each sample (data point) as a prerequisite for modifying the values differently for different groups (in the first strategy), or
measuring the loss of group-fairness regularisation term in iterations of training the classification or prediction models (in the second strategy).

The challenge of the availability of sensitive attributes has been partially addressed by \cite{oneto2020fairness,jiang2020wasserstein,quadrianto2017recycling,zafar2017fairnessBeyong}, who considered sensitive attributes at training time, and by \cite{elzayn2023estimating,chai2022self}, who leverages a small validation set with sensitive attributes to help guide training.
\cite{zhao2022towards,wang2020robust,amini2019uncovering,oneto2019taking} relaxed the strict requirement of the sensitive attribute being available, and use some noisy estimate instead.
\cite{sohoni2020no,yan2020fair} considered clustering to a similar effect. 
There is also methods on max-min fairness \citep{lahoti2020fairness,hashimoto2018fairness} that focus on improving the worst utility amongst all possible distributions. 
\cite{chai2022fairness} adjust the labels in the training data without knowledge of sensitive attributes, which may seriously dampen the utility of models.
Finally, \cite{liu2023group} aimed to mitigate bias and ensure fairness without relying on explicit group definitions.
Further discussion on the topic of fairness without demographics can be found in Section~\ref{sec:unvailable sa}.


\paragraph{Contributions}

We introduce novel methods for removing disparate impact in transfer learning (also known as domain adaptation and fairness repair), without requiring the sensitive attribute to be revealed.
Our methods are related to the ``total repair'' and ``partial repair'' schemes of \cite{feldman2015certifying} and  \cite{gordaliza2019obtaining}, which project the distributions of features for privileged and unprivileged groups to a target distribution interpolating between the groups.
However, the previous schemes \citep{gordaliza2019obtaining,feldman2015certifying}
require the sensitive attribute to define groups, while the sensitive attribute is either simply unavailable, or even illegal to collect
in many settings. 

Our work follows the idea \citep[Slide 23]{Quinn2023} to use marginals from another dataset in transfer learning towards a dataset whose bias is, at least in part, determined by an unavailable sensitive attribute. 
In the setting envisioned by the AI Act in the European Union, this additional dataset could be collected within so-called regulatory sandbox \cite{allen2019regulatory,morgan2023anticipatory}, wherein a regulator and the developer of an AI system agree to override data privacy protection
in a well-defined fashion.
In many other settings, the data could be collected in the census.
When no such data exists, such as when France bars the collection of data on the race and ethnicity of its citizens even in the census \citep{Ndiaye2020},
or India bars the collection of data on caste status in its census \citep{Bose2023}, 
it would seem difficult to address the bias, indeed.

We develop algorithms to realise this objective.  
The membership of each sample is used to calculate distinct projections for various groups and to modify the value of features via group-wise projections.
We extend these schemes to modifying the values of features via one group-blind projection map, and achieve equalised distributions of modified features between privileged and unprivileged groups.
The sample membership is not required for computing such group-blind projection map, or for modifying the values of features via this group-blind projection map.
We require only the population-level information regarding feature distributions for both privileged and unprivileged groups, as well as the assumptions that source data are collected via unbiased sampling from the population.
The target distribution in our framework is not necessarily the barycentre distribution.
Instead, since the modified or projected data will be used on a pre-trained classification or prediction model, the distribution of training data used to learn such a pre-trained model can be our target distribution, to preserve the classification or prediction performance.

This chapter is organised as follows: 
Section~\ref{sec:ot_movitation} gives the motivating example of school admission. Section~\ref{sec:ot_relatedwork} introduces related work and the state-of-the-art in entropic regularised \acrfull{OT}. 
Section~\ref{sec:ot_main} introduces our ``total repair'' and ``partial repair'' schemes with our algorithm. Sections~\ref{sec:ot_main}.1--\ref{sec:ot_main}.4 focus on a one-dimensional case (that is, only one unprotected feature is considered) and Section~\ref{sec:ot_main}.5 explains how to implement our schemes in higher-dimensional cases. 
Section~\ref{sec:ot_main}.6 explains the choices of target distribution.
Section~\ref{sec:ot_experiments} gives the numerical results on synthetic data and real-world data, as well as a comparison with two baselines.
The proofs of all lemmas and theorems can be found in Appendix~\ref{app:ot}.


\section{A Motivating Example}
\label{sec:ot_movitation}
\begin{figure}[htp]
\centering
\includegraphics[width=0.7\textwidth]{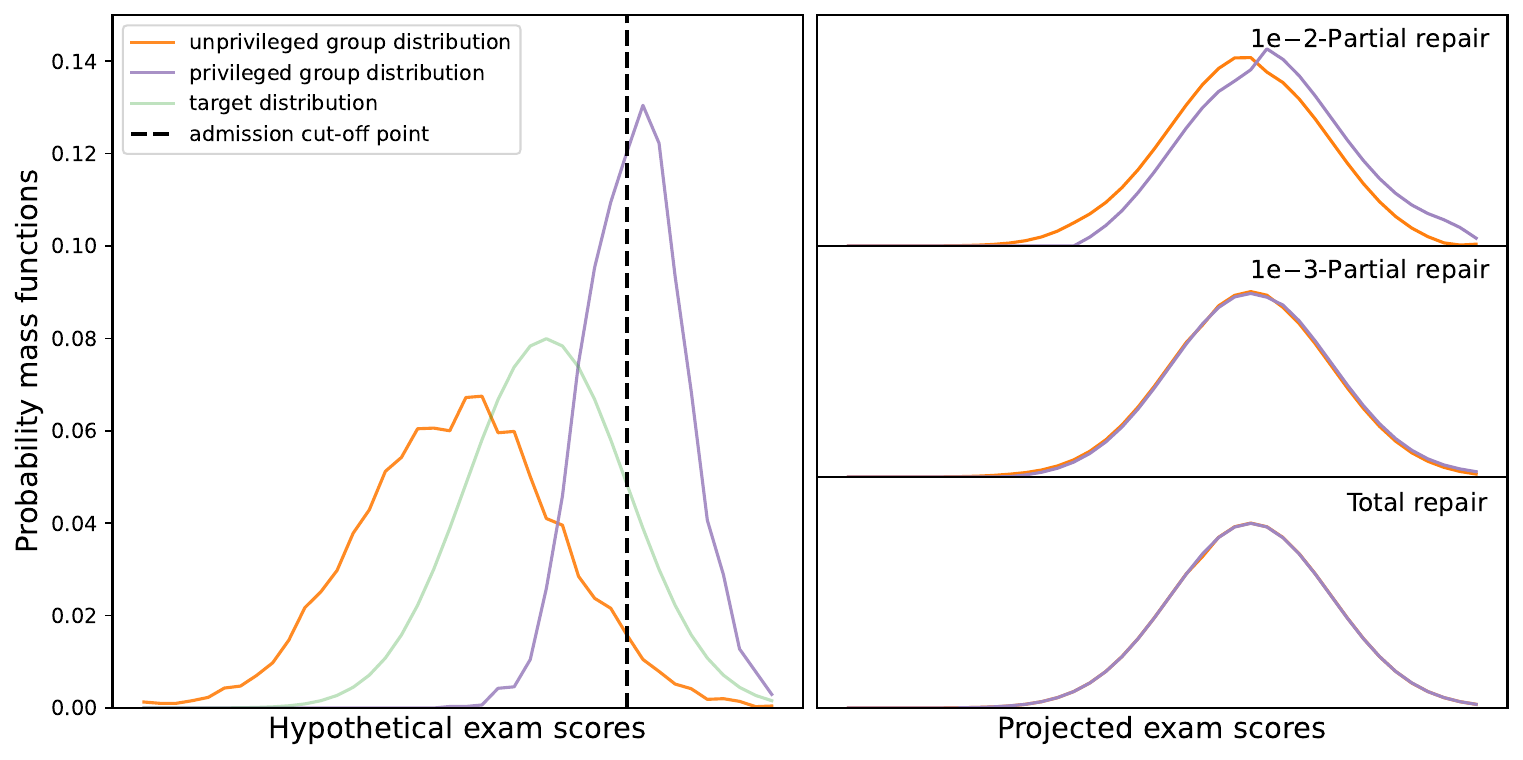}
\caption[A motivating example of school admission to illustrate the bias-repair framework]{\textbf{Left:} A cut-off point (vertical dark line) for an exam score is used to make school-admission decisions. With the hypothetical exam-score distributions of the unprivileged group (orange) and the privileged group (purple), this group-blind cut-off point would result in outcomes biased against the unprivileged group. 
If the score distributions of unprivileged group and the privileged group are projected closer to an arbitrary target distribution (green), the same cut-off point can achieve equalised admission rates.
\textbf{Right:} The partial repair (top two) and total repair (bottom) schemes proposed in this paper, move the score distributions of unprivileged group (orange) and the privileged group (purple) closer to the pre-defined target distribution, without access to the sensitive attribute of each sample.}
\label{fig:motivation}
\end{figure}

Let us consider an example of school admission to illustrate the bias repair schemes in our framework as well as in \cite{gordaliza2019obtaining,feldman2015certifying}.
Schools commonly use exam scores to make admission decisions with a cutoff point: students with scores higher than the cutoff point will be admitted.
In Figure~\ref{fig:motivation}, this is shown with the vertical dashed line. 
The applicants can be divided into two groups by a sensitive attribute, e.g., gender, race. 
The left subplot of Figure~\ref{fig:motivation} shows that the score distribution of the privileged group (denoted by a purple curve) concentrates somewhere above the cut-off point.
On the contrary, only a small portion of the unprivileged group (denoted by an orange curve) passes the cut-off point.
If the cutoff point is used in a straightforward manner, the admission results are strongly biased against the unprivileged group, although the admission-decision algorithm does not take into account the sensitive attribute.
Then, instead of training another fairness-aware admission-decision model, the bias repair schemes will project the score distributions of both groups into an arbitrary target distribution, for example, the green curve, such that the scores of both groups follow the same target distribution. The target distribution is a ``median'' distribution in \cite{feldman2015certifying}, and a barycentre distribution in \cite{gordaliza2019obtaining}, and are not necessarily the same as the green curve.

The right subplots of Figure~\ref{fig:motivation} illustrate the effects of bias repair schemes proposed in this paper, with the top two for ``partial repair'', and the bottom one for ``total repair''.
Now, the same admission cut-off point used on the projected scores will achieve equalised admission rates between the privileged and the unprivileged group.
Our extension eliminates the need for the sensitive attribute of each sample when computing the group-blind projection map and projecting both groups via this map.

\section{Related Work in Optimal Transport}\label{sec:ot_relatedwork}

We refer to \cite{peyre2019computational} for a comprehensive overview of the history of \acrfull{OT}.
It is a method to find a coupling that moves mass from one distribution to another at the smallest possible cost, known as the Monge problem, relaxed into so-called Kantorovich relaxation. 
The relaxation, with an entropic regularisation to promote smoothness of the map, leads to iterations of simple matrix-vector products of the Sinkhorn--Knopp algorithm \citep{sinkhorn1967concerning}. 
Often, the equality constraints in entropy-regularised \acrshort{OT} are softened by minimising the dual norm of the differences between both sides of equality constraints; see  \cite{chapel2021unbalanced,pham2020unbalanced,chizat2018scaling} for unbalanced OT and \cite{chizat2018unbalanced} for dynamic unbalanced \acrshort{OT}. 
Pioneered by \cite{gangbo1998optimal,agueh2011barycenters},
another direction is to explore barycentres of Wasserstein distance of multiple distributions with respect to some weights, which is a representation of the mean of the set of distributions, with several follow-up works in \cite{nath2020statistical,dard2016relevance,seguy2015principal}.
On the computational side, the simplest algorithm that exploits \acrfull{KL} divergence is the iterative Bregman projections \citep{bregman1967relaxation}, which is related to the Sinkhorn algorithm. 
The generic OT considering finitely many arbitrary convex constraints can be solved by Dykstra's algorithm with Bregman projections \citep{bauschke2000dykstras,benamou2015iterative}, or generalised scaling algorithm \citep{chizat2018scaling}. The mini-batch OT can be used to ease the computational burden  \citep{sommerfeld2019optimal}.

Next, we outline classic \acrshort{OT} formulation with entropic regularisation penalty, together with its generalisations. We start from introducing the common notation.
Let $N$ be the number of discretisation points, and $P\in\mathbb{R}^N$ be vectors in the probability simplex:
\begin{equation}
\Delta_N:=\left\{P\in\mathbb{R}^N_+\Bigg|\sum^N_{i=1} P_i=1\right\}.
\label{equ:simplex}
\end{equation}
The set of all admissible or feasible couplings between two probability distributions $P,Q\in\Delta_N$ is defined as 
\begin{equation}
\Pi(P,Q):=\{\gamma\in\mathbb{R}^{N\times N}_+\mid \gamma\mathbb{1}=P,\gamma'\mathbb{1}=Q\},
\label{equ:Pi-define}
\end{equation}
where $\mathbb{1}$ is the $N$-dimensional column vector of ones, $\gamma'$ is the transpose of $\gamma$.
\textcolor{black}{
Based on the definition, an arbitrary coupling $\gamma\in\Pi(P,Q)$ is a transport plan that moves the mass in a distribution $P$ to another distribution $Q$. The $(i,j)$ entry of such a coupling $\gamma_{i,j}$, represents the amount of mass transported from the entry $P_i$ to $Q_j$, with the total mass being $\sum_{i,j}\gamma_{i,j}=1$. As will be discussed later, a projection map can be derived from the transport plan, which maps the source data with distribution $P$ onto the target data with distribution $Q$.
Note that $\Pi(P,Q)$ is non-empty, as the outer product $P\otimes Q$ is always feasible (see Remark 2.13 in \cite{peyre2019computational}). This ensures that there is at least one projection map available.}

Further, we define the entropy of a coupling $\gamma$ as:
\begin{equation}
E(\gamma):=-\sum_{i,j=1}^N \gamma_{i,j}(\log(\gamma_{i,j})-1), \label{equ:entropy-define}
\end{equation}
which is a concave function and we use the convention $0\log 0=0$. The \acrshort{KL} divergence between the coupling $\gamma\in\mathbb{R}^{N\times N}_+$ and a positive reference matrix $\xi\in\mathbb{R}^{N\times N}_{++}$ (i.e., $\xi_{i,j}>0$) is defined as:
\begin{equation}
\kl{\gamma}{\xi}:=\sum_{i,j=1}^N \gamma_{i,j}\left(\log\left(\frac{\gamma_{i,j}}{\xi_{i,j}}\right) -1 \right).\label{equ:kl-define}
\end{equation}
Given a convex set $\mathcal{C}\in\mathbb{R}^{N\times N}$, the \acrshort{KL} projection is 
\begin{equation}
\prox^{KL}_{\mathcal{C}}(\xi):=\arg\min_{\gamma\in\mathcal{C}} \kl{\gamma}{\xi},\label{equ:kl-projec-define}
\end{equation}
which is uniquely defined since \acrshort{KL} divergence is a strictly convex and coercive\footnote{An (extended-valued) continuous function $f:\mathbb{R}^n\to\mathbb{R}\cup\{\pm\infty\}$ is coercive if $\lim_{\|\mathbf{x}\|\to\infty} f(\mathbf{x})=\infty$.} function and, $\mathcal{C}$ is a convex set. See Lemma~\ref{lem:subgradient-optimality} for more details. 

Referring to \cite{benamou2015iterative}, we introduce the state of the art in OT.
The discrete entropic regularisation OT problem has the form:
\begin{equation}
W_{\epsilon}(P,Q):=\min_{\gamma\in\Pi(P,Q)} \langle C,\gamma\rangle -\epsilon E(\gamma),
\label{equ:entropic-reg-OT-define}
\end{equation}
where $C\in\mathbb{R}^{N\times N}$ is the cost matrix, $\epsilon>0$ is the entropic regularisation parameter and $\langle\cdot,\cdot\rangle$ is the inner product. This formulation is equivalent to
\begin{equation}
\begin{split}
W_{\epsilon}(P,Q)&=\min_{\gamma\in\Pi(P,Q)} \sum^N_{i,j=1} C_{i,j}\gamma_{i,j}+\epsilon\gamma_{i,j}\left(\log (\gamma_{i,j}) -1 \right)\\
&=\min_{\gamma\in\Pi(P,Q)}\epsilon\sum^N_{i,j=1} \gamma_{i,j}\left(\log(\gamma_{i,j})-\log(\exp(-C_{i,j}/\epsilon)) -1 \right)\\
&=\min_{\gamma\in\Pi(P,Q)} \epsilon\;\kl{\gamma}{\xi}, \quad\textrm{where}\quad\xi=\exp{(-C/\epsilon)}.
\end{split} 
\label{equ:KL-equivalent}
\end{equation}

Further, the set of admissible (feasible) couplings in Equation~\eqref{equ:Pi-define} could be formulated as the intersection of two affine subspaces: $\Pi(P,Q)=\mathcal{C}_1\cap\mathcal{C}_2$, where
\begin{equation}
\mathcal{C}_1:=\{\gamma\in\mathbb{R}^{N\times N}_+\mid\gamma\mathbb{1}=P\},\quad\mathcal{C}_2:=\{\gamma\in\mathbb{R}^{N\times N}_+\mid\gamma'\mathbb{1}=Q\}.\label{equ:convex-set-1}
\end{equation}

Generally, when the convex sets $\mathcal{C}_{\ell},\ell\geq 1$ are affine subspaces,
initialise $\gamma^{(0)}:=\xi=\exp{(-C/\epsilon)}$, then conduct iterative \acrshort{KL} projections: for $n>0$
\begin{equation}
\gamma^{(n+1)}=\prox^{KL}_{\mathcal{C}_n} (\gamma^{(n)}),\label{equ:iterative-Bregman}
\end{equation} 
where $\mathcal{C}_n\equiv\mathcal{C}_{1+(n\mod 2)}$. 
The iterations in Equation~\eqref{equ:iterative-Bregman} can converge to the unique solution of $W_{\epsilon}(P,Q)$, as $n\to+\infty$ (Fact~1.4 in \cite{bauschke2020dykstra}, \cite{benamou2015iterative,bregman1967relaxation}). Note that the \acrshort{KL} projections used in this chapter, is a special case of Bregman projections.

\paragraph{Inequality constraints}
When the convex sets are not affine subspaces, iterative Bregman projections do not converge to $W_{\epsilon}(P,Q)$ (Fact~1.2 in \cite{bauschke2020dykstra}). Instead, Dykstra's algorithm\footnote{\textcolor{black}{Dykstra's algorithm calculates a point within the intersection of convex sets, whereas Dijkstra's algorithm, despite its similar name, is unrelated to the former. Instead, it determines the shortest path from one vertex to all other vertices.}} is known to converge when used in conjunction with Bregman divergences, for arbitrary convex sets (cf. \cite{bauschke2020dykstra,bauschke2000dykstras}).
Random Dykstra algorithm converges linearly in expectation for the general feasible sets satisfying Slater's condition \citep{necoara2022linear}.

It could be used to solve \acrshort{OT} problem with inequality constraints:
\begin{example}\label{exp:partial-OT}
The partial OT \citep{le2022multimarginal,caffarelli2010free,figalli2010optimal} only needs to transport a given fraction of mass and the two marginals $P,Q$ do not need to have the same total mass:
\begin{equation*}
\min_{\gamma\in\mathbb{R}^{N\times N}_+}\{\langle C,\gamma\rangle -\epsilon E(\gamma)\mid \gamma\mathbb{1}\leq P,\gamma'\mathbb{1}\leq Q, \mathbb{1}'\gamma\mathbb{1}=\eta\},
\end{equation*}
where the constant $\eta\in[0,\min\{P'\mathbb{1},Q'\mathbb{1}\}]$ denotes the fraction of mass needed to transport. Similarly, this problem can be reformulated as Equation~\eqref{equ:KL-equivalent} but with $\gamma$ falling into the intersection of three convex sets:
\begin{equation*}
\mathcal{C}_1:=\{\gamma\in\mathbb{R}^{N\times N}_+\mid\gamma\mathbb{1}\leq P\},\;\mathcal{C}_2:=\{\gamma\in\mathbb{R}^{N\times N}_+\mid\gamma'\mathbb{1}\leq Q\},\;\mathcal{C}_3:=\{\gamma\in\mathbb{R}^{N\times N}_+\mid\mathbb{1}'\gamma\mathbb{1}=\eta\}.
\end{equation*}
\end{example}

\begin{example}\label{exp:capacity-OT}
The capacity constrained OT, pioneered by \cite{korman2015optimal,korman2013insights}, is imposed by an extra upper bound on the amount of mass transported from $i$ to $j$. Its formulation reads:
\begin{equation*}
\min_{\gamma\in\mathbb{R}^{N\times N}_+}\{\langle C,\gamma\rangle -\epsilon E(\gamma)\mid \gamma\mathbb{1}= P,\gamma'\mathbb{1}=Q, \gamma\leq\Lambda \},
\end{equation*}
where $\Lambda \in\mathbb{R}^{N\times N}_+$ denotes the upper bound. We 
equivalently write the feasible set into the intersection of the following three convex sets:
\begin{equation*}
\mathcal{C}_1:=\{\gamma\in\mathbb{R}^{N\times N}_+\mid\gamma\mathbb{1}=P\},\;\mathcal{C}_2:=\{\gamma\in\mathbb{R}^{N\times N}_+\mid\gamma'\mathbb{1}=Q\},\;\mathcal{C}_3:=\{\gamma\in\mathbb{R}^{N\times N}_+\mid\gamma\leq\Lambda \}.
\end{equation*}
\end{example}

\paragraph{Computation of \acrshort{KL} projections}
In general, it is impossible to directly compute the \acrshort{KL} projections in closed form, so some form of subiterations are required. Notably, the convex set with the form in Equation~\eqref{equ:convex-set-1}, Example~\ref{exp:partial-OT}--\ref{exp:capacity-OT} can be computed by some matrix-vector multiplications \citep{benamou2015iterative}, with the $\min$, division operators being element-wise:
\begin{equation*}
\prox^{KL}_{\mathcal{C}}(\bar{\gamma})=
\begin{cases}
\diag{\frac{P}{\gamma\mathbb{1}}}\bar{\gamma} & \textrm{if } \mathcal{C}=\{\gamma\in\mathbb{R}^{N\times N}_+\mid\gamma\mathbb{1}=P\},\\
\bar{\gamma}\diag{\frac{Q}{\bar{\gamma}'\mathbb{1}}} & \textrm{if } \mathcal{C}=\{\gamma\in\mathbb{R}^{N\times N}_+\mid\gamma'\mathbb{1}=Q\},\\
\diag{\min\left\{\mathbb{1},\frac{P}{\bar{\gamma}\mathbb{1}}\right\}}\bar{\gamma} & \textrm{if } \mathcal{C}=\{\gamma\in\mathbb{R}^{N\times N}_+\mid\gamma\mathbb{1}\leq P\},\\
\bar{\gamma}\diag{\min\left\{\mathbb{1},\frac{Q}{\bar{\gamma}'\mathbb{1}}\right\}} & \textrm{if } \mathcal{C}=\{\gamma\in\mathbb{R}^{N\times N}_+\mid\gamma'\mathbb{1}\leq Q\},\\
\bar{\gamma}\frac{\eta}{\mathbb{1}'\bar{\gamma}\mathbb{1}} & \textrm{if } \mathcal{C}=\{\gamma\in\mathbb{R}^{N\times N}_+\mid\mathbb{1}'\gamma\mathbb{1}=\eta\},\\
\min\{\bar{\gamma},\Lambda \} & \textrm{if } \mathcal{C}=\{\gamma\in\mathbb{R}^{N\times N}_+\mid\gamma\leq\Lambda \},
\end{cases}
\end{equation*}
\label{page:define-diag}
where the symbol diag$(P)$ refers to an $N\times N$ diagonal matrix with its diagonal elements being the elements of $P$. 
The proof for the first two cases, and thus for Lemma~\ref{lem:c_1&c_2}, can be found in Appendix~\ref{app:lem:c_1&c_2}.

\section{The Bias-Repair Framework}\label{sec:ot_main}

Recall the school admission example introduced in Section~1.1. 
The school receive the exam score ($X$), but not the sensitive attribute ($S$) of each applicant. So, the source data would be all applicants' exam scores. Due to $S$ being not observed, the school wants to find a $S$-blind projection map ($\mathscr{T}$), to project these scores to the target data, where the two groups of applicants have the same (``total repair'') or equalised (``partial repair'') distributions of scores. 
In other words, the school wants to achieve equalised odds of admission between the privileged and unprivileged groups in the target data, using the pre-trained admission-decision policy.

\subsection{Definitions}

Let us introduce our ``total repair'' and ``partial repair'' schemes, starting with defining the sensitive attribute ($S$), distributions of the exam scores ($X$) and $S$-blind projection map ($\mathscr{T}$):

\begin{definition}[a sensitive attribute]\label{def:sensitive_attribute}
Let $S\in\mathbb{N}$ be an integer random variables of a  sensitive attribute (e.g., race, gender). Let $P^{S}_s:=\pp{S=s}$ be the probability of $S=s$. Its support is defined as 
\begin{equation}
\supp{S}:=\{s\in\mathbb{N}\mid \pp{S=s}>0\}.
\end{equation}
\end{definition}

\begin{definition}[Source variables of a neutral  attribute]\label{def:source_var}
Let $X,X_{s},s\in\supp{S}$ be scalar discrete random variables of a neutral attribute (e.g., income, credit scores), with their supports $\supp{X_s}\subseteq\supp{X}\subset\mathbb{R}$, for $s\in\supp{S}$.
Let $\supp{X}$ be $N$ discretisation points.
Their probability distributions are taken from the probability simplex $P^{X},P^{X_s}\in\Delta_N$:
\begin{equation}
P^{X}_i:=\pp{X=i},\quad P^{X_s}_i:=\pover{X=i}{S=s},
\end{equation}
where $\pp{X=i}$ is the probability of $X=i$ and $\pover{X=i}{S=s}$ is the conditional probability of $X=i$ when $S=s$.
\end{definition}

To an exploration, throughout the paper we only consider the case of two groups, i.e., $\supp{S}=\{s_0,s_1\}$.
we assume $X$ includes one neutral attribute till Section~\ref{sec:ot_main}.4 and the extension to multiple neutral attributes will be discussed in Section~\ref{sec:ot_main}.5.

Next, we define the target distribution, for instance, the ideal score distribution that we want to achieve in the target data. 
The choice of target distribution will be mentioned in Section~3.6.

\begin{definition}[Target variables of the neutral attribute]\label{def:target_var}
Let $\td{X},\td{X}_{s},s\in\supp{S}$ be scalar discrete random variables, with their supports $\supp{\td{X}_s}\subseteq\supp{\td{X}}\subset\mathbb{R}$, for $s\in\supp{S}$.
Let $\supp{\td{X}}$ be $N$ discretisation points.
Their probability distributions are taken from the probability simplex $P^{\td{X}},P^{\td{X}_s}\in\Delta_N$:
\begin{equation}
P^{\td{X}}_j:=\pp{\td{X}=j},\quad P^{\td{X}_s}_j:=\pover{\td{X}=j}{S=s},
\end{equation}
\end{definition}


In our setting, we would like to map the source data, i.e., samples from the tuple $(X,S)$ to the target data, i.e., samples from the tuple $(\td{X},S)$, via a projection map $\mathscr{T}$, that is defined to be $S$-blind, and induced by a coupling $\gamma\in\Pi(P^{X},P^{\td{X}})$:

\begin{definition}[Projection]\label{def:projection}

Once the coupling $\gamma\in\Pi(P^{X},P^{\td{X}})$ has been computed, to find the map that transports source data to target data, we define the (group-blind) projection $\mathscr{T}$:
\begin{equation*}
\begin{split}
\mathscr{T}:\supp{X}&\to\supp{\td{X}}\times\mathbb{R}_{+}\\
i&\mapsto (j,w_{i,j}),\forall j\in\supp{\td{X}}
\end{split}
,\quad\textrm{ where } w_{i,j}=\frac{\pp{X=i,\td{X}=j}}{\pp{X=i}}=\frac{\gamma_{i,j}}{P^X_{i}}.
\end{equation*}
A sample $(i,s)$ is hence split into a sequence of weighted samples $\{(j,w_{i,j},s)\}_{j\in\supp{\td{X}}}$.
Very importantly, we stress that the projection do not change $s$ and any other features even though $s$ is not observed. 
\end{definition}

\begin{example}\label{exp:projection-1}
Given a sample $(i,s)$ and
$\gamma_{i,j}=P^{X}_i$. Thus $\gamma_{i,j'}=0$ for $j'\in\supp{\td{X}}\setminus\{j\}$ and 
\begin{equation*}
\mathscr{T}(i)=(j,1).
\end{equation*}
Hence, this sample is transported to $(j,1,s)$. 
\end{example}
\begin{example}\label{exp:projection-2}
Given a sample $(i,s)$ and
$\gamma_{i,j}=\gamma_{i,j'}=P^{X}_i/2$.
Thus $\gamma_{i,j^{\dag}}=0$ for $j^{\dag}\in\supp{\td{X}}\setminus\{j,j'\}$ and
\begin{equation*}
\mathscr{T}(i)=
\begin{array}{c}
(j,1/2)\\
(j',1/2)
\end{array}.
\end{equation*}
Hence, this sample is split into $(j,1/2,s)$ and $(j',1/2,s)$.
\end{example}

\paragraph{Observations} 
So far, we have defined the projection map $\mathscr{T}$ in the formal way. We first observe something interesting:

Let $\supp{S}=\{s_0,s_1\},\supp{X}=\{i,i'\}$, $\supp{\td{X}}=\{j,j'\}$. 
\textcolor{black}{Then, the source data only include six samples, with orange (resp. purple) denoting samples from the group $s_0$ (resp. $s_1$): }
\begin{align*}
\textcolor{orange}{(i,s_0)},\textcolor{violet}{(i,s_1),(i,s_1)},\\
\textcolor{orange}{(i',s_0),(i',s_0)},\textcolor{violet}{(i',s_1)}.
\end{align*}
From the source data, we can directly compute
$P^{X}=[1/2,1/2]^{tr}$, $P^{X_{s_0}}=[1/3,2/3]^{tr}$ and $P^{X_{s_1}}=[2/3,1/3]^{tr}$.

If we set entries $\gamma_{i,j}=\gamma_{i',j'}=1/2$, and $\gamma_{i,j'}=\gamma_{i',j}=0$, following Example~\ref{exp:projection-1}, the projected data become:
\begin{align*}
\textcolor{orange}{(j,1,s_0)},\textcolor{violet}{(j,1,s_1),(j,1,s_1)},\\
\textcolor{orange}{(j',1,s_0),(j',1,s_0)},\textcolor{violet}{(j',1,s_1)},
\end{align*}
such that $P^{\td{X}}=P^{X}$, $P^{\td{X}_{s_0}}=P^{X_{s_0}}$, $P^{\td{X}_{s_1}}=P^{X_{s_1}}$. Nothing has changed.
However, if we set entries $\gamma_{i,j}=\gamma_{i',j'}=\gamma_{i,j'}=\gamma_{i',j}=1/4$, following Example~\ref{exp:projection-2}, the projected data become:
\begin{align*}
\textcolor{orange}{(j,1/2,s_0),(j,1/2,s_0),(j,1/2,s_0)},\textcolor{violet}{(j,1/2,s_1),(j,1/2,s_1),(j,1/2,s_1)},\\
\textcolor{orange}{(j',1/2,s_0),(j',1/2,s_0),(j',1/2,s_0)},\textcolor{violet}{(j',1/2,s_1),(j',1/2,s_1),(j',1/2,s_1)},
\end{align*}
such that $P^{\td{X}}=P^{X}=P^{\td{X}_{s_0}}=P^{\td{X}_{s_1}}$.
This is a trivial example indeed, but it hints at the possibility of manipulating $P^{\td{X}_{s_0}}$ and $P^{\td{X}_{s_1}}$ by a well-designed $S$-blind map.

Next, we explain how to extend the observation to nontrivial cases.

\subsection{Total Repair}

In the school admission example, even when the sensitive attribute $S$ is not observed, and the map is $S$-blind, we would like to achieve some parity between groups partitioned by $S$, in the exam score distributions.
Formally,


\begin{definition}[Total repair]\label{def:total-repair}
Following the definition of total repair in \cite{gordaliza2019obtaining}, 
we say that total repair is satisfied if 
\begin{equation}
P^{\td{X}_{s}}=P^{\td{X}_{s'}},\forall s,s'\in\supp{S}.
\end{equation}
Note that when total repair is satisfied, it holds $P^{\td{X}_{s}}=P^{\td{X}_{s'}}=P^{\td{X}}$.
\end{definition}
\begin{lemma}\label{lem:tx|us}
If the coupling $\gamma\in\Pi(P^{X},P^{\td{X}})$ is given, 
\begin{equation}
P^{\td{X_s}}=\gamma'\frac{P^{X_s}}{P^{X}}, \quad P^{\td{X}}=\gamma'\mathbb{1},
\end{equation}
where the division operator is element-wise.
\end{lemma}

\paragraph{Intuitive verification}
\label{page:define-M}
Alternatively, we can think about the source data where the probability equals to the proportion.
We select all samples that belong to group $s$: $\{(i_m,s)\}_{m=1,\dots,M}$. Since the projection does not change $s$, a sample $(j,s)$ in the target data only originates from sample $(i,s),i\in\supp{X}$ in the source data.
Now, we equivalently group the same samples and use the number of the same samples as the weight. For instance, if we have five samples of $(i,s)$, we only use one weighted sample $(i,w_i=5,s)$ to represent all of them. Since we assume that probability equals portion, the source data can be rewritten as
\begin{equation}
\{(i,w_i,s)\}_{i\in\supp{X_s}},\label{equ:weighted-source-data}
\end{equation}
where $w_i=M\times\pover{X=i}{S=s}$ for $i\in\supp{X_s}$. From the weighted source data, we can compute $\pover{X=i}{S=s}=w_i/M$. 
Recall Definition~\ref{def:projection}: a sample $(i,s)$, or equivalently a weighted sample $(i,w_i=1,s)$, is transported to a sequence of weighted samples $\{(j,1\times w_{i,j},s)\}_{j\in\supp{\td{X}}}$. 
Then the weighted sample $(i,w_i,s)$ is transported to $(j,w_i\times w_{i,j},s)$ for $j\in\supp{\td{X}}$. After projecting all samples in weighted source data (Equation~\eqref{equ:weighted-source-data}), the projected data become
\begin{equation*}
\{(j,w_i\times w_{i,j},s)\}_{i\in\supp{X_s}, j\in\supp{\td{X}}}.
\end{equation*} 
Hence, we can compute
\begin{equation*}
\pover{\td{X}=j}{S=s}=\sum_{i\in\supp{X_s}} w_{i,j}\times w_{i}/M=\sum_{i\in\supp{X_s}}\pover{X=i}{S=s}\frac{\pp{X=i,\td{X}=j}}{\pp{X=i}}.
\end{equation*}
Rewrite the above equation in matrix form, and setting the undefined conditional probability to $\pover{X=i}{S=s}=0$ for $i\in\supp{X}\setminus\supp{X_s}$, we get the same results as in Lemma~\ref{lem:tx|us}.

Next, we explore what kind of couplings can guarantee total repair in the target data, such that the total repair property is transformed into a constraint imposed on the coupling:
\begin{theorem}[Total repair to a constraint on the coupling $\gamma$]
\label{pro:binary_condition}
In the case of binary sensitive attributes, i.e., $\supp{S}=\{s_0,s_1\}$, if one wishes to achieve total repair, the coupling should satisfy
\begin{equation}
P^{\td{X_0}}-P^{\td{X_1}}=\gamma'\left(\frac{P^{X_{s_0}}-P^{X_{s_1}}}{P^{X}}\right)=\gamma' V=\mathbb{0},
\label{equ:difference2rv}
\end{equation}
where the division and subtraction operations are element-wise, and $\mathbb{0}\in\mathbb{R}^N$ is a zero vector.
In particular,
the vector $V\in\mathbb{R}^{N}$, is defined as
\begin{equation}
V:=\frac{P^{X_{s_0}}-P^{X_{s_1}}}{P^{X}},
\end{equation}
is the most important input in our framework, which only needs $P^{X_{s_0}},P^{X_{s_1}}$ and $P^{X}$ to be computed. Note that $P^{X}$ is already known due to observability of the feature $X$ in source data.
While $S$ is not observed, $P^{X_{s_0}},P^{X_{s_1}}$ can be obtained from the population-level information regarding the distributions of features for both groups ($s_0$ and $s_1$), given the crucial assumption that the source data represents unbiased samples from the broader population. This assumption of unbiased sampling ensures that statistical properties of the population carry over.
\begin{proof}
Total repair in Definition~\ref{def:total-repair} requires that for $ j\in\supp{\td{X}}$, the difference between $\pover{\td{X}=j}{s_0}$ and $\pover{\td{X}=j}{s_1}$ is zero. 
$\supp{X}$ could be divided into three disjoint subsets: $\supp{X_{s_0}}\cap\supp{X_{s_1}}$, $\supp{X_{s_0}}\setminus\supp{X_{s_1}}$ and $\supp{X_{s_1}}\setminus\supp{X_{s_0}}$. Let $\mathbb{X}^{0\wedge 1}$, $\mathbb{X}^{0-1}$ and $\mathbb{X}^{1-0}$ denote these three subsets.
Using Lemma~\ref{lem:tx|us}, we deduce that
\footnotesize
\begin{align*}
&\pover{\td{X}=j}{s_0}-\pover{\td{X}=j}{s_1}\\
&=\sum_{i\in\supp{X_{s_0}}}\pp{X=i,\td{X}=j}\frac{\pover{X=i}{S=s_0}}{\pp{X=i}}-\sum_{i\in\supp{X_{s_1}}}\pp{X=i,\td{X}=j}\frac{\pover{X=i}{S=s_1}}{\pp{X=i}}\\
&=\sum_{i\in\mathbb{X}^{0\wedge 1}}\pp{X=i,\td{X}=j}\frac{\pover{X=i}{S=s_0}}{\pp{X=i}}+\sum_{i\in\mathbb{X}^{0-1}}\pp{X=i,\td{X}=j}\frac{\pover{X=i}{S=s_0}}{\pp{X=i}}\\
&\quad -\sum_{i\in\mathbb{X}^{0\wedge 1}}\pp{X=i,\td{X}=j}\frac{\pover{X=i}{S=s_1}}{\pp{X=i}}-\sum_{i\in\mathbb{X}^{1-0}}\pp{X=i,\td{X}=j}\frac{\pover{X=i}{S=s_1}}{\pp{X=i}}\\
&=\sum_{i\in\mathbb{X}^{0\wedge 1}}\pp{X=i,\td{X}=j}\frac{\pover{X=i}{S=s_0}-\pover{X=i}{S=s_1}}{\pp{X=i}}\\
&\quad +\sum_{i\in\mathbb{X}^{0-1}}\pp{X=i,\td{X}=j}\frac{\pover{X=i}{S=s_0}}{\pp{X=i}}-\sum_{i\in\mathbb{X}^{1-0}}\pp{X=i,\td{X}=j}\frac{\pover{X=i}{S=s_1}}{\pp{X=i}}.
\end{align*}
\normalsize
Hence, if we want to achieve total repair, the coupling $\gamma$ must satisfy for $j\in\supp{\td{X}}$
\footnotesize
\begin{equation}
\begin{split}
&0=\sum_{i\in\mathbb{X}^{0\wedge 1}}\pp{X=i,\td{X}=j}\frac{\pover{X=i}{S=s_0}-\pover{X=i}{S=s_1}}{\pp{X=i}}\\
&+\sum_{i\in\mathbb{X}^{0-1}}\pp{X=i,\td{X}=j}\frac{\pover{X=i}{S=s_0}}{\pp{X=i}}-\sum_{i\in\mathbb{X}^{1-0}}\pp{X=i,\td{X}=j}\frac{\pover{X=i}{S=s_1}}{\pp{X=i}}.
\end{split}
\label{equ:coupling-condition-binary}
\end{equation}
\normalsize
Note that $\pover{X=i}{S=s}$ is not defined on $\supp{X}\setminus\supp{X_s}$.
Now, set $\pover{X=i}{S=s_0}=0$ for $i\in\mathbb{X}^{1-0}$ and $\pover{X=i}{S=s_1}=0$ for $i\in\mathbb{X}^{0-1}$, such that for all $i\in\supp{X}$, it holds $\pp{X=i}>0$ and $\pover{X=i}{S=s}\geq 0$ for $x\in\supp{X},s\in\supp{S}$.
Equation~\eqref{equ:coupling-condition-binary} can be simplified into 
\small
\begin{equation}
\sum_{i\in\supp{X}}\pp{X=i,\td{X}=j}\frac{\pover{X=i}{S=s_0}-\pover{X=i}{S=s_1}}{\pp{X=i}}=0, \forall  j\in\supp{\td{X}}.
\label{equ:coupling-condition-binary-simple}
\end{equation}
\normalsize
Rewrite it into the matrix form. Using the definition of the vector $V$, Equation~\eqref{equ:coupling-condition-binary-simple} is further simplified into
\begin{equation}
\gamma' V=\mathbb{0}.
\end{equation}

\end{proof}
\end{theorem}

\begin{remark}[Properties of the vector $V$]\label{rem:v-property}
The vector $V$, defined in Theorem~\ref{pro:binary_condition}, has the following properties.
\begin{enumerate}[(i)]
\item $(P^X)' V=0$.
\item Entries of the vector $V$ are neither all negative nor all positive.
\item For a vector $P\in\mathbb{R}^N$, let the $l_1$ norm of this vector be $\|P\|_1:=\sum_{i=1}^N |P_i|$.
If the norm $\|\frac{1}{P^X}\|_1$ is finite, then the norm $\|V\|_1$ is finite, with division being element-wise.
\end{enumerate}
\begin{proof}
As defined in the proof of Theorem~\ref{pro:binary_condition}, the conditional probability $\pover{X=i}{S=s}=0$ for $i\in\supp{X}\setminus\supp{X_s}$. 
we observe $\sum_{i\in\supp{X}}P^{x_{s_0}}=\sum_{i\in\supp{X}}P^{x_{s_1}}=1$.

The property (i) derives from the following:
\begin{equation*}
(P^X)'V=\sum_{i\in\supp{X}}P^X_i V_i= \sum_{i\in\supp{X}}P^X_i \frac{P^{X_{s_0}}_i-P^{X_{s_1}}_i}{P^X_i}=\sum_{i\in\supp{X}}P^{X_{s_0}}_i-P^{X_{s_1}}_i=0. 
\end{equation*}

The property (ii) is proved via contradiction. If $V_i>0$ for $i\in\supp{X}$, such that $P^{X_0}_i-P^{X_1}_i>0$ for $i\in\supp{X}$, hence $\sum_{i\in\supp{X}}(P^{X_0}_i-P^{X_1}_i)=\sum_{i\in\supp{X}}P^{X_0}_i-\sum_{i\in\supp{X}} P^{X_1}_i=1-1>0$. This contradiction shows that entries of $V$ cannot be all positive. The contradiction for all negative is similar.

The property (iii) derives from:
\begin{equation*}
\|V\|_1\leq\|P^{X_{s_0}}-P^{X_{s_1}}\|_1 \|\frac{1}{P^X}\|_1\leq (\|P^{X_{s_0}}\|_1+\|P^{X_{s_1}}\|_1 )\|\frac{1}{P^X}\|_1 =2\|\frac{1}{P^X}\|_1,
\end{equation*}
where the minus and division operators are element-wise. The first inequality uses H{\"o}lder's inequality, and the second inequality uses triangle inequality. The equality uses the fact that vectors $P^{X_{s_0}},P^{X_{s_1}}$ are from the set $\Sigma_N$. Hence, if the norm $\|\frac{1}{P^X}\|_1$ is finite, vector $V$ has finite $l_1$ norm, such that there are not infinite entries in $V$.
\end{proof}
\end{remark}

\subsection{Partial Repair}

The projection in Definition~\ref{def:projection} consist of changing the features and weights of samples in source data, while practical implementation may necessitate maintaining data distortion levels, i.e., the inner product $\langle C,\gamma\rangle$, within predefined acceptable thresholds.
From a practical point of view, we consider partial repair, as relaxations of the constraint in Theorem~\ref{pro:binary_condition}.


To measure how much we violate the total repair property, we first introduce: 
\begin{definition}[\acrfull{TV} distance]\label{def:tvdistance}
Given two discrete probability distributions $P$, $Q$ over $\supp{\td{X}}$, the TV distance $\tv{P}{Q}$ between $P$ and $Q$ is defined as:
\begin{align}
\tv{P}{Q}:=\frac{1}{2}\sum_{j\in\supp{\td{X}}}|P_j-Q_j|=\frac{1}{2}\|P-Q\|_1.
\end{align}
\end{definition}

Now let us analyse the TV distance between $P^{\td{X}_{s_0}}$ and $P^{\td{X}_{s_1}}$:
\begin{lemma}\label{lem:S-wise TV} 
Let the vector $V$ have finite $l_1$-norm, such that we can fine a nonnegative vector $\Lambda\in\mathbb{R}^N_+$ that satisfies 
\begin{equation}
-\Lambda_j\leq\sum_{i\in\overline{\supp{X}}} \gamma_{i,j}V_i\leq \Lambda_j,\quad \forall j\in\supp{\td{X}},
\end{equation}
where $\overline{\supp{X}}:=\{i\in\supp{X}\mid V_i\neq 0\}$.
Hence, $\|\gamma'V\|_1$ is bounded by $\|\Lambda\|_1$. Using Definition~\ref{def:tvdistance}, the TV distance between $P^{\td{X}_{s_0}}$ and $P^{\td{X}_{s_1}}$ is bounded by
\begin{equation}
\tv{P^{\td{X}_{s_0}}}{P^{\td{X}_{s_1}}}=\frac{\left\|\gamma'V\right\|_1}{2}\leq \frac{\|\Lambda\|_1}{2}.\end{equation}
\begin{proof}
Equation~\eqref{equ:difference2rv} shows that $P^{\td{X_0}}-P^{\td{X_1}}=\gamma' V$. 
See Appendix~\ref{app:lem:S-wise TV} for details.
\end{proof}
\end{lemma}

Now, we give the definition of partial repair:
\begin{definition}[Partial repair]
Given a vector $\Lambda\in\mathbb{R}^{N}_+$,
We say that $\Lambda$-repair is satisfied if $\Lambda_j\leq\sum_{i\in\overline{\supp{X}}} \gamma_{i,j}V_i\leq \Lambda_j$, for all $j\in\supp{\td{X}}$, such that
\begin{equation}
\tv{P^{\td{X}_{s_0}}}{P^{\td{X}_{s_1}}}\leq \frac{\|\Lambda\|_1}{2}.
\end{equation}
Note that $\mathbb{0}$-repair is equivalent to total repair. Also, computing the value of the upper bound for TV distance needs to consider the dimension of $\Lambda$, i.e, $N$.
\end{definition}


\subsection{Formulations and Algorithms}
Lemma~\ref{lem:S-wise TV} implies that
in the standard formulation of regularised OT, if we were able to add the extra constraint that limits $\|\gamma'V\|_1\leq\|\Lambda\|_1$, this optimal solution $\gamma^*$ would achieve $\Lambda$-repair or total repair if $\Lambda=\mathbb{0}$.
The revised formulation reads:
\begin{align}
\inf_{\gamma\in\Pi(P^{X},P^{\td{X}})} \left\{\langle C,\gamma\rangle-\epsilon E(\gamma)\Bigg|-\Lambda_j\leq \sum_{i\in\overline{\supp{X}}} \gamma_{i,j}V_i\leq\Lambda_j,\forall j\in\supp{\td{X}}\right\},\label{equ:our-formulation-1}
\end{align}
where $\epsilon>0$ and $\Lambda\in\mathbb{R}^{N}_+$.
\textcolor{black}{Note that the matrix variable $\gamma$ is of size $N \times N$, meaning that the number of variables in Equation~\eqref{equ:our-formulation-1} is related to the number of discretisation points $N$.}

\begin{lemma}[The feasible set is non-empty]\label{lem:existence}
We define a subset of all admissible couplings in Equation~\eqref{equ:our-formulation-1}:
\begin{equation}
\Pi_{\Lambda }(P^{X},P^{\td{X}}):=\{\gamma\in\Delta^2_N\mid \gamma\mathbb{1}=P^{X}, \gamma'\mathbb{1}=P^{\td{X}},\left|\gamma' V\right|_j\leq\Lambda_j,\forall j\in\supp{\td{X}}\},
\end{equation}
where $\left|\gamma' V\right|_j:=\left|\sum_{i\in\supp{X}}\gamma_{i,j}V_i\right|$. Let $0\leq\Lambda'_j\leq\Lambda_j$ for all $j\in\supp{\td{X}}$, the following sequence holds
\begin{equation}
\emptyset\neq\Pi_{\mathbb{0}}(P^{X},P^{\td{X}})\subseteq\Pi_{\Lambda'}(P^{X},P^{\td{X}})\subseteq\Pi_{\Lambda}(P^{X},P^{\td{X}})\subseteq\Pi(P^{X},P^{\td{X}}).
\end{equation}
\begin{proof}
The subset of all admissible couplings is
\begin{equation*}
\Pi_{\Lambda }(P^{X},P^{\td{X}}):=\{\gamma\in\Sigma^2_N\mid \gamma\mathbb{1}=P^{X}, \gamma'\mathbb{1}=P^{\td{X}},\left|\gamma' V\right|_j\leq\Lambda_j,\forall j\in\supp{\td{X}}\},
\end{equation*}
where $\left|\gamma' V\right|_j:=\left|\sum_{i\in\supp{X}}\gamma_{i,j}V_i\right|$. Recall the definition of feasible couplings in Equation~\eqref{equ:Pi-define}, we can equivalently write 
\begin{equation*}
\Pi_{\Lambda }(P^{X},P^{\td{X}})=\Pi(P^{X},P^{\td{X}})\cap\left\{ {\left|\gamma' V\right|_j\leq\Lambda_j,\forall j\in\supp{\td{X}}} \right\},
\end{equation*}
such that $\Pi_{\Lambda }(P^{X},P^{\td{X}})\subseteq\Pi(P^{X},P^{\td{X}})$.
Now for any two nonnegative vectors $\Lambda',\Lambda\in\mathbb{R}^{N}_+$, with their entries satisfying $0\leq\Lambda'_j\leq\Lambda_j$ for all $j\in\supp{\td{X}}$, the following holds
\begin{equation*}
\left\{ {\left|\gamma' V\right|_j\leq\Lambda'_j,\forall j\in\supp{\td{X}}} \right\} \subseteq \left\{ {\left|\gamma' V\right|_j\leq\Lambda_j,\forall j\in\supp{\td{X}}} \right\}.
\end{equation*}
Hence, we can deduce that
\begin{equation}
\begin{gathered}
\Pi(P^{X},P^{\td{X}})\cap\left\{ {\left|\gamma' V\right|_j\leq\Lambda'_j,\forall j\in\supp{\td{X}}} \right\} \subseteq \Pi(P^{X},P^{\td{X}})\cap\left\{ {\left|\gamma' V\right|_j\leq\Lambda_j,\forall j\in\supp{\td{X}}} \right\}\\
\iff  \Pi_{\Lambda'}(P^{X},P^{\td{X}})\subseteq\Pi_{\Lambda}(P^{X},P^{\td{X}}).    
\end{gathered}
\label{equ:theta-sequence}
\end{equation}
The set $\Pi_{\Lambda}(P^{X},P^{\td{X}})$ becomes $\Pi(P^{X},P^{\td{X}})$ when $\Lambda$ approaches infinity, and it becomes $\Pi_{\mathbb{0}}(P^{X},P^{\td{X}})$ when $\Lambda$ is zero. Along with Equation~\eqref{equ:theta-sequence}, we conclude the following sequence
\begin{equation}
\Pi_{\mathbb{0}}(P^{X},P^{\td{X}})\subseteq\Pi_{\Lambda'}(P^{X},P^{\td{X}})\subseteq\Pi_{\Lambda}(P^{X},P^{\td{X}})\subseteq\Pi(P^{X},P^{\td{X}}).
\label{equ:pi-sequence}
\end{equation}

Next, we show that the set $\Pi_{\mathbb{0}}(P^{X},P^{\td{X}})$ is non-empty.
Remark~2.13 in \cite{peyre2019computational} states that $\Pi(P^{X},P^{\td{X}})$ is not empty with an example of $P^{X}\otimes P^{\td{X}}\in\Pi(P^{X},P^{\td{X}})$, where $\otimes$ is the outer product. To verify this, we define a coupling $\gamma$ with its entries being $\gamma_{i,j}:=P^{X}_i P^{\td{X}}_j$. 
Using the fact $\sum_{i\in\supp{X}}P^{X}_i=\sum_{j\in\supp{\td{X}}}P^{\td{X}}_j=1$, we observe
\begin{align*}
\sum_{j\in\supp{\td{X}}}\gamma_{i,j}=\sum_{j\in\supp{\td{X}}}P^{X}_i P^{\td{X}}_j=P^{X}_i\sum_{j\in\supp{\td{X}}}P^{\td{X}}_j=P^{X}_i, \forall i\in\supp{X};\\
\sum_{i\in\supp{X}}\gamma_{i,j}=\sum_{i\in\supp{X}}P^{X}_i P^{\td{X}}_j=P^{\td{X}}_j\sum_{i\in\supp{X}}P^{X}_i =P^{\td{X}}_j,\forall j\in\supp{\td{X}};
\end{align*}
such that this coupling does belong to $\Pi(P^{X},P^{\td{X}})$.
Using Property (i) in Remark~\ref{rem:v-property}, we further observe 
\begin{equation*}
\sum_{i\in\supp{X}}\gamma_{i,j} V_i=P^{\td{X}}_j \sum_{i\in\supp{X}}P^{X}_i V_i=P^{\td{X}}_j\times \left((P^X)' V\right)=0,\forall j\in\supp{\td{X}},
\end{equation*}
such that this coupling also belongs to $\Pi_{\mathbb{0}}(P^{X},P^{\td{X}})$.
Since we found one coupling in the set $\Pi_{\mathbb{0}}(P^{X},P^{\td{X}})$, this set is not empty.
Using Equation~\eqref{equ:pi-sequence}, we know that for arbitrary nonnegative $\Lambda\in\mathbb{R}^N_+$, the feasible set of our formulation in Equation~\eqref{equ:our-formulation-1} is not empty, i.e., $\Pi_{\Lambda}(P^{X},P^{\td{X}})\neq\emptyset$, such that the solutions of our formulation exist.
\end{proof}
\end{lemma}

According to the same reasoning in Equation~\eqref{equ:KL-equivalent}, we can rewrite Equation~\eqref{equ:our-formulation-1} into Equation~\eqref{equ:our-formulation-2}. Note that the minimum exists because the feasible set is non-empty (c.f., Lemma~\ref{lem:existence}) and the objective function is coercive.
\begin{align}
\min_{\gamma\in\Pi_{\Lambda }(P^{X},P^{\td{X}})} \kl{\gamma}{\xi}, \textrm{ where }\xi=\exp{(-C/\epsilon)},
\label{equ:our-formulation-2}
\end{align}
and $\Pi_{\Lambda }(P^{X},P^{\td{X}})=\bigcap^{3}_{\ell=1}\mathcal{C}_{\ell}$ is the intersection of three convex sets:
\begin{equation}
\begin{split}
\mathcal{C}_1=\{\gamma\in\mathbb{R}^{N\times N}_+\mid\gamma\mathbb{1}=P^{X}\},&\mathcal{C}_2=\{\gamma\in\mathbb{R}^{N\times N}_+\mid\gamma'\mathbb{1}=P^{\td{X}}\},\\
\mathcal{C}_3=\{\gamma\in\mathbb{R}^{N\times N}_+&\mid -\Lambda\leq\gamma'V\leq\Lambda\}.
\end{split}
\label{equ:convex-sets}
\end{equation}
Note that $\mathcal{C}_3$ is equivalent to $\{\gamma\in\mathbb{R}^{N\times N}_+\mid \gamma'V=\mathbb{0}\}$ if $\Lambda=\mathbb{0}$.

\label{page:define-dykstra}
Then apply Dykstra's algorithm, we propose Algorithm~\ref{alg:Dykstra} to solve Equations~(\ref{equ:our-formulation-2}--\ref{equ:convex-sets}).
The inputs consist of the support $\supp{X}$ (resp. $\supp{\td{X}}$) and probability distribution $P^{X}$ (resp. $P^{\td{X}}$) of variables $X$ (resp. $\td{X}$); the number of discretised point of these supports $N$; the vectors $V$, $\Lambda$; the cost matrix $C$; entropic regularisation parameter $\epsilon>0$, the number of iterations $K$ and a very small number $\varepsilon$.
The algorithm is repeated \acrshort{KL} projections to $\mathcal{C}_1,\mathcal{C}_3,\mathcal{C}_3$, together with the computation of an auxiliary sequence $\{q_{k}\}_{k\geq 1}$.
Note that the very small number $\varepsilon\leq \|\Lambda\|_1$ is used to avoid arithmetic underflow problems, which could result in zero division errors in computing sequence $\{q_{k}\}_{k\geq 1}$.

\algrenewcommand\algorithmicrequire{\textbf{Input:}}
\algrenewcommand\algorithmicensure{\textbf{Output:}}

\begin{algorithm}[htp]
\caption{Our method}\label{alg:Dykstra}
\begin{algorithmic}
\Require $\supp{X},\supp{\td{X}},N$.
\Require $P^{X},P^{\td{X}},V$, $\Lambda,\epsilon,C,K,\varepsilon$.
\State Define three convex sets
$$\mathcal{C}_1=\{\gamma\in\mathbb{R}^{N\times N}_+\mid\gamma\mathbb{1}=P^{X}\},\;\mathcal{C}_2=\{\gamma\in\mathbb{R}^{N\times N}_+\mid\gamma'\mathbb{1}=P^{\td{X}}\},$$
$$\mathcal{C}_3=\{\gamma\in\mathbb{R}^{N\times N}_+\mid -\Lambda\leq\gamma'V\leq\Lambda\}.$$
\State Initialise $\gamma^{(0)}=\exp{(-C/\epsilon)}$ \Comment{Dykstra's Algorithm}
\For{$k=1,\dots,3$}
\State Compute $\gamma^{(k)}=\prox^{KL}_{\mathcal{C}_k}(\gamma^{(k-1)})$.
\EndFor
\For{$k=4,\dots,K$}
\State Set $\mathcal{C}_k=\mathcal{C}_{1+(k\mod 3)}$
\State Compute
\begin{equation*}
q_{k-3}:=
\begin{cases}
\frac{\gamma^{(k-4)}}{\gamma^{(k-3)}}& \textrm{ if }k=4,\dots,7\\
q_{k-7}\odot\frac{\gamma^{(k-4)}}{\gamma^{(k-3)}}&\textrm{ otherwise}.
\end{cases}
\end{equation*}
\State Compute $\gamma^{(k)}=\prox^{KL}_{\mathcal{C}_{k}}(\gamma^{(k-1)}\odot q_{k-3})$, using Lemma~\ref{lem:c_1&c_2} and~\ref{lem:prox_c3}.
\If{$\|(\gamma^{(k)})'V\|_1<\varepsilon$} \Comment{To avoid arithmetic underflow and zero division errors.}
\State \textbf{break}
\EndIf
\EndFor
\Ensure The solution of Equations~(\ref{equ:our-formulation-2}--\ref{equ:convex-sets}): $\gamma^{(K)}$.
\end{algorithmic}
\end{algorithm}

Now, we explain how to compute these \acrshort{KL} projections in our algorithm:
\begin{assumption}[Finite $l_1$ norm]\label{ass:finite}
The norm $\|\frac{1}{P^X}\|_1$ is finite, where the division operator is element-wise.
Then, property (iii) in Remark~\ref{rem:v-property} shows that the vector $V$ has finite $l_1$-norm without infinite entries.
\end{assumption}

\begin{lemma}[Subgradient optimality conditions of \acrshort{KL} projections]\label{lem:subgradient-optimality}
For a set $\mathcal{C}$ in $\mathbb{R}^{N\times N}$, we define its indicator function $\iota_{\mathcal{C}}$:
\begin{equation}
\iota_{\mathcal{C}}(x):=
\begin{cases}
0&\textrm{ if } x\in\mathcal{C},\\
+\infty&\textrm{ otherwise.}
\end{cases}
\label{equ:indicator}
\end{equation}
If $\mathcal{C}$ is a closed convex set, the \acrshort{KL} projection for $\mathcal{C}$ has a unique solution $\gamma^*$:
\begin{equation*}
\mathbb{0}\in\log\left(\frac{\gamma^*}{\bar{\gamma}}\right)+\partial\iota_{\mathcal{C}}(\gamma^*),
\end{equation*}
where $\partial\iota_{\mathcal{C}}(\gamma^*)$ denotes the set of all subgradients $\nu$ of $\iota_{\mathcal{C}}$ at $\gamma^*\in\dom\iota_{\mathcal{C}}$ satisfying Equation~\eqref{equ:subgradient-define}, with ``$\textrm{dom}$'' being the essential domain:
\begin{equation}
\iota_{\mathcal{C}}(\gamma)\geq \iota_{\mathcal{C}}(\gamma^*)+\langle \nu,(\gamma-\gamma^*)\rangle, \forall \gamma\in\mathbb{R}^{N\times N}.\label{equ:subgradient-define}
\end{equation}
\begin{proof}
See Appendix~\ref{app:lem:subgradient-optimality}.
\end{proof}
\end{lemma}

\begin{lemma}[\acrshort{KL} projections for $\mathcal{C}_1,\mathcal{C}_2$]
\label{lem:c_1&c_2}
The \acrshort{KL} projection for $\mathcal{C}_1,\mathcal{C}_2$ in Algorithm~\ref{alg:Dykstra},~\ref{alg:baseline} or~\ref{alg:barycentre} could be computed in closed form:
\begin{equation}
\prox^{KL}_{\mathcal{C}}(\bar{\gamma})=
\begin{cases}
\diag{\frac{P^{X}}{\bar{\gamma}\mathbb{1}}}\bar{\gamma} & \textrm{if } \mathcal{C}=\{\gamma\in\mathbb{R}^{N\times N}_+\mid\gamma\mathbb{1}=P^{X}\},\\
\bar{\gamma}\diag{\frac{P^{\td{X}}}{\bar{\gamma}'\mathbb{1}}} & \textrm{if } \mathcal{C}=\{\gamma\in\mathbb{R}^{N\times N}_+\mid\gamma'\mathbb{1}=P^{\td{X}}\},
\end{cases}
\end{equation}
where the division operator is element-wise.
\begin{proof}
See Appendix~\ref{app:lem:c_1&c_2}.
\end{proof}
\end{lemma}

\begin{lemma}[\acrshort{KL} projections for $\mathcal{C}_3$]
\label{lem:prox_c3}
Under Assumption~\ref{ass:finite}, the \acrshort{KL} projection for $\mathcal{C}_3$ in Algorithm~\ref{alg:Dykstra} could be computed in iterative manner:
\begin{equation}
\gamma^*=\prox^{KL}_{\mathcal{C}}(\bar{\gamma}), \textrm{ where }\mathcal{C}=\left\{\gamma\in\mathbb{R}^{N\times N}_+\Bigg|-\Lambda\leq\gamma' V\leq\Lambda\right\}.
\end{equation}
If $\bar{\gamma}\in\mathcal{C}$, $\gamma^*=\bar{\gamma}$. Otherwise, for all $i\in\supp{X},j\in\supp{\td{X}}$
\begin{equation}
\gamma^*_{i,j}:=
\begin{cases}
\bar{\gamma}_{i,j}\exp{(-V_i\nu_j )}\quad &\textrm{if }i\in\overline{\supp{X}}  \textrm{ and } [\bar{\gamma}'V]_j\notin[-\Lambda_j,\Lambda_j], \\
\bar{\gamma}_{i,j} \quad &\textrm{if }i\notin\overline{\supp{X}} \textrm{ or  } [\bar{\gamma}'V]_j\in[-\Lambda_j,\Lambda_j].
\end{cases}
\end{equation}
where $[\bar{\gamma}' V]_j$ is the $j^{th}$ entry of the vector $\bar{\gamma}'V$.
$\nu_j$ needs to satisfy Equation~\eqref{equ:nu_condition}. 
\begin{equation}
\sum_{i\in\overline{\supp{X}}}\bar{\gamma}_{i,j}V_i\exp{(-V_i\nu_j )}=
\begin{cases}
\Lambda_j &\textrm{if } [\bar{\gamma}'V]_j > \Lambda_j,\\
-\Lambda_j &\textrm{if } [\bar{\gamma}'V]_j < -\Lambda_j.
\end{cases}
\label{equ:nu_condition}
\end{equation}
\begin{proof}
See Appendix~\ref{app:lem:prox_c3}.
\end{proof}
\end{lemma}

\begin{remark}[Computation of $\nu_j$]
\label{rem:nu_j}
$\nu_j$ in Equation~\eqref{equ:nu_condition} is indeed the root of function $F(x):=\sum_{i\in\overline{\supp{X}}}\bar{\gamma}_{i,j}V_i\exp{(-V_i x)} + c$, where $c=\pm\Lambda_j$ is a constant. Its first derivative 
$\nabla F(x)=-\sum_{i\in\overline{\supp{X}}}\bar{\gamma}_{i,j}(V_i)^2\exp{(-V_i x)}\leq 0$,
such that $F(x)$ is non-increasing.
Under Assumption~\ref{ass:finite}, the root (i.e., $\nu_j$) can be found by root-finding algorithms e.g., Newton-Raphson algorithm, bisection method.
\end{remark}

\begin{lemma}[Convergence of Algorithm~\ref{alg:Dykstra}]\label{lem:convergence_algorithm1}
In the case of binary sensitive attributes, i.e., $\supp{S}=\{s_0,s_1\}$, under Assumption~\ref{ass:finite}, given the vector $\Lambda\in\mathbb{R}^N_+$, and \acrshort{KL} projections being computed as in Lemma~\ref{lem:c_1&c_2}--\ref{lem:prox_c3}, Algorithm~\ref{alg:Dykstra} converges to the unique coupling of Equation~(\ref{equ:our-formulation-2}--\ref{equ:convex-sets}). Then projecting the source data via the group-blind map induced from the unique coupling, we can achieve $\Lambda$-repair in projected data.
\begin{proof}
This proof relies on computation of \acrshort{KL} projections, which is explained in Lemma~\ref{lem:subgradient-optimality}--\ref{lem:prox_c3}, and the convergence result of Dykstra's algorithm in \cite{bauschke2000dykstras}, with Lemma~\ref{lem:existence} ensuring that our formulation satisfies the assumption of Dykstra's algorithm.
\end{proof}
\end{lemma}

\subsection{Higher Dimensions}
Now, we consider situations with more than one neutral attribute:
\begin{itemize}

    \item There are extra attribute(s) $U$ in the source data that need to stay the same during the projection, or $U$ is \textit{$S$-neutral}. In this case, a sample becomes $(i,u,s)$, where $i\in\supp{X}$, and $u,s$ are the values of attributes $U,S$.
    As mentioned in Definition~\ref{def:projection}, the projection map $\mathscr{T}$ would not change any attributes other than $X$. 
    Hence, this sample is split into a sequence of weighted samples $\{(j,w_{j},u,s)\}_{j\in\supp{\td{X}}}$. 
    
\item There are extra attribute(s) that need to be adjusted. Without loss of generality, we assume attributes $X_1,X_2\in\mathbb{R}$ need to be adjusted. Let $X=(X_1,X_2)$ and $\supp{X}=\supp{(X_1,X_2)}\in\mathbb{R}^2$, such that $\supp{X}$ includes all tuples $(x_1,x_2)$ present in the source data, where $x_1\in\supp{X_1},x_2\in\supp{X_2}$.
Then $N$ is the cardinality of the set $\supp{X}$. In the same manner, we can define $\supp{\td{X}}$ and the probability distributions:
\begin{equation}
\begin{split}
P^X_i:=\pp{(X_1,X_2)=i}&,\quad
P^{X_s}_i:=\pover{(X_1,X_2)=i}{S=s}, \forall i\in\supp{X},\\P^{\td{X}}_j:=\pp{(\td{X}_1,\td{X}_2)=j}&,\quad P^{\td{X}_s}_j:=\pover{(\td{X}_1,\td{X}_2)=j}{S=s}, \forall j\in\supp{\td{X}}.
\end{split}
\end{equation}
It is worth noting that the cost matrix $C$ needs to consider the range of each attribute that need to be adjusted. Suppose $\supp{X_1}=\{1,2\}$ and $\supp{X_1}=\{1,\dots,4\}$. While not necessarily appropriate in practice, if the entries of cost matrix $C_{i,j}:=\|i-j\|_1$, the cost of moving $X_1$ from $1$ to $2$ would be treated the same as moving $X_2$ from $1$ to $2$. Instead, we use $C_{i,j}:=\|\varrho\odot (i-j)\|_1$, where $\varrho\in\mathbb{R}^{2}_+$ denotes the weights for the cost of moving one unit of $X_1,X_2$, and $\odot$ is element-wise product. If $\varrho=[1,1/4]$, the cost of moving $X_1$ from $1$ to $2$ is the same as moving $X_2$ from $1$ to $4$.
\end{itemize}

\subsection{Choices of Target Distributions} 
There is no an explicit requirement regarding $P^{\td{X}}\in\Delta_N$ in Lemma~\ref{lem:convergence_algorithm1}. It leaves the choice of target distribution open.
\begin{itemize}
\item  Concerning with data distortion, one option would be the barycentre distribution, because it brings the least alteration to source data, from its definition in Equation~\eqref{equ:barycentre-define}, or in the original works of \cite{gordaliza2019obtaining,oneto2020fairness,jiang2020wasserstein}. 
In our setting, the requirement regarding the sensitive attribute $S$ is the vector $V$ only. If those $S$-wise distributions $P^{X_{s_0}},P^{X_{s_1}}$ are not explicitly given, we cannot compute the barycentre between them. But we include it as barycentre baseline in Section~\ref{sec:ot_experiments}.2.
\item The other option concerns with performance or utility of some pre-established machine learning models. 
Suppose the classification or prediction model is trained on a training set.
We set the target distribution the same as the the distribution of the training set, or other common representation space, to preserve the classification or prediction performance of this pre-trained model.
\end{itemize}

\section{Numerical Illustrations}\label{sec:ot_experiments}
Bias repair schemes consist of mapping source data into the pre-designed target data, and after the mapping, we obtain the projected data, which are expected to have equalised distributions of features in privileged and unprivileged groups.
During the mapping, a sample in the source data is divided into a sequence of weighted samples, leading to data distortion—specifically, a deviation from its original representation.
The data distortion may dampen accuracy or utility of some pre-trained prediction or classification models.
For example, recall the motivating example introduced in Section~\ref{sec:ot_movitation}. If all applicants are projected to the maximum exam score, the cut-off point loses its ability to distinguish between privileged and unprivileged groups but also significantly forfeits its capacity to classify qualified applicants.

This section first demonstrates the effects of our total repair and partial repair schemes using synthetic data, and then illustrates the classic trade-off between bias repair and data distortion, using the Adult Census Income dataset of \cite{misc_adult_2}.
Specifically,
Section~\ref{sec:ot_experiments}.1 explains how we use the sensitive attribute in experiments and introduce the performance indices.
Section~\ref{sec:ot_experiments}.2 introduces two baselines. 
Sections~\ref{sec:ot_experiments}.3 and \ref{sec:ot_experiments}.4 describe our experiments on synthetic data and real data, with comparison with these baselines. 
Our implementation is available online \footnote{\url{https://github.com/Quan-Zhou/OT_Debiasing}}.



\subsection{Input and Indices}

We have made the assumption that the sensitive attribute $S$ of each sample is not observed, but the vector $V$, as defined in Theorem~\ref{pro:binary_condition}, is given as the input for our algorithm.
The computation of $V$ requires the population-level information regarding the distributions of features for both groups ($s_0$ and $s_1$), given the crucial assumption that the source data represents unbiased samples from the broader population, such that groups in source data follow the same distributions.
In our experiments, since there is no such population-level information given, we directly compute $V$ from the source data. 
Apart from computing $V$, the sensitive attribute of each sample is not utilised for computing the coupling or for projecting source data.


Next, to introduce the performance indices, we start from defining computation of empirical distributions:
\begin{definition}[Empirical distributions]\label{def:empirical}
Let $\supp{Z}\in\mathbb{R}$.
Given data $(i,w_{i})_{i\in\supp{Z}}$, the empirical distribution $P^Z$ is defined as
\begin{equation}
P^Z_i:=\frac{w_i}{\sum_{i\in\supp{Z}} w_{i}}.
\label{equ:PY-define}
\end{equation}
For instance, Equation~\eqref{equ:PY-define} computes $P^{X}$ if $Z=X$, and $P^{\td{X}}$ if $Z=\td{X}$.
If the  sensitive attribute $S$ is known, such that the data become $(i,s,w_{i,s})_{i\in\supp{Z},s\in\supp{S}}$, we can compute the $S$-wise distributions $P^{Z_{s}}$ as follows:
\begin{equation}
P^{Z_{s}}_i:=\frac{w_{i,s}}{\sum_{i\in\supp{Z}} w_{i,s}},\;\forall\; s\in\supp{S}.
\label{equ:PY|s-define}
\end{equation}
\end{definition}

\begin{definition}[Performance indices]\label{def:performance}
Let $Y$ and $\hat{Y}$ denote the label that we need to estimate and the estimated label.
To measure the performance of prediction / classification models on the source or projected data, we introduce f1 scores as accuracy measures, disparate impact as the fairness measure, and $S$-wise TV distance:
\begin{align*}
\textrm{f1 micro}:=&\frac{\sum_{s\in\supp{S}} (2\times TP_s)}{\sum_{s\in\supp{S}} (2\times TP_s + FP_s + FN_s)},\\
\textrm{f1 macro}:=&\sum_{s\in\supp{S}}\frac{1}{2}\times\frac{2\times TP_s}{2\times TP_s + FP_s + FN_s},\\
\textrm{f1 weighted}:=&\sum_{s\in\supp{S}}{P^{S}_s}\times\frac{2\times TP_s}{2\times TP_s + FP_s + FN_s},\\
\textrm{DisparateImpact}:=&\frac{\pover{\hat{Y}=1}{S=s_0}}{\pover{\hat{Y}=1}{S=s_1}},\\
S\textrm{-wise TV distance}:=&\tv{P^{\td{X}_{s_0}}}{P^{\td{X}_{s_1}}},
\end{align*}
where $s_0$ denotes the unprivileged group and $Y=0$ is the negative label.
$TP_s,FP_s,FN_s$ are the numbers of true positive, false positive and false negative of the group $s$.
Empirical distributions $P^{S},P^{\td{X}_{s_0}},P^{\td{X}_{s_1}}$ are computed from the projected data, following Definition~\ref{def:empirical}.
Ideally, after the repair, f1 scores increase, disparate impact gets closer to $1$, and $S$-wise TV distance decreases.
\end{definition}


\subsection{Two Baselines}
\begin{definition}[The baseline of iterative Bregman projections]\label{def:baseline}
The first baseline is to solve the standard regularised OT formulation:
\begin{equation}
\min_{\gamma\in\Pi(P^{X},P^{\td{X}})} \langle C,\gamma \rangle -\epsilon E(\gamma)=\kl{\gamma}{\xi}, \textrm{ where }\Pi(P^{X},P^{\td{X}})=\bigcap^{2}_{\ell=1}\mathcal{C}_{\ell},\;\xi=\exp{(-C/\epsilon)}.\label{equ:formulation-baseline}
\end{equation}
The only difference between the baseline and our method is that we add one additional constraint $-\Lambda\leq\gamma'V\leq\Lambda$.
The optimal solution of Equation~\eqref{equ:formulation-baseline} can be founded by Dykstra's Algorithm, or computationally cheaply using the iterative Bregman projections, as suggested in Algorithm~\ref{alg:baseline}. 
Both algorithms converge at the same results, because $\mathcal{C}_1,\mathcal{C}_2$ are affine subspaces (See Fact~1.4 in \cite{bauschke2020dykstra}, or Theorem~4.3 in \cite{bauschke2000dykstras}).

\begin{center}
\begin{algorithm}[htp]
\caption{Baseline method}\label{alg:baseline}
\begin{algorithmic}
\Require $\supp{X},\supp{\td{X}}, N$, $P^{X},P^{\td{X}}$, $\epsilon,C,K$.
\State Set $$\mathcal{C}_1=\{\gamma\in\mathbb{R}^{N\times N}_+\mid\gamma\mathbb{1}=P^{X}\},\;\mathcal{C}_2=\{\gamma\in\mathbb{R}^{N\times N}_+\mid\gamma'\mathbb{1}=P^{\td{X}}\}.$$
\State Initialise $\gamma^{(0)}=\exp{(-C/\epsilon)}$. \Comment{Iterative Bregman Projections}
\For{$k=1,\dots,K$}
\State Set $\mathcal{C}_k=\mathcal{C}_{1+(k\mod 2)}$
\State Compute $\gamma^{(k)}=\prox^{KL}_{\mathcal{C}_k}(\gamma^{(k-1)})$, using Lemma~\ref{lem:c_1&c_2}.
\EndFor
\Ensure The solution of Equation~\eqref{equ:formulation-baseline}: $\gamma^{(K)}$.
\end{algorithmic}
\end{algorithm}
\end{center}
\end{definition}

We consider another baseline of barycentre projection: 
\begin{definition}[The barycentre projection in \cite{gordaliza2019obtaining}]\label{def:barycentre}
The second baseline is the total repair in Section~5.1.1 (B) of  \cite{gordaliza2019obtaining}, where each sample split its mass to be transported, the same as our setting.
The idea is that the two conditional distributions of the random variable $X$ by the sensitive attribute $S$ are going to be transformed into the Wasserstein barycentre $P^{B}$ between $P^{X_{s_0}}$ and $P^{X_{s_1}}$, with weight $\pi_0$ and $\pi_1$. The 1-Wasserstein barycentre is defined as
\begin{equation}
P^{B}=\arg\min_{P\in\Sigma_{N}}\{\pi_0 \mathcal{W}_1(P^{X_{s_0}},P)+\pi_1 \mathcal{W}_1(P^{X_{s_1}},P)\},
\label{equ:barycentre-define}
\end{equation}
where $\mathcal{W}_1(P^{X_{s_0}},P^{X_{s_1}}):=\sum_{i\in\supp{X}} |P^{X_{s_0}}_i - P^{X_{s_1}}_i|$ denotes the $1$-Wasserstein distance between two distributions $P^{X_{s_0}},P^{X_{s_1}}$ defined on $\supp{X}$. 
The nonnegative constants $\pi_0+\pi_1=1$ are barycentric coordinates.

The barycentre within two marginal distributions can be found via the optimal coupling between $P^{X_{s_0}},P^{X_{s_1}}$ (Remark~4.1 in \cite{gordaliza2019obtaining}, or Proposition~5.9 in \cite{villani2021topics}). 
The implementation is to find the coupling $\gamma^B\in\Pi(P^{X_{s_0}},P^{X_{s_1}})$, as in Algorithm~\ref{alg:barycentre}, and then transport the group of $s_0$ by the map corresponding to the coupling $\gamma^{0\to B}$ and the group of $s_1$ by the map corresponding to the coupling $\gamma^{1\to B}$, whose entries are defined by 
\begin{equation}
\gamma^{0\to B}_{i,k}:=\gamma^{B}_{i,\;\pi_0\cdot i + \pi_1\cdot k},\quad \gamma^{1\to B}_{i,k}:=\gamma^{B}_{\pi_1\cdot i + \pi_0\cdot k,\; i}.
\end{equation}
\begin{center}
\begin{algorithm}[htp]
\caption{Barycentre method}\label{alg:barycentre}
\begin{algorithmic}
\Require $\supp{X}, N$, $P^{X_{s_0}},P^{X_{s_1}}$, $\epsilon,C,K$, $\pi_0,\pi_1$.
\State Set $$\mathcal{C}_1=\{\gamma\in\mathbb{R}^{N\times N}_+\mid\gamma\mathbb{1}=P^{X_{s_0}}\},\;\mathcal{C}_2=\{\gamma\in\mathbb{R}^{N\times N}_+\mid\gamma'\mathbb{1}=P^{X_{s_1}}\}.$$
\State Initialise $\gamma^{(0)}=\exp{(- C/\epsilon)}$. \Comment{Iterative Bregman Projections}
\For{$k=1,\dots,K$}
\State Set $\mathcal{C}_k=\mathcal{C}_{1+(k\mod 2)}$
\State Compute $\gamma^{(k)}=\prox^{KL}_{\mathcal{C}_k}(\gamma^{(k-1)})$, using Lemma~\ref{lem:c_1&c_2}.
\EndFor
\Ensure The solution of $\gamma^B$ in Definition~\ref{def:barycentre}: $\gamma^{(K)}$.
\end{algorithmic}
\end{algorithm}
\end{center}
\end{definition}


\subsection{Revisiting Motivating Example}
In the first example, we compare with the baseline in Definition~\ref{def:baseline} on synthetic data, to demonstrate that an arbitrary $S$-blind projection map would not reach total repair.

Recall the motivating example introduced in Section~1.1, where there is only one neutral attribute (the exam score) and one binary sensitive attribute.
Let $\supp{X}=\supp{\td{X}}=\{-30,-29,\dots,10\}$. Suppose $\supp{S}=\{s_0,s_1\}$ and $\pp{s_0}=0.7$, $\pp{s_1}=0.3$. 
The random variable $X_{s_0}$ (resp. $X_{s_1}$) follow a discretised Gaussian distribution $\mathcal{N}(-10,6^2)$ (resp. $\mathcal{N}(1,3^2)$).

\begin{figure}[htp]
\centering\includegraphics[scale=0.6]{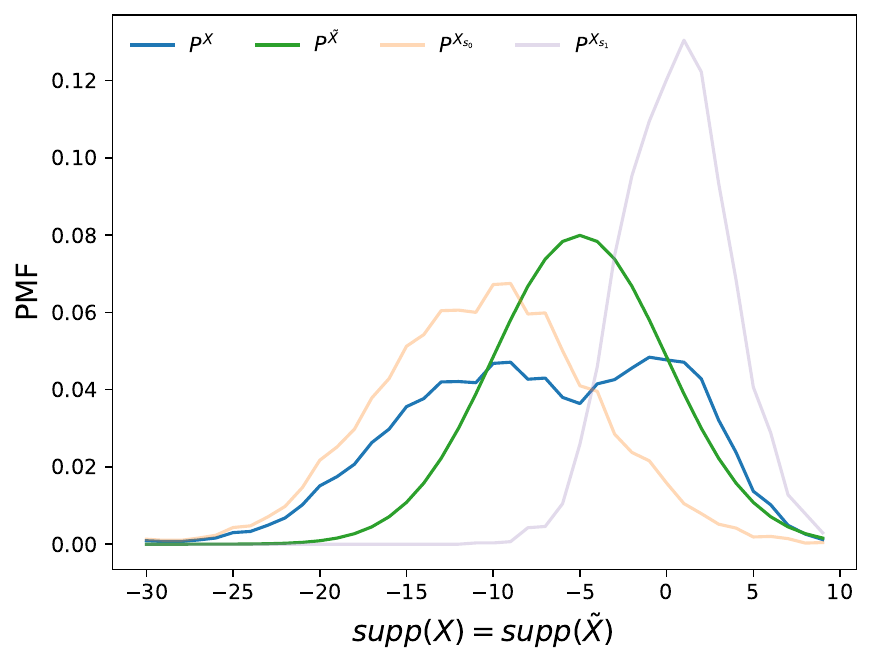}
\caption[The empirical distributions of generated source data and the target distribution]{Overview of the empirical distributions of $P^{X},P^{X_{s_0}},P^{X_{s_1}}$ from the generated source data and the target distribution $P^{\td{X}}$ that we set arbitrarily.}
\label{fig:exa_overview}
\end{figure}

We generate $M=10^{4}$ samples (source data) by repeating the procedure for $M$ times: first, we generate a uniform random variable within $[0,1]$. If this uniform variable is smaller than $\pp{s_0}$, we generate a Gaussian random variable $X_{s_0}\sim\mathcal{N}(-10,6^2)$ and the sample reads $(\lfloor X_{s_0}\rfloor,s_0)$, where $\lfloor \cdot\rfloor$ is the floor function. Otherwise, the sample should be $(\lfloor X_{s_1}\rfloor,s_1)$, where $X_{s_1}\sim\mathcal{N}(1,3^2)$. 
Then, we compute the empirical distributions of $P^{X},P^{X_{s_0}},P^{X_{s_1}},P^{S}$ from the source data.
For the target distribution, we arbitrarily set $P^{\td{X}}$ to the distribution of discretised Gaussian distribution $\mathcal{N}(-5,5^2)$.
In Figure~\ref{fig:exa_overview}, the discrete distributions $P^{X},P^{X_{s_0}},P^{X_{s_1}},P^{\td{X}}$ are displayed by blue, orange, purple, green curves, respectively.

\begin{figure}[htb]
\begin{minipage}{\textwidth}
\begin{minipage}{.24\textwidth}
\centering
\includegraphics[scale=0.5]{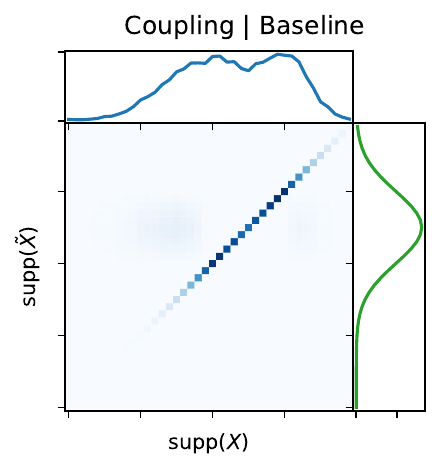}
\label{fig:prob1_6_2}
\end{minipage}%
\begin{minipage}{0.24\textwidth}
\centering
\includegraphics[scale=0.5]{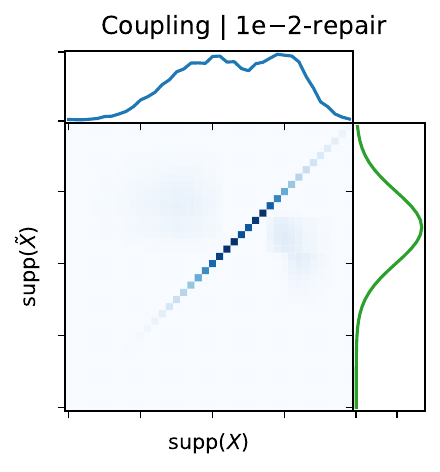}
\label{fig:prob1_6_1}
\end{minipage}
\begin{minipage}{.24\textwidth}
\centering
\includegraphics[scale=0.5]{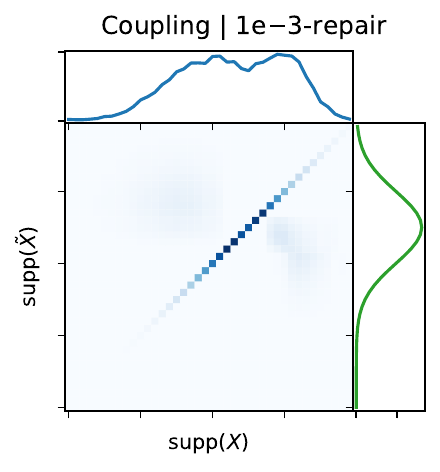}
\label{fig:prob1_6_3}
\end{minipage}%
\begin{minipage}{0.24\textwidth}
\centering
\includegraphics[scale=0.5]{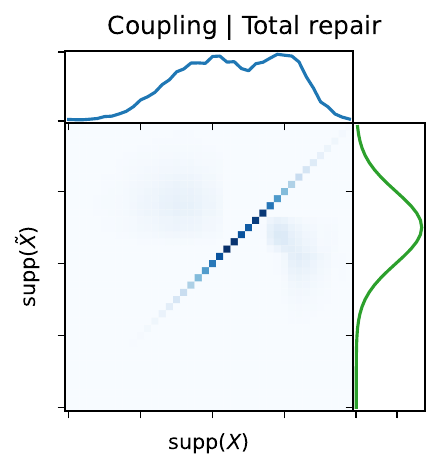}
\label{fig:prob1_6_4}
\end{minipage}
\caption[The couplings of our bias-repair schemes compared with the baseline]{From left to right: the baseline coupling of Equation~\eqref{equ:formulation-baseline}, 
the optimal coupling $\gamma^*$ of Equations~(\ref{equ:our-formulation-2}--\ref{equ:convex-sets}), when $\Lambda=10^{-2}\mathbb{1},10^{-3}\mathbb{1},\mathbb{0}$. The blue and green curves represent marginal distributions $P^{X},P^{\td{X}}$ that are the same across all couplings.}
\label{fig:exa_couplings}
\end{minipage}

\begin{minipage}{\textwidth}
\centering
\includegraphics[width=0.85\textwidth]{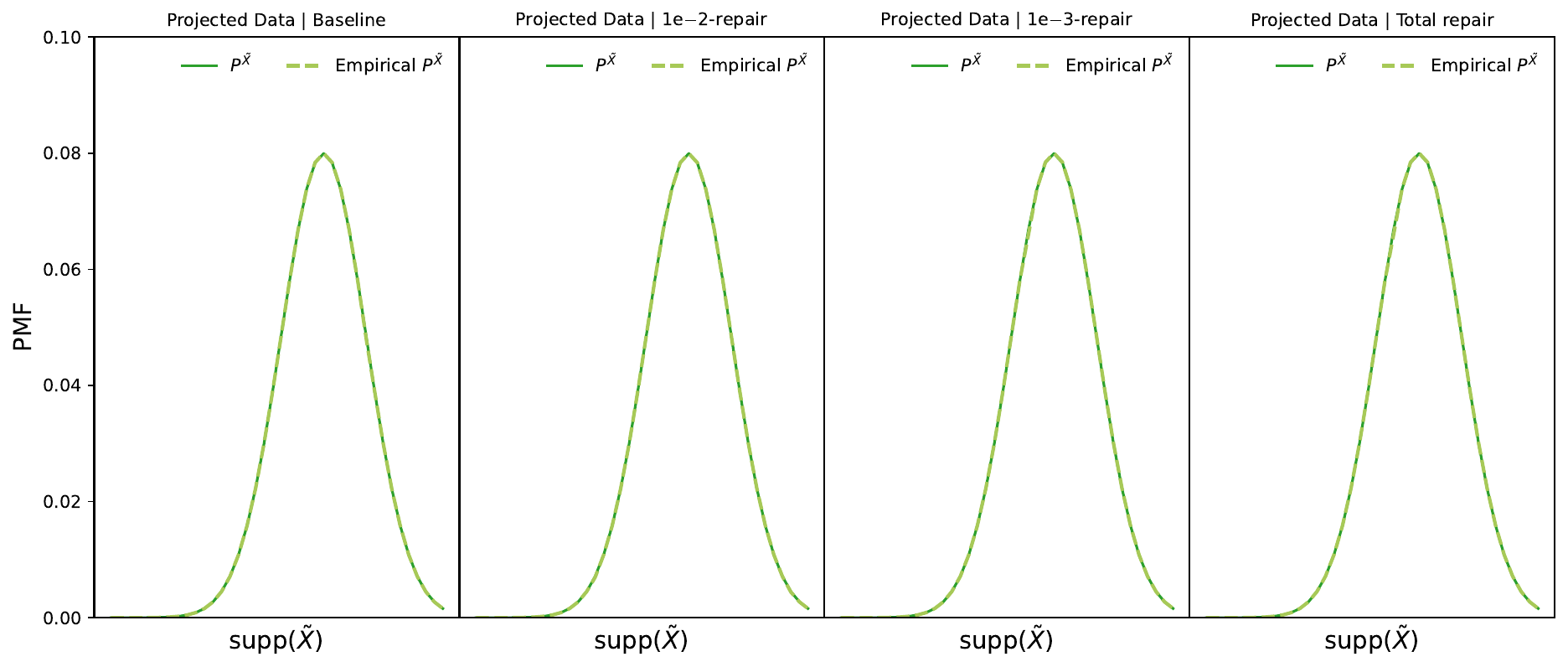}
\caption[The empirical group-blind distributions of our bias-repair schemes compared with the baseline]{\textbf{Group-blind distributions.}
Solid green curves are the $S$-blind target distributions $P^{\td{X}}$ used to compute couplings, which are the green curves in Figure~\ref{fig:exa_couplings}.
Dashed green curves (from left to right) are the $S$-blind empirical distributions of $P^{\td{X}}$ computed from projected data from baseline, from $10^{-2}\mathbb{1}$-repair, from $10^{-3}\mathbb{1}$-repair, and from total repair.
The overlap between dashed curves and solid curves shows the projected data follow the target distribution we design, and verifies that all couplings in Figure~\ref{fig:exa_couplings} are feasible and the projection method in Definition~\ref{def:projection} is correct.}
\label{fig:exa_results_groupbline}
\end{minipage}
\end{figure}

Compute $V=(P^{X_{s_0}}-P^{X_{s_1}})/P^{X}$, set the entry of cost matrix $C_{i,j}=|i-j|$, $\epsilon=0.01$, $\varepsilon=1e^{-4}$ and the number of iteration $K=400$ for baseline and  $K=600$ for our method. Along with $P^X,P^{\td{X}},V,\epsilon,C,K$ and $\Lambda=10^{-2}\mathbb{1},10^{-3}\mathbb{1},\mathbb{0}$ as input in Algorithm~\ref{alg:Dykstra}, we find the optimal coupling $\gamma^*$ of Equations~(\ref{equ:our-formulation-2}--\ref{equ:convex-sets}) 
and another baseline coupling of Equation~\eqref{equ:formulation-baseline}, shown in Figure~\ref{fig:exa_couplings}, where the blue and green curves represent $P^{X}$ and $P^{\td{X}}$. 

\begin{figure}[htp]
\centering
\includegraphics[width=0.85\textwidth]{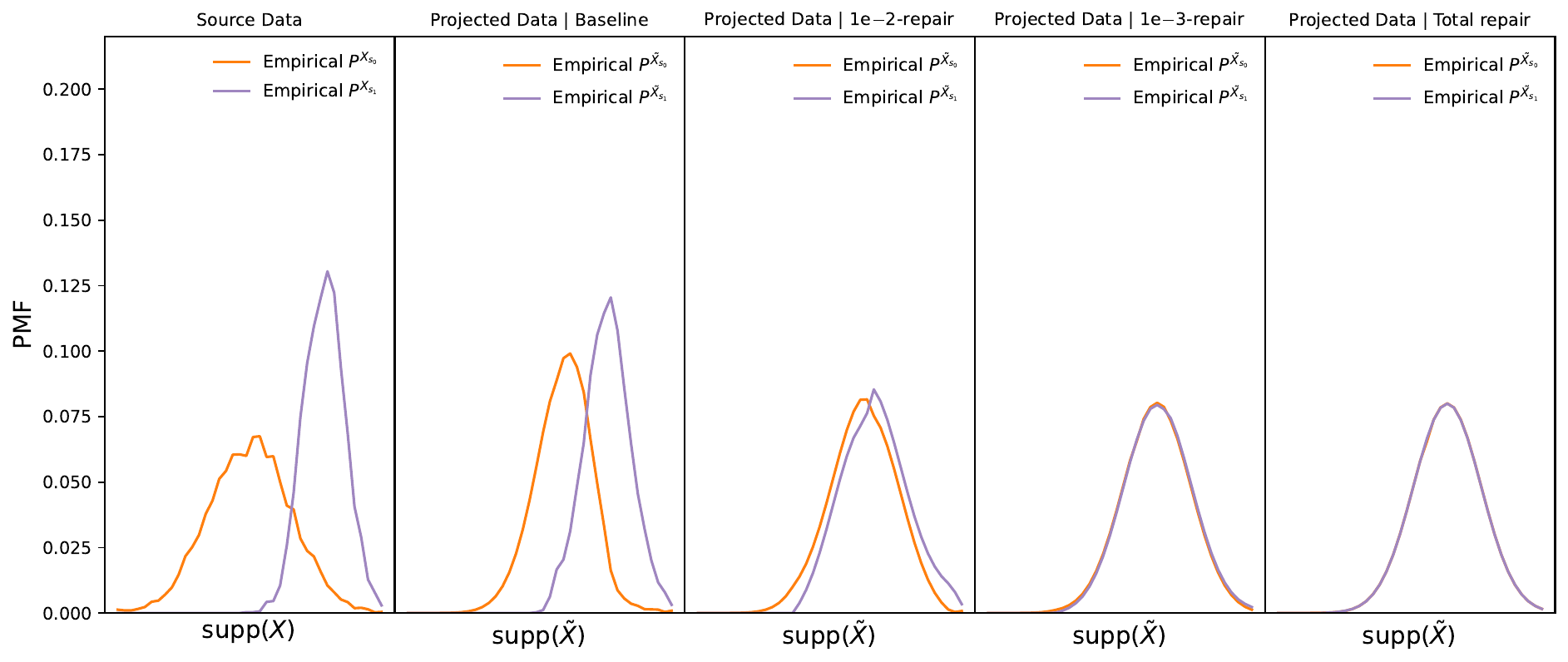}
\caption[The empirical group-wise distributions of our bias-repair schemes compared with the baseline]{\textbf{Group-wise distributions.} From left to right: the $S$-wise empirical distributions of $P^{X_{s_0}},P^{X_{s_1}}$ in source data, $P^{\td{X}_{s_0}},P^{\td{X}_{s_1}}$ in  projected data from baseline, from $10^{-2}\mathbb{1}$-repair, from $10^{-3}\mathbb{1}$-repair, and from total repair. $P^{\td{X}_{s_0}}$ is plotted orange and $P^{\td{X}_{s_1}}$ is plotted purple.}
\label{fig:exa_results}
\end{figure}

Then, for each coupling, we define a projection map following Definition~\ref{def:projection}. 
After applying the projections to source data, we compute the empirical distributions of $P^{\td{X}_{s_0}}$ and $P^{\td{X}_{s_1}}$ from the projected data. 
In Figure~\ref{fig:exa_results}, from left to right, we show the $S$-wise empirical distributions of source data, projected data from baseline, from $10^{-2}\mathbb{1}$-repair, from $10^{-3}\mathbb{1}$-repair, and from total repair, with $P^{\td{X}_{s_0}}$ plotted orange and $P^{\td{X}_{s_1}}$ plotted purple.
We can see that when $\Lambda$ gets closer to $\mathbb{0}$, the gap between purple and orange curves shrinks.

In order to demonstrate that all couplings in Figure~\ref{fig:exa_couplings} are feasible, we present the empirical $P^{\td{X}}$ (dashed green curves) computed from projected data and the input of target distribution $P^{\td{X}}$ used to compute all couplings (solid green curves) in Figure~\ref{fig:exa_results_groupbline}. The overlap of both green curves implies that source data are successfully projected to the target data we expect, regardless of the coupling used.



\subsection{Bias Repair vs. Data Distortion}\label{sec:trade-off}

We conduct random-forest classification models on the Adult Census Income dataset \citep{misc_adult_2}, to showcase the trade-off between bias repair and data distortion, compared with the total repair using barycentre projection in \cite{gordaliza2019obtaining}.

The Adult Census Income dataset comprises of 48842 samples of 14 features (e.g., gender, race, age, education level, marital-status, occupation) and a high-income indicator.
The high-income indicator denotes whether the annual income of a sample is lower than \$50K (label $Y=0$) or higher (label $Y=1$).
There are five numerical features out of these 14 features:
``education level'' (ranging from 1 to 16) and ``hours per week'', ``age'', ``capital gain'', ``capital loss'' (ranging from 0 to 4).
There are two commonly-recognised sensitive attributes, i.e., ``race'' and ``sex''.

If the sensitive attribute $S$ is ``race'', we select all samples ($M=46447$) whose race is black ($S=0$) or white ($S=1$).
If $S$ is ``sex'', we select all samples ($M=48842$) whose gender is female ($S=0$) or male ($S=1$).
For each numerical feature, we compute the total variation distance between race-wise (resp. gender-wise) marginal distributions in Table~\ref{tab:Race-wise TV}. They are used as reference to decide which features need to be adjusted, to avoid unnecessary alteration. 
We let $X$ include all features with $S$-wise TV distance higher than $0.08$, ensuring that only two features are selected when $S$ represents ``race'' or ``sex''.
Specifically, $X$ contains ``education level'', ``hours per week'' if $S$ is ``race'', and $X$ contains the ``hours per week'', ``age'' if $S$ is ``sex''. The other features are considered as $S$-neutral and denoted as $U$.
\begin{table}[!htp]
\centering
\begin{tabular}{|l|c|c|}
\hline
\textbf{Features} & \textbf{Race-wise TV distance}&\textbf{Gender-wise TV distance}  \\\hline
age              & 0.0415&\textbf{0.1010}\\
education level    & \textbf{0.1187}&0.0710\\
capital gain      & 0.0268&0.0369\\
capital loss      & 0.0142&0.0201\\
hours per week     & \textbf{0.1222}&\textbf{0.1819}\\\hline
\end{tabular}
\caption[The total variation distance between race-wise and gender-wise marginal distributions of five numerical features in the Adult Census Income dataset]{The TV distance between race-wise and gender-wise marginal distributions of five numerical features in the Adult Census Income dataset \citep{misc_adult_2}. Highlighted features, whose $S$-wise TV distance higher than $0.08$ are chosen as $X$ and would be adjusted. The rest are considered as $S$-neutral and denoted as $U$.}
\label{tab:Race-wise TV}
\end{table}

We have performed $30$-fold cross-validation. 
For each trial, we randomly select $60\%$ of $M$ samples as the training set, and use these five numerical features and the high income indicator $Y$ to build a random-forest model $\mathcal{M}(X,U)$, with the sensitive attribute $S$ ignored. The outputs of the model $\hat{Y}=\mathcal{M}(X,U)$ is the prediction of $Y$.

The rest $40\%$ of $M$ samples are used as the test set.
From the test set, we compute the empirical marginal distributions of the feature $X$ (i.e., $P^X,P^{X_0},P^{X_1}$) and $V$.
Next, we handle the test set in five different ways as outlined below.
\begin{itemize}
\item Do nothing (denoted ``Origin''). 
\item For each sample in test set, project its $X$ features with a baseline coupling computed as in Definition~\ref{def:baseline} (denoted ``Baseline''). The other features and $Y$ stay the same.
\item Project its $X$ features with a barycentre coupling computed as in Definition~\ref{def:barycentre} (denoted ``Barycentre''). Features $U$ and the label $Y$ stay the same.
\item Project its $X$ features with a coupling computed by Algorithm~\ref{alg:Dykstra} with $\Lambda=1e^{-2}\mathbb{1}$ (denoted ``$1e^{-2}$-repair''). Features $U$ and the label $Y$ stay the same.
\item Project its $X$ features with a coupling computed by Algorithm~\ref{alg:Dykstra} with $\Lambda=1e^{-3}\mathbb{1}$ (denoted ``$1e^{-3}$-repair''). Features $U$ and the label $Y$ stay the same.
\end{itemize}

The number of iterations is $K=400$ for ``Baseline'', ``Barycentre'', and is $K=600$ for our partial repair schemes, i.e., ``$1e^{-2}$-repair'', ``$1e^{-3}$-repair''.
The entropic regularisation parameter is $\epsilon=0.01$. The very small number is $\varepsilon=1e^{-5}$.
The entry of cost matrix is $C_{i,j}:=\|g\odot(i-j)\|_1$, where the entry $g_i$ is the reciprocal of the range of the $i^{th}$ feature in $X$.
Both source and target distributions are $P^{X}$ in ``Basline'' and our partial repair schemes ``$1e^{-2}$-repair'', ``$1e^{-3}$-repair'', such that the target distribution is simply $P^{\td{X}}=P^X$ to avoid unnecessary data distortion.
``Barycentre'' use $P^{X_{s_0}}$ as source and $P^{X_{s_1}}$ as target, with $\pi_0$ being the portion of $s_0$ group in source data, the same setting as in \cite{gordaliza2019obtaining}.

We test the five test sets on a random-forest classification model built from the training set as follows.
Given a sample $(x,u,s,y)$ in the test set, for ``Origin'', the prediction of its high income indicator is $\hat{Y}=\mathcal{M}(x,u)$.
With exception of ``Origin'', this sample is split into a sequence of weighted samples $\{(\tx,u,s,y,w_{\tx})\}_{\tx\in\supp{\td{X}}}$. In this case, the repaired prediction would be $0$ if $\sum_{\tx\in\supp{\td{X}}} w_{\tx} \hat{Y}_{\tx}<\hat{Y}^{th}$ and $1$ otherwise, where the prediction for a single sample is $\hat{Y}_{\tx}=\mathcal{M}(\tx,u)$, and the threshold is $\hat{Y}^{th}=0.1$, chosen by grid search.

Once the predictions for the five test sets are computed, we measure the performance by disparate impact, f1 scores and $S$-wise TV distance, as in Definition~\ref{def:performance}. Note that for ``Origin'', the $S$-wise TV distance is $\tv{P^{X_{s_0}}}{P^{X_{s_1}}}$.
After $30$ trials, where in each trial we generate new training set and test set, we can further compute the mean and standard deviation of these performance indices of each model.

\begin{figure}[!htp]
\centering
\begin{minipage}{.8\textwidth}
\centering
\includegraphics[width=0.8\textwidth]{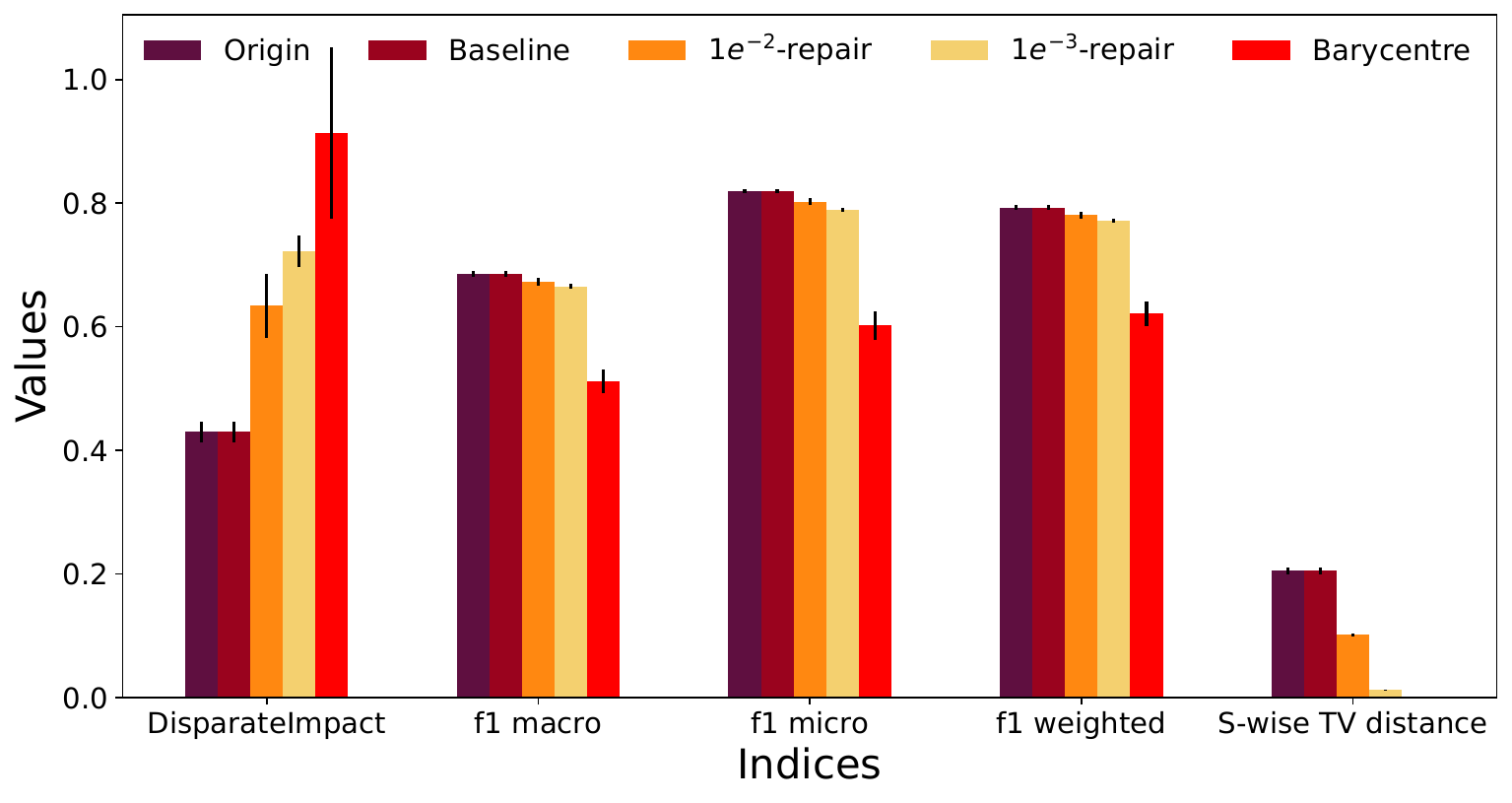}
\end{minipage}%

\begin{minipage}{0.8\textwidth}
\centering
\includegraphics[width=0.8\textwidth]{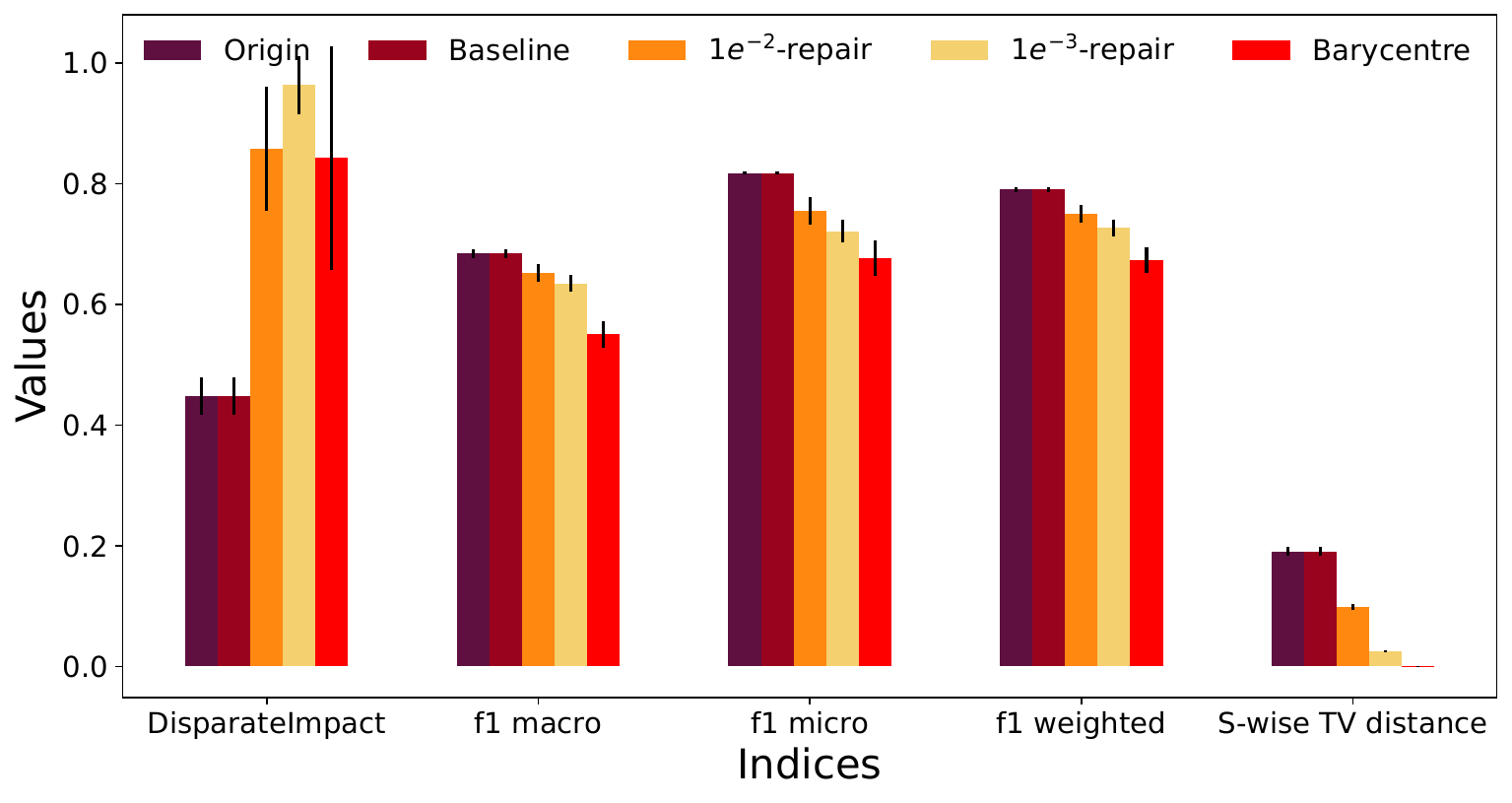}
\end{minipage}
\caption[Prediction performance of our bias-repair schemes compared against baselines, for the Adult Census Income dataset]{Prediction performance for the Adult Census Income dataset \citep{misc_adult_2}, with $S$ being ``sex'' (upper) and ``race'' (lower). The mean and mean $\pm$ one standard deviation of these performance indices, across $30$ trials are displayed by bars and dark vertical lines. `Origin'' method without any bias repair is plotted in brown, ``Baseline'' is plotted in dark red, ``Barycentre'' in \cite{gordaliza2019obtaining} is plotted in light red. Our partial repair schemes ``$1e^{-2}$-repair'' is plotted in orange and ``$1e^{-3}$-repair'' is plotted in yellow. 
An ideal method should have the disparate impact close to $1$, f1 scores as high as possible and $S$-wise TV distance close to $0$.
}
\label{fig:adult_S}
\end{figure}

In Figure~\ref{fig:adult_S}, we present disparate impact, f1 scores and $S$-wise TV distance of the prediction for these five test sets as the mean (bars) and mean $\pm$ one standard deviation (dark vertical lines) across the $30$ trials. To distinguish the test sets, ``Origin'' is plotted in brown,  ``Baseline'' is plotted in dark red, ``$1e^{-2}$-repair'' is plotted in orange, ``$1e^{-3}$-repair'' is plotted in yellow and ``Barycentre'' is plotted in light red.
Note that an ideal method should have a disparate impact close to 1, F1 scores as high as possible, and the $S$-wise TV distance close to 0. 

\textcolor{black}{
Figure~\ref{fig:adult_S} shows that our methods ("$1e^{-2}$-repair", "$1e^{-3}$-repair") significantly improve fairness performance with very little sacrifice in prediction performance.
Notably, it is expected that the $S$-wise TV distance of "Barycentre" is nearly zero, as it is one of the total repair schemes in \cite{gordaliza2019obtaining}. However, despite the "Barycentre" method utilising the group membership (i.e., the sensitive attribute) of each datapoint and achieving very similar projected data for both groups (with the S-wise TV distance nearly zero), it results in a significant decrease in prediction performance. This is likely because the target distribution of the "Barycentre" method (i.e., a barycentric distribution) differs from the distribution of the training data used by the pre-trained random forest model.
}

\section{Conclusion}
\textcolor{black}{
We have provided a new bias-repair framework to mitigate disparate impact, without access to each datapoint's sensitive attribute, via the technologies of \acrshort{OT}.
To the best of our knowledge, this is the first work that entirely removes the necessity for each datapoint's sensitive attribute while ensuring the upper bound of distance between the projected data of both groups, which are divided by a binary sensitive attribute.
}

\textcolor{black}{
The contribution is two-fold. First, we design a novel algorithm to compute a group-blind coupling that can project two distinct datasets into similar ones, including its relaxed versions, with a bounded distance between the projected data of both groups using generalised \acrshort{OT}. The computation of such a coupling only requires the distributions of each group, and since this coupling is group-blind, its usage does not necessitate each datapoint's group membership. 
Then, due to these properties, we build a bias-repair framework based on this new algorithm to address the problem of unavailable sensitive attributes in machine learning fairness.
}

The limitation is that only one binary attribute is allowed.
We could further extend these schemes to cases where the sensitive attribute has multiple classes.
Alternatively, we can consider the accelerated Dykstra’s algorithm \citep{chai2022self}.
\cleardoublepage
\addtocontents{toc}{\protect\setcounter{tocdepth}{0}}
\chapter{Conclusion}
\label{cha:conclusions}

Ensuring fairness is one of the most significant challenges within the domain of \acrshort{AI}.
This thesis begins by recognising the urgent demand for fairness in AI and the ongoing efforts of governments and institutions to ensure fairness across various AI applications.

Within academia, the field of machine learning fairness has emerged in response to this challenge. The central question is how to formulate fairness. There is not a single correct or universal answer, as it strongly depends on societal opinions and significantly impacts people's lives. What was considered fair in the past may not be accepted today, and what is acceptable in one context may not work in another. The pursuit of fairness is an ongoing process, as people's views on fairness continuously evolve and never fully converge.

Thereby, multiple fairness formulations have been developed over time, each within a specific context. With these formulations in hand, how should they be implemented? What types of fairness can be reliably achieved? How can fairness be maintained when practical circumstances differ from ideal conditions? The thesis then provides a comprehensive overview of the primary methodologies developed for machine learning fairness, identifying key open questions and commonly adopted strategies.

The remainder of the thesis addresses:
\begin{itemize}
    \item the prevalent utilisation of black-box models in AI, posing challenges in comprehending their effects, particularly in high-stake applications such as school admissions;
    \item the discriminatory or biased data commonly used in training these black-box models.
\end{itemize}

To address these concerns, two frameworks have been formalised.

\section{Fairness-Aware Optimisation}

Optimisation plays a crucial role in various scientific domains, encompassing machine learning, \acrshort{AI}, and economics.
Contemporary practice of computational optimisation is, however, too often divorced from its classical mathematical roots.
Practitioners often rely on black-box algorithms. However, the lack of interpretability and transparency of these models can impede efforts to diagnose and rectify errors, as well as optimise the model's performance.
The initial framework, spanning Chapters~\ref{cha:tac} through \ref{cha:jair}, aims to bridge this disciplinary gap by integrating \acrshort{NCPOP} (cf. Appendix~\ref{cha:ncpop}) into machine learning applications, while developing algorithms to ensure fairness in high-stack applications.


In Chapters~\ref{cha:tac}, we propose a new algorithm for system identification of \acrshort{LDS}, utilising \acrshort{NCPOP} and min-max objectives in working with time-series problems, without assumptions on the dimension of the hidden state. 
As discussed in Chapter~\ref{cha:overview-MLfairness}, most commonly used fairness concepts are linear or convex formulations. This new algorithm can accommodate a variety of ``polynomial-like'' objectives and constraints, thereby broadening the types of fairness formulations that can be achieved with guarantees.

In Chapter~\ref{cha:jair}, we introduce two natural notions of fairness in forecasting, as an extension of the method in Chapter~\ref{cha:tac}.
When the corresponding optimisation problems are solved to global optimality, we show that this produces the present-best results on the \acrshort{COMPAS} benchmark dataset.

\color{black}

\section{Bias-Repair without Demographics}

The framework, introduced in Chapter~\ref{cha:ot}, aims to adjust input data from both privileged and unprivileged groups to align the feature distributions of the two groups. It incorporates lower bounds for the effects and guarantees convergence for computation.

In this framework, we propose a group-blind optimal transport method that maps two distinct datasets into similar ones. After this projection, both groups will have similar input data, resulting in similar outputs from AI systems. 
Importantly, individual-level demographics (i.e., each data point's sensitive attribute) are not required in these fairness-repair schemes. This group-blind projection map is computed via generalised \acrshort{OT}, and its results are illustrated on the Adult Census Income dataset.

One limitation is that only one binary sensitive attribute is allowed, but there is potential to further extend these schemes to cases where the sensitive attribute has multiple classes. Additionally, we could explore relaxing the requirements of group-wise feature distributions (e.g., exam score distributions for rich and poor students) in these schemes, by evaluating the effects when only distributions with noise are provided. Alternatively, we can explore how to utilise scenarios where some individual-level demographics are observed, while certain groups of members are more likely to disclose their demographics.

\section{Future Work and Outlook}

In the context of non-deterministic decision-making systems, fairness and randomness are somehow related \citep{jaeger2009fairness}. This relationship may involve incorporating randomness into the decision-making process to prevent systematic discrimination against certain groups. Hence, there are potential avenues for exploring ``entropy-like'' fairness definitions, such as maximum entropy voting systems \citep{sewell2009probabilistic}, for future investigations.


\cleardoublepage
\begin{appendices}
\chapter{Notation Lists}
\label{app:notation}


\begin{table}[htp]
\centering
\begin{tabular}{llc}
\hline 
$n$ & the number of scalar variables or n.c. variables & p. \pageref{cha:mom_app} \& \pageref{cha:ncpop}\\
$d$ & the moment degree& p. \pageref{def:monomial_basis}\\
$\mathbf{x}$                                        & a $n$-tuple of commutative variables & p. \pageref{cha:mom_app} \\
$\mathbf{X}$                                        & a $n$-tuple of non-commutative variables & p. \pageref{cha:ncpop} \\
$\omega,\nu,u,z$ & monomials in $\mathbf{x}$ or $\mathbf{X}$   & Definition~\ref{def:monomial_basis} \& p. \pageref{def:monomial_basis_nc} \\
$|\omega|$ & the degree of $\omega$ & Definition~\ref{def:monomial_basis} \\
$\mathbf{W},\mathbf{W}_d$                           & a (truncated) monomial basis & Definition~\ref{def:monomial_basis} \&~\ref{def:monomial_basis_nc} \\
$\sigma(d)$  & cardinality of $\mathbf{W}_d$ in commutative case   & Definition~\ref{def:monomial_basis} \\
$\sigma_{nc}(d)$  & cardinality of $\mathbf{W}_d$ in non-commutative case  & Definition~\ref{def:monomial_basis_nc} \\
$\overrightarrow{\mathbf{W}_d}$                     & a column vector of all elements in $\mathbf{W}_d$ & p. \pageref{def:monomial_basis} \\
$\mathbb{R}[\mathbf{x}]$     & the ring of polynomials in $\mathbf{x} \in \mathbb{R}^n$ & p. \pageref{cha:mom_app}   \\
$\mathbb{R}[\mathbf{x}]_d$ & the ring of polynomials with degree at most $d$ & p. \pageref{def:monomial_basis} \\
$\mathbb{R}[\mathbf{X}]$     & the ring of n.c. polynomials in $\mathbf{X}$  & p. \pageref{sec:ncpop} \\
$\Sym\mathbb{R}[\mathbf{X}]$  & all symmetric elements in $\mathbb{R}[\mathbf{X}]$ & \eqref{equ:SymR[X]} \\
$f,g,g_j,h_j$  & real polynomials & p. \pageref{def:monomial_basis} \& \pageref{def:semi-algebraic}\\
$\deg(f)$ & the polynomial degree of $f$ & p. \pageref{def:loca_matrix}\\
$\overrightarrow{f}$                                & a column vector consisting of coefficients of $f$  & p. \pageref{def:monomial_basis} \\
$\mathbf{g}$ & a finite subset of polynomials in $\mathbb{R}[\mathbf{x}]$ or $\Sym\mathbb{R}[\mathbf{X}]$ & Definition~\ref{def:semi-algebraic} \&~\ref{def:semi-algebraic-nc} \\
$\mathscr{D}_{\mathbf{g}}$                          & a semi-algebraic positive domain & Definition~\ref{def:semi-algebraic} \&~\ref{def:semi-algebraic-nc} \\
$\Sigma[\mathbf{x}],\Sigma[\mathbf{X}]$             & the space of SOS \& SOHS polynomials                         & Definition~\ref{def:sos} \&~\ref{def:quadratic_module_nc} \\
$Q(\mathbf{g})$                                     & the quadratic module & Definition~\ref{def:quadratic_module} \&~\ref{def:quadratic_module_nc} \\
$y_{\omega}$                                        & a moment associated with a monomial $\omega$  & Definition~\ref{def:sequence_y} \\
$\mathbf{y}$                                        & a (truncated) moment sequence                                 & Definition~\ref{def:sequence_y} \\
$L_{\mathbf{y}},L$                                  & a linear functional associated with $\mathbf{y}$                & Definition~\ref{def:functional_L} \\
$\mathbf{M}_d(\mathbf{y}),\mathbf{M}_d$             & moment matrices associated with $\mathbf{y}$                  & Definition~\ref{def:moment_matrix} \&~\ref{def:moment_matrix_nc} \\
$\mathbf{M}_d(g\mathbf{y}),\mathbf{M}_d(g)$         & localising matrices associated with $\mathbf{y}$ \& $g$      &Definition~\ref{def:loca_matrix} \&~\ref{def:moment_matrix_nc}\\\hline
\end{tabular}
\caption{Common notation in Appendices~\ref{cha:mom_app} \& \ref{cha:ncpop}}
\end{table}


\begin{table}[htp]
\centering
\begin{tabular}{llc}
\hline 
$\Gamma$ & a set of indices &  p. \pageref{equ:GMP} \\
$\mathscr{M}_+(\mathscr{D})$ & the space of finite positive Borel measures on $\mathscr{D}$  & p. \pageref{equ:GMP}\\
$\delta_{\mathbf{x}}$ & the Dirac measure at a point $\mathbf{x}$ & p. \pageref{the:optimality}\\
$\rho_{mom}$ & optimum of GMP &  \eqref{equ:GMP}, p. \pageref{equ:GMP} \\
$\rho_{pop}$ & optimum of POP  &  \eqref{equ:GMP-simplex}\\
$\rho_d$ & optimum of SDP relaxation at degree $d$ &  \eqref{equ:GMP-prime-sdp}\\
\hline
\end{tabular}
\caption{Notation exclusive for moment problems in Appendix~\ref{sec:GMP}}
\end{table}

\begin{table}[htp]
\centering
\begin{tabular}{llc}
\hline
$\mathbb{S}^{n}$ & the space of $n$-tuple symmetric matrices & p. \pageref{equ:SymR[X]} \\
$\Sym\mathbb{R}[\mathbf{X}]$ & symmetric elements in $\mathbb{R}[\mathbf{X}]$                   &  \eqref{equ:SymR[X]}               \\
$\mathscr{D}^{\II}_{\mathbf{g}}$  & von Neumann semi-algebraic positive domain & \makecell[c]{
Definitions~1.59 of\\ \cite{burgdorf2016optimization} }  \\
$\stackrel{\cyc}{\sim}$  & cyclic equivalence  & Definition~\ref{def:cyclic_equivalence}    \\
$\Theta(\mathbf{g}),\Theta_{d}(\mathbf{g})$  & (truncated) cyclic quadratic modules  & Definition~\ref{def:cyclic quadratic_modules} \\
$\left(\Theta_{d}(\mathbf{g})\right)^{\vee}$ & the dual cone  & Definition~\ref{def:dual cones}\\
$\tr_{\min}(f,\mathbf{g})$& trace minimum & \eqref{trace-NCPO}\\
$\tr^{\II}_{\min}(f,\mathbf{g})$ & optimum on von Neumann semi-algebraic set & \eqref{equ:trace-NCPO-con-sup}, p. \pageref{equ:trace-NCPO-con-sup}\\ 
$\tr_{\Theta_{d}}(f,\mathbf{g})$ & optimum of primal SDP relaxation & \eqref{equ:trace-NCPO-sup-Theta}, p. \pageref{equ:trace-NCPO-sup-Theta}\\ 
$\trd_{\Theta_{d}}(f,\mathbf{g})$ & optimum of dual SDP relaxation& \eqref{equ:trace-NCPO-dual}, p. \pageref{equ:trace-NCPO-dual}\\\hline
\end{tabular}
\caption{Notation exclusive for trace minimisation in Appendix~\ref{sec:ncpop}}
\end{table}

\begin{table}[htp]
\centering
\begin{tabular}{llc}
\hline 
$I_1,\dots,I_p$ & subsets of indices of variables & Assumption~\ref{ass:rip}\\
$\mathbf{X}(I_q)$ & a subset of variables $\mathbf{X}$ & Assumption~\ref{ass:rip}\\
$J_1,\dots,J_p$ & a partition of indices of $\mathbf{g}$ & Assumption~\ref{ass:rip}\\
$Q^q(\mathbf{g}),Q^q_{d}(\mathbf{g})$  & (truncated) quadratic modules for $\mathbf{X}(I_q)$ &  \eqref{equ:truncated_quadratic_sparse} \\
$\Theta^q(\mathbf{g}),\Theta^q_{d}(\mathbf{g})$  & (truncated) cyclic quadratic modules for $\mathbf{X}(I_q)$ &  \eqref{equ:truncated_cyc_quadratic_sparse} \\
\makecell[l]{
$\Theta^{sparse}(\mathbf{g})$\\$\Theta^{sparse}_d(\mathbf{g})$}  & (truncated) sparse cyclic quadratic modules & Theorem~\ref{the:positivetrace2cyclic_sparse}\\
\hline
\end{tabular}
\caption{Notation exclusive for sparse representation in Appendix~\ref{sec:sparse_ncpop}}
\end{table}


\begin{table}[htp]
\centering
\begin{tabular}{llc}
\hline
$t$ & the time step & \eqref{equ:LDS}, p. \pageref{equ:LDS} \\
$k$ & the actual dimension of hidden states & \eqref{equ:LDS}, p. \pageref{equ:LDS} \\
$T$ & the length of the time window & \multicolumn{1}{c}{Section~\ref{sec:settings}} \\
$d$ & the moment degree & \eqref{equ:LDS}, p. \pageref{equ:LDS}\\
$G,F,V,W$ & system matrices of a LDS & \eqref{equ:LDS}, p. \pageref{equ:LDS} \\
$Y_t,\hat{Y}_t$ & the (estimated) observation/ output at time $t$ & \eqref{equ:LDS}, p. \pageref{equ:LDS} \& (\ref{NFF_1}--\ref{NFF_2}) \\
$\theta_t,\hat{\theta}_t$ & the (estimated) hidden state at time $t$ & \eqref{equ:LDS}, p. \pageref{equ:LDS} \& (\ref{NFF_1}--\ref{NFF_2}) \\
$\omega_t$ & the (estimated) process noise at time $t$ & \eqref{equ:LDS}, p. \pageref{equ:LDS} \& (\ref{NFF_1}--\ref{NFF_2})  \\
$\nu_t$ & the (estimated) observation noise at time $t$ & \eqref{equ:LDS}, p. \pageref{equ:LDS} \& (\ref{NFF_1}--\ref{NFF_2})  \\
$\lambda_1,\lambda_2,\lambda_3$ & multipliers & \multicolumn{1}{c}{\eqref{obj_tac}} \\\hline
\end{tabular}
\caption{Common notation in Chapters~\ref{cha:tac} \& \ref{cha:jair}}
\end{table}

\begin{table}[htp]
\centering
\begin{tabular}{llc}
\hline
nrmse & normalised root mean squared error fitness value & \multicolumn{1}{c}{\eqref{NRMSE}} \\
$\hat{k}$ & the estimated dimension of hidden states & \multicolumn{1}{c}{Section~\ref{sec:settings}} \\\hline
\end{tabular}
\caption{Notation exclusive for Chapter~\ref{cha:tac}}
\end{table}

\begin{table}[htp]
\centering
\begin{tabular}{llc}
\hline
$s$ & the index of a subgroup & p. \pageref{page:define-s} \\
$\mathcal{S}$ & the set of subgroups & p. \pageref{page:define-s}\\
$\mathcal{I}^{(s)}$ & the set of trajectories in subgroup $s$ & p. \pageref{page:define-s}\\
$\mathcal{T}^{(i,s)}$ & \makecell[l]{the set comprises all time indices during which\\ observations in trajectory $i\in\mathcal{I}^{(s)}$ are available.} & p. \pageref{page:define-s}\\
$\mathcal{T}^+$ & $\cup_{i\in\mathcal{I}^{(s)},s\in\mathcal{S}}\mathcal{T}^{(i,s)}$ & p. \pageref{equ:loss}\\
$\mathcal{L}^{(s)}$ & the model of subgroup $s$ &  p. \pageref{page:define-s} \\
$\mathcal{L}$ & a subgroup-blind model & p. \pageref{page:define-L}\\
$Y^{(i,s)}_t$ & the observation of the trajectory $i\in\mathcal{I}^{(s)}$ at time $t$ & p. \pageref{page:define-s}\\
$\loss^{(i,s)}(\hat{Y}_t)$ & the loss function& \eqref{equ:loss} \\
$a$ & an auxiliary scalar variable & p. \pageref{page:define-a}\\
$\mathcal{\beta}^{(s)}$ & used to adjust the degree of bias & p. \pageref{page:define-beta} \\
$\mathrm{\nrmse^{(s)}}$ & the nrmse fitness value for subgroup $s$ & \eqref{equ:NRMSE_s} \\
$p$ & \makecell[l]{used to define a threshold as \\$p^{\textrm{th}}$ percentile of all outputted scores} & p. \pageref{page:define-yhat} \\
$\hat{y}^{(i,s)}$ & the outputted score & p. \pageref{page:define-yhat} \\
$A^{(s)},X^{(i,s)},e^{(s)}$ & coefficients, explanatory variables and noise& \eqref{equ:post-process-constraints} \\
IND(rw) & unfairness metrics of independence (p. \pageref{sec:group-fairness})  & p. \pageref{equ:indices} \\
SP(rw) & unfairness metrics of separation (p. \pageref{sec:group-fairness}) & p. \pageref{equ:indices} \\
SF(rw) & unfairness metrics of sufficiency (p. \pageref{sec:group-fairness}) & p. \pageref{equ:indices} \\
INA(rw) & inaccuracy metrics & p. \pageref{equ:indices}\\\hline
\end{tabular}
\caption{Notation exclusive for Chapter~\ref{cha:jair}}
\end{table}

\clearpage
\begin{table}[htp]
\centering
\begin{tabular}{llc}
\hline
$N$ & the number of discretisation points & \eqref{equ:simplex} \\
$M$ & the number of samples & p. \pageref{page:define-M}\\
$K$ & the maximal number of iterations & p. \pageref{page:define-dykstra}\\
$C$ & the $N\times N$ cost matrix & p. \pageref{equ:entropic-reg-OT-define} \\
$\Delta_N$ & probability simplex in $\mathbb{R}^N$ & \eqref{equ:simplex} \\
$\Pi(P,Q)$ & the set of couplings 
between $P,Q\in\Delta_N$ in $\mathbb{R}^{N\times N}$ & \eqref{equ:Pi-define} \\
$\gamma,\gamma^*,\gamma^{(k)}$ & couplings in $\Pi(P,Q)$ & \eqref{equ:Pi-define} \\
$\mathbb{1},\mathbb{0}$ & the $N$-dimensional column vector of ones and zeros & \eqref{equ:Pi-define} \& \eqref{equ:difference2rv}\\
$E(\gamma)$ & the entropy of a coupling & \eqref{equ:entropy-define} \\
$\xi$ & a positive $N\times N$ reference matrix& \eqref{equ:kl-define} \\
$\kl{\gamma}{\xi}$ & Kullback-Leibler divergence between $\gamma$ and $\xi$ & \eqref{equ:kl-define} \\
$\prox^{KL}_{\mathcal{C}}(\xi)$ & Kullback-Leibler projection of $\xi$ to a convex set $\mathcal{C}$ & \eqref{equ:kl-projec-define} \\
$\epsilon>0$ & the entropic regularisation parameter & \eqref{equ:entropic-reg-OT-define}\\
$W_{\epsilon}(P,Q)$ & the objective value of entropic regularisation OT & \eqref{equ:entropic-reg-OT-define}\\
diag & the diagonal operator & p. \pageref{page:define-diag} \\
supp & the support of a random variable & Definition~\ref{def:sensitive_attribute} \\
$S$ & a sensitive attribute & Definition~\ref{def:sensitive_attribute} \\
$X,X_s$ & group-blind/ group-wise source variable & Definition~\ref{def:source_var} \\
$\td{X},\td{X}_{s}$ & group-blind/ group-wise target variable & Definition~\ref{def:target_var} \\
$P^{X},P^{X_s},P^{\td{X}},P^{\td{X}_s}$ & probability distributions of variables $X,X_s,\td{X},\td{X}_s$ & Definitions~\ref{def:source_var}--\ref{def:target_var} \\
$\mathscr{T}$ & projection induced from a coupling & Definition~\ref{def:projection} \\
$w_{i,j},w_{i},w_{j}$ & weights of original/ projected samples & Definition~\ref{def:projection} \\
$\tv{P}{Q}$ & total variation distance between $P,Q\in\Delta_N$ & Definition~\ref{def:tvdistance} \\
$V$ & a vector defined via $\frac{P^{X_{s_0}}-P^{X_{s_1}}}{P^{X}}$ & Theorem~\ref{pro:binary_condition} \\
$\Lambda$ & a nonnegative vector & Lemma~\ref{lem:S-wise TV} \\
$\overline{\supp{X}}$ & $\{i\in\supp{X}\mid V_i\neq 0\}$ & Lemma~\ref{lem:S-wise TV} \\
$\mathcal{C}_1,\mathcal{C}_2,\mathcal{C}_3$ & convex sets & \eqref{equ:convex-sets} \\
$\{q_{k}\}_{k\geq 1}$ & an auxiliary sequence &Algorithm~\ref{alg:Dykstra} \\
$\varepsilon$ & a small number to prevent arithmetic underflow & Algorithm~\ref{alg:Dykstra} \\
$\iota_{\mathcal{C}}$ & the indicator function of a set $\mathcal{C}$ & \eqref{equ:indicator} \\
$\partial$ & subdifferential & \eqref{equ:subgradient-define} \\
$U$ & the group-neutral features & Section~\ref{sec:ot_main}.5 \\
$Y,\hat{Y}$ & the label and the estimated label & Section~\ref{sec:ot_experiments}.2 \\
$\mathcal{M}$ & the group-blind model used to estimate label & Section~\ref{sec:ot_experiments}.2 \\
$P^{B}$ & 1-Wasserstein barycentre & \eqref{equ:barycentre-define} \\
$\pi_0,\pi_1$ & barycentric coordinates & \eqref{equ:barycentre-define}
\\\hline
\end{tabular}
\caption{Notation in Chapter~\ref{cha:ot}}
\end{table}
\chapter{Semi-Definite Programming}
\label{app:sdp}


The following content is based on \cite{vandenberghe1996semidefinite}.
In semi-definite programming (SDP), one minimises a linear objective function subject to the constraint that an affine combination of symmetric matrices is positive semi-definite. 
The formulation reads:

\begin{equation}
\begin{split}
\min_{x\in\mathbb{R}^n} \quad & c' x\\
\textrm{s.t.} \quad & F(x)\succeq 0,
\end{split}\tag{Primal SDP}\label{app-sdp:prime}
\end{equation}

where $F(x):=F_0 + \sum_{i=1}^n x_i F_i$, and $F_0,\dots,F_n$ are symmetric matrices in $\mathbb{R}^{n\times n}$. $c$ is a vector in $\mathbb{R}^n$ with its transpose being $c'$. The inequality sign in $F(x)\succeq 0$ means that $F(x)$ is positive semi-definite.

\section{Duality}

Let us derive the dual formulation. 
The semi-definiteness of $F(x)$ implies that $F(x)$ is a symmetric matrix. Then, it could be decomposed into 
\begin{equation}
F(x)=Q' \Lambda Q,\label{app-sdp:decomposition}
\end{equation}
where $Q$ is an orthogonal matrix, i.e., $Q'=Q^{-1}$, and $\Lambda$ is a diagonal matrix whose entries are the eigenvalues of $F(x)$. 


The semi-definiteness of $F(x)$ is equivalent to the diagonal elements of $\Lambda$ being non-negative.
The Lagrangian function is defined as

\begin{align*}
c' x - \sum_{j=1}^n y_j(\Lambda_{jj}) &= c' x - \tr\left(\Lambda\diagg(y)\right) = c' x - \tr\left(Q F(x) Q'\diagg(y)\right) \\ 
&= c' x - \tr\left(F(x) Q'\diagg(y) Q\right) \\
&= c' x - \tr\left(F_0 Q'\diagg(y) Q\right) - \sum_{i=1}^n  x_i \tr\left(F_i Q'\diagg(y) Q\right)\\
&= \sum_{i=1}^n x_i\left(  c_i- \tr\left(F_i Q'\diagg(y) Q\right) \right) - \tr\left(F_0 Q'\diagg(y) Q\right),
\end{align*}

where $y_1,\dots,y_n\geq 0$ are Lagrangian multipliers. $\diagg(y)$ forms a diagonal matrix whose entries are $y$. $\tr(\cdot)$ denotes the trace.
The second equation uses Equation~\eqref{app-sdp:decomposition}. The third equation uses the cyclic property of trace. The forth equation uses the definition of $F(x)$.

Since $x\in\mathbb{R}^n$ are free variables, in order to maximise the Lagrangian function, the multipliers $y\in\mathbb{R}^n_+$ need to make sure $ c_i= \tr\left(F_i Q'\diagg(y) Q\right)$ for all $i=1,\dots,n$. Let $Z:=Q'\diagg(y) Q$ be a symmetric matrix, and obviously $Z\succeq 0$ because $Z$ is similar to $\diagg(y)$. Hence, the dual formulation reads:
\begin{equation}
\begin{split}
\max_{Z\in\mathbb{R}^{n\times n}} \quad & -\tr\left(F_0 Z\right)\\
\textrm{s.t.} \quad & \tr\left(F_i Z\right)=c_i,\forall i=1,\dots,n,\\
\quad & Z\succeq 0,
\end{split}\tag{Dual SDP}\label{app-sdp:dual}
\end{equation}

Next, we shows that weak duality holds. 
Suppose that $x$ is primal feasible and $Z$ is dual feasible, then we compute the duality gap associated with $x$ and $Z$
\begin{align*}
\eta(x,Z):=c'x + \tr(F_0 Z) = \sum_{i=1}^n \tr\left(F_i Z\right) x_i + \tr(F_0 Z) = \tr\left(F(x) Z\right) \geq 0,
\end{align*}
where the inequality comes from $F(x)\succeq 0$ and $Z\succeq 0$. Hence weak duality holds.

Let $p^*$ denote the optimal value of Equation~\eqref{app-sdp:prime}, and $d^*$ denote the optimal value of Equation~\eqref{app-sdp:dual}.
Further, we give the conditions for strong duality. 
\begin{theorem}[strong duality] Suppose that there exists a strictly feasible solution for Equation~\eqref{app-sdp:prime}, i.e., $F(x)\succ 0$, and that there exists a strictly feasible solution for Equation~\eqref{app-sdp:dual}, i.e., $Z\succ 0$. Then strong duality hold, i.e., $p^*=d^*$.
\end{theorem}

\section{Interior Point Method}

SDP can be solved via interior point methods, and most of its applications can usually be solved very efficiently in practice as well as in theory.

In interior point methods, a barrier function is employed to ensure that the iterates of $F(X)$  remain in the interior of the feasible set, i.e., the semi-definite cone, for instance
\begin{equation*}
\phi(x):=\left(\sum_{j=1}^n \ln(\Lambda_{jj})\right)^{-1}=\ln\left(\prod_{j=1}^n\Lambda_{jj}\right)^{-1} = \ln(\det F(x))^{-1},
\end{equation*}
where $\Lambda_{jj}$ is the $j^{th}$ eigenvalue of $F(x)$, and $\det$ denotes the determinant. This function is strictly convex, and it is finite if and only if $F(x)\succ 0$.

Suppose that the space $\{x\in\mathbb{R}^n\mid F(x)\succ 0\}$ is bounded and strong duality holds. Consider the analytic centre
\begin{equation}
\begin{split}
x^*(\gamma):=\arg\min_{x\in\mathbb{R}^n} \quad & \phi(x)\\
\textrm{s.t.} \quad & c'x=\gamma,
\end{split}\label{app-sdp:analytic}
\end{equation}
where $p^*<\gamma<\sup\{c'x\mid F(x)\succ 0\}$.
The curve described by $x^*(\gamma)$ is called the \textit{central path}. It passes through a point of $\arg\min_{x\in\mathbb{R}^n}\phi(x)$, and converges to the optimal solution of Equation~\eqref{app-sdp:prime} as $\gamma$ approaches $p^*$ from above.
Equation~\eqref{app-sdp:analytic} can be solved by Newton's method.

The optimality conditions for Equation~\eqref{app-sdp:analytic} implies that 
\begin{equation*}
\left(\nabla\phi(x^*(\gamma))\right)_i=-\tr F(x^*(\gamma))^{-1} F_i = \delta c_i,
\end{equation*}
for $i=1,\dots,n$, and $\delta$ is a Lagrangian multiplier. Therefore, $F(x^*(\gamma))^{-1}/\delta$ is dual feasible when $\delta>0$, such that the points on the primal central path yield dual feasible solutions.

We can interpret $x^*(\gamma)$ as a sub-optimal point that gives the upper bound $p^*\leq c'x^*(\gamma)$ and the induced dual feasible solution as certificate that proves the lower bound $p^*\geq -\tr F_0 F(x^*(\gamma))^{-1}/\delta$. We can bound how sub-optimal the point $x^*(\gamma)$ is in terms of the duality gap associated with the feasible pair $x^*(\gamma),F(x^*(\gamma))^{-1}/\delta$:
\begin{equation*}
c'x^*(\gamma) - p^*\leq \eta(x^*(\gamma),F\left(x^*(\gamma))^{-1}/\delta\right)= \tr F(x^*(\gamma)) F(x^*(\gamma))^{-1}/\delta = n/\delta.
\end{equation*}


\chapter{Moment Problems \& Polynomial Optimisation}
\label{cha:mom_app}

\begin{quote}
\textit{As a prequel to our work on learning for linear dynamic systems, this chapter provides background materials on polynomial optimisation \citep{lasserre2001global}.
It commences with the development of relaxations for moment problems \citep{lasserre2001global,lasserre2009moments}, which will later be demonstrated to be useful in polynomial optimisation. This appendix follows \cite{lasserre2009moments}.}
\end{quote}

A major technical building block for the work in this thesis is non-commutative polynomial optimisation.
We now provide readers with essential background materials on non-commutative polynomial optimisation.
For ease of exposition, we split this background materials into two chapters.
The first chapter deals with the so-called moment problem, which is in fact equivalent to polynomial optimisation.
The second chapter then moves on to discuss the non-commutative analog.

Before stepping into non-commutative polynomial optimisation, we start from its commutative analog, in which we consider variables to be real numbers.
Let $\mathscr{D}$ denote a Borel subset of $\mathbb{R}^n$, referred to as the positive domain.
Let us denote the vectors in the positive domain in bold face, in contrast to the elements thereof: $\mathbf{x}=(x_1,\dots,x_n)\in\mathscr{D}$. 
Let us denote $\mathbb{R}[\mathbf{x}]$ the ring of real polynomials in the variable $\mathbf{x}$. The \acrfull{POP} involves finding the minimum of a polynomial $f\in\mathbb{R}[\mathbf{x}]$ over the positive domain $\mathscr{D}$:
$$\inf_{\mathbf{x}\in\mathscr{D}} f(\mathbf{x}).$$

Indeed, as we will see later, this formulation of POP is equivalent to
\begin{equation*}
\begin{aligned}
\inf_{\mu\in\mathscr{M}_+(\mathscr{D})} &\int_{\mathscr{D}} f d\mu\\
\textrm{s.t.} &\int_{\mathscr{D}} d\mu =1,
\end{aligned}
\end{equation*}
where $\mathscr{M}_+(\mathscr{D})$ is the space of finite positive Borel measures on $\mathscr{D}$.
The second formulation is a particular case of moment problems \citep{lasserre2009moments}, which have a rich history in functional analysis. Let us consider the following more general variant: Given a set of indices $\Gamma$ and polynomials $f, h_i \in \mathbb{R}[\mathbf{x}]$ for $i \in \Gamma$, which are integrable with respect to every measure $\mu\in\mathscr{M}_+(\mathscr{D})$, the formulation of the \acrfull{GMP} is as follows:
\begin{equation}
\begin{aligned}
\rho_{mom}=\sup_{\mu\in\mathscr{M}_+(\mathscr{D})} &\int_{\mathscr{D}} f d\mu\\
\textrm{s.t.} &\int_{\mathscr{D}} h_i d\mu \leq 0,\; \forall i\in\Gamma.
\end{aligned}
\tag{GMP}
\label{equ:GMP}
\end{equation}
Handling GMP seems challenging due to the integration and polynomial functions in the formulation. 
To get rid of them, and also to ``linearise'' the above formulation, consider the idea that given a polynomial, e.g., $f(\mathbf{x}):=x_1^2+3x_1 x_2$, if we define two auxiliary variables (called monomials) $\omega_1:=x_1^2$ and $\omega_2:=x_1 x_2$, such a polynomial can be written as a linear combination of these two auxiliary variables:
$$f=x_1^2+3x_1 x_2=\omega_1+3\omega_2.$$

Then, given a measure $\mu\in\mathscr{M}_+(\mathscr{D})$,
if we define two extra auxiliary variables (called moments) $y_{\omega_1}:=\int_{\mathscr{D}} \omega_1 d\mu$ and $y_{\omega_2}:=\int_{\mathscr{D}} \omega_2 d\mu$, the integration of such a polynomial can also be written as a linear combination:
$$\int_{\mathscr{D}} f d\mu=\int_{\mathscr{D}} x_1^2  d\mu+ \int_{\mathscr{D}} 3x_1 x_2 d\mu=y_{\omega_1} + 3 y_{\omega_2}.$$

While GMP has been ``linearised'' in terms of the moment sequence (i.e., $\{y_{\omega_1},y_{\omega_2}\}$ in the above example), a new challenge arises in guaranteeing the existence of a representing measure $\mu$ for the moment sequence, and, more importantly, in formulating such a guarantee. 
As we will see later, 
the celebrated Putinar's positivstellensatz, demonstrates that such a guarantee can be equivalently expressed as an infinite number of linear constraints in the variables of an infinite moment sequence. (This formulation is outlined later in Equation~\eqref{equ:GMP-prime-sdp}.)
Although finite-dimensional linear programming is well-understood, a new challenge arises in this infinite-dimensional setting.

When considering only measures with a finite number of moments, rather than the infinite moment sequence, in the so-called truncated moment problems, the resulting relaxation can be formulated as the \acrfull{SDP} \citep{vandenberghe1996semidefinite}, with a \textbf{finite} number of linear matrix inequalities in the \textbf{finite} truncated moment sequence. By iteratively enlarging the truncated moment sequence, the sequence of truncated problems forms a hierarchy of SDP relaxations for GMP. Then, the Curto-Fialkow’s theorem, 
ensures convergence of this hierarchy of SDP relaxations after a finite number of iterations.
(This truncated formulation is presented later in Equation~\eqref{equ:GMP-prime-sdp-relaxation} on p.~\pageref{equ:GMP-prime-sdp-relaxation}.)
Notably, 
since POP is equivalent to a special case of GMP (cf. Theorem~\ref{the:pop-duality}), it can also be addressed through this hierarchy of SDP relaxations.

In the following Sections~\ref{sec:GMP}--\ref{sec:GMP-SDP}, we will formally introduce these definitions and demonstrate how to transform the GMP formulation into a hierarchy of SDP relaxations, step by step. Then, Section~\ref{sec:pop} demonstrates that GMP is, in fact, the dual problem of POP, allowing the hierarchy of SDP relaxations for GMP to be readily applied to POP.


\section{Generalised Moment Problem}\label{sec:GMP}

Suppose we are given a polynomial $f(\mathbf{x})=x_1^2+3x_1 x_2$, which is nonlinear if $\mathbf{x}=(x_1,x_2)$ are the variables. If we regard $x_1^2$ and $x_1 x_2$ as variables, this polynomial becomes linear.
With this reasoning, all polynomials can be considered as linear combinations of elements in the monomial basis, defined as follows.

\begin{definition}[monomial basis]\label{def:monomial_basis}
We define a monomial with respect to a multi-index $\mathbf{\alpha}=(\alpha_1,\dots,\alpha_n)\in\mathbb{N}^n$:
\begin{equation*}
\omega:=x_1^{\alpha_1}x_2^{\alpha_2}\dots x_n^{\alpha_n}.
\end{equation*}

Let the degree of the monomial $|\omega|:=\sum_{i=1}^n \alpha_i$. We define the monomial basis of all monomials with degree less than or equal to $d$:
\begin{equation*}
\mathbf{W}_d:=\{\omega\mid |\omega|\leq d\}=\{1,x_1,\dots,x_n,x_1^2,x_1x_2,\dots,x_1^d,\dots,x_n^d\},
\end{equation*}
The cardinality of $\mathbf{W}_d$ is $\sigma(d):=\binom{n+d}{d}$. 
Let $\mathbf{W}:=\cup_{d=0}^{\infty} \mathbf{W}_d$.
\end{definition}

Let $\mathbb{R}[\mathbf{x}]_d$ be the vector space of polynomials of degree at most $d$. 
Now, a polynomial $f\in\mathbb{R}[\mathbf{x}]_d$ can be written as
\begin{equation*}
\mathbf{x}\mapsto f(\mathbf{x})=\sum_{\omega\in\mathbf{W}_d} f_{\omega} \omega=\langle\overrightarrow{f},\overrightarrow{\mathbf{W}_d}\rangle,
\end{equation*}
where $\overrightarrow{f}=(f_{\omega})\in\mathbb{R}^{\sigma(d)}$ denotes the column vector of coefficients in the monomial basis $\mathbf{W}_d$, and $\overrightarrow{\mathbf{W}_d}$ denotes the column vector consisting of all elements of $\mathbf{W}_d$.
Further, we use the same reasoning to linearise the integral of a polynomial, i.e., $\int_{\mathscr{D}} f d\mu$, by associating a moment sequence with the monomial basis, defined as follows.

\begin{definition}[moment sequence $\mathbf{y}$]\label{def:sequence_y}
Let $\mathbf{y}=\{y_{\omega}\}\subset\mathbb{R}$ be an infinite moment sequence
\begin{equation*}
y_{\omega}=\int_{\mathscr{D}} \omega d\mu,\;\forall \omega\in\mathbf{W}.
\end{equation*}
\end{definition}
\begin{definition}[linear functional $L_{\mathbf{y}}$]\label{def:functional_L}
Let $L_{\mathbf{y}}:\mathbb{R}[\mathbf{x}]\to\mathbb{R}$ be the linear functional:
\begin{equation*}
f(\mathbf{x})=\sum_{\omega\in\mathbf{W}} f_{\omega} \omega\mapsto L_{\mathbf{y}}(f)=\sum_{\omega\in\mathbf{W}} f_{\omega} y_{\omega}=\int_{\mathscr{D}} f d\mu.
\end{equation*}
\end{definition}
Then, using Definitions~\ref{def:sequence_y}--\ref{def:functional_L},
Equation~\eqref{equ:GMP} can be written in the equivalent form:
\begin{equation}
\begin{aligned}
\rho_{mom}=\sup_{\mathbf{y}}& L_{\mathbf{y}}(f)\\
\textrm{s.t.}& L_{\mathbf{y}}(h_i) \leq 0,\; \forall i\in\Gamma\\
&\exists \mu\in\mathscr{M}_+(\mathscr{D}): y_{\omega}=\int_{\mathscr{D}} \omega\; d\mu,\;\forall \omega\in\mathbf{W}.
\end{aligned}
\label{equ:GMP-prime}
\end{equation}
Equation~\eqref{equ:GMP} has been transformed into a formulation of the infinite moment sequence $\mathbf{y}$, instead of the measure $\mu$, with the condition that the real sequence $\mathbf{y}$ should be the moment sequence of some measures $\mu\in\mathscr{M}_+(\mathscr{D})$.
In other words, there should be a representing measure $\mu\in\mathscr{M}_+(\mathscr{D})$ for the moment sequence $\mathbf{y}$.

\subsection{Existence of Measures on a Semi-Algebraic Set}

In Equation~\eqref{equ:GMP-prime}, the final constraint imposes that the real sequence $\mathbf{y}$ should be the moment sequence of certain measures $\mu\in\mathscr{M}_+(\mathscr{D})$. 
As we will see in Theorem~\ref{the:exist_measure}, this constraint can be expressed as linear matrix inequalities when $\mathscr{D}$ is a compact semi-algebraic set, defined as follows.

\begin{definition}[semi-algebraic set $\mathscr{D}_{\mathbf{g}}$]\label{def:semi-algebraic}
Let $\mathscr{D}_{\mathbf{g}}$ be a semi-algebraic set, corresponding to a finite subset of polynomials $\mathbf{g}\subset\mathbb{R}[\mathbf{x}]$ of the form:
\begin{equation*}
\mathscr{D}_{\mathbf{g}}:=\left\{\mathbf{x}\in\mathbb{R}^n\;|\;g_j(\mathbf{x})\geq 0,\;\forall g_j\in\mathbf{g} \right\}.
\end{equation*}
\end{definition}

We rely on the Archimedean assumption\footnote{It assumes that the set of polynomials $\mathbf{g}$, used to define the semi-algebraic set $\mathscr{D}_{\mathbf{g}}$, contains a ball constraint, for example, $g_i(\mathbf{x})=N-\|\mathbf{x}\|^2$ for some real number $N>0$.} to ensure the compactness of this semi-algebraic set.
To delve into this, we must first introduce the concepts of sum of squares polynomials and quadratic modules.

\begin{definition}[sum of squares]\label{def:sos}
A polynomial $f\in\mathbb{R}[\mathbf{x}]$ is a sum of squares (SOS) if can be written as
\begin{equation*}
\mathbf{x}\mapsto f(\mathbf{x})=\sum_{i} f_i(\mathbf{x})^2, \;\forall \mathbf{x}\in\mathbb{R}^n,
\end{equation*}
for some finite family of polynomials $f_i\in\mathbb{R}[\mathbf{x}]$. 
Let $\Sigma[\mathbf{x}]\subset \mathbb{R}[\mathbf{x}]$ be the space of all SOS polynomials. 
Clearly, 
$f(\mathbf{x})\geq 0$, for all $\mathbf{x}\in\mathbb{R}^n$.
\end{definition}

\begin{definition}[quadratic module]\label{def:quadratic_module}
Given a finite subset of polynomials $\mathbf{g}\subset\mathbb{R}[\mathbf{x}]$, the quadratic module generated by $\mathbf{g}$ is
\begin{equation*}
    Q(\mathbf{g}):=\left\{f_0+\sum_{j} f_j g_j\;|\; f_j\in\Sigma[\mathbf{x}],g_j\in\mathbf{g}\right\}.
\end{equation*}
\end{definition}

\begin{assumption}[Archimedean]{\cite{lasserre2009moments}, Theorem~2.15.}\label{ass:archimedean}
With a finite subset of polynomials $\mathbf{g}\subset\mathbb{R}[\mathbf{x}]$, and quadratic module $Q(\mathbf{g})$ as Definition~\ref{def:quadratic_module},
there exists $f\in Q(\mathbf{g})$ such that the level set $\{\mathbf{x}\in\mathbb{R}^n:f(\mathbf{x})\geq 0\}$ is compact. This assumption is equivalent to
\begin{equation*}
    \exists N>0: \mathbf{x}\mapsto N-\|\mathbf{x}\|^2\in Q(\mathbf{g}).
\end{equation*}
\end{assumption}

Suppose we know some $N>0$ such that $\mathscr{D}_{\mathbf{g}}\subset\{\mathbf{x}\in\mathbb{R}^n:\|\mathbf{x}\|^2\leq N\}$. 
By adding the redundant constraint $N-\|\mathbf{x}\|^2\geq 0$ to $\mathscr{D}_{\mathbf{g}}$ in Definition~\ref{def:semi-algebraic}, the Archimedean assumption is satisfied without changing $\mathscr{D}_{\mathbf{g}}$.

With the Archimedean assumption, guaranteeing the compactness of the semi-algebraic set $\mathscr{D}_{\mathbf{g}}$, we introduce Putinar's renowned representation for positive polynomials on compact semi-algebraic sets. This representation is pivotal for reshaping the final constraint in Equation~\eqref{equ:GMP-prime}.

\begin{theorem}[Putinar's positive polynomials]\label{the:positive_polynomial}
With a finite subset of polynomials $\mathbf{g}\subset\mathbb{R}[\mathbf{x}]$, let $\mathscr{D}_{\mathbf{g}}$ as in Definition~\ref{def:semi-algebraic}, and $Q(\mathbf{g})$ as in Definition~\ref{def:quadratic_module}. Let the Archimedean assumption hold. If $f\in\mathbb{R}[\mathbf{x}]$ is \textbf{strictly} positive on $\mathscr{D}_{\mathbf{g}}$, then $f\in Q(\mathbf{g})$, that is 
\begin{equation*}
    f=f_0+\sum_{j} f_j g_j,
\end{equation*}
where $f_0,f_j\in\Sigma[\mathbf{x}]$, and $g_j\in\mathbf{g}$.
\end{theorem}

\begin{theorem}[existence of a representing measure $\mu$ of the sequence $\mathbf{y}$]{\cite{lasserre2009moments}, Theorem~3.8.}\label{the:exist_measure}
Let $\mathbf{y}=\{y_{\omega}\}$ be an infinite sequence. Let $L_{\mathbf{y}}$ be as in Definition~\ref{def:functional_L} and $\mathscr{D}_{\mathbf{g}}$ as in Definition~\ref{def:semi-algebraic}, with a finite subset of polynomials $\mathbf{g}\subset\mathbb{R}[\mathbf{x}]$ and $g_0=1$. Let the Archimedean assumption hold. Then, the condition of the real sequence $\mathbf{y}$ having a finite Borel representing measure $\mu\in\mathscr{M}_+(\mathscr{D}_{\mathbf{g}})$ is equivalent to some linear constraints:
\begin{equation*}
\exists \mu\in\mathscr{M}_+(\mathscr{D}_{\mathbf{g}}): y_{\omega}=\int_{\mathscr{D}_{\mathbf{g}}} \omega d\mu,\;\forall \omega\in\mathbf{W} \iff L_{\mathbf{y}}(f g_j)\geq 0, \;\forall\; f\in\Sigma[\mathbf{x}],\;g_j\in\mathbf{g}.
\end{equation*}
\begin{proof}
The forward direction stems from the definitions of positive domain $\mathscr{D}_{\mathbf{g}}$ and SOS polynomials. 
This proof only tackles the backward direction.
Theorem~3.1 (Riesz-Haviland) in \cite{lasserre2009moments} states that a finite Borel measure $\mu$ on $\mathscr{D}_{\mathbf{g}}$ exists if and only if $L_{\mathbf{y}}(f)\geq 0$ for all polynomials $f\in\mathbb{R}[\mathbf{x}]$ nonnegative on $\mathscr{D}_{\mathbf{g}}$.
We first consider positive polynomials $f(\mathbf{x})>0$ for $\mathbf{x}\in\mathscr{D}_{\mathbf{g}}$. Under Assumption~\ref{ass:archimedean}, Theorem~\ref{the:positive_polynomial} shows that positive polynomials can be written as $f=f_0+\sum_{j} f_j g_j$ for some SOS polynomials $f_0,f_j\in\Sigma[\mathbf{x}]$ (cf. Definition~\ref{def:sos}). 
By linearity of $L_{\mathbf{y}}$, we only need to satisfy $L_{\mathbf{y}}(f_j g_j)\geq 0$ for any $f_j\in\Sigma[\mathbf{x}]$, to ensure that $L_{\mathbf{y}}(f)\geq 0$ for all polynomials $f\in\mathbb{R}[\mathbf{x}]$ positive on $\mathscr{D}_{\mathbf{g}}$.
Further, let $f\in\mathbb{R}[\mathbf{x}]$ be nonnegative on $\mathscr{D}_{\mathbf{g}}$. For arbitrary $\epsilon>0$, we have $f+\epsilon>0$ on $\mathscr{D}_{\mathbf{g}}$, such that $f+\epsilon$ is positive on $\mathscr{D}_{\mathbf{g}}$.
By linearity of $L_{\mathbf{y}}$ and  $L_{\mathbf{y}}(f)\geq 0$ holds for all polynomials $f\in\mathbb{R}[\mathbf{x}]$ positive on $\mathscr{D}_{\mathbf{g}}$, we have $L_{\mathbf{y}}(f+\epsilon)=L_{\mathbf{y}}(f)+\epsilon y_{\mathbf{0}}\geq 0$, where $y_{\mathbf{0}}=\int_{\mathscr{D}_{\mathbf{g}}} d\mu$. Since $\epsilon$ is arbitrary, we can ensure 
$L_{\mathbf{y}}(f)\geq 0$ for all polynomials nonnegative on $\mathscr{D}_{\mathbf{g}}$, such that a representing measure $\mu\in\mathscr{M}_+(\mathscr{D}_{\mathbf{g}})$ exists.
\end{proof}
\end{theorem}

Now, according to Theorem~\ref{the:exist_measure},
let the Archimedean assumption hold,
Equation~\eqref{equ:GMP-prime} can be written in the equivalent form:
\begin{equation}
\begin{aligned}
\rho_{mom}=\sup_{\mathbf{y}}& L_{\mathbf{y}}(f)\\
\textrm{s.t.}& L_{\mathbf{y}}(h_i) \leq 0,\; \forall i\in\Gamma\\
&L_{\mathbf{y}}(f_j g_j)\geq 0, \;\forall f_j\in\Sigma[\mathbf{x}],\; g_j\in\mathbf{g}.
\end{aligned}
\label{equ:GMP-prime-sdp}
\end{equation}
Under the Archimedean assumption, Equation~\eqref{equ:GMP} has been equivalently reformulated in terms of the infinite moment sequence, subject only to linear inequalities.
Notably, in the final constraint, determining whether a polynomial $f_j$ can be represented as SOS, at a specific polynomial degree, is an SDP problem.
However, there is not an upper bound on the polynomial degree of $f_j\in\Sigma[\mathbf{x}]$, such that the number of linear inequalities is infinite. 

\section{SDP Relaxations and Finite Convergence}
\label{sec:GMP-SDP}

The infinite moment sequence and the infinite number of linear inequalities in Equation~\eqref{equ:GMP-prime-sdp} prompt the need for the truncated moment problem. We now focus on the truncated sequence $\mathbf{y}=\{y_{\omega}\}_{\omega\in \mathbf{W}_{2d}}$, where the cardinality is denoted by $\sigma(2d)$. Subsequently, we introduce the crucial concepts of moment matrices and localising matrices.

\begin{definition}[moment matrix]\label{def:moment_matrix}
Given a truncated sequence $\mathbf{y}=\{y_{\omega}\}_{\omega\in \mathbf{W}_{2d}}$, and the functional $L_{\mathbf{y}}$ in Definitions~\ref{def:sequence_y}--\ref{def:functional_L},
let $\mathbf{M}_d(\mathbf{y})\in\mathbb{R}^{\sigma(d)\times \sigma(d)}$ be the moment matrix, with row and columns labelled by elements in the monomial basis $\mathbf{W}_d$.
Its entries are defined as follows:
\begin{equation*}
\mathbf{M}_d(\mathbf{y})(\omega,\nu):=y_{\omega\nu}=L_{\mathbf{y}}(\omega\nu),\;\forall \omega,\nu\in\mathbf{W}_d.
\end{equation*}
Let $\overrightarrow{\mathbf{W}_d}'$ denote the transpose of $\overrightarrow{\mathbf{W}_d}$.
The $(\omega,\nu)$-entry of the outer product $\overrightarrow{\mathbf{W}_d}\overrightarrow{\mathbf{W}_d}'$ is exactly $\omega\nu$. Hence, we can  re-write $\mathbf{M}_d(\mathbf{y})=L_{\mathbf{y}}\left(\overrightarrow{\mathbf{W}_d}\overrightarrow{\mathbf{W}_d}'\right)$, where $L_{\mathbf{y}}$ is applied entry-wise.
\end{definition}

\begin{definition}[localising matrix]\label{def:loca_matrix}
Given a polynomial $f\in\mathbb{R}[\mathbf{x}]$ with polynomial degree $\deg(f)$, it can be expressed as the linear combination of monomials $f=\sum_{u\in\mathbf{W}_{\deg(f)}}f_{u}u$.
Then, given a truncated sequence $\mathbf{y}=\{y_{\omega}\}_{\omega\in\mathbf{W}_{2d+\deg(f)}}$, the functional $L_{\mathbf{y}}$,
let $\mathbf{M}_d(f\mathbf{y})\in\mathbb{R}^{\sigma(d)\times \sigma(d)}$ be the localising matrix with row and columns labelled by elements in the monomial basis $\mathbf{W}_d$.
Its entries are defined as follows:
\begin{equation*}
\mathbf{M}_d(f\mathbf{y})(\omega,\nu):=\sum_{u\in\mathbf{W}_{\deg(f)}}f_{u}y_{\omega\nu u}=L_{\mathbf{y}}(f\omega\nu),\;\forall \omega,\nu\in\mathbf{W}_d.
\end{equation*}
Equivalently, we have $\mathbf{M}_d(f\mathbf{y})=L_{\mathbf{y}}\left(f\;\overrightarrow{\mathbf{W}_d}\overrightarrow{\mathbf{W}_d}'\right)$, where $L_{\mathbf{y}}$ is applied entry-wise.
\end{definition}

With the moment matrix and localising matrices, we further derive the equivalent form of the last constraint in Equation~\eqref{equ:GMP-prime-sdp}.

\begin{remark}[truncated form of Theorem~\ref{the:exist_measure}]\label{cor:exist_measure}
With the same information in Theorem~\ref{the:exist_measure}, 

\begin{equation*}
L_{\mathbf{y}}(f g_j)\geq 0, \;\forall\; f\in\Sigma[\mathbf{x}],\;g_j\in\mathbf{g}\iff\mathbf{M}_d(g_j\mathbf{y})\succeq 0,\;\forall g_j\in\mathbf{g},\;d\in\mathbb{N}.
\end{equation*}
\begin{proof}
Without loss of generality, we assume $f\in\Sigma[\mathbf{x}]_{2d}$.
Since $f$ is a SOS polynomial,
it can be expressed as 
\begin{equation*}
f=\sum_{i} f_i^2=\sum_{i}\left(\overrightarrow{f_i}'\overrightarrow{\mathbf{W}_d}\right)\left(\overrightarrow{f_i}'\overrightarrow{\mathbf{W}_d}\right)'=\sum_{i} \overrightarrow{f_i}' \overrightarrow{\mathbf{W}_d}\overrightarrow{\mathbf{W}_d}' \overrightarrow{f_i},
\end{equation*}
for some finite family of polynomials $f_i\in\mathbb{R}[\mathbf{x}]_d$ and $\overrightarrow{f_i}$ is the column vector of coefficients.
When $j=0$ and $g_0=1$, the condition on the left is equivalent to $L_{\mathbf{y}}(f)=\sum_{i}\overrightarrow{f_i}' \mathbf{M}_d(\mathbf{y}) \overrightarrow{f_i}\geq 0$, for all $f_i\in\mathbb{R}[\mathbf{x}]$, such that $\mathbf{M}_d(\mathbf{y})\succeq 0$. When $j>0$, this condition is equivalent to $L_{\mathbf{y}}(fg_j)=\sum_{i} \overrightarrow{f_i}' \mathbf{M}_d(g_j\mathbf{y}) \overrightarrow{f_i}\geq 0$, for all $f_i\in\mathbb{R}[\mathbf{x}]$, such that $\mathbf{M}_d(g_j\mathbf{y})\succeq 0$. Since $d$ is arbitrary, this condition is for all $d\in\mathbb{N}$.
\end{proof}
\end{remark}

In Remark~\ref{cor:exist_measure}, note that at a given moment degree $d\in\mathbb{N}$, the set of linear matrix inequalities $\mathbf{M}_d(g_j\mathbf{y})\succeq 0,\;\forall g_j\in\mathbf{g}$ is of finite cardinality. If we further specify a polynomial $g_j\in\mathbf{g}$, the corresponding linear matrix inequality $\mathbf{M}_d(g_j\mathbf{y})\succeq 0$ is expressed in terms of the truncated moment sequence $\mathbf{y}=\{y_{\omega}\}_{\omega\in\mathbf{W}_{2d+\deg(f)}}$.
Hence, by combining Theorem~\ref{the:exist_measure} and Remark~\ref{cor:exist_measure}, the condition of the real sequence $\mathbf{y}$ having a representing measure $\mu\in\mathscr{M}_+(\mathscr{D}_{\mathbf{g}})$ is equivalent to finite number of linear matrix inequalities at each moment degree $d$, expressed in terms of the truncated moment sequence.

Following this reasoning and still under the Archimedean assumption, Equation~\eqref{equ:GMP-prime-sdp} can be transformed into a hierarchy of SDP relaxations, each indexed by a moment order $d$:

\begin{equation}
\begin{aligned}
\rho_d=\sup_{\mathbf{y}=\{y_{\omega}\}_{\omega\in\mathbf{W}_{2d}}}& L_{\mathbf{y}}(f)\\
\textrm{s.t.}& L_{\mathbf{y}}(h_i) \leq 0,\; \forall i\in\Gamma,\\
&\mathbf{M}_d(\mathbf{y})\succeq 0,\\
&\mathbf{M}_{d-\lceil\deg(g_j)/2\rceil}(g_j\mathbf{y})\succeq 0,\;\forall g_j\in\mathbf{g},
\end{aligned}
\tag{GMP-SDP}
\label{equ:GMP-prime-sdp-relaxation}
\end{equation}
for $d\geq \lceil d_0/2 \rceil$, where $d_0:=\max\{\deg(f);\deg(h_i),i\in\Gamma;\deg(g_j),g_j\in\mathbf{g}\}$, and $\deg(\cdot)$ denote the degree of a polynomial, $\lceil\cdot\rceil$ denote the ceiling function.
Clearly, $\rho_d\geq \rho_{mom}$ for all $d\geq \lceil d_0/2 \rceil$, and $\{\rho_d\}_{d\geq \lceil d_0/2 \rceil}$ forms a monotone non-increasing sequence.

Next, Curto-Fialkow's theorem guarantees the existence of a representing measure for a sequence $\mathbf{y}$ in the \textbf{truncated} moment problem through the flatness condition, ensuring finite convergence for this hierarchy of SDP relaxations.

\begin{theorem}[global optimality condition]{\cite{curto2005truncated};\cite{lasserre2009moments}, Theorem~4.1.}\label{the:optimality}
Let the Archimedean assumption hold.
Let $\rho_{mom}$ be the optimal values of Equation~\eqref{equ:GMP} and be finite.
Let $\{\rho_d\}_{d\geq \lceil d_0/2 \rceil}$ be the sequence of optimal values of Equation~\eqref{equ:GMP-prime-sdp-relaxation}.
Let $\eta:=\max\{\lceil\deg(g_i)/2\rceil,\forall g_i\in\mathbf{g}; 1\}$. If for some $d\geq \lceil d_0/2 \rceil$, Equation~\eqref{equ:GMP-prime-sdp-relaxation} has an optimal solution $\mathbf{y}$ satisfying
\begin{equation*}
\rk\mathbf{M}_{d}(\mathbf{y})=\rk \mathbf{M}_{d+\eta}(\mathbf{y}),
\end{equation*}
Then, $\rho_d=\rho_{mom}$ and Equation~\eqref{equ:GMP} has an (global) optimal solution $\mu\in\mathcal{M}_{+}(\mathscr{D}_{\mathbf{g}})$, which is finitely supported on $\rk\mathbf{M}_{d}(\mathbf{y})$-points of $\mathscr{D}_{\mathbf{g}}$.
\end{theorem}

Suppose the flatness condition in Theorem~\ref{the:optimality} holds at the moment degree $d$. Let $\mu\in\mathcal{M}_{+}(\mathscr{D}_{\mathbf{g}}) $ be the optimal solution of GMP in Equation~\eqref{equ:GMP}, supported on $r:=\rk\mathbf{M}_{d}(\mathbf{y})$ points of $\mathscr{D}_{\mathbf{g}}$. We denote these points by $\mathbf{x}(k)\in\mathscr{D}_{\mathbf{g}}$, for $k=1,\dots,r$, and the measure $\mu$ can be written as
\begin{equation*}
\mu=\sum_{k=1}^{r} \lambda_k \delta_{\mathbf{x}(k)}, \; \lambda_k> 0,\;k=1,\dots,r;\quad  \sum_{k=1}^{r}\lambda_k=1;
\end{equation*}
where $\delta_{\mathbf{x}(k)}$ is the Dirac measure at the point $\mathbf{x}(k)$. 
Let $\overrightarrow{\mathbf{W}_d}(\mathbf{x}(k))$ denote the column vector of monomial basis corresponding to the point $\mathbf{x}(k)\in\mathscr{D}_{\mathbf{g}}$.
Following Definition~\ref{def:moment_matrix}, the moment matrix associated with the measure $\mu$ is
\begin{eqnarray*}
\mathbf{M}_d(\mathbf{y})&=&\sum_{k=1}^{r} \lambda_k \overrightarrow{\mathbf{W}_d}(\mathbf{x}(k)) \overrightarrow{\mathbf{W}_d}(\mathbf{x}(k))',\\
&=&\underbrace{\begin{bmatrix}\overrightarrow{\mathbf{W}_d}(\mathbf{x}(1))&\cdots&\overrightarrow{\mathbf{W}_d}(\mathbf{x}(r))\end{bmatrix}}_{=:\mathbf{\bar{L}}\in\mathbb{R}^{\sigma(d)\times r}}\begin{bmatrix}
\lambda_1&&\\&\ddots&\\&&\lambda_r
\end{bmatrix} \begin{bmatrix}\overrightarrow{\mathbf{W}_d}(\mathbf{x}(1))\\\vdots\\\overrightarrow{\mathbf{W}_d}(\mathbf{x}(r))\end{bmatrix}.
\end{eqnarray*}
Conduct Cholesky decomposition on the moment matrix, such that $\mathbf{M}_d(\mathbf{y})=\mathbf{LL}'$. Both matrices $\mathbf{\bar{L}},\mathbf{L}\in\mathbb{R}^{\sigma(d)\times r}$ span the same linear subspaces. The optimal solutions $\mathbf{x}(k)$, for $k=1,\dots,r$, can be extracted via transforming $\mathbf{\bar{L}}$ from $\mathbf{L}$ by column operations.

\section{Polynomial Optimisation}
\label{sec:pop}

GMP and POP form a pair of dual problems, and strong duality holds. To illustrate this, let us consider a special case of Equation~\eqref{equ:GMP}:
\begin{equation}
\begin{aligned}
\rho_{mom}=\inf_{\mu\in\mathscr{M}_+(\mathscr{D}_{\mathbf{g}})} &\int_{\mathscr{D}_{\mathbf{g}}} f d\mu\\
\textrm{s.t.} &\int_{\mathscr{D}_{\mathbf{g}}} d\mu =1,
\end{aligned}
\label{equ:GMP-simplex}
\end{equation}
where $f\in\mathbb{R}[\mathbf{x}]$.
Note that the equality constraint $\int_{\mathscr{D}_{\mathbf{g}}} d\mu=1$ is equivalent to two opposite inequality constraints $\int_{\mathscr{D}_{\mathbf{g}}} d\mu-1\geq 0$ and $1-\int_{\mathscr{D}_{\mathbf{g}}} d\mu\geq 0$. 

\begin{theorem}[polynomial optimisation is equivalent to GMP]{\cite{lasserre2009moments}, Theorem~1.1.}\label{the:pop-duality}
The dual problem of Equation~\eqref{equ:GMP-simplex} is POP
\begin{equation}
\rho_{pop}=\inf_{\mathbf{x}\in\mathscr{D}_{\mathbf{g}}} f(\mathbf{x}),
\label{equ:GMP-simplex-dual}
\tag{POP}
\end{equation}
and strong duality holds, i.e., $\rho_{pop}=\rho_{mom}$.
\begin{proof}
The Lagrangian of Equation~\eqref{equ:GMP-simplex} is 
\begin{align*}
\int_{\mathscr{D}_{\mathbf{g}}} f d\mu-\lambda\left(\int_{\mathscr{D}_{\mathbf{g}}} d\mu -1\right)
=
\lambda +
\int_{\mathscr{D}_{\mathbf{g}}} \left( f-\lambda \right)d\mu,
\end{align*}
where the Lagrangian multiplier $\lambda\in\mathbb{R}$ is a free variable.
To maximise the Lagrangian, the multiplier needs to make sure $\left( f-\lambda \right)$ is nonnegative for all $\mathbf{x}\in\mathscr{D}_{\mathbf{g}}$. The Lagrangian dual problem reads:
\begin{equation*}
\sup \lambda \quad\textrm{s.t.}\;\lambda\leq f(\mathbf{x}),\;\forall\mathbf{x}\in\mathscr{D}_{\mathbf{g}},
\end{equation*}
which is to find the maximum lower bound of $f(\mathbf{x})$ for $\mathbf{x}\in\mathscr{D}_{\mathbf{g}}$, and is equivalent to Equation~\eqref{equ:GMP-simplex-dual}. 
Next, we show Equation~\eqref{equ:GMP-simplex-dual} is equivalent to Equation~\eqref{equ:GMP-simplex}.
Since Equation~\eqref{equ:GMP-simplex-dual} is the dual problem of Equation~\eqref{equ:GMP-simplex}, it holds that $\rho_{pop}\leq\rho_{mom}$ due to weak duality.
With every $\mathbf{x}\in\mathscr{D}_{\mathbf{g}}$, we associate the Dirac measure $\delta_{\mathbf{x}}$, which is a feasible solution of Equation~\eqref{equ:GMP-simplex}. For every $\mathbf{x}\in\mathscr{D}_{\mathbf{g}}$, it holds that
\begin{equation*}
f(\mathbf{x})=\int_{\mathscr{D}_{\mathbf{g}}} f d \delta_{\mathbf{x}} \geq \rho_{mom}, \forall \mathbf{x}\in\mathscr{D}_{\mathbf{g}} \quad\Rightarrow \rho_{pop}\geq\rho_{mom}.
\end{equation*}
Due to weak duality, we can conclude that strong duality holds. 
\end{proof}
\end{theorem}

According to Theorem~\ref{the:pop-duality}, by solving the SDP relaxations of the moment problem in Equation~\eqref{equ:GMP-simplex} to the optimum, we find the minimum of polynomial optimisation.
We first write down the Lasserre's hierarchy of SDP relaxations, with $d\geq \lceil d_0/2 \rceil$ and $\mathscr{D}_{\mathbf{g}}$ in Definition~\ref{def:semi-algebraic}. 
\begin{equation}
\begin{aligned}
\rho_d=\inf_{\mathbf{y}=\{y_{\omega}\}_{\omega\in\mathbf{W}_{2d}}}& L_{\mathbf{y}}(f)\\
\textrm{s.t.}& y_0=1,\\
&\mathbf{M}_d(\mathbf{y})\succeq 0,\\
&\mathbf{M}_{d-\lceil\deg(g_j)/2\rceil}(g_j\mathbf{y})\succeq 0,\;\forall g_j\in\mathbf{g}.
\end{aligned}
\tag{POP-SDP}
\label{equ:pop-sdp-relaxation}
\end{equation}

Next, we show that the minimisers of the moment problem and the polynomial optimisation problem are correlated.

\begin{remark}
Let $\mathbf{x}^*$ be a global minimiser of Equation~\eqref{equ:GMP-simplex-dual}, and $\delta_{\mathbf{x}^*}$ be the Dirac measure at $\mathbf{x}^*$. From strong duality in Theorem~\ref{the:pop-duality}, we know that 
\begin{equation*}
\rho_{pop}=f(\mathbf{x}^*)=\int_{\mathscr{D}_{\mathbf{g}}} f d \delta_{\mathbf{x}^*}= \rho_{mom},
\end{equation*}
such that $ \delta_{\mathbf{x}^*}$ is an optimal solution of Equation~\eqref{equ:GMP-simplex}.
\end{remark}


\begin{theorem}{\cite{lasserre2009moments}, Theorem~5.7.}\label{the:optimisers-equivalent}
Let $\mu^*$ be the minimiser of Equation~\eqref{equ:GMP-simplex}.
Suppose optimality condition in Theorem~\ref{the:optimality} is satisfied at certain $\Bar{d}$.
Let the moment sequence $\mathbf{y}^*$ be the optimal solution of Equation~\eqref{equ:pop-sdp-relaxation} at moment degree $\Bar{d}$, with objective value of $\rho_{\Bar{d}}=\rho_{pop}$ by strong duality in Theorem~\ref{the:pop-duality}.
Then $\mathbf{y}^*$ is the moment sequence of some $\rk\mathbf{M}_{\Bar{d}}(\mathbf{y}^*)$-atomic probability measure $\mu^*$ on $\mathscr{D}_{\mathbf{g}}$. Each of the $\rk\mathbf{M}_{\Bar{d}}(\mathbf{y}^*)$ atoms of $\mu^*$ is a global minimiser of Equation~\eqref{equ:GMP-simplex-dual}.
\end{theorem}
By Theorem~\ref{the:optimisers-equivalent}, the minimiser $\mu^*$ of the moment problem can be used to  extract the minimiser(s) of the polynomial optimisation problem.

\chapter{Operator-Valued Polynomial Optimisation}
\label{cha:ncpop}

\begin{quote}
\textit{As a further prequel to our work on learning for linear dynamic systems, this chapter provides important background materials on non-commutative polynomial optimisation \citep{pironio2010convergent}, building upon the previous appendix. 
In particular, we focus on non-commutative trace minimisation \citep{burgdorf2016optimization}, and one sparsity-exploiting  variant \citep{klep2022sparse}.}
\end{quote}

Lasserre's method for polynomial optimisation \citep{lasserre2001global}, introduced in Appendix~\ref{sec:pop}, has been extended to operator-valued variables \citep{pironio2010convergent,burgdorf2016optimization}, following Helton and McCullough's certification of non-commutative (n.c.) positive polynomials. 
There, new variables $\mathbf{X}=\{X_1,\dots, X_n\}$ are not simply real numbers but n.c. variables, for which, in general, $X_i X_j\neq X_j X_i$. 
This technology could be applied to optimisation problems over matrices as well as operators, as we will see in the following chapters.



The global convergence of the \acrfull{NCPOP} algorithms was initially demonstrated within the \acrfull{NPA} hierarchy of SDP relaxations \citep{pironio2010convergent,navascues2012sdp}. 
The NPA hierarchy can be used for eigenvalue optimisation, and further extended to trace optimisation, facilitated by Wedderburn's theorem  \citep{burgdorf2016optimization}.
In Section~\ref{sec:ncpop}, we introduce a hierarchy of SDP relaxations for NCPOP, using the example of trace optimisation. 

Notably, the extension of polynomial optimisation to consider n.c. variables introduces a significantly larger monomial basis in SDP relaxations. To reduce the computational burden of NCPOP, sparsity-exploiting variants have been developed \citep{klep2022sparse,wang2021tssos,wang2021tssos}. In Appendix~\ref{sec:sparse_ncpop}, we introduce one of the sparsity-exploiting variants, still using the example of trace optimisation.


\section{Non-Commutative Polynomial Optimisation}\label{sec:ncpop}

We now consider the generalisation of polynomial optimisation.
Let $\mathbb{S}^{n}$ be the space of (bounded) $n$-tuple real symmetric matrices of the same order. Suppose $\mathbf{X}=\{X_1,\dots, X_n\}\in\mathbb{S}^{n}$, then $X_1,\dots, X_n$ have the same order despite that the specific order is not given.
We introduce $\dag$ as conjugate transpose, with definition of $(X^{\dag})^{\dag}=X$, on an operator $X$; $(X_1 X_2 X_3)^{\dag}=X_3^{\dag} X_2^{\dag} X_1^{\dag}$ on a monomial; and $f^{\dag}=\sum_{\omega\in\mathbf{W}} f_{\omega}\omega^{\dag}$ on the moment representation of a polynomial $f(\mathbf{X})=\sum_{\omega\in\mathbf{W}} f_{\omega}\omega$. 
We consider $\mathbb{R}[\mathbf{X}]$ as the polynomial ring in these n.c. variables $\mathbf{X}$.
The elements, e.g., $f(\mathbf{X})\in\mathbb{R}[\mathbf{X}]$, are defined by replacing the variables $\mathbf{x}$ with operators $\mathbf{X}$ from the previous expression $f(\mathbf{x})$ in the last chapter.\footnote{
Initially, the NPA hierarchy \citep{pironio2010convergent} considers the variables $\mathbf{X} = {X_1, \dots, X_n}$ from a set of bounded operators on a Hilbert space, and the polynomials from $\mathbb{R}[\mathbf{X}, \mathbf{X}^{\dag}]$.
In contrast, the hierarchy for trace optimisation \citep{burgdorf2016optimization} considers the self-adjoint elements, i.e., $X_i^{\dag} = X_i$, within these bounded operators on a Hilbert space, and polynomials from $\mathbb{R}[\mathbf{X}]$.}


Let $\Sym\mathbb{R}[\mathbf{X}]$ denote all symmetric elements in $\mathbb{R}[\mathbf{X}]$, such that
\begin{equation}
\Sym\mathbb{R}[\mathbf{X}]:=\{f\in\mathbb{R}[\mathbf{X}]\;|\;f=f^{\dag}\}.
\label{equ:SymR[X]}
\end{equation}

The semi-algebraic set considered here is slightly different from the commutative case.
\begin{definition}[n.c. semi-algebraic set]{\cite{burgdorf2016optimization}, Definitions~1.23}\label{def:semi-algebraic-nc}
Given a subset of n.c. polynomials $\mathbf{g}\subseteq\Sym\mathbb{R}[\mathbf{X}]$, the semi-algebraic set generated by $\mathbf{g}$ is the class of $n$-tuples $\mathbf{X}$ of real symmetric matrices of the same order such that $g_i(\mathbf{X})\succcurlyeq 0$, for all $g_i\in\mathbf{g}$. Formally,
\begin{equation}
\mathscr{D}_{\mathbf{g}}:=\{\mathbf{X}\in\mathbb{S}^n\;|\; g_i(\mathbf{X})\succcurlyeq 0,\;\forall g_i\in \mathbf{g}\}.
\end{equation}
Note that if $\mathbf{g}=\varnothing$, the semi-algebraic set is just $\mathbb{S}^n$.
\end{definition}

As considered in \cite{burgdorf2016optimization}, trace optimisation of polynomials in non-commutative variables, with respect to a polynomial $f\in\Sym\mathbb{R}[\mathbf{X}]$, has the form in \eqref{trace-NCPO}:
\begin{subequations}
\begin{align}
\tr_{\min}(f)&:=\inf\{\tr f(\mathbf{X})\;|\;\mathbf{X}\in\mathbb{S}^{n} \},\label{trace-NCPO-uncon}\\
\tr_{\min}(f,\mathbf{g})&:=\inf\{\tr f(\mathbf{X})\;|\;\mathbf{X}\in\mathscr{D}_{\mathbf{g}}\},\label{trace-NCPO-con}
\end{align}
\label{trace-NCPO}
\end{subequations}
where the former is the unconstrained trace minimisation, and the latter is the constrained version subject to the positive semidefiniteness of all polynomials $\mathbf{g}=\{g_i\}_{i=1}^m\subseteq\Sym\mathbb{R}[\mathbf{X}]$.

Since the algebra of all bounded operators on the infinite dimensional Hilbert space does not admit a trace, we focus on the von Neumann semi-algebraic set $\mathscr{D}^{\II}_{\mathbf{g}}$ (cf. Definitions~1.59 of \cite{burgdorf2016optimization}), with $\mathscr{D}_{\mathbf{g}}\subseteq\mathscr{D}^{\II}_{\mathbf{g}}$ (cf. Remark~1.61 of \cite{burgdorf2016optimization}).
The following formulation $\tr^{\II}_{\min}(f,\mathbf{g})$ provides a lower bound for $\tr_{\min}(f,\mathbf{g})$:
\begin{equation}
    \tr^{\II}_{\min}(f,\mathbf{g}):=\inf\{\tr f(\mathbf{X})\;|\;X\in\mathscr{D}^{\II}_{\mathbf{g}}\}\leq \tr_{\min}(f,\mathbf{g}).\label{trace-NCPO-con-II}
\end{equation}

 

In terms of \eqref{trace-NCPO-uncon}, it is easy to see that $\tr (f-a)(\mathbf{X})\geq 0, \forall \mathbf{X}\in\mathbb{S}^n$ if $a\leq\tr_{\min}(f)$. Hence, \eqref{trace-NCPO-uncon} is equivalent to \eqref{trace-NCPO-uncon-sup}. Following the same reasoning, we can rewrite \eqref{trace-NCPO-con-II} as \eqref{equ:trace-NCPO-con-sup}.
\begin{subequations}
\begin{align}
\tr_{\min}(f)&:=\sup\{a\;|\;\mathbf{X}\in\mathbb{S}^{n};\tr (f-a)(X)\geq 0 \},\tag{Unconstrained-NCPOP}\label{trace-NCPO-uncon-sup}\\
\tr^{\II}_{\min}(f,\mathbf{g})&:=\sup\{a\;|\;X\in\mathscr{D}^{\II}_{\mathbf{g}};\tr (f-a)(X)\geq 0 \}\leq\tr_{\min}(f,\mathbf{g}).\tag{NCPOP}\label{equ:trace-NCPO-con-sup}
\end{align}
\label{trace-NCPO-sup}
\end{subequations}

The unconstrained trace minimisation can be relaxed to a single SDP; however, the details fall outside the scope of our current discussion.
The subsequent content in this section exclusively focuses on the constrained case of trace minimisation, i.e., $\mathbf{g}\neq\varnothing$ (where $\varnothing$ denotes the empty set), which is particularly relevant to the forthcoming chapters. 

\subsection{SDP Relaxations of the Primal Problem}

The fundamental results in n.c. polynomial optimisation exploit the positiveness of n.c. polynomials on a tuple $\mathbf{X}$ of (symmetric) bounded operators in \cite{helton2004positivstellensatz}, which is the n.c. extension of Putinar's results in Theorem~\ref{the:positive_polynomial}.
 
\begin{definition}[n.c. quadratic module]{\cite{burgdorf2016optimization}, Definitions~1.22.}\label{def:quadratic_module_nc}
Given a subset of polynomials $\mathbf{g}=\{g_i\}_{i=1}^m\subseteq\Sym\mathbb{R}[\mathbf{X}]$.
The quadratic module generated by $\mathbf{g}$ is
\begin{equation*}
    Q(\mathbf{g}):=\left\{\sum^{N}_{i=1} f_i^{\dag}f_i+\sum^{m}_{i=1} \sum^{N_i}_{j=1}  h_{ij}^{\dag}g_i h_{ij} \;|\; g_i\in \mathbf{g};N,N_i\in\mathbb{N};  f_i,h_{ij}\in\mathbb{R}[\mathbf{X}]\right\}.
\end{equation*}
Note that when $\mathbf{g}$ is empty, the quadratic module is the space of sum of Hermitian squares (SOHS) polynomials, denoted by $\Sigma[\mathbf{X}]$. Each element, e.g., $f\in\Sigma[\mathbf{X}]$ can be written as $f=\sum_{i} f_i^{\dag}f_i$, for some finite family of n.c. polynomials $f_i\in\mathbb{R}[\mathbf{X}]$.
We define the truncated quadratic module of order $d$:
\begin{eqnarray*}
Q_{d}(\mathbf{g}):=Q(\mathbf{g})\cap\mathbb{R}[\mathbf{X}]_{2d}.
\end{eqnarray*}
\end{definition}

Now, following \cite{helton2004positivstellensatz} or Proposition~1.25 of \cite{burgdorf2016optimization}, for all $\mathbf{X}\in\mathscr{D}_{\mathbf{g}}$ if $f\in Q(\mathbf{g})$, then $f(X)\succcurlyeq 0$, thus $\tr f(X)\geq 0$. However, does it hold conversely? As we will show below, there are other polynomials out of the quadratic module $Q(\mathbf{g})$ but still have nonnegative trace, i.e., $\tr f(X)\geq 0$.

A commutator of two polynomials $f,g\in\mathbb{R}[\mathbf{X}]$ is defined as $fg-gf$.
Since operators can be cyclically interchanged inside a trace, i.e., $\tr(fg)=\tr(gf)$, the trace of a commutator is zero. Conversely, trace zero matrices are (sums of) commutators, using the proof in \cite{albert1957matrices}.
This fact allows us to define an equivalence relation between two polynomials when they have the same trace. 

\begin{definition}[cyclic equivalence]{\cite{burgdorf2016optimization}, Definition~1.50.}\label{def:cyclic_equivalence}
Two polynomials $f,g\in\mathbb{R}[\mathbf{X}]$ are cyclic equivalent if $f-g$ is a sum of commutators, denoted by 
\begin{equation*}
    f\stackrel{\cyc}{\sim} g.
\end{equation*}
\end{definition}

\begin{definition}[cyclic quadratic modules]{\cite{burgdorf2016optimization}, Definitions~1.56.}\label{def:cyclic quadratic_modules}
Given a subset of polynomials $\mathbf{g}=\{g_i\}_{i=1}^m\subseteq\Sym\mathbb{R}[\mathbf{X}]$.
As a tracial variant of $Q(\mathbf{g})$ in Definition~\ref{def:quadratic_module_nc}, the cyclic quadratic module $\Theta(\mathbf{g})$ includes $Q(\mathbf{g})$ and other n.c. polynomials that are cyclically equivalent to elements in $Q(\mathbf{g})$:
\begin{equation*}
    \Theta(\mathbf{g}):=\left\{f\in\Sym\mathbb{R}[\mathbf{X}] \;|\; \exists g_i\in Q(\mathbf{g}): f\stackrel{\cyc}{\sim} g_i \right\}.
\end{equation*}
We define the truncated cyclic quadratic module of order $2d$ generated by $\mathbf{g}$:
\begin{eqnarray*}
\Theta_{d}(\mathbf{g}):=\left\{f\in\Sym\mathbb{R}[\mathbf{X}] \;|\; \exists g_i\in Q_{d}(\mathbf{g}): f\stackrel{\cyc}{\sim} g_i \right\}.
\end{eqnarray*}
\end{definition}


We know that two polynomials are cyclic equivalent when they have the same trace, and polynomials in the quadratic module $Q(\mathbf{g})$ have nonnegative trace.
Directly from the definition of the cyclic quadratic module, the certificate of $f\in\Theta(\mathbf{g})$ means the nonnegative trace of $f$, as in Theorem~\ref{the:cyclic2positivetrace}.
Further, if the Archimedean assumption\footnote{See Assumption~\ref{ass:archimedean} in Appendix~\ref{sec:pop} with the variables $\mathbf{x}$ replaced by the operators $\mathbf{X}$.} holds, a n.c. polynomial with nonnegative trace belongs to or is very close to the cyclic quadratic module, as in Theorem~\ref{the:positivetrace2cyclic}.


\begin{theorem}[forward]{\cite{burgdorf2016optimization}, Proposition~1.62.}\label{the:cyclic2positivetrace}
Let $\mathbf{g}\subseteq\Sym\mathbb{R}[\mathbf{X}]$.
If $f\in\Theta(\mathbf{g})$, then $\tr f(X)\geq 0, \;\forall X\in\mathscr{D}_{\mathbf{g}}$, and further, $\tr f(X)\geq 0,\;\forall X\in\mathscr{D}^{\II}_{\mathbf{g}}$.
\end{theorem}
\begin{theorem}[backward]{\cite{burgdorf2016optimization}, Proposition~1.63.}\label{the:positivetrace2cyclic}
Let $\mathbf{g}\cup\{f\}\subseteq\Sym\mathbb{R}[\mathbf{X}]$.
Suppose $Q(\mathbf{g})$ is Archimedean, such that $\exists N>0: N-\|\mathbf{X}\|^2\in Q(\mathbf{g})$.
If $\tr f(X)\geq 0,\;\forall X\in\mathscr{D}^{\II}_{\mathbf{g}}$, then $ f+\epsilon\in\Theta(\mathbf{g})$, for all $\epsilon>0$.
\end{theorem}

Along the lines of Theorems~\ref{the:cyclic2positivetrace}--\ref{the:positivetrace2cyclic}, and then let the Archimedean assumption hold, the constrained formulation in \eqref{trace-NCPO-sup} can be relaxed to \eqref{equ:trace-NCPO-sup-Theta}, with $2d\geq\min_{g\in\mathbb{R}[\mathbf{X}], g\stackrel{\cyc}{\sim} f}\deg g$: 
\begin{equation}
\tr_{\Theta_{d}}(f,\mathbf{g}):=\sup\{a\;|\;f-a\in\Theta_{d}(\mathbf{g})\} \leq
\tr^{\II}_{\min}(f,\mathbf{g})\leq\tr_{\min}(f,\mathbf{g}).
\tag{NCPOP-SDP}
\label{equ:trace-NCPO-sup-Theta}
\end{equation}

Although the above formulation is labelled as NCPOP-SDP, we postpone proving it being an SDP problem.
Since we are working with n.c. variables, we first define the n.c. analog of the monomial basis, utilising the same notation $\mathbf{W}$ as in the previous chapter, because commutative monomial basis can be regarded as a special case of the n.c. analog.

\begin{definition}[n.c. monomial basis]\label{def:monomial_basis_nc}
The n.c. monomial basis consists all possible words of finite degree (denoted $\mathbf{W}$), or of at most $d$ (denoted $\mathbf{W}_d$), in these symmetric letters $X_1,\dots, X_n$.
Importantly, the n.c. variables $\mathbf{X}=\{X_1,\dots, X_n\}$ do not commute. For instance, $X_1 X_2$ and $X_2 X_1$ are two different words. 
The cardinality of the n.c. monomial basis $\mathbf{W}_d$ is $\sigma_{nc}(d):=\sum_{k=0}^d n^k=\frac{n^{d+1}-1}{n-1}$, which is much larger than the commutative case.
\end{definition}

In this chapter, the symbols $\omega,\nu,v,z$ denote the words (or monomials) in n.c. monomial basis, with the degree of a monomial $\omega$ being $|\omega|$.
Note that a n.c. polynomial $f\in\mathbb{R}[\mathbf{X}]$ is also a linear combination of elements in the n.c. monomial basis:
\begin{equation*}
f=\sum_{\omega\in\mathbf{W}} f_{\omega} \omega.
\end{equation*}

Using the n.c. monomial basis, we show that that cyclic equivalence is linear, as in Theorem~\ref{prop:cyclic equivalence is linear}, and hence, checking if a polynomial belongs to the cyclic quadratic modular, can be formulated as some linear matrix inequalities (LMI), as in Theorem~\ref{prop:constrained LMI}.

\begin{theorem}[cyclic equivalence is linear]{\cite{burgdorf2016optimization}, Proposition~1.51.}\label{prop:cyclic equivalence is linear}
Let $\mathbf{W}$ be the set of monomials as in Definition~\ref{def:monomial_basis_nc}.
Given two monomials $\omega,\nu\in \mathbf{W}$,
\begin{equation*}  
\omega\stackrel{\cyc}{\sim}\nu \iff \exists u_1,u_2\in \mathbf{W} :\omega=u_1 u_2,\;\nu=u_2 u_1.
\end{equation*}
Given two polynomials $f,g\in\mathbb{R}[\mathbf{X}]$, which can be written as $f(\mathbf{X})=\sum_{|\omega|\leq \deg(f)} f_{\omega} \omega$ and $g(\mathbf{X}) = \sum_{|\omega|\leq \deg(g)} g_{\omega}\omega$, with $f_{\omega},g_{\omega}\in\mathbb{R}$. Then
\begin{equation*}
    f\stackrel{\cyc}{\sim} g \iff \sum_{\omega\in\mathbf{W},\omega\stackrel{\cyc}{\sim}\nu} p_{\omega}=\sum_{\omega\in\mathbf{W},\omega\stackrel{\cyc}{\sim}\nu} q_{\omega},\;\; \forall \nu\in\mathbf{W}.\label{equ:cyclic equivalence is linear}
\end{equation*}
\end{theorem}

\begin{theorem}[certificate of a cyclic quadratic module is some LMI]{\cite{burgdorf2016optimization}, Proposition~5.7.}
\label{prop:constrained LMI}
In the constrained case, $\varnothing\neq \mathbf{g}\subset\Sym\mathbb{R}[\mathbf{X}]$. Each element $g_{i}\in\mathbf{g}$ could be written as $g_{i}=\sum_{u\in\mathbf{W}} g_{i,u} u$.
Then, the certificate of the cone $\Theta_{d}(\mathbf{g})$ could be formulated as a set of LMI:
\begin{eqnarray*}
f\in \Theta_{d}(\mathbf{g}) &\iff& \exists\; A\succcurlyeq 0,B^{i} \succcurlyeq 0: 
f_{z}=\sum_{
\substack{\omega,\nu\in\mathbf{W}_{d},\\\omega^{\dag}\nu\stackrel{\cyc}{\sim}z}}
A_{\omega,\nu} +\sum_i \sum_{\substack{\omega,\nu\in\mathbf{W}_{d_i},u\in\mathbf{W}_{\deg g_i},\\\omega^{\dag}u\nu\stackrel{\cyc}{\sim}z}} g_{i,u} B^{i}_{\omega,\nu},
\end{eqnarray*}
where $z\in\mathbf{W}$, $d_i=\lfloor 
d-\deg(g_i)/2\rfloor$, and $i\in\{i\;|\;g_i\in\mathbf{g}\}$. The matrix $A$ is of order $\sigma_{nc}(d)$ and for each $g_i$, its corresponding matrix $B^{i}$ is of order $\sigma(d_i)$.
\end{theorem}

From Theorem~\ref{prop:constrained LMI}, such that the constrain in Equation~\eqref{equ:trace-NCPO-sup-Theta} can be formulated as some LMI, it becomes apparent that Equation~\eqref{equ:trace-NCPO-sup-Theta} at a given moment degree is indeed an SDP problem.


Hence, we conclude that for an increasing sequence of moment degrees, the relaxations of trace minimisation in Equation~\eqref{equ:trace-NCPO-sup-Theta} form a hierarchy of SDP relaxations.

\subsection{SDP Relaxations of the Dual Problem}

In Appendix~\ref{sec:pop}, we have shown that the dual of polynomial optimisation is the moment problem. 
Loosely speaking, the dual of n.c. polynomial optimisation is also a moment problem with a much larger monomial basis $\mathbf{W}$, as in Definition~\ref{def:monomial_basis_nc}.


Similarly, we can associate a sequence $y_{\omega}=\langle\phi,\omega\phi\rangle$ for $\omega\in\mathbf{W}$, with $\phi$ being a unit vector, i.e., $\langle\phi,\phi\rangle=1$. 
From the sequence $(y_{\omega})_{\omega\in\mathbf{W}}$, we can define a linear functional $L(f)=\sum_{\omega\in\mathbf{W}}f_{\omega}y_{\omega}$. With this functional $L$ (the sequence $(y_{\omega})_{\omega\in\mathbf{W}}$ is skipped), we give the n.c. analogical definition of moment and localising matrix.
\begin{definition}[n.c. moment matrix and localising matrix]\label{def:moment_matrix_nc}
Consider the finite dimensional case, a symmetric linear functional $L:\mathbb{R}[\mathbf{X}]_{2d}\to\mathbb{R}$ is in bijective corresponse to a Hankel matrix $\mathbf{M}_d\in\mathbb{R}^{\sigma_{nc}(d)\times\sigma_{nc}(d)}$, indexed by monomials $u,\nu\in\mathbf{W}_{d}$, such that 
\begin{equation*}
\mathbf{M}_d(u,\nu):=L(u^{\dag}\nu),\forall\; u,\nu\in\mathbf{W}_{d}.
\end{equation*}
Given $\mathbf{g}\subseteq\Sym\mathbb{R}[\mathbf{X}]$, we define a localising matrix $\mathbf{M}_{d}(g_i)\in\mathbb{R}^{\sigma_{nc}(d)\times\sigma_{nc}(d)}$ for $g_i\in\mathbf{g}$, which is indexed by monomials $u,\nu\in\mathbf{W}_{d}$, such that
\begin{equation*}
\mathbf{M}_{d}(g_i)(u,\nu):=L(u^{\dag}g_i\nu),\forall\;u,\nu\in\mathbf{W}_{d}.
\end{equation*}
\end{definition}



\begin{definition}[The dual cones]{\cite{burgdorf2016optimization}, Lemma~1.44 and Remark~1.64.}\label{def:dual cones}
If given $\mathbf{g}\subseteq\Sym\mathbb{R}[\mathbf{X}]$, we define the dual cone $\left(\Theta_{d}(\mathbf{g})\right)^{\vee}$ to consist of symmetric linear functionals which are nonnegative on $\Theta_{d}(\mathbf{g})$ and zero on commutators.
Using the Hankel matrix $\mathbf{M}_d,\mathbf{M}_{d-\lceil\deg(g_i)/2\rceil}(g_i)$, for $g_i\in\mathbf{g}$, the dual cone can also be rewritten as a set of LMI.
\begin{eqnarray*}
\left(\Theta_{d}(\mathbf{g})\right)^{\vee}:&=&\{L:\mathbb{R}[\mathbf{X}]_{2d}\to\mathbb{R}\ \;|\; L \textrm{ linear}, L(f)=L(f^{\dag}),L(f)\geq 0, \forall f\in\Theta_{d}(\mathbf{g})\}.\nonumber\\
&\cong&\{\mathbf{M}_d\;|\;\mathbf{M}_d\succcurlyeq 0; \mathbf{M}_{d-\lceil\deg(g_i)/2\rceil}(g_i)\succcurlyeq 0,\;\forall g_i\in\mathbf{g}; (\mathbf{M}_d)_{u,v}=(\mathbf{M}_d)_{\omega,z},\; \forall u^{\dag}v \stackrel{\cyc}{\sim} \omega^{\dag}z \}.
\end{eqnarray*}
\end{definition}

Next, we derive the dual problem of \eqref{equ:trace-NCPO-sup-Theta} for $2d\geq\min_{g\in\mathbb{R}[\mathbf{X}], g\stackrel{\cyc}{\sim} f}\deg g$: 
\begin{eqnarray}
\tr_{\Theta_{d}}(f,\mathbf{g})&:=&
\begin{array}{cl}
\sup_{a} & a  \\
\textrm{s.t.} & f-a\in\Theta_{d}(\mathbf{g})
\end{array} \nonumber \\
&=&\sup_a \inf_{L\in\left(\Theta_{d}(\mathbf{g})\right)^{\vee}} \left\{ a + L(f-a) \right\}
\label{equ:cone-transfer} \\
&\leq &\inf_{L\in\left(\Theta_{d}(\mathbf{g})\right)^{\vee}}\sup_a \left\{ a + L(f-a)\right\} \label{equ:weak duality}\\
&=& \inf_{L\in\left(\Theta_{d}\right)^{\vee}}\sup_a \left\{a + L(f)-a L(1) \right\}\label{equ:linearity of L}\\
&=& \inf_{L\in\left(\Theta_{d}(\mathbf{g})\right)^{\vee}}  \left\{ L(f) +\sup_a \{a(1- L(1))\} \right\}\label{equ:inner problem}\\
&=&\begin{array}{cl}
\inf & L(f)  \\
\textrm{s.t.} & L\in\left(\Theta_{d}(\mathbf{g})\right)^{\vee}\\ & L(1)=1.
\end{array}
=:\trd_{\Theta_{d}}(f,\mathbf{g}).
\tag{NCPOP-DSDP}\label{equ:trace-NCPO-dual}
\end{eqnarray}
Equation~\eqref{equ:cone-transfer} uses the Definition~\ref{def:dual cones} to transfer the cone constraint. Equation~\eqref{equ:weak duality} utilises the weak duality, since the variable $a$ is unbounded. Equation~\eqref{equ:linearity of L} is due to the linearity of the functional $L$. Finally, the inner problem in Equation~\eqref{equ:inner problem} is bounded only if $L(1)=1$, such that we obtain the dual problem in Equation~\eqref{equ:trace-NCPO-dual}, which is an SDP due to Definition~\ref{def:dual cones}.

\subsection{Flatness, Finite Convergence and Optimiser Extraction}

So far, we have defined the problems $\trd_{\Theta_{d}}(f,\mathbf{g})$, $\tr_{\Theta_{d}}(f,\mathbf{g})$, $\tr^{\II}_{\min}(f,\mathbf{g})$ and $\tr_{\min}(f,\mathbf{g})$. What are the relationships among them at a finite moment order $d$ such that $2d\geq\min_{g\in\mathbb{R}[\mathbf{X}], g\stackrel{\cyc}{\sim} f}\deg g$? 
We first show the strong duality between $\trd_{\Theta_{d}}(f,\mathbf{g})$, $\tr_{\Theta_{d}}(f,\mathbf{g})$, whose objective values form a monotonically increasing sequence and bounded above by $\tr^{\II}_{\min}(f,\mathbf{g})$.

\begin{lemma}{\cite{burgdorf2016optimization}, Proposition 1.58.}\label{lem:closed_module}
Given $\mathbf{g}\subset\Sym\mathbb{R}[\mathbf{X}]$.
Assume that the semi-algebraic set $\mathscr{D}_{\mathbf{g}}$ in Definition~\ref{def:semi-algebraic-nc} contains an $\epsilon$-neighbourhood of $0$. Then the cyclic quadratic module $\Theta_d(\mathbf{g})$ is a closed convex cone in the finite dimensional real vector space $\mathbb{R}[\mathbf{X}]$.
In particular, if 
$f\in\Sym\mathbb{R}[\mathbf{X}]_{2d}\setminus\Theta_d(\mathbf{g})$, then there exists a linear functional $L:\mathbb{R}[\mathbf{X}]\to\mathbb{R}$ that is zero on commutators, nonnegative on $\Theta_d(\mathbf{g})$, positive on $\Sigma[\mathbf{X}]_{2d}\setminus\{0\}$ and $L(f)<0$.
\end{lemma}

\begin{theorem}[strong duality]{\cite{burgdorf2016optimization}, Proposition~5.10 and Theorem~4.1.}\label{the:strong_duality_ncpop} 
Assume that the semi-algebraic set $\mathscr{D}_{\mathbf{g}}$ in Definition~\ref{def:semi-algebraic-nc} contains an $\epsilon$-neighbourhood of $0$.
For every $d\geq 1$, it holds that $\trd_{\Theta_{d}}(f,\mathbf{g})=\tr_{\Theta_{d}}(f,\mathbf{g})$.
\begin{proof}
From weak duality, we have $\trd_{\Theta_{d}}(f,\mathbf{g})\geq\tr_{\Theta_{d}}(f,\mathbf{g})$.
The dual problem $\trd_{\Theta_{d}}(f,\mathbf{g})$ is always feasible, with the moment matrix $\mathbf{M}_d$ being one in the $(1,1)$-entry and zero otherwise. Hence $\trd_{\Theta_{d}}(f,\mathbf{g})<\infty$.
When the primal problem $\tr_{\Theta_{d}}(f,\mathbf{g})$ is feasible, i.e., $\tr_{\Theta_{d}}(f,\mathbf{g})>-\infty$. From the definition of $\trd_{\Theta_{d}}(f,\mathbf{g})$, the following holds
\begin{equation*}
L(f)-\trd_{\Theta_{d}}(f,\mathbf{g})=L\left(f-\trd_{\Theta_{d}}(f,\mathbf{g})\right)\geq 0,\;\forall L\in\left(\Theta_{d}(\mathbf{g})\right)^{\vee},
\end{equation*}
such that $f-\trd_{\Theta_{d}}(f,\mathbf{g})$ belongs to the closure of $\Theta_{d}(\mathbf{g})$. By Lemma~\ref{lem:closed_module}, the cyclic quadratic module $\Theta_{d}(\mathbf{g})$ is a closed convex cone, such that $f-\trd_{\Theta_{d}}(f,\mathbf{g})\in\Theta_{d}(\mathbf{g})$. Then, from the definition of $\tr_{\Theta_{d}}(f,\mathbf{g})$, it holds that $\tr_{\Theta_{d}}(f,\mathbf{g})\geq \trd_{\Theta_{d}}(f,\mathbf{g})$. Combined with weak duality, the strong duality holds.
When the primal problem $\tr_{\Theta_{d}}(f,\mathbf{g})$ is infeasible. Then for any $a\in\mathbb{R}$, $f-a$ is not an element of $\Theta_{d}(\mathbf{g})$. By Lemma~\ref{lem:closed_module}, there is a functional $L$ satisfying the constraints in the dual problem $\trd_{\Theta_{d}}(f,\mathbf{g})$ and $L(f-a)<0$, such that $L(f)<a$ for an arbitrary $a\in\mathbb{R}$. Hence, the dual problem is unbounded.
\end{proof}
\end{theorem}

\begin{theorem}[convergence]{ \cite{burgdorf2016optimization}, Corollary~5.5.}\label{the:convergence-ncpop}
Given $\mathbf{g}\cup\{f\}\subset\Sym\mathbb{R}[\mathbf{X}]$. Let the Archimedean assumption hold. Then the following holds
\begin{equation*}
\lim_{d\to\infty}\trd_{\Theta_{d}}(f,\mathbf{g})=\lim_{d\to\infty}\tr_{\Theta_{d}}(f,\mathbf{g})=\tr^{\II}_{\min}(f,\mathbf{g}).
\end{equation*}
\begin{proof}
Following Equation~\eqref{equ:trace-NCPO-sup-Theta} and strong duality in Theorem~\ref{the:strong_duality_ncpop}, we know that 
\begin{equation}
\trd_{\Theta_{d}}(f,\mathbf{g})=\tr_{\Theta_{d}}(f,\mathbf{g})\leq\tr^{\II}_{\min}(f,\mathbf{g}),\label{equ:convergence-ncpop-forward}
\end{equation}
where the first inequality comes from Theorem~\ref{the:cyclic2positivetrace}. According to Theorem~\ref{the:positivetrace2cyclic}, we know that for all $\epsilon>0$, there exists $d(\epsilon)\in\mathbb{N}$, such that 
\begin{equation*}
f-\tr^{\II}_{\min}(f,\mathbf{g})+\epsilon\in\Theta_{d(\epsilon)}(\mathbf{g}).
\end{equation*}
Using the definition of $\tr_{\Theta_{d}}(f,\mathbf{g})$ in Equation~\eqref{equ:trace-NCPO-sup-Theta}, for all $\epsilon>0$, there exists $d(\epsilon)\in\mathbb{N}$, such that
\begin{equation}
\tr^{\II}_{\min}(f,\mathbf{g})-\epsilon\leq\tr_{\Theta_{d}}(f,\mathbf{g}).\label{equ:convergence-ncpop-backward}
\end{equation}
Combing Equations~(\ref{equ:convergence-ncpop-forward}--\ref{equ:convergence-ncpop-backward}) yields the desired conclusion.
\end{proof}
\end{theorem}

In general, the convergence is not finite. However, the n.c. analog of Curto-Fialkow's theorem shows that when a flatness condition is satisfied, at a finte order $d$, the objective values of these problems are the same. Then, the \acrfull{GNS} construction could be applied to extract the solutions for NCPOP.

\begin{definition}[$\eta$-flatness]{\cite{burgdorf2016optimization}, Definition 1.49.}\label{def:flat}
Given an integer $\eta\geq 1$. 
Let $\mathbf{M}_{d}$ and $\mathbf{M}_{d+\eta}$ be the optimal solutions of problems $\trd_{\Theta_{d}}(f,\mathbf{g})$ and $\trd_{\Theta_{2d+2\eta}}(f,\mathbf{g})$. Then $\eta$-flatness is reached if
\begin{equation*}
\rk\mathbf{M}_{d}=\rk\mathbf{M}_{d+\eta}.
\end{equation*}
\end{definition}

\begin{theorem}[\acrshort{GNS} construction]{\cite{burgdorf2016optimization}, Theorem 1.69.}\label{the:GNS}
Given $\mathbf{g}\subset\Sym\mathbb{R}[\mathbf{X}]$, semi-algebraic set $\mathscr{D}_{\mathbf{g}}$ in Definition~\ref{def:semi-algebraic-nc}.
Let the Archimedean assumption hold.
Let $\eta:=\max\{\lceil\deg(g_i)/2\rceil,\forall g_i\in\mathbf{g}; 1\}$. Let $\mathbf{M}_{d}$ and $\mathbf{M}_{d+\eta}$ be the optimal solutions of problems $\trd_{\Theta_{d}}(f,\mathbf{g})$ and $\trd_{\Theta_{d+\eta}}(f,\mathbf{g})$. Suppose $\eta$-flatness in Definition~\ref{def:flat} is reached, i.e., $r:=\rk\mathbf{M}_{d}=\rk\mathbf{M}_{d+\eta}$. Let $L$ be the linear functional associated with the moment matrix $\mathbf{M}_{d+\eta}$. There exists a tuple of matrices $\mathbf{X}\in\mathscr{D}_{\mathbf{g}}$, with the dimension of each matrix being $r$, and a unit vector $\phi$ such that for $f\in\Sym\mathbb{R}[\mathbf{X}]_{2d}$:
\begin{equation*}
L(f)=\langle\phi,\;f(\mathbf{X})\phi\rangle.
\end{equation*}
\begin{proof}
In finite dimensional cases, the linear functional $L$ induces a positive definite inner product $\langle f,g\rangle\mapsto L(f^{\dag} g)$ on the Hilbert space $\mathscr{H}=\mathbb{R}[\mathbf{X}]_{2d}$. 
These multiplication maps are well defined, symmetric and linear. If we fix an orthonormal basis of $\mathscr{H}$ and let $A_i$ denote the representative of the left multiplication by $X_i$ in the algebra of bounded operators $\mathscr{B(H)}$ with respect to this basis.
Since the Archimedean assumption holds, there exists an $N\in\mathbb{N}$, such that $N-\|\mathbf{X}\|^2\in Q(\mathbf{g})$, and
$L(f^{\dag}(N-X^2_i)f)\geq 0$ for all $i=1,\dots,n$. Hence
\begin{equation*}
0\leq \langle X_i f,X_i f\rangle=L(f^{\dag}X^2_i f)\leq NL(f^{\dag}f),
\end{equation*}
such that these maps are bounded. Thus it extends uniquely to a bounded operator on $\mathscr{H}$, which we denote by the same symbol $A_i$.
Further, 
\begin{equation*}
\langle X_i f,X_i\rangle=L((X_i f)^{\dag}f)=L(f^{\dag}(X_i f))=\langle X_i,X_i f\rangle,
\end{equation*}
such that these maps are symmetric, $\mathbf{A}=\{A_1,\dots,A_n\}\in\mathscr{D}_{\mathbf{g}}$.
Now, given a n.c. polynomial $f\in\Sym\mathbb{R}[\mathbf{X}]_{2d}$, 
\begin{equation*}
    L(f)=\langle f,1\rangle=\langle \sum_{\omega}f_{\omega}\omega,1\rangle=\sum_{\omega}f_{\omega}\langle\omega,1\rangle=\sum_{\omega}f_{\omega}\langle\phi,\omega(\mathbf{A})\phi\rangle=\langle\phi, f(\mathbf{A})\phi\rangle,
\end{equation*}
where $\phi$ denotes the vector corresponding to the identity polynomial $1$. The fact that $L(1)=1$ implies that the vector is normalised, $\langle\phi,\phi\rangle=1$.

\end{proof}
\end{theorem}

\begin{theorem}[finite convergence]{\cite{burgdorf2016optimization}, Theorem 5.12 and 5.4.}\label{the:finite-convergence-ncpop}
Given the same setting in Theorem~\ref{the:GNS}. Given the objective function $f\in\Sym\mathbb{R}[\mathbf{X}]_{2d}$.
There are finitely many tuples $\mathbf{X}(j)$ of symmetric matrices in $\in\mathscr{D}_{\mathbf{g}}$ with positive scalars $\lambda_j>0$ with $\sum_{j}\lambda_j=1$, such that
\begin{equation*}
L(f)=\sum_{j}\lambda_j \tr f(\mathbf{X}(j)).
\end{equation*}
In particular, $\trd_{\Theta_{d}}(f,\mathbf{g})=\tr_{\Theta_{d}}(f,\mathbf{g})=\tr^{\II}_{\min}(f,\mathbf{g})=\tr_{\min}(f,\mathbf{g})$.
\begin{proof}
Suppose $\eta$-flatness in Definition~\ref{def:flat} is reached at moment degree $d$, with the optimal solution $L$ of $\trd_{\Theta_{d}}(f,\mathbf{g})$.
Following Equation~\eqref{equ:trace-NCPO-sup-Theta} and strong duality in Theorem~\ref{the:strong_duality_ncpop}, we know that 
\begin{equation*}
\trd_{\Theta_{d}}(f,\mathbf{g})=\tr_{\Theta_{d}}(f,\mathbf{g})\leq\tr^{\II}_{\min}(f,\mathbf{g})\leq\tr_{\min}(f,\mathbf{g}),
\end{equation*}
where the first inequality comes from Theorem~\ref{the:cyclic2positivetrace} and the second inequality comes from the fact that $\mathscr{D}_{\mathbf{g}}\subseteq\mathscr{D}^{\II}_{\mathbf{g}}$. 
Using Gelfand-Naimark-Segal construction
in Theorem~\ref{the:GNS}, and Wedderburn theorem, the functional $L$ has the tracial representation $L(f)=\sum_{j}\lambda_j \tr f(\mathbf{X}(j))$ for $f\in\mathbb{R}[\mathbf{X}]_{2d}$,
with the scalars $\lambda_j>0$, $\sum_{j}\lambda_j=1$ and finitely many tuples $\mathbf{X}(j)\in\mathscr{D}_{\mathbf{g}}$. Hence,
\begin{equation*}
\trd_{\Theta_{d}}(f,\mathbf{g})=L(f)=\sum_{j}\lambda_j \tr f(\mathbf{X}(j)) \geq \tr_{\min}(f,\mathbf{g}).
\end{equation*}
Combined the above equations, we conclude that $\trd_{\Theta_{d}}(f,\mathbf{g})=\tr_{\Theta_{d}}(f,\mathbf{g})=\tr^{\II}_{\min}(f,\mathbf{g})=\tr_{\min}(f,\mathbf{g})$.
\end{proof}
\end{theorem}

\section{Sparse Representation of NCPOP}\label{sec:sparse_ncpop}

Suppose the number of variables is $n$.
At a given moment order $d$, the cardinality of monomial basis, is $\sigma(d)=\binom{n+d}{d}$ for GMP or POP, and $\sigma_{nc}(d)=\frac{n^{d+1}-1}{n-1}$ for NCPOP.
Thus, it grows exponentially with the moment degree $d$. To lower the computational burden of these hierarchies of SDP relaxations, we can exploit the sparsity patterns in these SDP problems, cf. \cite{magron2023sparse,klep2022sparse,wang2021exploiting}. This section is based on \cite{klep2022sparse} for trace optimisation exploiting correlative sparsity.

Sparse representations of GMP, POP, NCPOP typically rely on the running intersection property. In terms of trace minimisation in Section~\ref{sec:ncpop}, this assumption is defined as follows. 
\begin{assumption}[running intersection property]{\cite{klep2022sparse}, Assumption~2.4.}\label{ass:rip}
Define the index set of variables $I_0:=\{1,\dots,n\}$. For $p\in\mathbb{N}$, consider $I_1,\dots,I_p\subset I_0$ satisfying $\cup^p_{q=1}I_q=I_0$. We denote by $\mathbb{R}[\mathbf{X}(I_q)]$ the set of polynomials in the variables $\{X_i\in\mathbf{X} |i\in I_q\}$. Note that $\mathbb{R}[\mathbf{X}(I_0)]=\mathbb{R}[\mathbf{X}]$. 
Given $\mathbf{g}=\{g_j\}^m_1\subset\Sym\mathbb{R}[\mathbf{X}]$. Define the index set of constraints $J:=\{1,\dots,m\}$. Let $J$ be partitioned into $p$ disjoint sets $J_1,\dots,J_p$. The running intersection property holds if the two collections $\{I_1,\dots,I_p\}$ and $\{J_1,\dots,J_p\}$ satisfy the following conditions.
\begin{itemize}
\item For all $j\in J_q$, $g_j\in\mathbb{R}[\mathbf{X}(I_q)]\cap\Sym\mathbb{R}[\mathbf{X}]$.
\item The objective function can be decomposed as $f=f_1+\dots +f_p$, with $f_q\in\mathbb{R}[\mathbf{X}(I_q)]$, for $q=1,\dots,p$.
\item For all $q=1,\dots,p-1$, it satisfies $I_{q+1}\cap \bigcup_{j\leq q} I_j\subseteq I_{\ell}$, for some $\ell\leq q$.
\end{itemize}
\end{assumption}

Under this property, the absence of coupling of variables can be preserved in a sparse representation \citep[Section~2.7]{lasserre2009moments}. 
Let the two collections $\{I_1,\dots,I_p\}$ and $\{J_1,\dots,J_p\}$ as in Assumption~\ref{ass:rip}.
Given $\mathbf{g}\subset\Sym\mathbb{R}[\mathbf{X}]$.
We define the sparse variant of quadratic modules for $q\in 1,\dots,p$
\begin{equation*}
Q^{q}(\mathbf{g}):=\left\{\sum^{N}_{\ell=1} f_{\ell}^{\dag}f_{\ell}+\sum_{i} \sum^{N_i}_{j=1}  h_{ij}^{\dag}g_i h_{ij} \;|\; g_i\in \mathbf{g}\cap\mathbb{R}[\mathbf{X}(I_q)];N,N_i\in\mathbb{N};  f_{\ell},h_{ij}\in\mathbb{R}[\mathbf{X}(I_q)]\right\},
\end{equation*}
with the truncated version
\begin{equation}
Q_d^{q}(\mathbf{g}):=Q^q(\mathbf{g})\cap\mathbb{R}[\mathbf{X}(I_q)]_{2d}.
\label{equ:truncated_quadratic_sparse}
\end{equation}

We define the sparse variant of truncated cyclic quadratic modules
\begin{equation}
\Theta_d^{q}(\mathbf{g}):=\{f\in\Sym\mathbb{R}[\mathbf{X}]_{2d}\;|\:\exists h\in Q_d^{q}(\mathbf{g}): f\stackrel{\cyc}{\sim} h\},
\label{equ:truncated_cyc_quadratic_sparse}
\end{equation}
as well as the sum
\begin{equation*}
\Theta_d^{sparse}(\mathbf{g}):=\Theta_d^{1}(\mathbf{g})+\dots+\Theta_d^{p}(\mathbf{g}).
\end{equation*}

\begin{theorem}[sparse version of Theorem~\ref{the:cyclic2positivetrace}--\ref{the:positivetrace2cyclic}]{\cite{klep2022sparse}, Proposition~6.5.}\label{the:positivetrace2cyclic_sparse}
Let $\mathbf{g}\cup\{f\}\subseteq\Sym\mathbb{R}[\mathbf{X}]$.
Let $\Theta^{q}(\mathbf{g}):=\cup_{d\in\mathbb{N}} \Theta_d^{q}(\mathbf{g})$ and $\Theta^{sparse}(\mathbf{g})=\cup_{d\in\mathbb{N}} \Theta_d^{sparse}(\mathbf{g})$.
Let the running intersection property assumption hold. Suppose for each $k=1,\dots,p$,   $Q^q(\mathbf{g})$ in  \eqref{equ:truncated_quadratic_sparse} is Archimedean, that is $\exists N>0: N-\|\mathbf{X}(I_q)\|^2\in Q^q(\mathbf{g})$.
$\tr f(X)\geq 0,\;\forall X\in\mathscr{D}^{\II}_{\mathbf{g}}$, if and only if $ f+\epsilon\in\Theta^{sparse}(\mathbf{g})$, for all $\epsilon>0$.
\end{theorem}

We define the sparse variant of SDP relaxations of trace minimisation
\begin{eqnarray*}
\tr^{sparse}_{\Theta_{d}}(f,\mathbf{g})&:=&\sup\{a\;|\;f-a\in\Theta^{sparse}_{d}(\mathbf{g})\},\\
\trd^{sparse}_{\Theta_{d}}(f,\mathbf{g})&:=&\inf\{L(f)\;|\;L\in\left(\Theta^{sparse}_{d}(\mathbf{g})\right)^{\vee},L(1)=1\}.
\end{eqnarray*}

Under the Archimedean assumption and the running intersection property, Theorem~\ref{the:positivetrace2cyclic_sparse} implies convergence
\begin{equation*}
\lim_{d\to\infty}\trd^{sparse}_{\Theta_{d}}(f,\mathbf{g})=\lim_{d\to\infty}\tr^{sparse}_{\Theta_{d}}(f,\mathbf{g})=\tr^{\II}_{\min}(f,\mathbf{g}),
\end{equation*}
where the proof is similar to Theorem~\ref{the:convergence-ncpop}, or see Proposition~6.6 in \cite{klep2022sparse}. Further, when flatness condition is satisfied at moment order $d$, and the irreducibility condition (cf. Theorem 4.2 in \cite{klep2022sparse}) holds, the finite convergence is reached
\begin{equation*}
\trd^{sparse}_{\Theta_{d}}(f,\mathbf{g})=\tr^{sparse}_{\Theta_{d}}(f,\mathbf{g})=\tr^{\II}_{\min}(f,\mathbf{g})=\tr_{\min}(f,\mathbf{g}),
\end{equation*}
where the proof is similar to Theorem~\ref{the:finite-convergence-ncpop}, or see Proposition~6.7 in \cite{klep2022sparse}.

\chapter{Proofs for Chapter~\ref{cha:ot} Claims}
\label{app:ot}

\section{Proof of Lemma~\ref{lem:tx|us}}
\label{app:lem:tx|us}
The right equation follows from the definition of coupling, as in Equation~\eqref{equ:Pi-define}. We only prove the left equation.
Since the projection $\mathscr{T}$ from $X$ to $\td{X}$ is irrelevant to $S$, the following holds
\begin{equation}
\pover{\td{X}=j}{X=i,S=s}=\pover{\td{X}=j}{X=i}, \forall i\in\supp{X_s}.
\end{equation}
Using Bayes theorem, for $i\in\supp{X_s}$:
\begin{align*}
\pp{X=i,\td{X}=j,S=s}&=\pover{\td{X}=j}{X=i,S=s}\pp{X=i,S=s}\\
&=\pover{\td{X}=j}{X=i}\pp{X=i,S=s}\\
&=\pp{X=i,\td{X}=j}\frac{\pp{X=i,S=s}}{\pp{X=i}}\\
&=\pp{X=i,\td{X}=j}\pover{S=s}{X=i}.
\end{align*}
Further, 
\begin{align*}
\pover{\td{X}=j}{S=s}&=\frac{\pp{\td{X}=j,S=s}}{\pp{S=s}}=\frac{\sum_{i\in\supp{X_s}}\pp{X=i,\td{X}=j,S=s}}{\pp{S=s}}\\
&=\frac{\sum_{i\in\supp{X_s}}\pp{X=i,\td{X}=j}\pover{S=s}{X=i}}{\pp{S=s}}\\
&=\sum_{i\in\supp{X_s}}\pp{X=i,\td{X}=j}\frac{\pover{S=s}{X=i}}{\pp{S=s}}\\
&=\sum_{i\in\supp{X_s}}\pp{X=i,\td{X}=j}\frac{\pp{S=s,X=i}}{\pp{X=i}\pp{S=s}}\\
&=\sum_{i\in\supp{X_s}}\pp{X=i,\td{X}=j}\frac{\pover{X=i}{S=s}}{\pp{X=i}}.
\end{align*}
We obtain 
\begin{equation*}
\pover{\td{X}=j}{S=s}=\sum_{i\in\supp{X_s}}\pp{X=i,\td{X}=j}\frac{\pover{X=i}{S=s}}{\pp{X=i}},\forall  j\in\supp{\td{X}},s\in\supp{S}.\label{equ:tx_from_coupling}
\end{equation*}
Note that $\pover{X=i}{S=s}$ is not defined on $\supp{X}\setminus\supp{X_s}$.
If we set the undefined conditional probability be $\pover{X=i}{S=s}=0$ for $i\in\supp{X}\setminus\supp{X_s}$, and rewrite the equation above in matrix form, we complete the proof.

\section{Proof of Lemma~\ref{lem:S-wise TV}}
\label{app:lem:S-wise TV}
 Using TV distance in Definition~\ref{def:tvdistance}, we can deduce that
\begin{align}
\tv{P^{\td{X}_{s_0}}}{P^{\td{X}_{s_1}}}&=\frac{1}{2}\|P^{\td{X}_{s_0}}-P^{\td{X}_{s_1}} \|_1=\frac{1}{2}\left\|\gamma'\left(\frac{P^{X_{s_0}}-P^{X_{s_1}}}{P^{X}}\right) \right\|_1=\frac{1}{2}\left\|\gamma'V\right\|_1,\label{equ:tv-equivalent}
\end{align}
where the second equation follows from Lemma~\ref{lem:tx|us} such that $P^{\td{X}_{s}}=\frac{P^{X_{s}}}{P^X}$, with the division operator being element-wise.
The third equation follows from the definition of vector $V$ in Theorem~\ref{pro:binary_condition}.
Note that when $V_i=0$, $\gamma_{i,j},j\in\supp{\td{X}}$ could be any value without affecting the distance $\tv{P^{\td{X}_{s_0}}}{P^{\td{X}_{s_1}}}$, such that
\begin{align}
\tv{P^{\td{X}_{s_0}}}{P^{\td{X}_{s_1}}}
&=\frac{1}{2}\sum_{ j\in\supp{\td{X}}}\left|\sum_{i\in\supp{X}}\gamma_{i,j}V_i\right|\\
&\leq\frac{N}{2}\max_{ j\in\supp{\td{X}}} \left|\sum_{i\in\supp{X}}\gamma_{i,j}V_i\right|=\frac{N}{2}\max_{ j\in\supp{\td{X}}} \left|\sum_{i\in\overline{\supp{X}}}\gamma_{i,j}V_i\right|,
\end{align}
where $\overline{\supp{X}}:=\{i\in\supp{X}\mid V_i\neq 0\}$.
Given the assumption that $V$ has finite $l_1$ norm, we can find a nonnegative vector $\Lambda$ with entries $\Lambda_j:=\left|\sum_{i\in\overline{\supp{X}}}\gamma_{i,j}V_i\right|$, the following holds 
\begin{equation}
-\Lambda_j\leq \sum_{i\in\overline{\supp{X}}} \gamma_{i,j}V_i\leq\Lambda_j,\quad \forall j\in\supp{\td{X}},
\end{equation}
such that the vector $V$ is bounded by $\Lambda$, i.e., $\|\gamma' V\|_1\leq \|\Lambda\|_1$.
Using Equation~\eqref{equ:tv-equivalent}, we know that $\tv{P^{\td{X}_{s_0}}}{P^{\td{X}_{s_1}}}=\left\|\gamma'V\right\|_1\leq\|\Lambda\|_1$.

\section{Proof of Lemma~\ref{lem:subgradient-optimality}}
\label{app:lem:subgradient-optimality}
The proof uses subgradient optimality conditions that can be found in Proposition~5.4.7 in \cite{bertsekas2009convex}.
We start from restating the constrained KL projection as an unconstrained problem:
\begin{equation}
\prox^{KL}_{\mathcal{C}}(\bar{\gamma})=\arg\min_{\gamma\in\mathcal{C}}\kl{\gamma}{\bar{\gamma}}=\arg\min_{\gamma\in\mathbb{R}^{N\times N}_+}\kl{\gamma}{\bar{\gamma}} + \iota_{\mathcal{C}}(\gamma).
\end{equation}
The indicator function of a convex set $\mathcal{C}$ is a convex function, see Section~2 in \cite{ekeland1999convex}. Further, $\iota_{\mathcal{C}}$ is lower semi-continuous if and only if $\mathcal{C}$ is closed. Since $\mathcal{C}$ is a closed convex set (e.g., affine hyperplanes, closed half-spaces bounded by affine hyperplanes), $\iota_{\mathcal{C}}$ is a convex, lower semi-continuous coercive function.
Also, $\kl{\cdot}{\bar{\gamma}}$ is a strictly convex and coercive function.
This operator is uniquely defined by strict convexity.

As $\kl{\cdot}{\bar{\gamma}}$ is a Gateaux differential function with derivative $\nabla\kl{\gamma}{\bar{\gamma}}=\partial\kl{\gamma}{\bar{\gamma}}=\log(\gamma/\bar{\gamma})$.
Then the necessary and sufficient condition for $\gamma^*$ being the minimiser is
\begin{equation}
\mathbb{0}\in\partial\left(\kl{\gamma^*}{\bar{\gamma}}+\iota_{\mathcal{C}}(\gamma^*)\right)=\log\left(\frac{\gamma^*}{\bar{\gamma}}\right)+\partial\iota_{\mathcal{C}}(\gamma^*),
\end{equation}
where $\partial\iota_{\mathcal{C}}(\gamma^*)$ is the set of all subgradients at $\gamma^*$, and $\mathbb{0}$ is an $N\times N$ zero matrix.
The equation comes from $\dom\; \kl{\cdot}{\bar{\gamma}}\cap \textrm{dom}\;\iota_{\mathcal{C}}\neq\emptyset$ and Moreau-Rockafellar theorem.
To prove this optimality condition, define a function $J(\gamma)=\kl{\gamma}{\bar{\gamma}}+\iota_{\mathcal{C}}(\gamma)$ on $\mathbb{R}^{N\times N}_+$.
Due to $J(\gamma)\geq J(\gamma^*)$ for all $\gamma\in\mathbb{R}^{N\times N}_+$, if $\mathbb{0}\in\partial J(\gamma^*)$, then the definition of subgradients in Equation~\eqref{equ:subgradient-define} implies $J(\gamma)\geq J(\gamma^*)+\langle\mathbb{0},(\gamma-\gamma^*)\rangle$ for all $\gamma\in\mathbb{R}^{N\times N}_+$.

\section{Proof of Lemma~\ref{lem:c_1&c_2}}
\label{app:lem:c_1&c_2}
The proof is based on \cite{peyre2015entropic} and \cite{benamou2015iterative}.

Let $\gamma^*=\prox^{KL}_{\mathcal{C}_{\ell}}(\bar{\gamma})$.
For the first case, i.e., $\ell=1$, note that 
\begin{equation}
\iota_{\mathcal{C}_1}(\gamma)=\iota_{P^X}(\gamma\mathbb{1}).    
\end{equation}
Let $u\in\partial\iota_{P^X}(P^*)\subset\mathbb{R}^N$, where $P^*=\gamma^*\mathbb{1}$. According to Lemma~\ref{lem:subgradient-optimality}, the following holds
\begin{equation}
\mathbb{0}=\log\left(\frac{\gamma^*}{\bar{\gamma}}\right)+u\mathbb{1}' \Rightarrow \gamma^*=\diag{\exp{(-u)}}\bar{\gamma}.
\end{equation}
Plug in $\gamma^*\mathbb{1}=P^X$,
\begin{equation}
\gamma^*\mathbb{1}=\diag{\exp(-u)}\bar{\gamma}\mathbb{1}=P^X \Rightarrow
\gamma^*=\diag{\frac{P^{X}}{\bar{\gamma}\mathbb{1}}}\bar{\gamma}.
\end{equation}
For the second case, i.e., $\ell=2$, the optimality condition becomes
\begin{equation}
\mathbb{0}=\log\left(\frac{\gamma^*}{\bar{\gamma}}\right)+\mathbb{1}\nu' \Rightarrow \gamma^*=\bar{\gamma}\diag{\exp{(-\nu)}},
\end{equation}
where $\nu\in\partial\iota_{P^{\td{X}}}((\gamma^*)'\mathbb{1})\subset\mathbb{R}^N$.
Using $(\gamma^*)'\mathbb{1}=P^{\td{X}}$,
\begin{equation}
(\gamma^*)'\mathbb{1}=\diag{\exp(-u)}\bar{\gamma}'\mathbb{1}=P^{\td{X}} \Rightarrow
\gamma^*=\bar{\gamma}\diag{\frac{P^{\td{X}}}{\bar{\gamma}'\mathbb{1}}}.
\end{equation}

\section{Proof of Lemma~\ref{lem:prox_c3}}
\label{app:lem:prox_c3}
Let $\gamma^*=\prox^{KL}_{\mathcal{C}_3}(\bar{\gamma})$. 
First notice that if some parts of $\bar{\gamma}$ already satisfy the constraint in $\mathcal{C}_3$, these parts should carry over to $\gamma^*$ in order to minimise the term $\kl{\gamma^*}{\bar{\gamma}}$ in the objective function of $\prox^{KL}_{\mathcal{C}_3}$.
If $\bar{\gamma}\in\mathcal{C}_3$, it is straightforward $\gamma^*=\bar{\gamma}$. Further, since $\mathcal{C}_3$ has not relevance to these entries $i\notin\overline{\supp{X}}$, $j\in\supp{\td{X}}$ of $\bar{\gamma}$, we know that $\gamma^*_{i,j}=\bar{\gamma}_{i,j}$, for $i\notin\overline{\supp{X}}$, $j\in\supp{\td{X}}$. 
Also, if the constraint $-\Lambda_j\leq\sum_{i\in\overline{\supp{X}}}\bar{\gamma}_{i,j}V_i\leq\Lambda_j$ is satisfied for certain $j^{\dag}\in\supp{\td{X}}$, we should set these entries $\gamma^*_{i,j}=\bar{\gamma}_{i,j}$, for all qualified $j^{\dag}$ and $i\in\supp{X}$.

Therefore, we only discuss the case that $\bar{\gamma}\notin\mathcal{C}_3$ and the entries $(i,j)$ that $i\in\overline{\supp{X}}$ and $j\in\{\supp{\td{X}}\mid[\bar{\gamma}'V]_j\notin[-\Lambda_j,\Lambda_j]\}$.

According to Lemma~\ref{lem:subgradient-optimality}, the optimality condition becomes
\begin{equation}
\mathbb{0}=\log\left(\frac{\gamma^*}{\bar{\gamma}}\right)+V{\nu}' \Rightarrow \gamma^*=\bar{\gamma}\odot\exp{(-V{\nu}')},
\end{equation}
where $\odot$ is element-wise product and $\nu\in\partial\iota_{\Lambda}(P^*)\subset\mathbb{R}^N$, $P^*=(\gamma^*)'V$. 
The indicator function $\iota_{\Lambda}(P)$ is 0 if $-\Lambda\leq P\leq\Lambda$ and infinity otherwise.
Recall the definition of subgradient in Lemma~\ref{lem:subgradient-optimality}. The following holds
\begin{equation}
\langle \nu, (\gamma-\gamma^*)'V\rangle + \iota_{\Lambda}((\gamma^*)'V)\leq \iota_{\Lambda}(\gamma'V), \forall \gamma\in\mathbb{R}^{N\times N}_+.\label{equ:subgradient-character}
\end{equation}
Now, we discuss the conditions imposed on existence of (at least) one subgradient, such that the subdifferential $\partial\iota_{\Lambda}(P^*)\neq\emptyset$, where $P^*=(\gamma^*)'V$.
If $\gamma^*\notin\mathcal{C}_3$, $ \iota_{\Lambda}((\gamma^*)'V)=+\infty$, there is not such $\nu$ can satisfy Equation~\eqref{equ:subgradient-character}. Hence, we must have $\gamma^*\in\mathcal{C}_3$ to ensure the existence of (at least) one subgradient $\nu$.
Further, the right part of Equation~\eqref{equ:subgradient-character} is $\iota_{\Lambda}(\gamma'V)=0$ if the arbitrary matrix $\gamma\in\mathcal{C}_3$, and $\iota_{\Lambda}(\gamma'V)=\infty$ if the arbitrary matrix $\gamma\notin\mathcal{C}_3$, while this inequality in
Equation~\eqref{equ:subgradient-character} holds for any arbitrary matrix $\gamma$.
We equivalently simplify the condition in Equation~\eqref{equ:subgradient-character} into for all $\gamma\in\mathcal{C}_3$, the following holds
\begin{equation*}
\langle \nu, (\gamma-\gamma^*)'V\rangle\leq 0\Rightarrow \nu' \gamma' V \leq \nu'{(\gamma^*)}' V.
\end{equation*}

Now, we gather these conditions found so far: 
\begin{align}
\boxed{
\nu' \gamma' V \leq \nu'{(\gamma^*)}' V,\forall \gamma\in\mathcal{C}_3;\quad
\gamma^*=\bar{\gamma}\odot\exp{(-V{\nu}')};\quad
\gamma^*\in\mathcal{C}_3.}
\label{equ:c_3_conditions}
\end{align}

\begin{itemize}
\item 
If $\Lambda=\mathbb{0}$, 
given the third condition in Equation~\eqref{equ:c_3_conditions}, 
the first condition holds for any $\nu\in\mathbb{R}$ because $\gamma'V={(\gamma^*)}' V=\mathbb{0}$. The last two conditions require that $\nu$ needs to satisfy
\begin{equation*}
{(\gamma^*)}' V=[\bar{\gamma}\odot\exp{(-V{\nu}')} ]'V =\mathbb{0}.
\end{equation*}
\item
If $\Lambda\neq\mathbb{0}$, to make sure there exists (at least) one subgradient $\nu$ that satisfies the first condition $\nu'\gamma' V \leq \nu'{(\gamma^*)}' V$, for any arbitrary matrix $\gamma\in\mathcal{C}_3$,
the optimal solution $\gamma^*$ must be on the edge of $\mathcal{C}_3$:
\begin{equation}
[{(\gamma^*)}' V]_j=\Lambda_j \textrm{ or } -\Lambda_j, \forall j\in\supp{\td{X}}.\label{equ:nu-condition-0}
\end{equation}
It is obvious that the sign of $\nu_j$ should be the same as the sign of $[{(\gamma^*)}' V]_j$.

Recall that we only focus on the cases that $\bar{\gamma}\notin\mathcal{C}_3$. Hence, there are some $j\in\supp{\td{X}}$:
\begin{equation*}
\sum^N_{i,j=1} \bar{\gamma}_{i,j} V_i > \Lambda_j \textrm{ or } \sum^N_{i,j=1} \bar{\gamma}_{i,j} V_i <-\Lambda_j.
\end{equation*}
As the exponential function is positive, i.e., $\exp{(-V_i\nu_j)}>0$, we cannot change the sign of $[{(\gamma^*)}' V]_j$ by adjusting $\nu$. Hence, the last two conditions in Equation~\eqref{equ:c_3_conditions} imply that the subgradient $\nu$ should satisfy
\begin{equation}
[{(\gamma^*)}' V]_j=\sum^N_{i,j=1} \bar{\gamma}_{i,j}\exp{(-V_i\nu_j)}V_i=
\begin{cases}
\Lambda_j & \textrm{if } \sum^N_{i,j=1} \bar{\gamma}_{i,j} V_i > \Lambda_j\\
-\Lambda_j & \textrm{if } \sum^N_{i,j=1} \bar{\gamma}_{i,j} V_i < -\Lambda_j
\end{cases}
,\forall j\in\supp{\td{X}}.
\label{equ:nu_j-app}
\end{equation}
\end{itemize}

Once the subgradient $\nu$ is found, plug in the second condition in Equation~\eqref{equ:c_3_conditions}, we find the optimal solution $\gamma^*$.

Please be aware that the first and third conditions in Equation~\eqref{equ:c_3_conditions} guarantees the existence of subgradient $\nu$, i.e., the subdifferential is not empty, while we also need (at least) one $\nu$ in the subdifferential to satisfy the second condition (i.e., the optimality condition).

\end{appendices}

\addcontentsline{toc}{chapter}{Bibliography}
\bibliographystyle{apalike}
\bibliography{optimisation/opt,fairness/ref}

\end{document}